\documentclass[12pt]{article}
\usepackage{graphicx}
\usepackage{amssymb} 
\graphicspath{ {./images/} }

\begin{document}

\author{Mark A. Atkins, Ph.D.\\
aginik@outlook.com}
\title{Tumbug: A pictorial, universal knowledge representation method}
\maketitle

\begin{abstract}
Since the key to artificial general intelligence (AGI) is commonly believed to be commonsense reasoning (CSR) or, roughly equivalently, discovery of a knowledge representation method (KRM) that is particularly suitable for CSR, the author developed a custom KRM for CSR. This novel KRM called Tumbug was designed to be pictorial in nature because there exists increasing evidence that the human brain uses some pictorial type of KRM, and no well-known prior research in AGI has researched this KRM possibility. Tumbug is somewhat similar to Roger Schank's Conceptual Dependency (CD) theory, but Tumbug is pictorial and uses about 30 components based on fundamental concepts from the sciences and human life, in contrast to CD theory, which is textual and uses about 17 components (= 6 Primitive Conceptual Categories + 11 Primitive Acts) based mainly on human-oriented activities. All the Building Blocks of Tumbug were found to generalize to only five Basic Building Blocks that exactly correspond to the three components \{O, A, V\} of traditional Object-Attribute-Value representation plus two new components \{C, S\}, which are Change and System. Collectively this set of five components, called "SCOVA," seems to be a universal foundation for all knowledge representation.
\end{abstract}

\tableofcontents

\section{Introduction}

A major underlying problem with all natural language translation is that there does not exist any universally accepted knowledge representation method (KRM) for all natural languages. This problem in turn appears to be based on a more fundamental problem: No known method exists, whatsoever, of representing semantics (meaning) of sentences in any natural language.

\begin{figure}
	\begin{center}
	\includegraphics[width=0.75\textwidth]{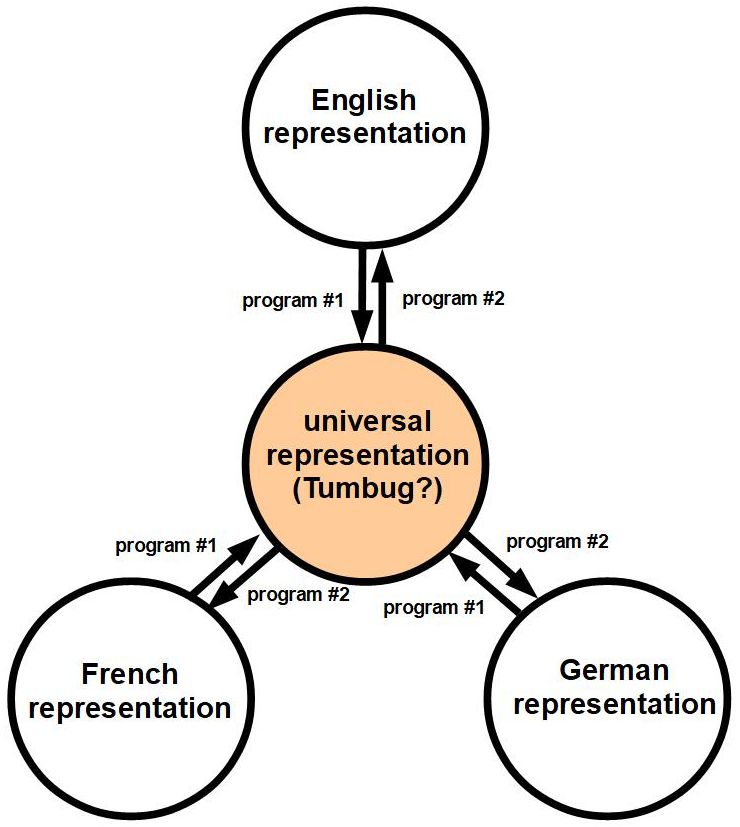}
	\caption{First alternative for language translation: This strategy uses one universal representation, and would need only two programs for three languages.}
	\label{fig:translation-common-rep}
	\end{center}
\end{figure}

Figure~\ref{fig:translation-common-rep} shows that a universal representation would require exactly two steps for any language conversion: one step from the source language representation to the universal representation, then one more step to the target language representation. Only 1-2 types programs would be needed, namely 1-2 program(s) that converted between the universal representation and any specific language representation. This is a diagrammatic way to express an "interlingual conceptual base," as Roger Schank described it (Schank 1972, pp. 553-554).

\begin{figure}
	\begin{center}
	\includegraphics[width=0.75\textwidth]{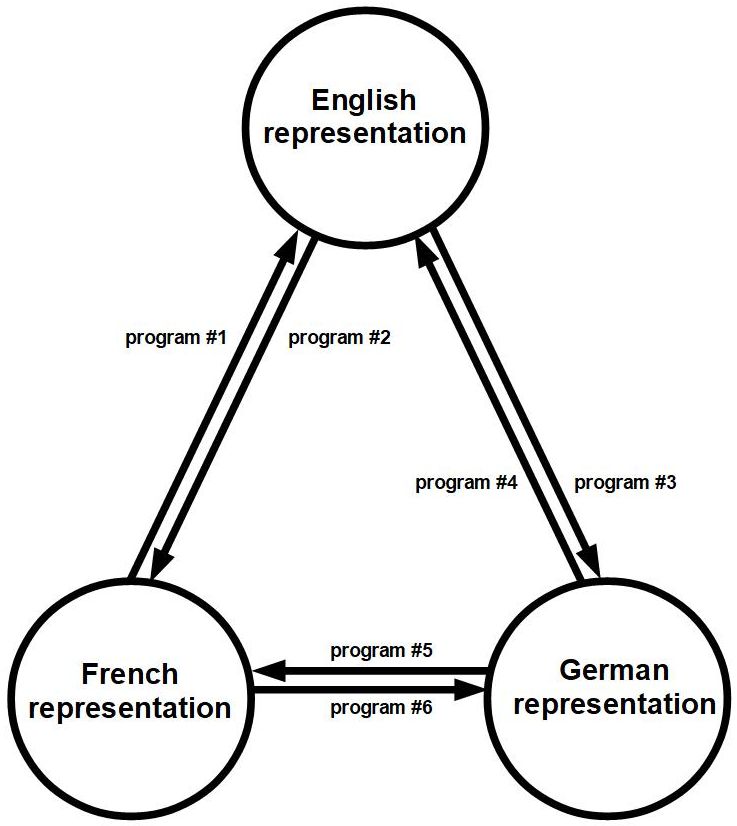}
	\caption{Second alternative for language translation: This strategy uses no universal representation, and would need six programs for three languages.}
	\label{fig:translation-separate-programs}
	\end{center}
\end{figure}

Figure~\ref{fig:translation-separate-programs} shows an extreme alternative: only one step would be needed to convert from the source language representation to the target language representation, but six total programs would be needed for arbitrary conversion between three languages. In the first alternative, n languages would always require only 1-2 types of programs, whereas in the second alternative, n languages would always require $n^{2} - n$ separate programs. The first type of program would be concerned with only one type of mapping, the second type would be concerned with n types of mappings, which suggests the first type scales better and generalizes better, and is therefore ultimately a better solution that saves on computer coding effort. Another motivation for the universal approach is that it would likely lead to insights into the nature of language in general, which in turn might also lead to insights into the relationships between language and other fields such as physics, math, computer programming, and neuroscience.

This article presents a new KRM named "Tumbug" that appears to be universal for representing semantics, and therefore can represent any given sentence in any natural language as a single data structure with appropriately associated values. This structure-with-values combination appears to be invariant for every language used to write a sentence that has the same meaning, provided that the corresponding words have exactly the same meaning across the two languages, and provided that all attributes of the corresponding words across the two languages are the same. Thus Tumbug may implement the "universal representation"--the center circle of the three language circles shown in Figure~\ref{fig:translation-common-rep}.

\textit{Emphasized clarification: This is a theoretical study, not an empirical study. Misinterpretation of the nature of this study probably arises from the common misinterpretation of the term "artificial intelligence" to mean "machine learning," but machine learning is only one subfield within AI, and the study in this document is not about the subfield of machine learning.}

\section{Motivation}

\begin{figure}
	\begin{center}
	\includegraphics[width=0.75\textwidth]{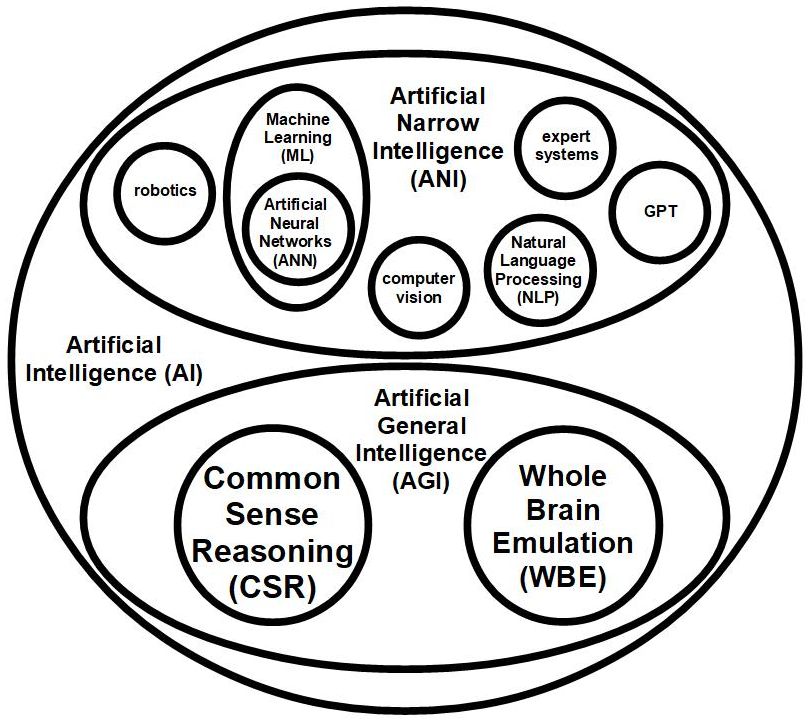}
	\caption{The field of AI is divided into ANI and AGI, and in turn the field of AGI contains CSR. There is currently little or no overlap between ANI and AGI except in trivial ways such as hierarchies or the massive training of large language models (LLMs).}
	\label{fig:venn-big-font}
	\end{center}
\end{figure}

The overall goal of this research is to advance progress in artificial general intelligence (AGI). AGI is the ultimate goal of the larger field of AI in general, although currently only the practical, applied aspects of AI--the subfield of AI called artificial narrow intelligence (ANI)--is well understood, already implemented, and making obvious progress. In contrast, the subfield of AI called AGI is currently languishing and awaiting new ideas that will vitalize it so that it too can be applied to real-life problems. Since AGI progress is usually claimed to be based on Common Sense Reasoning (CSR) progress (e.g., Davis 1990, p. 1; Devlin 1997, p. 167), the most promising subfield of AGI to tackle appears to be CSR. Within CSR, one well-known test of machine understanding of written language is the Winograd Schema (WS). Therefore example sentences from the WS will be used in this study to generate generalities, insights, and examples into the data structures needed for such a universal semantic KRM, and these insights will be implemented as a KRM called "Tumbug." This KRM therefore has the potential to advance the field of AGI. Figure~\ref{fig:venn-big-font} shows the relationship of the main subfields of AI.

\textit{Emphasized clarification: The author has never had any involvement with the Winograd Schema Challenge (WSC), a competition that no longer even exists. The WS is used here only as a convenient, well-known list of benchmark problems in CSR, and its presence in this document should not be interpreted to mean that this research refers to some past, present, or future submission to the WSC by the author.}

A comment about notation: There exist various versions of the Winograd Schema, each of which has a designation "WS" or "WSC" followed by the number of sample problems in that schema (list). The WS version used in this document is called "WS150" because it has 150 problems (Davis 2018), though that designation has not been used elsewhere. Some other named versions of the WS are WSC273 (Kocijan et al. 2022, p. 2), which has 273 problems, and WSC285 (Kocijan et al. 2022, p. 6), which has 285 problems.

\textit{This article's convention: WS150 problems quoted in this document have the second word of the word pair omitted, for simplicity, clarity, and readability. For example, the WS150 sentence "The city councilmen refused the demonstrators a permit because they [feared/advocated] violence" would appear in this document as "The city councilmen refused the demonstrators a permit because they feared violence."}

\section{Overview and background of Tumbug}

The author developed Tumbug from 2021-2023 in an attempt to universalize all the grammars of all the natural languages for the purpose of making foreign language learning easier. The word "Tumbug" is a simpler, more readable version of the acronym "TUMBVG" (Temporal Universal Model-Based Visual Grammar). The term "visual" in the acronym means that diagrams are used in this KRM instead of text, numbers, logic symbols, or statistics, which makes Tumbug an extremely unique KRM of a type that no one appears to have tried yet. The term "model-based" means that this KRM is intended to represent arbitrary models of systems, whether a mechanical system of gears and pulleys, a biological system of organs and bodily fluids, a physical system such as planets orbiting a star, or any other type of system. The term "temporal" means that the models represented also include time, as opposed to static diagrams drawn on paper. Although Tumbug diagrams are typically drawn on paper, Motion Arrows and Time Arrows in Tumbug diagrams show where and how the components are moving relative to each other, which in a good software implementation would literally appear as moving images, such as via simulations or animated GIF files. The word "grammar" originates from Tumbug's originally intended use, namely language translation based on sentence grammatical structure, but Tumbug can be applied to any system whatsoever.

\begin{figure}
	\begin{center}
	\includegraphics[width=0.50\textwidth]{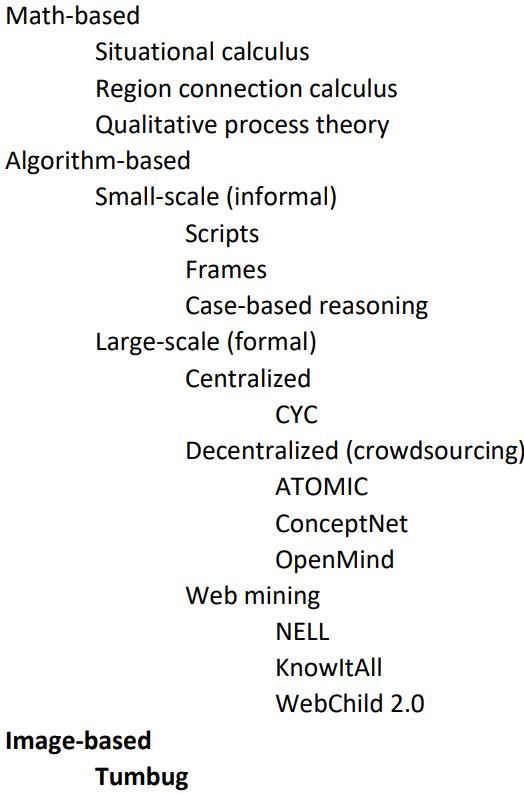}
	\caption{A summary of knowledge-based approaches to CSR. Tumbug is so unique that it has its own category.}
	\label{fig:csr-approaches-list-snap}
	\end{center}
\end{figure}

Figure~\ref{fig:csr-approaches-list-snap} is a taxonomy of the main knowledge-based approaches to CSR, based mostly on a list by Davis and Marcus (2015, p. 11) and a very similar list with more details by Huang et al. (2020, p. 93), but organized here by KRM: mathematics versus algorithm versus images. The highlighted category "Image-based" that includes "Tumbug" is new to this diagram and is the author's approach promoted in this article. Note that Tumbug is not only a completely new approach, but is also an entirely new class of approach for CSR, one that has never been explored yet. This situation suggests that Image-based CSR may contain much new ground for exploration.

So far all code-based approaches have been described as "informal." Tumbug might also be considered "informal" because there is some flexibility in how the visual structures are pieced together, although not much.

\section{Derivation of Tumbug}

\subsection{Knowledge representation methods}

\subsubsection{The importance of KRMs in AI}

The fact that KRMs are extremely important for AI is known to surprisingly few researchers in the field. Related to this, very few AI researchers are working on KRMs.

\begin{itemize}
    \item
    		Marvin Minsky said, 'Well, I'd like to have lots of people thinking about how to \textit{combine} different approaches. It's not the different approaches themselves: it's 'Why aren't there more people making a machine that uses three different representations of knowledge and crosses over?' That's a very specific kind of research project, and I see no one doing it. So that's, to me, that's the missing link." (GBH Archives 1990)
    \item
    		Herbert L. Dreyfus wrote, "A second difficulty shows up in connection with representing the problem in a problem-solving system. It reflects the need or essential/inessential discrimination. Feigenbaum, in discussing problems facing artificial intelligence research in the second decade, calls this problem 'the most important through not the most immediately tractable.'" (Dreyfus 1979, p. 298)
    	\item
    		Han Reichgelt wrote, "Knowledge representation is a major concern in Artificial Intelligence and its importance cannot be overestimated." (Reichgelt 1991, p. 1).
    	\item
    		John Sowa wrote, "A knowledge representation is a fragmentary theory of intelligent reasoning". (Sowa 2000, p. 134) This suggests that a KRM itself, even without being used in a computer program, is already a partial model of intelligent reasoning.
    	\item
    		Laurence Harris and Michael Jenkin wrote, "The representation of knowledge is fundamental to human cognition and machine understanding" (Harris and Jenkin 1997, p. 2).
    	\item
    		Heather Jenkin wrote, "Fundamental to all human cognition is the representation of knowledge" (Jenkin 1997, p. 268). It is therefore not surprising that the field of knowledge representation arose from the field of AI (Sowa 2000, p. XI; Cercone 1987, p. 3).
     \item
    		Haugeland wrote: "In other words, Artificial Intelligence must start by trying to understand knowledge (and skills and whatever else is acquired) and then, on that basis, tackle learning." (Haugeland 1985, p. 11).
\end{itemize}

There have been cases in the history of science when merely switching to a different KRM was the key to major insights and rapid progress, if not immediate solution to a single difficult problem. Some examples are: (1) The switch from the awkward rules of Roman numbers to Arabic numerals, especially when augmented by use of the placeholder digit "0." (2) Use of Feymann diagrams to ease visualization of particle interactions in physics (Fischler and Firschein 1987, p. 69). (3) Standardization of business information in data base format for consistent, non-redundant storage and ease of access through query languages. (4) The switch from a 1D list of elements to the shaped 2D periodic table of the elements. (5) Use of phase portraits to visualize chaotic systems (Gleick 1987, p. 317). (6) Discovery that the KRM of genetics was a long chain of paired nucleobases from the set {A, C, G, T}, where A pairs only with T, and C pairs only with G, within each link of the chain.

As for some history on the terminology, the term "knowledge-representation method" dates back to at least 1974 with Marvin Minsky (Minsky 1974, p. 28). The abbreviation "KRM" for this term is still occasionally found on the Internet (in 2023). More recent terminology includes the term "knowledge representation language" (KRL) (e.g., Reichgelt 1991, p. 3; Luger and Stubblefield 1998, p. 293), but that term is avoided here because "languages" are usually assumed to be textual whereas Tumbug is not, which makes the term somewhat misleading when applied to visual KRMs such as Tumbug.

The concept of knowledge representation is very close to the concept of data structures in computer science. Technically a data structure is a structured collection of data and a set of routines to create and manipulate that data, whereas discussion of knowledge representation methods tends to ignore any associated routines. Associated routines are generally ignored in this document, as well.

\textit{Emphasized clarification: Tumbug is a knowledge representation method, not an algorithm or program, therefore the system is inherently incapable of learning in its current, non-software form. Misinterpretation of the nature of this study probably arises from the common misinterpretation of the term "artificial intelligence" to mean "machine learning," but machine learning is only one subfield within AI, and the study in this document is not about the subfield of machine learning.}

\subsubsection{Typical tradeoffs of KRMs}

\textbf{1. General properties}

Numerous KRMs abound in the sciences, especially in computer science, and even in games such as chess. In general, KRMs tend to share the following general properties:

\begin{enumerate}
	\item
		Each KRM has at least one competing KRM that can represent the same knowledge.
	\item
		Any given KRM will be inefficient in some way that one of its competing KRMs will not.
	\item
		When a system is rendered in a KRM that is particularly suitable for a given type of problem, algorithms become highly simplified in that KRM. In other words, a good KRM can solve the hardest part of any given problem due to high efficiency arising from the closeness of the system's structure with the system's description.
	\item
		A KRM that attempts to be suitable for solving too many problems at once becomes impractically large in its notational system. This is an example of a time-space tradeoff: increased speed comes at the cost of increased representation size.
	\item
		KRMs that use "exotic isomorphisms" (\textit{verbum} Hofstadter 1979, p. 159) are typically so unwieldy that they are impractical for computational systems. Some examples of exotic isomorphisms are: (1) DNA structure to represent phenotype structure (Hofstadter 1979, p. 160), (2) cellular automata to represent numerical computations (Wolfram 2002, p. 640), and (3) an energy function to "program" a continuous Hopfield network to perform the appropriate optimization. Related to exotic isomorphisms is the desirability of a representation to remain stable when the represented situation changes slightly (Fischler and Firschein 1987, p. 69), since violation of this desideratum results in discontinuous mappings. This strongly suggests that complexity is another key constraint on system efficiency, in addition to the better known constraints of time, space, and energy. The opposite of "exotic isomorphism" is "prosaic isomorphism" (\textit{verbum} Hofstadter 1979, p. 159). In other words, human designers should design only systems that use prosaic isomorphisms.
\end{enumerate}

More specifically, applied to the question of which type of KRM the brain uses, these general properties can be interpreted as:

\begin{enumerate}
	\item
		The brain's KRM could be any one of multiple possibilities, including a non-numerical KRM. This is exactly what Tumbug proposes: a non-numerical KRM that the brain might realistically use.
	\item
		A non-numerical KRM will likely be efficient at solving non-numerical problems but inefficient at solving numerical problems. This is exactly the difference that has been noticed since the early days of computer science: humans are fast at spatial reasoning but slow and error-prone at numerical reasoning, whereas digital computers are slow and error-prone at spatial reasoning but fast and accurate at numerical reasoning.
	\item
		A KRM that is a good candidate for the brain's KRM might solve the hardest part of CSR, which might result in the associated algorithms being quite simple. This is exactly what the author claims: that Tumbug renders WS problems easily solvable via very simple algorithms. In other words, the hardest part of the CSR problem might be easily solvable by a suitable KRM.
	\item
		A universal computing machine is the wrong approach for AGI. This is exactly what the author claims: that digital computers will always have their useful niche of solving large or intricate discretized problems (such as graph problems, data bases, and number crunching) whereas a non-numerical machine would likely have its niche in non-numerical problems with noisy data. In other words, a non-computational/non-computer-like architecture is likely the solution to AGI problems.
	\item
		Exotic isomorphisms should be eschewed as a design heuristic when developing intelligent machines, otherwise such machines will be too difficult to program and to understand. This is one reason why the author believes that that first practical, intelligent machines cannot be based on cellular automata, chemistry, or quantum computers. In other words, the first intelligent machines should use prosaic isomorphisms.
\end{enumerate}

\textbf{2. Examples of pairs of competing KRMs}

In each of the three domain examples of this section, note that there exist two main, competing KRMs, the "best" of which depends on the type of information elicited, and that both KRMs fail to be useful on the last question, which means there likely exists a third KRM that is applicable and efficient. There may exist yet more applicable KRMs beyond these three that are not discussed.

\textbf{2.1. Chess moves: algebraic notation versus descriptive notation}

\begin{figure}
	\begin{center}
	\includegraphics[width=0.75\textwidth]{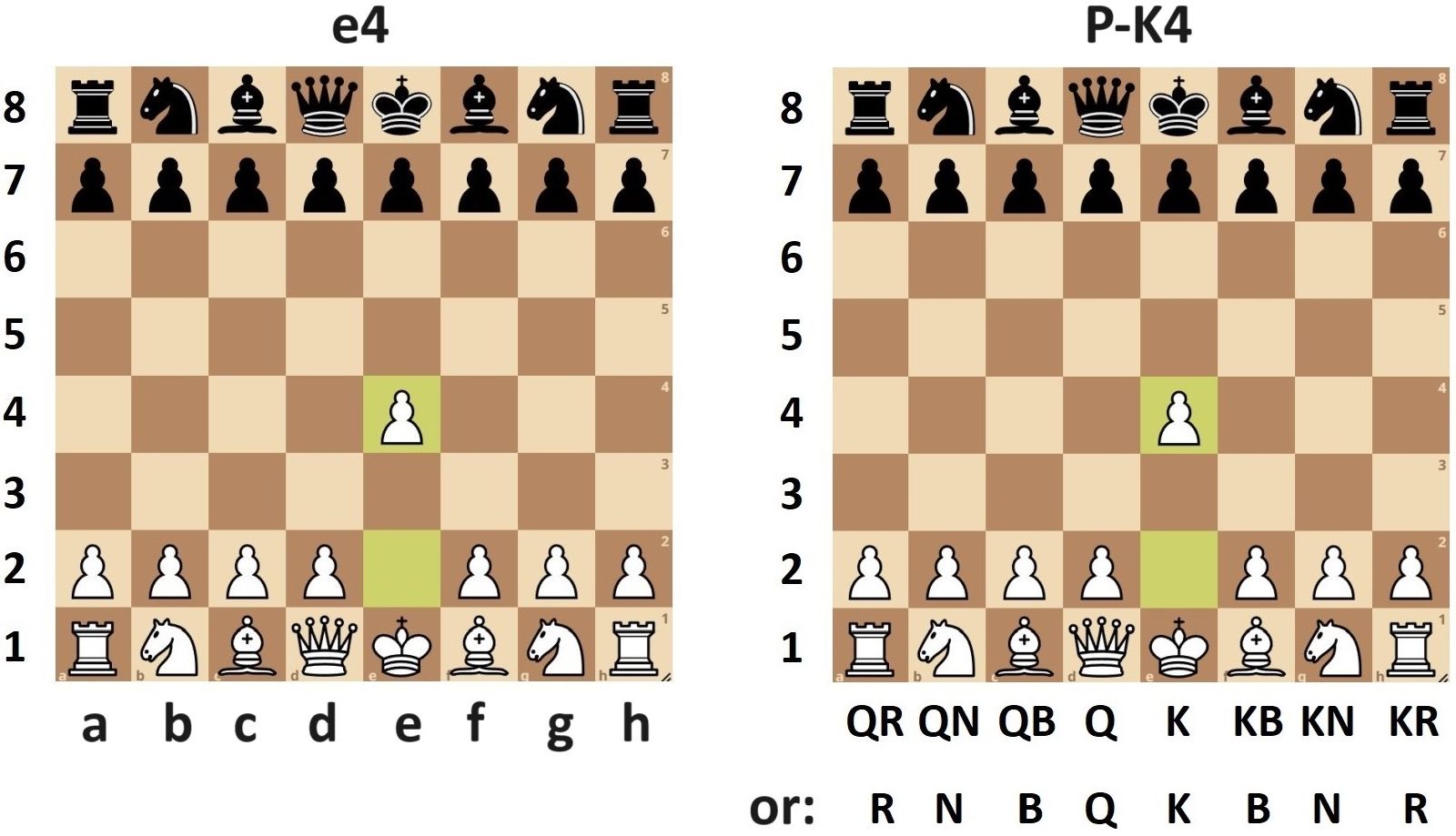}
	\caption{Examples of two KRMs for chess: The shown chess move can be represented as "e4" in algebraic notation, based on the indices on the left, or as "P-K4" in descriptive notation, based on the indices on the right. Which is "best" depends on the type of information elicited.}
	\label{fig:chess-lichess-e4-ANN-algebraic-and-descriptive-CRO}
	\end{center}
\end{figure}

\begin{figure}
	\begin{center}
	\includegraphics[width=0.60\textwidth]{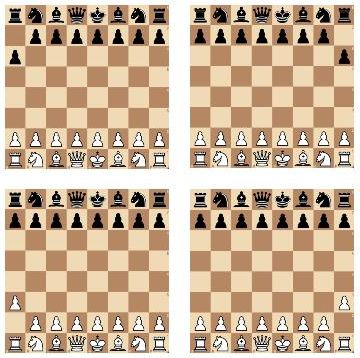}
	\caption{The more general chess move called "P-R3" can refer to either side of the board (kingside or queenside) and to either player (White or Black). This diagram shows all four possibilities.}
	\label{fig:chess-all-p-r3}
	\end{center}
\end{figure}

The earliest successful symbolic KRM for chess moves is called "descriptive notation." For example, the most common first move on the chess board would be called "P-K4," or "pawn to king 4," which means that the pawn in front of the king (this pawn starts on the 2nd row of the chess board) moves to the 4th row of the chess board. In contrast, the modern, favored KRM for chess moves is "algebraic notation." For example, the move that is called "P-K4" in descriptive notation would be called "e4" in algebraic notation, which means that the pawn in question (a pawn is implied because no piece letter precedes the pair of characters) moves to the square represented by "e" (= the 5th column from the left) and "4" (= the 4th row from the bottom). See Figure~\ref{fig:chess-lichess-e4-ANN-algebraic-and-descriptive-CRO} and Figure~\ref{fig:chess-all-p-r3}.

\begin{figure}
	\begin{center}
	\includegraphics[width=0.80\textwidth]{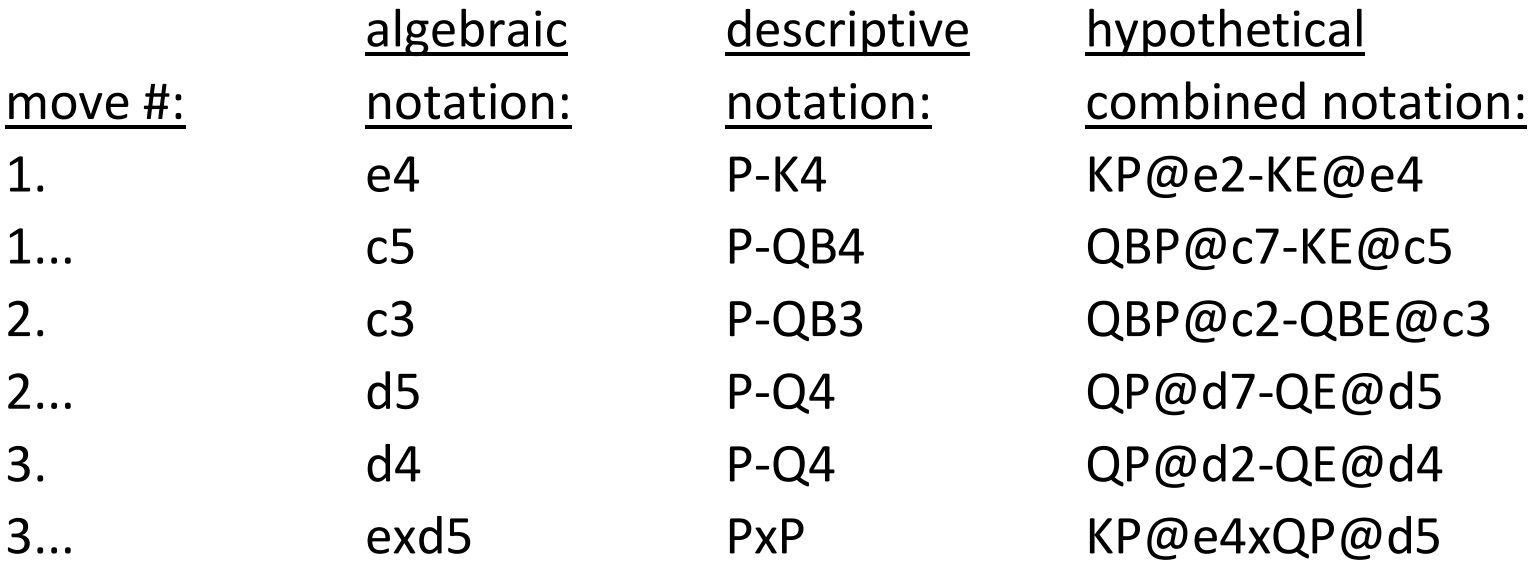}
	\caption{The mentioned hypothetical combined notation clearly shows that increasing the power of the representation (in this case by including more information) tends to increase the size of the representation, in this case in the form of a longer textual description.}
	\label{fig:chess-combined-notation-snap}
	\end{center}
\end{figure}

\begin{figure}
	\begin{center}
	\includegraphics[width=0.75\textwidth]{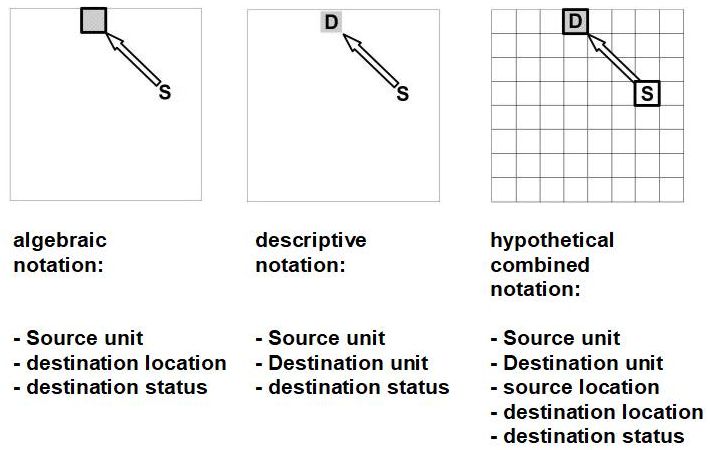}
	\caption{The mentioned hypothetical combined notation approaches a detailed diagram of an arbitrary object's trajectory.}
	\label{fig:chess-trajectory-combined-all}
	\end{center}
\end{figure}

\begin{enumerate}
	\item
		\textbf{Question:} In the move by White, KPxQP (or exd5), what is the destination square of the capturing pawn?\\
		\textbf{The best KRM to generate the answer:} Algebraic notation, since the answer d5 is already provided in that notation, and this answer cannot be determined by descriptive notation, which can only list all five possibilities, namely {exd3, exd4, exd5, exd6, exd7}.\\
		\textbf{The simple algorithm used:} For move Xij, where X = the unit letter (N, B, R, Q, K, or nil for pawn moves) with a possible capture sign (= "x"), and where ij are the last two characters in the string, the destination square is (i, j).\\
		\textbf{Answer, and how it was generated:} For the move Xij = exd5, ij = d5, which is the answer.\\
	\item
		\textbf{Question:} How would one describe the moves P-QR3 and P-KR3 in general, regardless of the player and the side of the board?\\
		\textbf{The best KRM to generate the answer:} Descriptive notation, since the answer is the intersection of the two move representations, namely P-R3, whereas algebraic notation can only list all four possibilities, namely {a3, h3, ...a6, ...h6}.\\
		\textbf{The simple algorithm used:} For moves X and Y, the general description of that move is X$\cap$Y, where the result treats any nils as empty strings that will not show when concatenated.\\
		\textbf{Answer, and how it was generated:} For the two moves P-QR3 and P-KR3, we calculate P-QR3 $\cap$ P-KR3 = P-R3, which is the answer. (See Figure~\ref{fig:chess-all-p-r3}.)\\
	\item
		\textbf{Question:} In the move P-K4 (or e4), from which square did the pawn originate?\\
		\textbf{The best KRM to generate the answer:} Neither, since neither notation carries this information. Descriptive notation can only list all two possibilities, namely {K2-K4, K3-K4}, and algebraic notation can only list all four possibilities, namely {e2-e4, e3-e4, ...e7-e5, ...e6-e5}. Another, more verbose type of notation would be needed, such as long algebraic notation (e.g., Eade 1996, p. 331).\\
		\textbf{An example of both representations combined into a third, hypothetical representation:} For a given player (White or Black), a combined notation that would carry all this information could be $S_1$ $U_1$@$X_1$A$S_2$ $U_2$@$X_2$, where $S_1$ = side of board (K or Q) of the moving unit, $U_1$ = originating unit (P, N, B, R, Q, or K), $X_1$ = originating square, A = action (- or x, for "to" or "takes", respectively), $S_2$ = side of board of the destination square, $U_2$ = the unit on the destination square ("E" if the destination square is empty), $X_2$ = destination square, and the square coordinates can have any representation desired (such as from algebraic or descriptive notation). (See Figure~\ref{fig:chess-combined-notation-snap}.) For example, for the move e4 (or P-K4) of the Ruy Lopez opening, this generalized notation would be KP@e2-KE@e4, and for the move exd5 (or KPxQP) of the Center Counter Defense, this generalized notation would be KP@e4xQP@d5.\\
		\textbf{Answer, and how it was generated:} The move e4 would be represented as $S_1$ $U_1$@$X_1$A$S_2$ $U_2$@$X_2$ = KP@e2-KE@e4, and the originating square is the variable $X_1$ in this string, therefore via matching, $X_1$ = e2, which is the answer.\\
\end{enumerate}

\textbf{2.2. Elements: list versus periodic table}

\begin{figure}
	\begin{center}
	\includegraphics[width=0.75\textwidth]{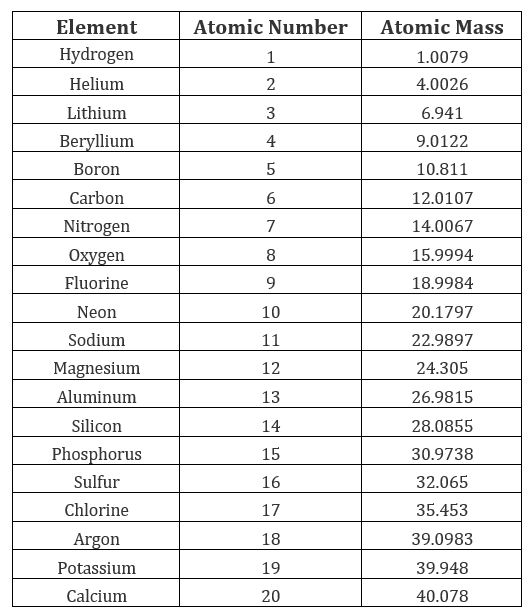}
	\caption{A list of elements. The elements are listed here in order of atomic number. This is basically a 1D data structure where each cell contains an embedded 1D list of attributes.}
	\label{fig:elements-list-atomic-mass-number}
	\end{center}
\end{figure}

\begin{figure}
	\begin{center}
	\includegraphics[width=1.00\textwidth]{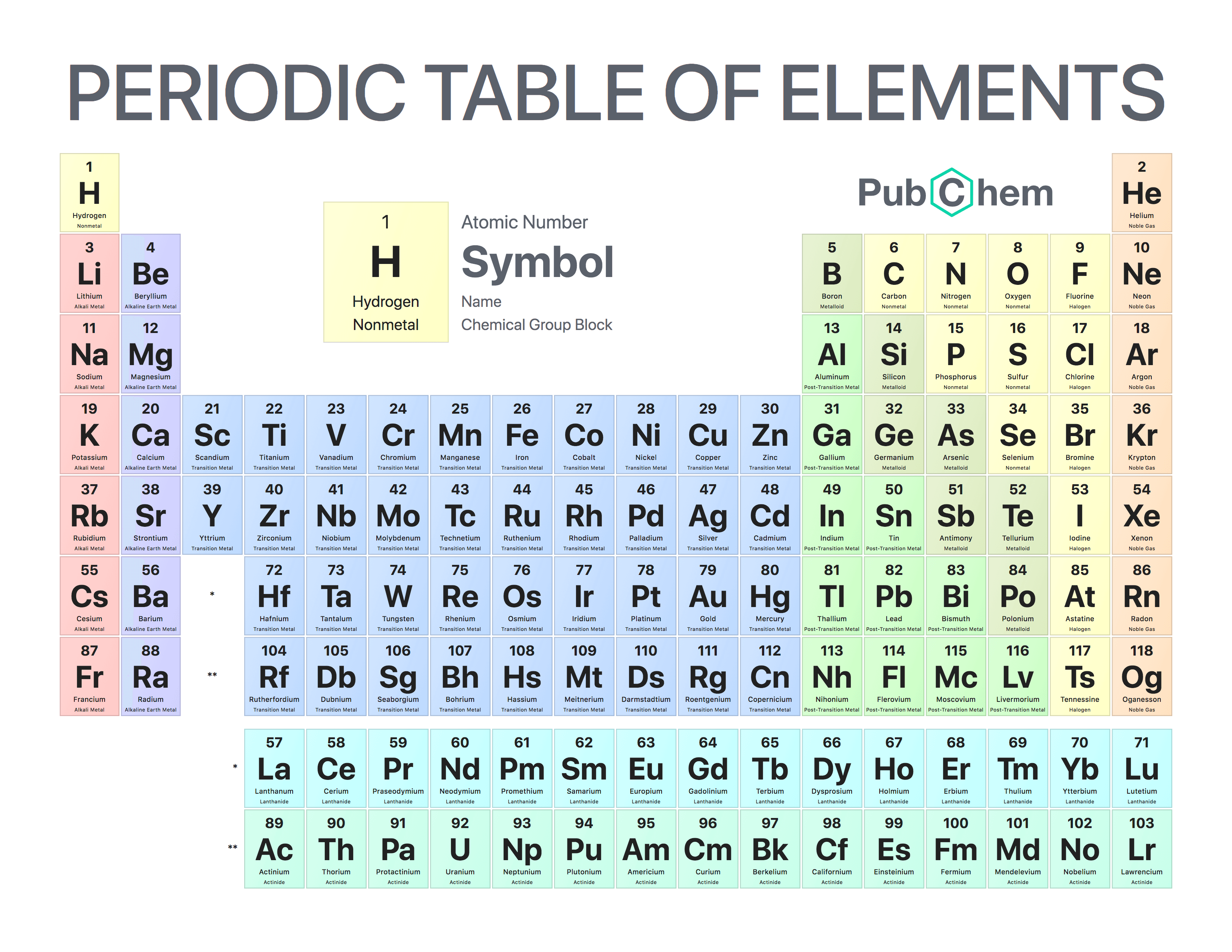}
	\caption{A periodic table. Note that the metals form a clear-cut, roughly rectangular region (colored in blue here, which are: Sc through Zn, Y through Cd, Hf through Hg, and Rf through Cn) when elements are organized into this periodic table.  This table is roughly a 2D data structure where each cell contains an embedded 1D list of attributes. (Source: Public Domain, PubChem via NIH.)}
	\label{fig:elements-periodic-table-britannica}
	\end{center}
\end{figure}

\begin{figure}
	\begin{center}
	\includegraphics[width=0.95\textwidth]{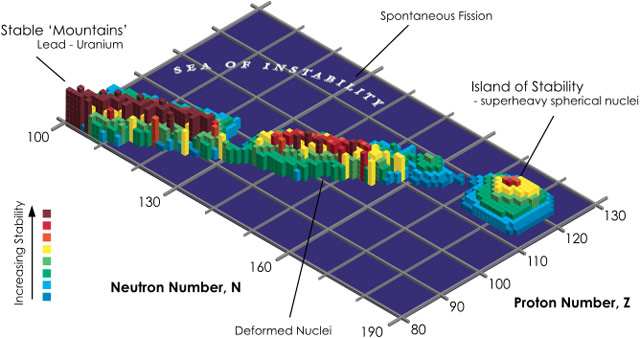}
	\caption{A stability graph. Knowledgeable prediction of stable heavy elements requires spatial assessment of nuclides of the basic elements. This is a 3D data structure. (Source: unknown.)}
	\label{fig:elements-island-of-stability}
	\end{center}
\end{figure}

\begin{figure}
	\begin{center}
	\includegraphics[width=0.75\textwidth]{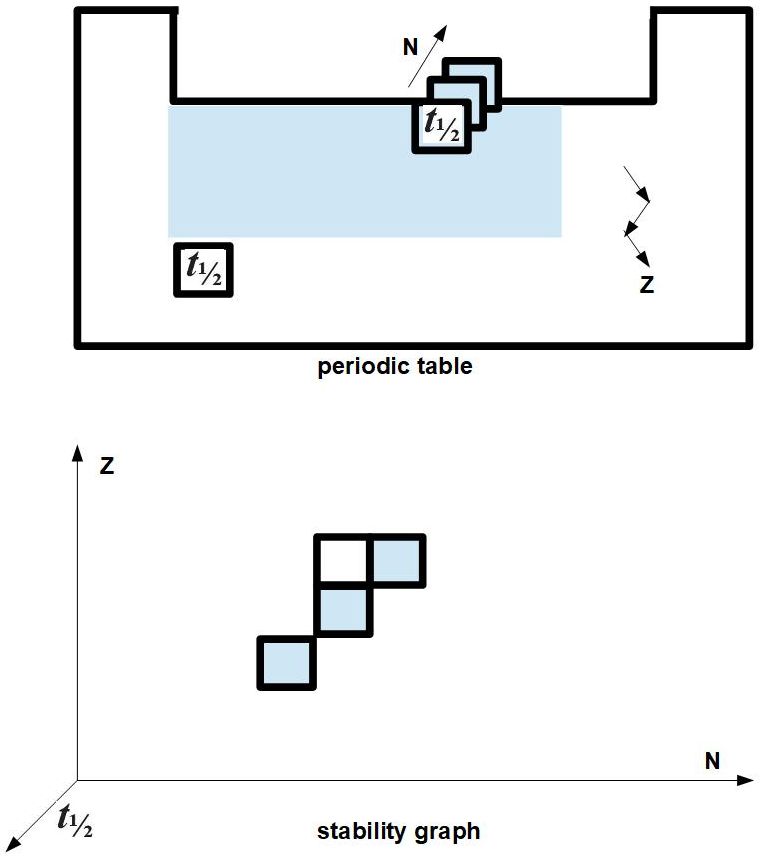}
	\caption{The stability graph has the same information as a periodic table that includes nuclides and half-life ($t_\frac{1}{2}$) information, but the information has been spatially rearranged.}
	\label{fig:elements-3d-rep}
	\end{center}
\end{figure}

\begin{figure}
	\begin{center}
	\includegraphics[width=0.75\textwidth]{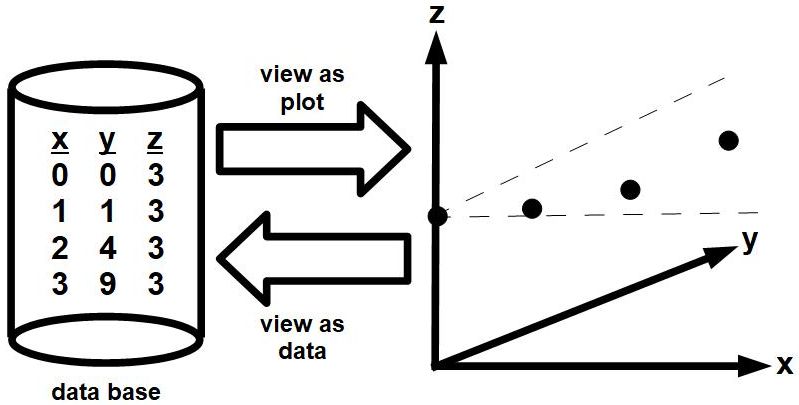}
	\caption{For numerical data, plots carry the same information as a data base, and vary only in the way the data is viewed.}
	\label{fig:data-vs-plot}
	\end{center}
\end{figure}

\begin{figure}
	\begin{center}
	\includegraphics[width=0.50\textwidth]{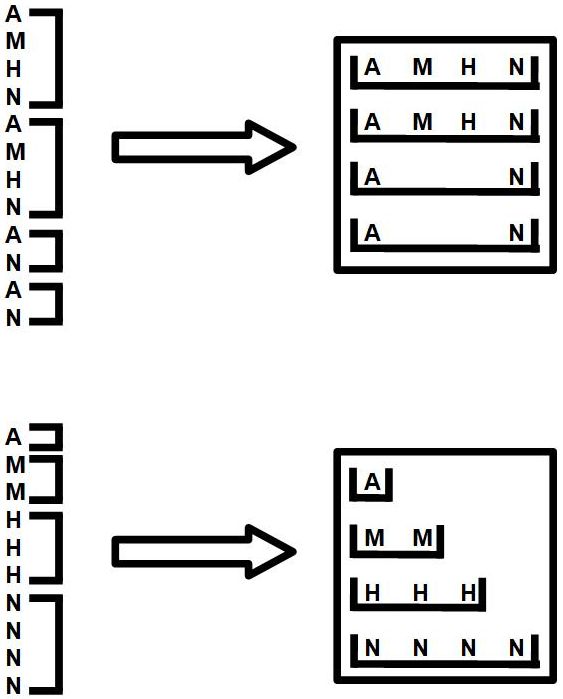}
	\caption{A list in which attributes form a pattern can be easily folded into a 2D table, but the resulting table may not be purely rectangular. The reverse operation of unfolding such a table into a list is even easier.}
	\label{fig:elements-list-folding}
	\end{center}
\end{figure}

\begin{enumerate}
	\item
		\textbf{Question:} What is the name of the element that has 18 protons?\\
		\textbf{The best KRM to generate the answer:} A list, since the list equivalent of the periodic table would be ordered by atomic number, which is the number of protons in an element, and a list can be easily stored on paper (as a written list) or on computer (as a data base or a narrow array), and simply traversed with that structure, either by eye or by computer.\\
		\textbf{The simple algorithm used:} For a given, specified number of protons i, look up the name of the element at item \#i in a list ordered by atomic number.\\
		\textbf{Answer, and how it was generated:}  The element at position \#18 in the list has the name argon, which is the answer. See Figure~\ref{fig:elements-list-atomic-mass-number}.\\
	\item
		\textbf{Question:} Is element \#49 (indium) a metal?\\
		\textbf{The best KRM to generate the answer:} The periodic table, since a clearly defined region occupying about the left 3/4 of the periodic table is metals, and all other regions are nonmetals.\\
		\textbf{The simple algorithm used:} If the column of the element in question is $\geq$ 3 and $\leq$ 12, and if its row is $\geq$ 4 and $\leq$ 7, then the element is a metal, else it is not a metal.\\
		\textbf{Answer, and how it was generated:} Indium is located at column 13, row 5, and since column 13 is $\geq$ 12, this location is outside of the metal region, indium is not a metal. See Figure~\ref{fig:elements-periodic-table-britannica}.\\
	\item
		\textbf{Question:} Which superheavy elements, not yet seen, are most likely to be stable?\\
		\textbf{The best KRM to generate the answer:} Neither, since neither notation carries enough information to make such a prediction. Not only are several more pieces of information (attributes of the element) needed that are not carried in smaller periodic tables, but nuclides (alternative physical forms of each element) must also be listed, in contrast to the regular periodic table that shows only the most common nuclide. The additional information needed includes: (1) the number of neutrons, not just the number of protons, (2) half-life information for measuring stability. Also, a spatial arrangement of the number of protons and neutrons is needed for efficiency so that estimates can be made about regions of high stability by using spatial direction and clustering information that results, and the range of proton-neutron values of such nuclides. Ultimately the best way to show the result is by a 3D plot called a "stability graph" of proton number, neutron number, and half-life, as in Figure~\ref{fig:elements-island-of-stability}.\\
		\textbf{An example of both representations combined into a third representation:} See Figure~\ref{fig:elements-3d-rep}. The same information can be carried by both the periodic table and stability graph, but there are deep visual alterations. For example, the periodic table's 2D regions, such as the region of metals that is shown in blue, disappears because a stability graph "straightens out" the 2D periodic table that was folded along the Z (= number of protons) axis into a straight Z axis, so any 2D region becomes scattered across the 1D Z axis. Also, the half-life information ($t_h$) that was textual in the periodic table becomes plotted as a point in a stability graph since that value is of the highest concern in the stability graph. Also, typical periodic tables do not include more than one nuclide per element, though the figure shown illustrates how a periodic table would logically include multiple nuclides per element, a scheme that causes the periodic table to become 3D with nuclides stored along the N (= number of protons) axis. In this case a generic representation of both KRMs at once would require a data base (or spreadsheet) + sort and subsort operation + list folding (see Figure~\ref{fig:elements-list-folding}) + plot ability. With such a foundation, an extended periodic table and stability graph could be converted to one another in either direction. Whether numerical data is represented as a data in a data base or a plot is only a preference in how it is viewed, not as an important distinction in KRM. (See Figure~\ref{fig:data-vs-plot}.)\\
		\textbf{Answer, and how it was generated:} Visually it can be seen that the center of the cluster called the "island of stability" is at the coordinates Z = 114 protons and N = 184 neutrons, which corresponds to element 114, or nuclides of copernicum (= element 112) and flerovium (= element 114).\\
\end{enumerate}

\textbf{2.3. Series notation: infinite series versus infinite product}

\begin{figure}
	\begin{center}
	\includegraphics[width=0.75\textwidth]{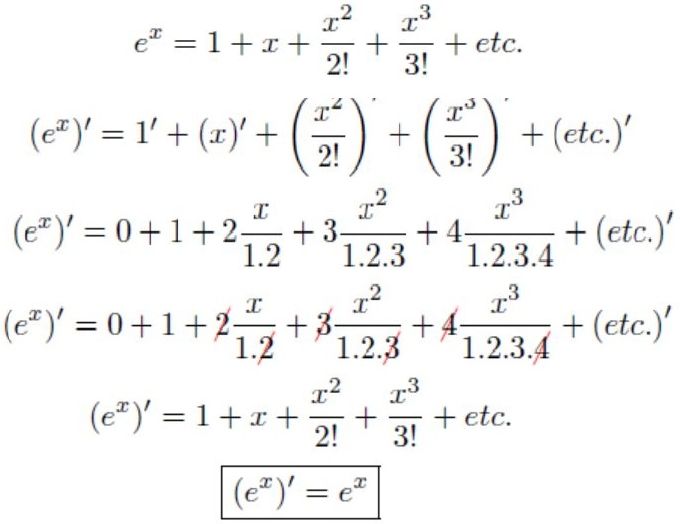}
	\caption{An infinite series (= sum) can be differentiated easily, per term.}
	\label{fig:math-infinite-series-derivative-e-to-x-1-compressed}
	\end{center}
\end{figure}

\begin{figure}
	\begin{center}
	\includegraphics[width=0.50\textwidth]{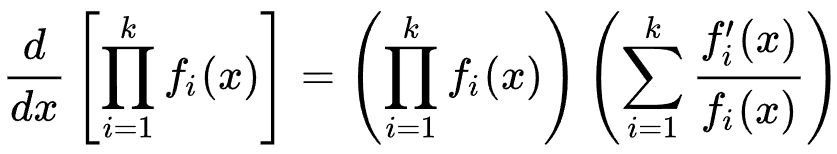}
	\caption{An infinite product cannot be differentiated easily, since the Product Rule requires an indefinite number (k) of products be multiplied by an indefinite number (k) of sums as k approaches infinity.}
	\label{fig:math-product-rule-k-functions-CRO2}
	\end{center}
\end{figure}

\begin{figure}
	\begin{center}
	\includegraphics[width=0.75\textwidth]{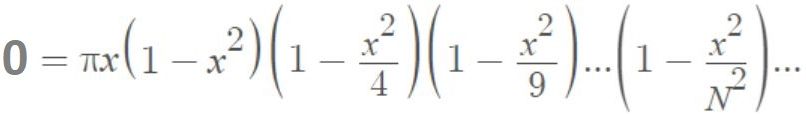}
	\caption{An infinite product with an infinite number of roots, all easy to find.}
	\label{fig:math-infinite-product-sin-pi-x-MOD}
	\end{center}
\end{figure}

\begin{figure}
	\begin{center}
	\includegraphics[width=0.50\textwidth]{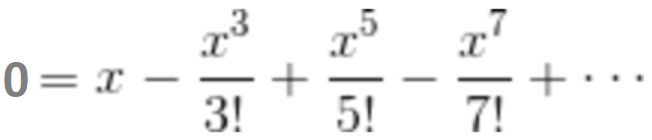}
	\caption{An infinite series with an infinite number of roots, which are difficult to find, in general.}
	\label{fig:math-sine-series-MOD}
	\end{center}
\end{figure}

\begin{figure}
	\begin{center}
	\includegraphics[width=0.50\textwidth]{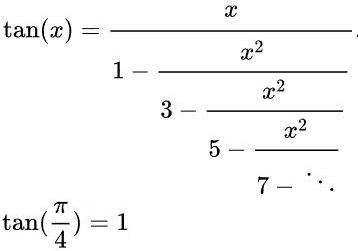}
	\caption{This continued fraction representation for tan(x) and the accompanying identity were used in the first proof that $\pi$ is an irrational number.}
	\label{fig:math-continued-fraction-tan-x-MOD}
	\end{center}
\end{figure}

\begin{figure}
	\begin{center}
	\includegraphics[width=0.50\textwidth]{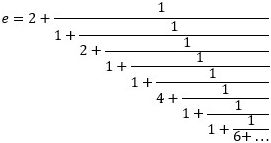}
	\caption{This continued fraction representation for e was used in the first proof that e is an irrational number.}
	\label{fig:math-e-cf-to-6-appearance-CRO}
	\end{center}
\end{figure}

\begin{enumerate}
    \item 
		\textbf{Question:} Is it easier to take the derivative of an infinite series, as in Figure~\ref{fig:math-infinite-series-derivative-e-to-x-1-compressed}, or an infinite product, as in Figure~\ref{fig:math-infinite-product-sin-pi-x-MOD}?\\
		\textbf{The best KRM to generate the answer:} Infinite series.\\
		\textbf{The simple algorithm used:} Take the derivative of each term. At the end, add all the terms. If this were an infinite product expression, it would instead be necessary to use the product rule (fg)' = f'g + fg', which requires taking two derivatives, two multiplications, and one addition per factor, as well as adding all the terms at the end. See Figure~\ref{fig:math-product-rule-k-functions-CRO2}.\\
		\textbf{Answer, and how it was generated:} The infinite series in Figure~\ref{fig:math-infinite-series-derivative-e-to-x-1-compressed}. See the steps of differentiation in the lower levels that figure.\\
    \item 
		\textbf{Question:} Is it easier to find the roots of an infinite series, as in Figure~\ref{fig:math-sine-series-MOD}, or an infinite product, as in Figure~\ref{fig:math-infinite-product-sin-pi-x-MOD}?\\
		\textbf{The best KRM to generate the answer:} Infinite products.\\
		\textbf{The simple algorithm used:} Set each factor in the product to 0, and solve each such equation for x.\\
		\textbf{Answer, and how it was generated:} The infinite series in Figure~\ref{fig:math-infinite-product-sin-pi-x-MOD}. Set 1-($x^{2}$) = 0; 1-($x^{2}$/4) = 0; 1-($x^{2}$/9) = 0, etc., or equivalently: set $x^{2}$ = 1, which has solution x = 1; set $x^{2}$/4 = 1, which has solution x = 2; set $x^{2}$/9 = 1, which has solution x = 3, etc.\\
    \item 
		\textbf{Question:} Which representation is likely to be the easiest for proving that a number is irrational: infinite series or infinite products?\\
		\textbf{The best KRM to generate the answer:} Neither, considering the history of the first proofs of irrationality for the numbers $\pi$ and e. Both of those first proofs were based on continued fractions, which are neither infinite series nor infinite products.\\
		\textbf{An example of both representations combined into a third representation:} Infinite sums and infinite products can be artificially generalized by introducing an auxiliary function "a" that can be set to a binary operator such as addition, multiplication, or exponentiation: a(x, y) = (x + y) or (x * y) or ($x^{y}$). An infinite sequence of simple expressions to be combined can then be described by using recursion in conjunction with this function "a" as follows: $P_{n}$ = a($P_{n-1}$, $t_{n}$), where $P_{n}$ is the current partially accumulated value to be computed, $P_{n-1}$ is the previous partially accumulated value computed, and $t_{n}$ is the current term to be included. For example, the infinite series 1 + (1/2) + (1/4) + (1/8) + ... = 1/$2^{0}$ + (1/$2^{1}$) + (1/$2^{2}$) + (1/$2^{3}$) + ... could be represented with this new notation as $P_{n}$ = a($P_{n-1}$, (1/2)$t_{n-1}$), where a(x, y) = (x + y) to designate summation, $t_{0}$ = 1, $P_{0}$ = 1, $t_{i}$ = (1/2)$t_{i-1}$ = (1/2)1 = 1/2. Therefore $P_{1}$ = a($P_{0}$, (1/2)$t_{0}$) = 1 + (1/2)1 = 1 + 1/2, and $t_{1}$ = (1/2)$t_{0}$ = (1/2)1 = 1/2; $P_{2}$ = a($P_{1}$, (1/2)$t_{1}$) = (1 + 1/2) + (1/2)(1/2) = 1 + 1/2 + 1/4, and $t_{2}$ = (1/2)$t_{1}$ = (1/2)(1/2) = 1/4; $P_{3}$ = a($P_{2}$, (1/2)$t_{2}$) = (1 + 1/2 + 1/4) + (1/2)(1/4) = 1 + 1/2 + 1/4 + 1/8, and $t_{3}$ = (1/2)$t_{2}$ = (1/2)(1/4) = 1/8; etc. Some simpler continued fractions can also be represented with this recursive general notation if a(x, y) is set to k + 1/a(x, y), but more complicated continued fractions may need an additional auxiliary function. In general, a pair of coupled recursive formulas are needed, one for the numerator and one for the denominator (Khinchin 1964, p. 4). For a simple example, $\sqrt {2}$ = 1 + (1/(2+1/(2+1/(2+1/...)))) in continued fraction form, which recursively would use k = 2, or a(x, y) = 2 + 1/a(x, y).\\
		\textbf{Answer, and how it was generated:} Partial fraction representation was historically the basis of the first proof that $\pi$ is irrational, proven by Lambert in 1767, and by Legendre in 1794 (Beckmann 1971, pp. 170-171) (Figure~\ref{fig:math-continued-fraction-tan-x-MOD}), as well as the first proof that e is irrational, proven by Euler in 1737 (Maor 1994, p. 192) (Figure~\ref{fig:math-e-cf-to-6-appearance-CRO}). Since these two numbers are the most famous and most commonly occurring natural mathematical constants, history suggests that partial fractions are the best representation for proving that a given natural mathematical constant is irrational.\\
\end{enumerate}

\textbf{3. Generalizations from the specific KRM examples}

Further examples beyond the three specific KRM examples would be prohibitively long for this document, but in general the following observations seem to hold for all KRMs:

\begin{enumerate}
	\item
		The best KRM to use for a given domain depends on which information needs to be made more explicit, though in making that information more explicit, some other information is usually made less explicit to compensate for the increase in representation size. This is already a very well-known heuristic in computer science.
	\item
		It is possible to retain both types of information in the KRM as one type needs to be made more explicit, but the cost will be the increase in representation size.
	\item
		No matter which new KRM is used for each increase in types of information retained, there will always be some application that needs yet another, different KRM, until a KRM with maximum complexity is reached for that domain.
	\item
		As the representation size becomes large, the KRM approaches a maximum complexity in a way that suggests some  very general form of KRM, such as a generic data base without specifying any specific tables, records, or fields, or a generic neural network without specifying any specific number of layers, number of neurons, or connection pattern.
    \item
    		Systems with shaped objects in motion reach maximum complexity in the representation form of simulation type systems (i.e., SCOVA representation). Examples of such systems include board games, puzzles, folding molecules, and the piano mover's problem.
    \item
    		Systems without shaped objects, motion, aggregation, or rules for directly combining objects reach maximum complexity in the representation form of data base type systems (i.e., OAV representation). Examples of such systems include data bases, data sets, and the periodic table of the elements.
    \item
		Systems with aggregation but without shaped objects or motion, but with rules or heuristics for directly combining objects reach maximum complexity in the representation form of formulas or grammars (i.e., OAV representation), where a large number of relatively simple components of limited types combine according to simple rules or heuristics to form very impressive structures or descriptions. Examples of such systems include mathematics, music, language, chemistry, and genetics.
\end{enumerate}

\textbf{4. Problem decomposition into KRM + algorithms}

\begin{figure}
	\begin{center}
	\includegraphics[width=0.50\textwidth]{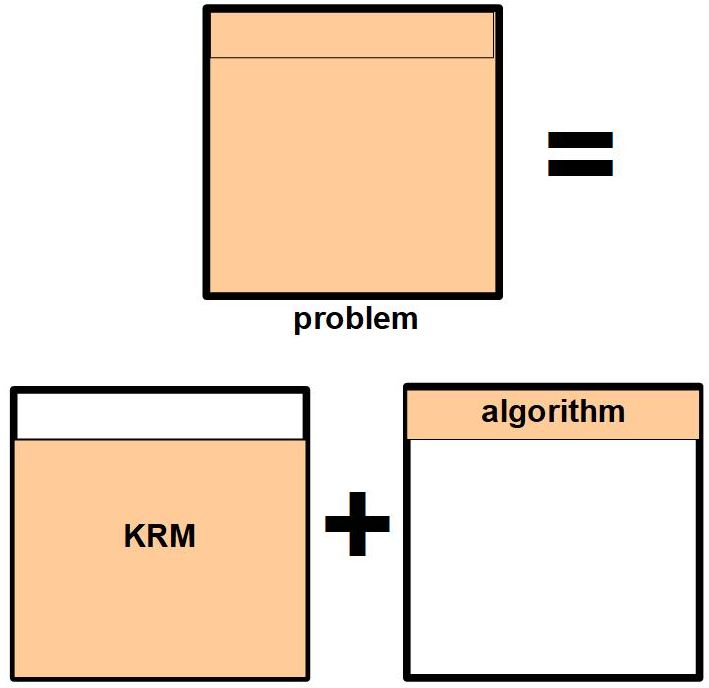}
	\caption{A difficult problem can often be solved by decomposing it first into an appropriate KRM, which then leads to a simple algorithm. The bulk of the effort in solving the problem therefore goes into development of the KRM.}
	\label{fig:decomposition-general-1x2}
	\end{center}
\end{figure}

The aforementioned generalizations about KRMs can be viewed as a method of decomposing a difficult problem into two parts: the KRM and the algorithm that works on pieces that use that KRM. See Figure~\ref{fig:decomposition-general-1x2}.

Recent evidence from modern large language models (LLMs) suggests that statistical learning works impressively well for giving the illusion of intelligence operating behind chatbots based on LLMs. Although such systems have no understanding of what any of the words mean, and cannot do binding or explicit spatial reasoning, an important lesson learned from their performance and popularity is that massive statistical training can work extremely well in providing smooth, real-time, plausible, default assumptions, and can occasionally even give rise to spontaneously emergent behavior such as arithmetic abilities (Rorvig 2021, p. 39).

\begin{figure}
	\begin{center}
	\includegraphics[width=0.50\textwidth]{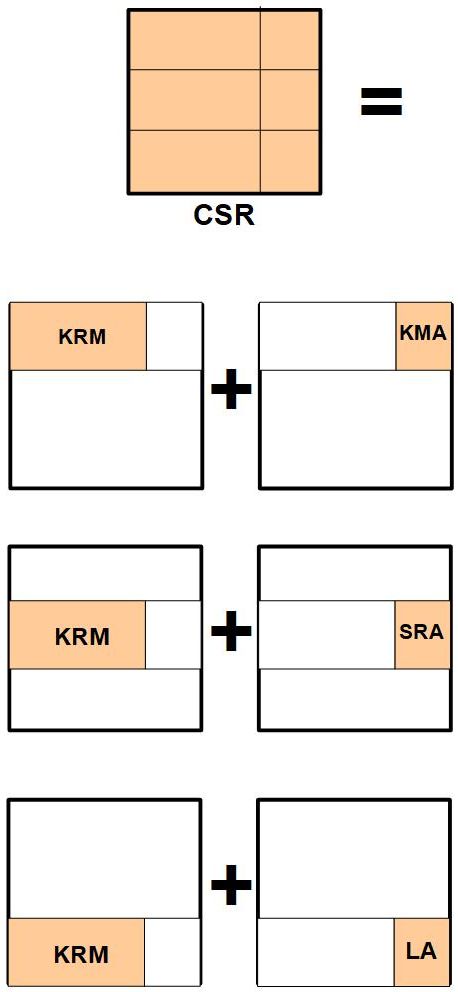}
	\caption{The author's prediction: CSR will require three algorithms in its decomposition, and all will be based on a KRM like Tumbug. The bulk of the effort in solving the problem therefore goes into development of the KRM.\\
	KRM = knowledge representation method (viz. Tumbug)\\
	KMA = knowledge matching algorithm (probably a special case of the SRA)\\
	SRA = spatial reasoning algorithm\\
	LA = learning algorithm}
	\label{fig:decomposition-csr-3x2}
	\end{center}
\end{figure}

The author believes that the aforementioned behavior from LLMs suggests that CSR will decompose into six instead of two main components, where the first pair of components deals with knowledge matching, and the last pair of components deals with learning. The training time for learned CSR knowledge is necessarily very lengthy, essentially of the same order of magnitude as for human CSR learning, and although this is a complication, time is not shown in the algorithm decomposition diagram of Figure~\ref{fig:decomposition-csr-3x2}. If this proposed decomposition strategy holds true for CSR then the implication is that Phase 1 alone of this project will be highly incomplete because it address only two components--KRM and algorithm--not lengthy learning time. This is where the fusion of explicit and implicit memory merge: this current study uses explicit memory in the form of connected groups of icons, but true general intelligence will need to have both forms of memory, including statistical fusion of huge numbers of Tumbug icon structures in order to immediately default to proper associations and temporal continuations. This current direction of study must start somewhere, however, so the issues of training on Tumbug icon structures are deferred for future work. Hopefully the follow-up work will be greatly simplified due to the ability of the Tumbug KRM to represent knowledge in a useful and profound way, which echoes Haugeland's statement: "It may even happen that, once the fundamental structures are worked out, acquisition and adaptation will be comparatively easy to include" (Haugeland 1985, p. 11). On the positive side, conventional neural network training \textit{per se} is mostly straightforward, although generating quality training data is a challenge. Also, a certain amount of Tumbug programming can substitute for naive training, especially with Correlation Boxes, whose function mappings can resemble the mappings of neural networks.

As most highly experienced people will confirm, especially teachers of enormous topics such as a given foreign language or chess, there is no substitute for experience. General intelligence, just like large academic topics, requires an very large number of training experiences over large spans of time in order to generalize and predict properly. Some of this burden can probably be eased by explicit programming of some heuristics, but the amount, quality, and detail of knowledge learned by the human brain is incredibly vast, too vast to encode, even when summarized with explicitly coded heuristics. This implies that there is no shortcut to overcome the need for vast amounts of training, which usually involves vast amounts of time.

\subsubsection{Clues to the type of KRM the brain uses}

Not many AI researchers have directly conjectured what type of KRM the human brain might use, but a few researchers have provided clues:

\begin{enumerate}
    \item
		Per Roger Schank, predicate calculus is insufficient (Schank 1976, p. 172). Prior to Schank's CD theory, predicate calculus was the only suggested canonical form of propositional information (Schank 1976, p. 172), which is one indication of the paucity of research into KRMs for AGI, a situation that still persists today (as of 2023). Implication: The brain's KRM is more powerful in representational power than predicate logic.
    \item
		Roger Schank reasons that a canonical language representational scheme could not have natural-language words as its elements since natural language is too ambiguous (Schank 1976, pp. 171-172). In fact, this limitation was Schank's motivation for creating CD theory. Implication: The brain's KRM is more powerful in representational power than the tree data structure, which is used in sentence diagramming.
    \item
		William A. Woods wrote, "Viewing the knowledge base of an intelligent agent as a model of the external world focuses attention on a number of problems that are not normally addressed in database systems or knowledge-based expert systems." (Woods 1986, p. 1323) Implication: The brain's KRM is more powerful in representational power than data bases or rule-based expert systems, which means more powerful than tables or trees, even when outfitted with search and matching routines, even when outfitted with an uncertainty management system such as confidence factors, fuzzy logic, or Dempster-Shafer theory (e.g., Coppin 2004, p. 483).
    \item
		Ben Coppin wrote, "The human mind uses some form of representation for all concepts, which enables us to understand such abstract ideas as "happiness," "lateness," and "common sense." . . . Clearly this internal representation has a lot to do with our ability to think, to understand, and to reason, and it is no surprise, therefore, that much of Artificial Intelligence research is concerned with finding suitable representations for problems" (Coppin 2004, p. 466). Implication: The brain's KRM needs to represent abstract concepts as well as concrete concepts.
\end{enumerate}

In summary, it is clear the human brain's KRM is much more complex and expressive than any of the commonly used KRMs of AI. At the least, the human brain's KRM covers any form of symbols. Jerry A. Fodor wrote, "I suspect that the representational system with which we think, if that's the right way to describe it, is so rich that if you think up any form of symbolism at all, it probably plays some role in thinking" (Fischler and Firschein 1987, p. 308). Similarly, John Haugeland wrote, "AI has discovered that knowledge itself is extraordinarily complex and difficult to implement--so much so that even the general structure of a system with common sense is not yet clear" (Haugeland 1985, p. 11).

\textit{Side conjecture: The KRM used by the human brain may be at or above the pictorial level of representation in a list of KRMs ordered by their representational power that has the most powerful KRMs at the top.}

One possibility is that the human brain's KRM involves images, or some something like images (= "imageoids"), which a possibility that has been suggested by several authors over the years, and even has an entire book devoted to the topic, viz. Kosslyn et al. 2006:

\begin{enumerate}
    \item
		Martin Fischler and Oscar Firschein believe the mind uses two, major representations, namely propositions and images (Fischler and Firschein 1987, p. 308), though they admit that, when considering images, some representations are isomorphic to images but would not be considered images \textit{per se}. Fischler and Firschein mention some important advantages and disadvantages of using images as a KRM.
    \item
		John Haugeland notes, "The beauty of images is that (spatial) side effects take care of themselves. If I imagine myself astride a giraffe, my feet 'automatically' end up near its belly and my head near the middle of its neck. If I have a scale model of my living room, and I put the model couch in the front alcove, it 'automatically' ends up by the bay window and opposite the door. I don't have to arrange for these results deliberately or figure them out; they just happen, due to the shapes of the images themselves" (Haugeland 1985, p. 229).
    \item
		James Hogan wrote, "Vision and language involve what are perhaps our most complex and abstract representations of the external and internal realities we experience. Comprehending how they operate would get close to the core of what true intelligence is all about" (Hogan 1997, p. 179).
    \item
    		Kenneth Haase came up with an idea of visual memory that influenced Marvin Minsky's idea of K-lines (Minsky 1986, p. 82), which is a theory of memory. "You want to repair a bicycle. Before you start, smear your hand with red paint. Then every tool you need to use will end up with red marks on it. When you're done, just remember that red means 'good for fixing bicycles.' Next time you fix a bicycle, you can save time by taking out all the red-marked tools in advance." He then generalized this idea to different colors for different jobs.
	\item
		Rodney Brooks wrote: "There is certainly no AI vision program which can find arbitrary chairs in arbitrary images; they can at best find one particular type of chair in carefully selected images. This characterization, however, is perhaps the correct AI representation of solving certain problems; e.g., a person sitting on a chair in a room is hungry and can see a banana hanging from the ceiling just out of reach. Such problems are never posed to AI systems by showing them a photo of the scene. A person (even a young child) can make the right interpretation of the photo and suggest a plan of action. For AI planning systems, however, the experimenter is required to abstract away most of the details to form a simple description in terms of atomic concepts such as PERSON, CHAIR and BANANAS." (Brooks 1992, p. 143)
	\item
		Donald R. Tveter wrote: "The fact that people do store many memories as pictures and do at least some of their reasoning about the world using pictures ought to be one of the most obvious principles of all, and yet it has been neglected, in part due to the predominance of symbol processing and in part due to the fact that processing pictures is still underdeveloped and has not been used in conjunction with representing the real world in programs where the goal of the program is to reason about the real world." (Tveter 1998, p. 18)
\end{enumerate}

The fact that imageoids are such an obvious candidate for the brain's KRM and yet are still not being researched to any great extent is puzzling. If a breakthrough in AGI is going to be made, this is likely the subject area in which it will occur. This is why the Tumbug KRM has been based on imageoids.

\subsubsection{Ranking of KRM methods}

Can the "best" KRM be determined by its properties alone? Probably, but this has not been done yet. Only a few authors have mentioned ways to measure the relative power of KRMs. One is Arthur B. Markman, who suggested three attributes (Markman 1999, p. 14):

\begin{enumerate}
    \item
		the duration of the representational states
    \item
		the presence of discrete symbols
    \item
		the abstractness of representations
\end{enumerate}

The following previously unpublished, additional criteria would likely be useful, as well:

\begin{enumerate}
    \item
		analog representation capability (e.g., for representing time/space with indefinite precision)
    \item
		data structure intricacy or nonlinearity (e.g., tree structures are more intricate than lists)
    \item
		ability to represent time (e.g., for representing moving images, or other changes over time)
\end{enumerate}

Another such author is Han Reichgelt (Reichgelt 1991, p. 5), who divided KRMs into two practical considerations:
\begin{enumerate}
    \item
		syntactic aspects - "the naturalness and the expressiveness" of the KRM
    \item
		inferential aspects - "the power of the underlying inference machinery"
\end{enumerate}

Despite the lack of consensus on this topic, the following hierarchy attempts to show a few well-known KRMs, and the author's estimate of their approximate representation power relative to each other, where \#1 is the most powerful KRM listed here.

\begin{enumerate}
    \item
		pictures or illustrations
    \item
		directed graphs (e.g., semantic nets, knowledge graphs, state diagrams, DFAs, NFAs)
    \item
		directed acyclic graphs
    \item
		trees
    \item
		binary trees
    \item
		jagged arrays
    \item
		arrays or adjacency graphs
    \item
		Venn diagrams
    \item
		lists
    \item
		vectors
    \item
		scalars
\end{enumerate}

A few other KRMs such as predicate calculus, pushdown automata, CD theory, and Petri nets are not on this list since their advantages and disadvantages involve attributes that are distributed across multiple dimensions and are distributed excessively across even the levels of a 1D list.

The "pictures" type of KRM mentioned at \#1 occurs in Tumbug only inside of a Location Box of some type, which is a Tumbug Building Block icon for representing space. Such pictures may be represented verbatim, in which case a Verbatim Box is used, or they can be abstracted into general regions and general objects without being rigidly specified, in which case a Descriptive Box is used. Together these types of Location Boxes allow concrete images and abstracted images to be represented, or both types mixed together. The fact that Tumbug uses the most powerful KRM, viz. images, suggests that Tumbug could be the most powerful KRM developed so far, provided that its Building Blocks prove to be strong and consistent.

\textit{Side conjecture: In theory, it is likely that ultimately no canonical form of meaning exists, any more than a canonical description of 2D shapes exists, since any given description can be based on different conceptual foundations that can be combined in uniquely different ways that produces unmatched forms. However, in practice there usually does exist a description that is more intuitive or natural than others, and usually also simpler than others, which in practice serves as the canonical form.}

\subsubsection{Why dependency on learning is misguided}

A common fallacy is that all that is needed to produce AGI is to let a system learn. Haugeland (1985, p. 11) described this misconception well:\\

"Learning is the acquisition of knowledge, skills, etc. The issue is typically conceived as: given a system capable of knowing, how can we make it capable of acquiring? Or: starting from a static knower, how can we make an adaptable or educable knower? This tacitly assumes that knowing as such is straightforward and that acquiring or adapting it is the hard part; but that turns out to be false. AI has discovered that knowledge itself is extraordinarily complex and difficult to implement--so much so that even the general structure of a system with common sense is not yet clear. Accordingly, it's far from apparent what a learning system needs to acquire; hence the project of acquiring some can't get off the ground."

\begin{figure}
	\begin{center}
	\includegraphics[width=0.75\textwidth]{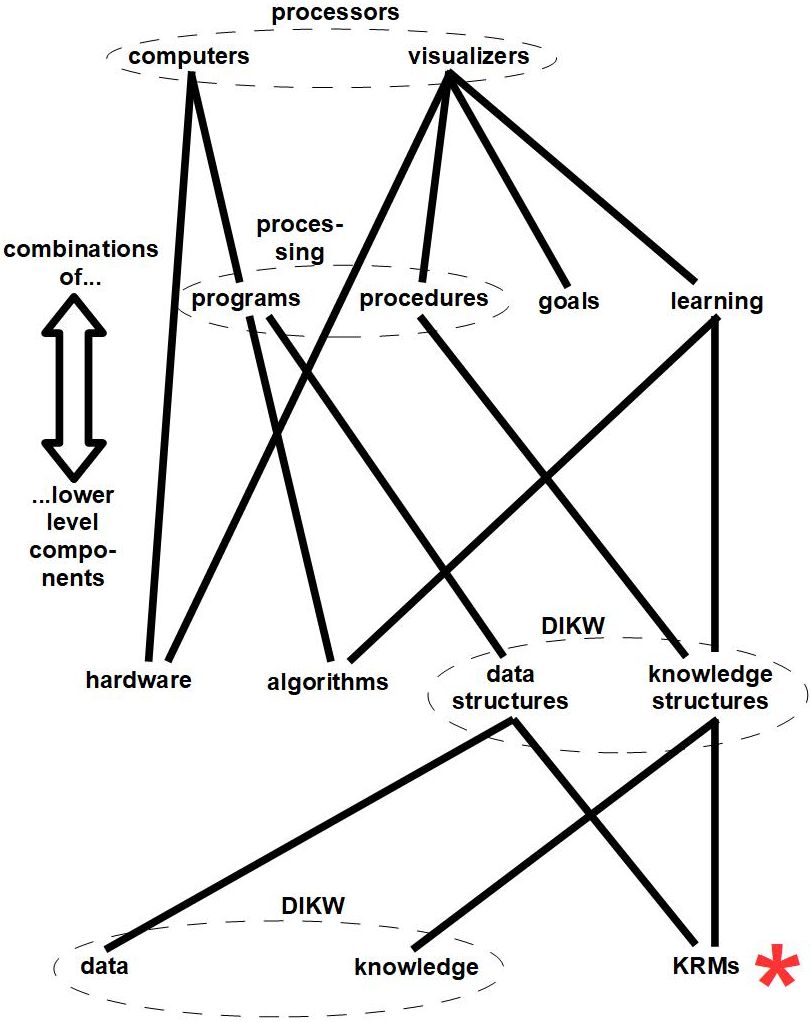}
	\caption{Conceptual dependencies in processors. Note that KRMs underlie both types of processors.}
	\label{fig:dependencies}
	\end{center}
\end{figure}

Figure~\ref{fig:dependencies} shows basic conceptual components of processors (both for IT and AGI) and their dependencies. Note that learning is dependent upon data structures or knowledge structures, or in general learning is dependent of \textit{structured} data/information/knowledge/wisdom (DIKW), which in turn are based on KRMs. Therefore there exists no easy way to avoid KRMs because they are basic to every processor of which humans can perceive.

\subsubsection{Criteria and heuristics that make Tumbug seem particularly promising}

\textbf{1. The-only-move that-doesn't-lose Principle}

\begin{figure}
	\begin{center}
	\includegraphics[width=1.00\textwidth]{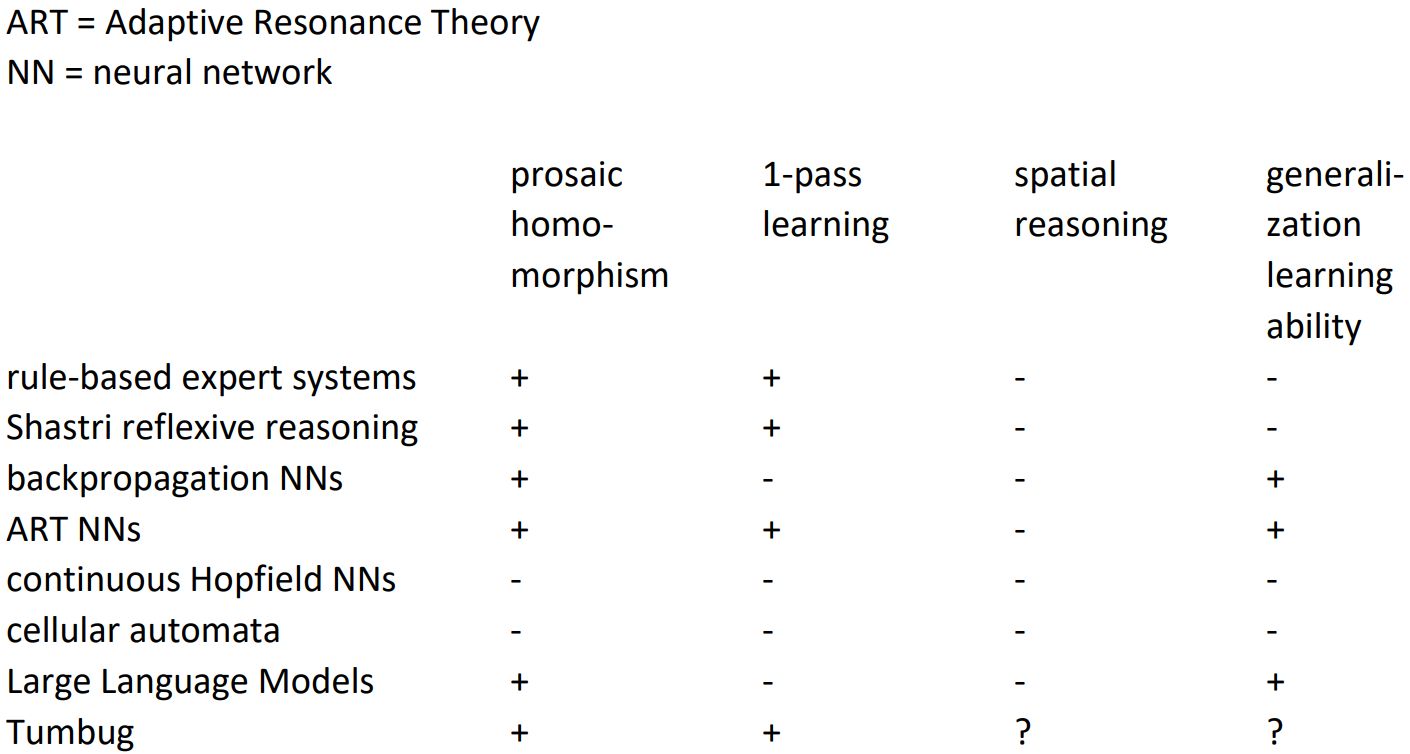}
	\caption{Some of the most promising current approaches to AGI.}
	\label{fig:ai-approaches-table-snap}
	\end{center}
\end{figure}

Every current AI design appears to have at least one fatal flaw that prevents it from every achieving true general intelligence status. An overview is shown in Figure~\ref{fig:ai-approaches-table-snap}. A question mark in the table means "unknown." Merely by virtue of being new, Tumbug has a particularly promising design because most of its table cells are still blank because not enough research has been done on Tumbug to identify any fatal flaws. Although many blank cells seems to be weak justification of a promising design, Tumbug was specifically designed to overcome the flaws of every major AI paradigm. Also, by analogy, a row of the table is like an evaluation of physical paths through a rugged landscape, where the traveler knows in advance that at least one navigable path exists: many paths can be tentatively ruled out because they appear to have impasses, whereas a path where nothing is known has a greater likelihood of being one of the paths that are navigable. This type of ignorance evaluation logic also appears in chess, where the proper move can sometimes be chosen only because all the well-known alternatives are known to lose, therefore the best candidate for the best move is a move of which little is known.

\textbf{2. Organization is more important than performance}

Some of the deepest insights into the nature of intelligence and its relation to AI come from well-known authors who emphasize that what is not important for producing AGI is not speed (therefore faster computers, parallel computers, and quantum computers will not be the key to AGI) or superficially impressive results, but rather the organization of the system. AGI will clearly need a clever type of underlying organization that has not yet been publicly considered. Below are some quotes to this effect.

Marvin Minsky: "Hardware is not the limiting factor for building an intelligent computer. We don’t need supercomputers to do this; the problem is that we don’t know what’s the software to use with them. A 1 MHz computer probably is faster than the brain and would do the job provided that it has the right software." (Sabbatini 1998)

Jeff Hawkins: "According to functionalism, being intelligent or having a mind is purely a property of organization and has nothing inherently to do with what you're organized out of. A mind exists in any system whose constituent parts have the right causal relationship with each other, but those parts can just as validly be neurons, silicon chips, or anything else." (Hawkins 2004, p. 36)

\textbf{3. Pei Wang's tables}

\begin{figure}
	\begin{center}
	\includegraphics[width=0.90\textwidth]{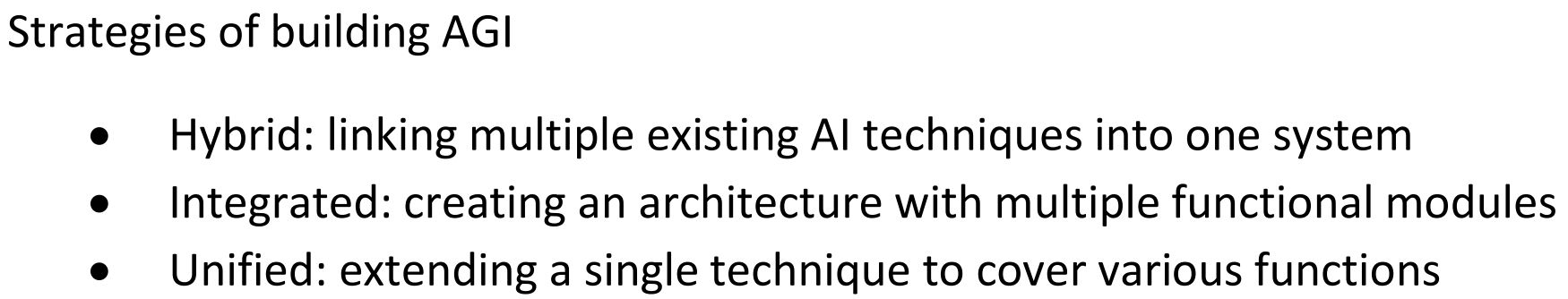}
	\caption{Wang's three types of combined systems are hybrid, integrated, and unified.}
	\label{fig:wang-strategies}
	\end{center}
\end{figure}

\begin{figure}
	\begin{center}
	\includegraphics[width=0.90\textwidth]{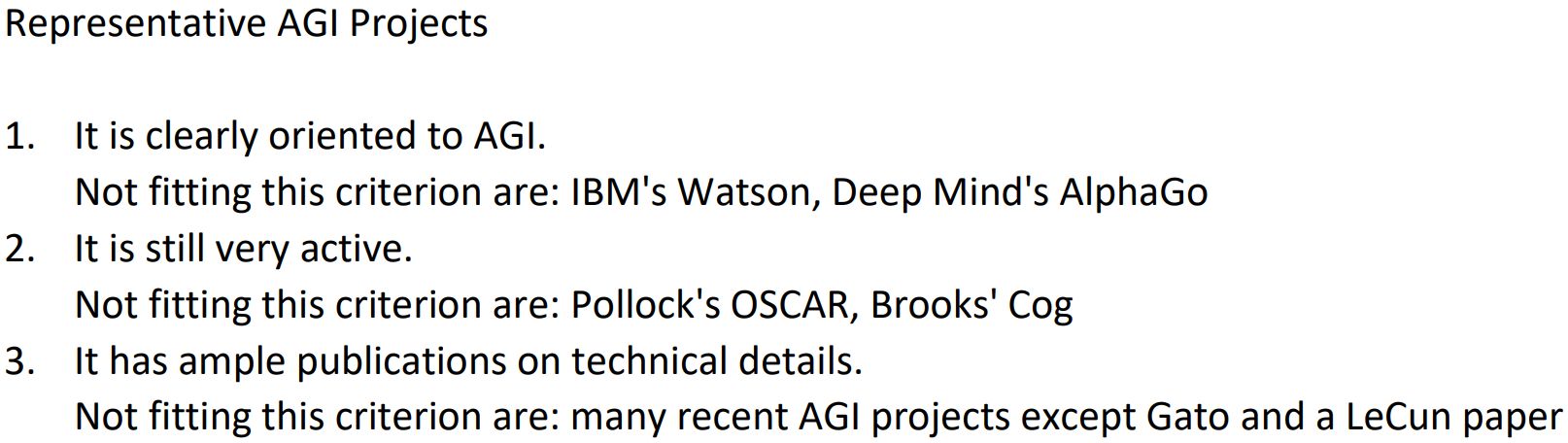}
	\caption{Wang's table of criteria of promising AGI approaches.}
	\label{fig:wang-criteria}
	\end{center}
\end{figure}

\begin{figure}
	\begin{center}
	\includegraphics[width=0.90\textwidth]{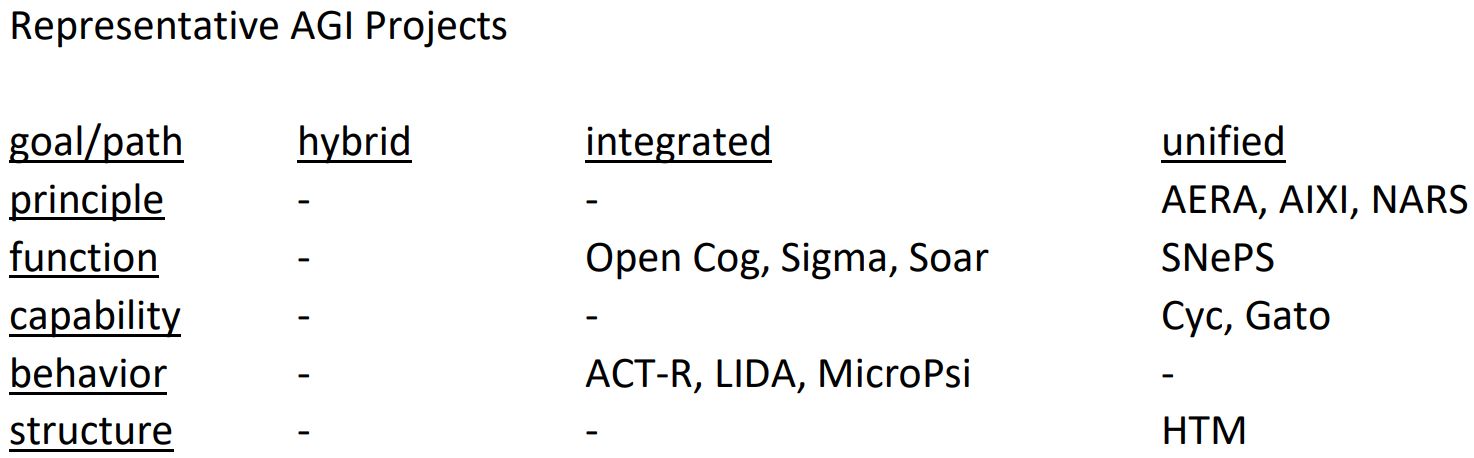}
	\caption{Wang's table of some of the most promising AGI approaches.}
	\label{fig:wang-representative-table-snap}
	\end{center}
\end{figure}

Professor Pei Wang, in his course on AGI at Temple University, Philadelphia, collected and summarized the approaches to AGI that he considers the most promising, which are listed in Figure~\ref{fig:wang-representative-table-snap}. His list of criteria is shown in Figure~\ref{fig:wang-criteria}. The architectures are allowed to come from any of three types of integration, which are listed in Figure~\ref{fig:wang-strategies}. The "integrated" type is the same concept that the author called "fusion" type in 2000 (Atkins 2000), which is the type that the author favors. The author's Visualizer Project that includes Tumbug fits two out of three of Wang's criteria for a promising AGI architecture:

\begin{enumerate}
    \item
		The Visualizer Project is not application-specific, therefore Tumbug fits criteria \#1.
    \item
		The Visualizer Project is still currently being worked on, therefore Tumbug fits criteria \#2.
    \item
		The Visualizer Project is too new to be documented well, therefore Tumbug does not fit criteria \#3.
\end{enumerate}

Wang's last two criteria are based on the current worldly success of a given architecture, so if an architecture fails to meet either of these criteria it does not necessarily mean that the architecture is deficient \textit{per se}, only that it is not popular for some reason.

\begin{figure}
	\begin{center}
	\includegraphics[width=0.60\textwidth]{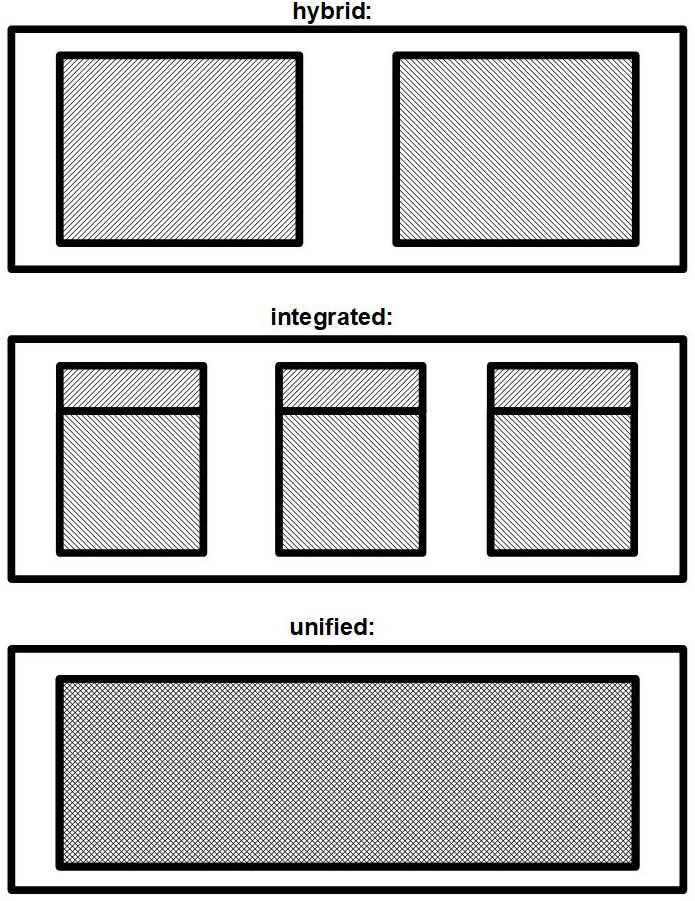}
	\caption{A diagrammatic view of Wang's list of three strategies for combining existing disparate architectures. Per Wang, the ideal strategy for AGI is unified, which happens to be the type that Tumbug is.}
	\label{fig:unified}
	\end{center}
\end{figure}

Regarding Wang's three categories of combined architectures (viz., hybrid, integrated, and unified), see Figure~\ref{fig:unified}. Some examples of two different architectures being combined in these three ways in a single system are:

\begin{itemize}
    \item
		rule-based expert system + neural network (NN):
		\begin{enumerate}
		    \item
				hybrid: each type of architecture can be called from a program running in the system, as needed
		    \item
				integrated: a heuristic search can use the expert system for the search, and the NN for the heuristics
		    \item
				unified: the rule-based expert system is implemented with artificial neurons
		\end{enumerate}
    \item
    		digital computer + analog computer:
		\begin{enumerate}
		    \item
				hybrid: each type of computer can be called from a program running in the system, as needed
		    \item
				integrated: the digital computer performs only algebraic operations, and the analog computer performs only calculus operations
		    \item
				unified: each calculation is done in analog, but is iterated digitally to confirm accuracy
		\end{enumerate}
    \item
    		object-oriented program + data base program:
		\begin{enumerate}
		    \item
				hybrid: each type of program can be called from a program running in the system, as needed
		    \item
				integrated: the data base holds the initial values of each object, and the object-oriented program performs simulations with these objects
		    \item
				unified: an object-oriented data base
		\end{enumerate}
\end{itemize}

\textbf{4. Kocijan's criteria}

Kocijan et al. (2022) assessed different approaches used so far to solve WSC problems, and mentioned the following five desiderata. Each desideratum below is listed with a comment about how Tumbug (or more generally, a visualizer) is believed to obviate the mentioned pitfall. Page numbers in this list refer to the aforementioned reference.

\begin{enumerate}
    \item
		The inference should be carried out automatically, before the question points out the anomaly (p. 4).\\
		A visualizer conforms to this criterion, but demonstration of this will require a knowledge matching algorithm, which will not be fully investigated and documented until the end of Phase 2. Tumbug is quite simple, so an architecture such as a visualizer using Tumbug does not require ”commonsense reasoning of some depth and complexity” (p. 4). A visualizer is a novel implementation of a ”reflexive reasoning” architecture (cf. Shastri and Ajjanagadde 1990).
    \item
		The approach should not be easily solvable using word correlation (p. 16).\\
		As used so far, the Tumbug approach completely ignores the statistics of word frequency or word correlation, so obviously Tumbug bypasses this pitfall.
    \item
		The approach should demonstrate some semblance of CSR, not merely technical tricks (pp. 7-8).\\ 
Because Tumbug uses a visual grammar that the author believes is universal across all natural languages, and because Tumbug deals with the actions of real-world objects, at the very least each scenario that Tumbug visualizes could be run as a video simulation, which would confirm that the correct corporeal actions on the correct corporeal objects are being used.
    \item
		The system should have reasoning capability, not just general pattern mapping as in neural networks (p. 23).\\
Tumbug is mostly symbolic/iconic, not a neural network, so it avoids naive pattern mapping of the neural network type.
    \item
		The system should be able to generalize, so that good performance is attained across any subset of WC, not just one selected subset (p. 26).\\
		If this statement refers to learning, then a visualizer should conform to this criterion, but research on this will not begin until Phase 4. If this statement refers to programmable representation of a concept, that a Tumbug diagram can already do this.
\end{enumerate}

\subsection{Applicable systems from the real world}

\subsubsection{Physics}

Obviously physics is intended to provide explanations for events that happen in our physical universe, so a KRM that is trying to represent those same events should do well by basing itself on physics. Less reliable alternatives would be to cover only topics in a narrow domain, cover only encountered topics in some convenient corpus of text, or to guess at which foundations that might span all known possibilities. Physics is the study of space, time, matter, motion, energy, and any other related phenomena (Baez 1967, p. 3), and certainly a large percentage of English sentences describe objects (which consist of matter) and motion, so physics is one of the most obvious foundations of descriptions of real-world events. Also, many authors (e.g., Pezzulo 2007; Berthoz 2000, p.22) have noted that brains of higher animals seem to simulate real-world events, especially for purposes of prediction, planning, and CSR, and to simulate is to represent events partly by relying on physics formulas, therefore physics is probably the best field on which to found a set of Building Blocks.

The most basic concepts in physics are often listed at the beginnings of physics textbooks (e.g., Baez 1967, p. 3), and include especially the following:

\begin{itemize}
    \item
		spacetime: a single structure that includes space and time as separate dimensions
		\begin{itemize}
	  		\item
				SUBCOMPONENTS: space, time
		\end{itemize}
    \item
		mass-energy: physical objects
		\begin{itemize}
	   	 	\item
	    			SUBCOMPONENTS: mass (m), energy (E)
	    		\item
	    			CONSERVATION: approximately yes, for total energy of a system in classical mechanics
			\item
	    			INTERCHANGABILITY: with energy via the equation $E=mc^{2}$
	    	\end{itemize}
    \item
momentum: a property of a physical object
	\begin{itemize}
   	 	\item
    			SUBCOMPONENTS: mass (m), velocity (v)
	    \item
	    		CONSERVATION: yes, for an isolated system
	    \item
    			INTERCHANGABILITY: with mass and velocity via the equation p=mv; with force via the equation $F=\Delta p / \Delta t$
	\end{itemize}
    \item
force: deals with the interaction of objects, not so much with single objects
	\begin{itemize}
	    \item
	   	 	SUBTYPES of force: contact force, non-contact force
	    \item
	    		SUBTYPES of force: strong nuclear, weak nuclear, electromagnetic, gravity (a pseudo-force)
	    \item
	    		CONSERVATION: not in general, but depends on the type of force
	\end{itemize}
    \item
charge: deals more with single objects, not so much multiple objects
	\begin{itemize}
	    \item
	    	CONSERVATION: yes, for an isolated system
	\end{itemize}
    \item
probability: used in quantum mechanics where waves consist of probabilities, not energy
	\begin{itemize}
	    \item\
	    		CONSERVATION: no
	\end{itemize}
\end{itemize}

Since Tumbug is intended to model primarily the human-centric view of the world, especially that of human intuition, Tumbug currently does not make a distinction between true forces (viz., electromagnetic force, strong nuclear force, weak nuclear force) and pseudo-forces (viz., gravitational force). Tumbug also uses its own version of spacetime, a much simpler concept than Minkowski space, largely because no quantified interactions (e.g., the formulas of relativity) between space and time are considered in Tumbug unless explicitly modeled. Also, Minkowski space is limited to four dimensions, which Tumbug space is not. Tumbug is able to model those more complex versions of physical phenomena, but such constraints are not built into Tumbug.

\subsubsection{Mathematics}

Physics is not the sole and ultimate science for modeling the real world, however, since the laws of physics are in turn described with mathematics; mathematics is assumed to be the descriptive foundation of physics. A few mathematical concepts used in physics that are seldom found in basic physics books but that are concepts that involve useful mathematics for modeling the real world are:

\begin{itemize}
    \item
		combinatorics (from probability)
    \item
		randomness (from probability)
    \item
		connectivity (from topology)
\end{itemize}

A few of the more fundamental mathematical concepts needed to represent natural language concepts have been mentioned and included with Tumbug concepts in this document, but there may exist many more mathematical concepts, possibly infinite in number, that may be needed to describe real-world situations. In particular, more concepts from first-order logic, which is a subset of math, may well be needed in Tumbug since predicate logic is one of the oldest and soundest KRMs for AI systems (Cercone and McCalla 1987, p. 3). Already a form of the NOT concept (viz., XOR) and a form of the existential qualifier (viz., C-A Aggregation Boxes) from predicate logic have needed to be added to Tumbug in order to represent WS150 sentences.

\subsubsection{Other systems and the Big Picture}

In the big picture, humankind has recently begun to struggle with its ancient systems of representation such as mathematics and language to cope with the extreme advancements of science. Several authors have noted that some new type of mathematics is likely needed to cope with the very complicated concepts that modern science is uncovering (e.g., Bailey 1996, pp. 9, 26, 28; Devlin 1997, pp. 282-283; Wolfram 2002, p. 627; O'Neill 1981, pp. 48-49). Presumably the need for new KRMs is correlated with the need for new mathematics, since mathematics is itself a type of KRM.

\begin{figure}
	\begin{center}
	\includegraphics[width=0.90\textwidth]{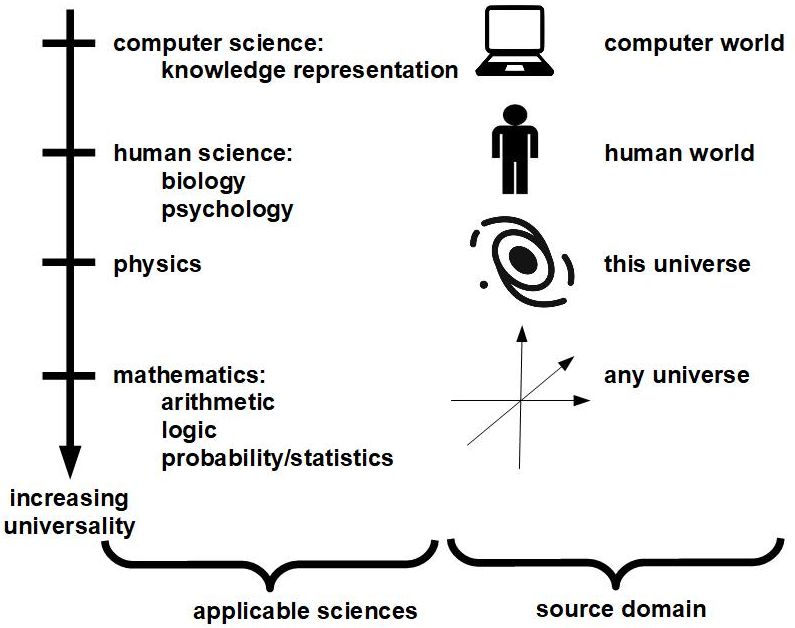}
	\caption{A summary of the originating fields of Tumbug icons, in order of increasing universality.}
	\label{fig:big-field-priority}
	\end{center}
\end{figure}

Figure~\ref{fig:big-field-priority} shows the most basic fields from which each concept arises, where "most basic" means "most universal": (1) mathematics, (2) logic, (3) physics, (4) human world, (5) computer science. This justification is based on the observations that: (1) mathematics is applicable to any universe because it is based on nonphysical abstractions, (2) physics is universal for this universe, regardless of whether physical objects exist or not, (3) human perception and mental organization can detect natural patterns not covered by physics, (4) computer science is a human-created system of abstraction not covered naturally by the human mind.

\begin{figure}
	\begin{center}
	\includegraphics[width=0.91\textwidth]{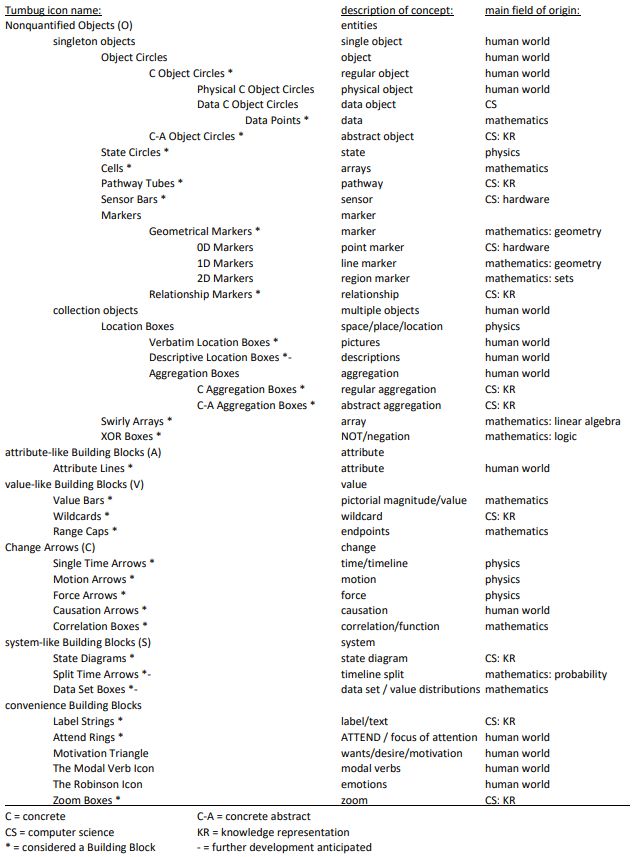}
	\caption{The current Building Blocks of Tumbug and their originating fields.}
	\label{fig:building-block-hierarchy-list-snap}
	\end{center}
\end{figure}

See Figure~\ref{fig:building-block-hierarchy-list-snap} includes the same list with more detail, but with the corresponding Building Blocks of Tumbug, in all their higher and lower generalizations. The specific icons that implement these concepts have names that are capitalized. These icons visually tend to be a type of Box, Arrow, Circle, Bar, Line, or String. A few of these icons have not yet been used for any encountered sentences, but are believed to be essential for certain future extensions and modifications of Tumbug. Since those unused Building Blocks are untested, those are the Building Blocks most likely to have unforeseen drawbacks in their current form. All these Building Blocks of Tumbug are discussed in detail later in this document.

Relative to the importance of the five overall categories (S, C, O, V, A) that are clearly defined by SCOVA (see the section on SCOVA), the particular choices of categories to be considered Building Blocks of Tumbug are relatively unimportant, other than for historical value in that those icons gave rise to the more general SCOVA insight and are more frequently used than other categories with other levels of generality. For example, since a Physical C Object Circle and a Data C Object Circle can be used interchangeably with ease, with possibly nobody even noticing the difference in appearance of their respective icons (viz., a solid line versus a dashed line), it is not very useful to consider those different Building Blocks. At the other extreme, a Change Arrow is too general to be of practical value in translating sentences into Tumbug because, for example, time and cause-and-effect are very different concepts, so the Change Arrow icon is not considered a Building Block, only a Basic Building Block of theoretical value for categorization purposes. In turn, the exact count of Building Blocks in Tumbug is not particularly important, although an approximate count is useful for comparing Tumbug to CD theory, or in estimating the amount of time it would require to code Tumbug, for example.

Regardless of topic, the concepts and relationships between concepts must be represented somehow, and some system or systems must bear the burden of performing that task. Ideally there would exist a universal standard representation system, a system that Roger Schank sought for language. Since Tumbug largely follows this ideal, and since Tumbug shows signs of exceeding the representative power of CD theory, Tumbug may be the representation system needed to replace math in certain areas such as AI.
 
\begin{figure}
	\begin{center}
	\includegraphics[width=0.75\textwidth]{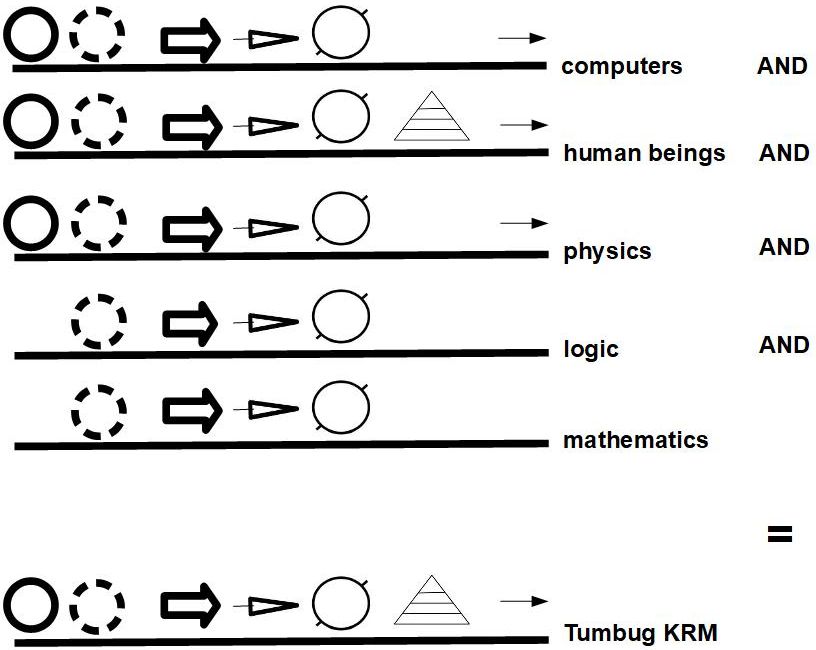}
	\caption{The Tumbug KRM considers this collection of the fundamental entities from several systems from the real world, all AND-ed together for to cover the basics of every one of those fields.}
	\label{fig:big-picture}
	\end{center}
\end{figure}

The goal of Tumbug is to collect all concepts of the real world that humans would reasonably want to express in language, and then to represent those concepts with images. Mathematics and its branches are obvious candidates for sources of concepts, and those fields have been used in AI for decades. Physics and model-based reasoning (e.g., Luger and Stubblefield 1998, p. 231) have also been used in AI, though to a lesser extent. The world of human beings, which includes psychology and sociology, contains concepts (especially the concept of "object" and "motivation") not covered in the hard sciences of inanimate objects, so Tumbug needs to include those concepts, also. Several knowledge representation concepts (especially general aggregations and labels) do not occur in their general form in the more general fields so computer science must be included. (For example, math has sets and parentheses, but does not generalize these notions into a general "aggregation" concept.) (See Figure~\ref{fig:big-picture}.) Because of Tumbug's eclectic nature and the infinite nature of the real world, Tumbug is not guaranteed to cover every basic concept in the real world, but in practice its span is so extreme that the author has found no lacunas despite extensive perusal of several fields. If a lacuna does exist, it would probably be found in some of the less frequent concepts of mathematics, possibly from group theory or topology.

\section{Object-like Building Blocks of Tumbug (O)}

The term "Building Blocks" comes from other authors' descriptions of Roger Schank's CD theory (Schank 1975), evidently not from Schank himself. More specifically, the term "building blocks" is used to describe all the components of CD theory, including the Primitive Acts and the Primitive Conceptual Categories. The term "Building Blocks" is used in this document in the same way as in CD theory: to describe Tumbug's basic components, especially since CD theory and Tumbug have roughly the same intent: to form a foundation on which any action can be described. Tumbug's Building Blocks were derived from generalizing many examples of sentence representation, primarily from the following sources: (1) WS150 problems, (2) CD theory, (3) Kolln grammatical examples, and (4) the author's own insights.

Terminology used in this document for CD theory:\\

\setlength\parindent{0pt}
Primitive Conceptual Categories = \{PP, ACT, PA, AA, LOC, T\}\\
Primitive Acts = \{ATRANS, PTRANS, PROPEL, MTRANS, MBUILD, SPEAK, ATTEND, MOVE, GRASP, INGEST, EXPEL\}\\
building blocks = Primitive Conceptual Categories $\cup$ Primitive Acts
\setlength\parindent{24pt}

\subsection{Singleton objects}

\subsubsection{Object Circles}

The preliminary letter "C" is a formality that stands for "concrete." Usually C Object Circles are called simply "Object Circles."

\textbf{1. C Object Circles}

\textbf{1.1. Physical C Object Circles}

\textbf{1.1.1. General}

\begin{figure}
	\begin{center}
	\includegraphics[width=0.15\textwidth]{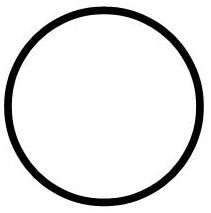}
	\caption{Tumbug's icon for a Physical C Object Circle.}
	\label{fig:icon-object-circle-plain}
	\end{center}
\end{figure}

A Physical C Object Circle is shown in Figure~\ref{fig:icon-object-circle-plain}.

Computers ultimately represent everything with numbers. This is because computers evolved from calculating machines that work with only numbers, and because the foundation of all science is mathematics, which primarily uses numbers. (Some exceptions are group theory and topology.) Even images are represented in a computer by matrix-like file formats such as BMP, JPG, and GIF, which are ultimately only numbers arranged in an orderly manner in space. Mathematics, however, is only an abstraction of the real world. In the real world, the fundamental entities of importance that are perceived and mentally manipulated by animals are objects, which are different than both numbers and pictures.

It is probably this last observation that was primarily responsible for the development of object-oriented programming (OOP), which groups data into an abstract data structure called an "object" whose simulated behavior is similar to that of its corresponding real-world object. Tumbug likewise treats objects as the fundamental entity of importance. The Tumbug icon used to represent an object is usually a plain one-lined circle called a "C Object Circle."

\begin{figure}
	\begin{center}
	\includegraphics[width=0.50\textwidth]{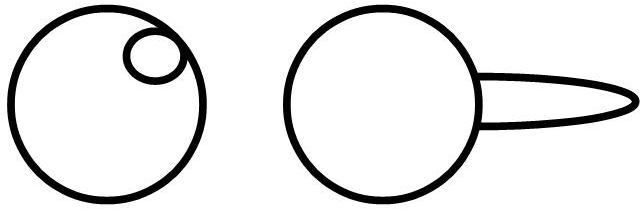}
	\caption{Left: Example of a physical object with interior structure. Right: Example of an object with exterior structure.}
	\label{fig:icon-object-circle-internal-and-external}
	\end{center}
\end{figure}

Any objects where more detailed anatomy (e.g., taste buds on the object's interior, or an arm on the object's exterior) is important to represent can be drawn as bumps or protrusions on the perimeter of the C Object Circle, protruding in the appropriate direction, and can have any desired size, length, or shape, though typically a single truncated or elongated circle or ellipse is sufficient for the typical level of detail in WS problems. See Figure~\ref{fig:icon-object-circle-internal-and-external}.

\begin{figure}
	\begin{center}
	\includegraphics[width=0.50\textwidth]{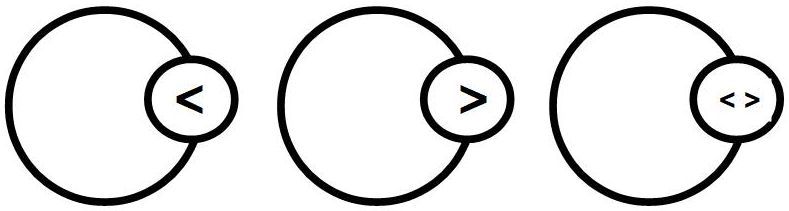}
	\caption{Left: Object with input sensor. Middle: Object with output generator. Right: Object with input-output.}
	\label{fig:icon-object-circle-sensors}
	\end{center}
\end{figure}

Very commonly an object, especially a person, needs to be represented as generating output (e.g., speaking, shouting, gesturing) or actively processing input (e.g., watching, looking, listening). Logically this would involve an anatomical structure that is partly inside and partly outside the body, so Tumbug uses a smaller C Object Circle on the border of the entity's C Object Circle to represent such an input-output device. The symbol "$\langle$" can be placed on this structure to designate it as an input device (sensor), "$\rangle$" for output, and "$\langle$ $\rangle$" for either input or output, as shown in Figure~\ref{fig:icon-object-circle-sensors}.

Why are Tumbug representations so simplistic and abstracted, instead of being more detailed and realistic, for example by showing arms and mouths? The answer is that Tumbug in a future software implementation would eventually need to view and interpret its own icons, therefore the simpler the icons that Tumbug must interpret, the better. This is especially true in language translation, where the Tumbug representation would need to be read after it had been stored, in order to make the final translation to the target language.

C Object Circles are allowed to overlap, although such a need tends to be uncommon. For example, two circular spots of light would logically be represented as two C Object Circles, and since spots of light can overlap, C Object Circles should also be allowed to overlap. Similar logic would apply to clusters of objects, where each cluster could be represented by one C Object Circle or by one Aggregation Box, and would apply to set boundaries as seen in Venn diagrams, where each set boundary could be represented by one C Object Circle or by one Aggregation Box.

\textit{This article's convention: Tumbug object icons are kept as simple as possible, as often as possible. For example, details such as hands on arms or taste buds in the mouth are typically never shown unless relevant to the problem, or unless explicitly stated in the sentence to be represented.}

\textbf{1.1.2. Icon variations}

\begin{figure}
	\begin{center}
	\includegraphics[width=1.00\textwidth]{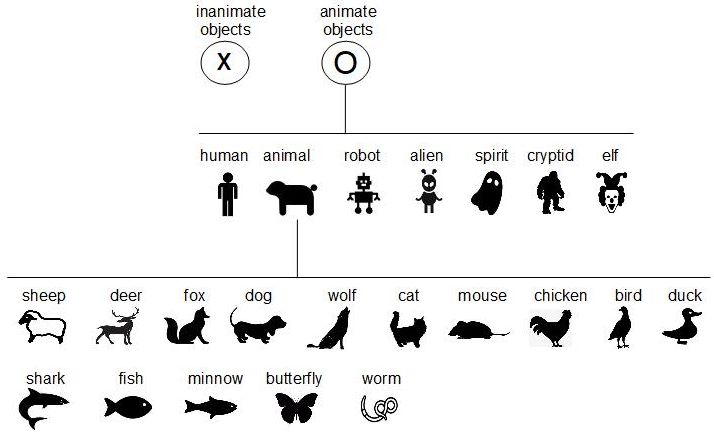}
	\caption{Elaborate hierarchies of types of animate objects can be created with icons for a Tumbug/WS library.}
	\label{fig:animate-entity-hierarchy}
	\end{center}
\end{figure}

For the purposes of WS problem representation, Tumbug can be easily and optionally supplied with icons for any encountered concepts whatsoever, especially for physical objects. A predefined class of intelligent, human-like entities seems particularly useful for high-level, cross-ideology-set discussions in philosophy, science, politics, and religion: {human, generic animal, robot, alien, spirit, cryptid, elf}. At the lower-level end of the hierarchy, the set of animals {sheep, deer, fox, dog, wolf, cat, mouse, chicken, bird, duck, shark, fish, minnow, butterfly, worm} would be particularly useful because those are all the animals specifically mentioned in WS150 problems, the solution of which one Tumbug-based algorithm was originally designed to provide. The hierarchy of Figure~\ref{fig:animate-entity-hierarchy} shows animate object types and their icons is only one example of the endless elaborate hierarchies that can be created. 

The exact icons used are not critical, though the existence of such a hierarchy of objects is useful. Such a library is useful because the more specific the Tumbug icons are, the more readily a Tumbug diagram can be understood. One downside of any hierarchy, however, is that a hierarchy is necessarily rigid, therefore any unexpected item (such as a spork) is automatically excluded from fitting into any existing classes (such as spoon or fork). In order to keep this presentation simple, the circle with "X" and circle with "O" are not used elsewhere in this document, but in other documents such a distinction could be quite useful.

Roger Schank referred to objects in CD theory as "PP," which stands for "picture producers," meaning objects that have a physical appearance that typically comes to mind upon hearing or seeing the word for that concept. This term suggests that use of icons would be particularly beneficial to use in place of Physical C Object Circles in Tumbug whenever possible. Throughout this document, icons may appear in place of Physical C Object Circles.

As an aside, the icons of Tumbug can be grouped into different types:

\begin{enumerate}
	\item
		geometrical icons - very simple circles, rectangles, arrows, lines, etc.
	\item
		composite  geometrical icons - organized groups of geometrical icons
		\begin{enumerate}
				\item
					composite tangible geometrical icons - for articulators, sensory organs, etc.
				\item
					composite intangible geometrical icons - for emotions, verb modality, etc.
		\end{enumerate}
	\item
		realistic icons - for cats, dogs, computers, bottles, etc.
\end{enumerate}

\textbf{1.2. Data C Object Circles}

\textbf{1.2.1. General}

\begin{figure}
	\begin{center}
	\includegraphics[width=0.15\textwidth]{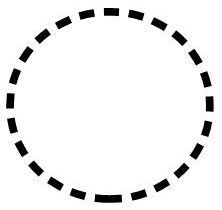}
	\caption{Tumbug's icon for a Data C Object Circle.}
	\label{fig:icon-data-object-circle}
	\end{center}
\end{figure}

A complication in the real world, especially in the modern world, is that many of the "objects" being used are collections of data, not physical objects. This distinction can be fairly important in the meaning and implications of sentences, therefore Tumbug modifies the Physical C Object Circle icon so that the circle has a dotted border (or sometimes a dashed border) when it represents a label-worthy chunk of data, as shown in Figure~\ref{fig:icon-data-object-circle}. As with physical objects, data objects can have attributes, such as topic, format, size in number of bytes, truthfulness, source, and so on.

Data C Object Circles can be thought of as envelopes or flash drives that hold their contents inside the circle. For example, a digital photo is a data package whose contents could easily be represented verbatim inside a Data C Object Circle by placing the photo inside a Verbatim Box inside the C Object Circle. 

\textbf{1.2.2. WS150 example: \#35 (subway)}

\begin{figure}
	\begin{center}
	\includegraphics[width=0.50\textwidth]{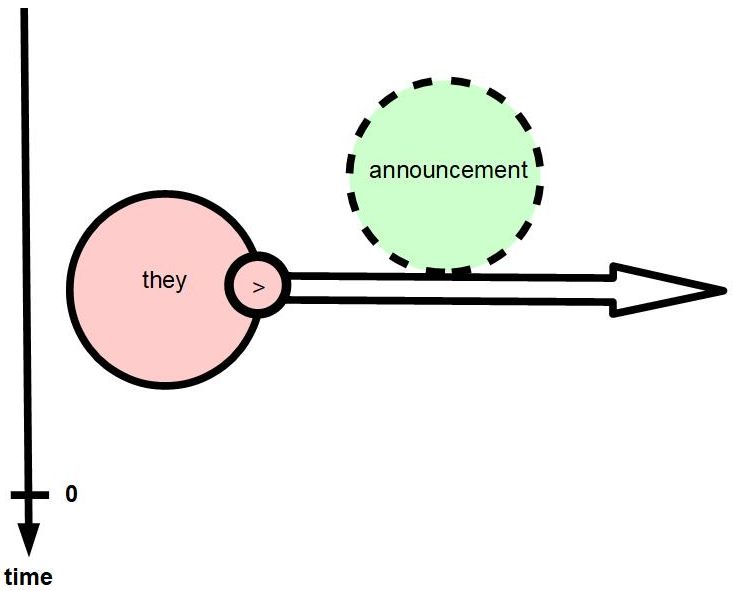}
	\caption{[35] Tumbug for "They broadcast an announcement."}
	\label{fig:ws-035}
	\end{center}
\end{figure}

The example in Figure~\ref{fig:ws-035} is from a portion of WS150 question \#35. A Data C Object Circle is necessary because data is implied via the words "broadcast" and "announcement."

"[35] They broadcast an announcement, but a subway came into the station and I couldn't hear it. What couldn't I hear? POSSIBLE ANSWERS: \{the announcement, the subway\}"

As with physical objects, data objects have direction of travel, although their transmission pattern can be more flexible since data without a carrier path they can either be transmitted in a straight line (as with a laser beam modulated to carry a message) or in a spherical plane (as with a loudspeaker broadcasting a message). Dashed planes are more difficult to draw than dashed lines, so a Tumbug diagram uses three representative lines with a single Data C Object Circle and a single Motion Arrow to help convey the concept of an emanating plane wave using simpler icons that are already used in simpler scenarios.

Conceivably a system of icons could be developed for data objects as well as physical objects. For example, speech (which is particularly common in WS examples and grammatical examples) could have its own icon, as could vision.

\begin{figure}
	\begin{center}
	\includegraphics[width=0.60\textwidth]{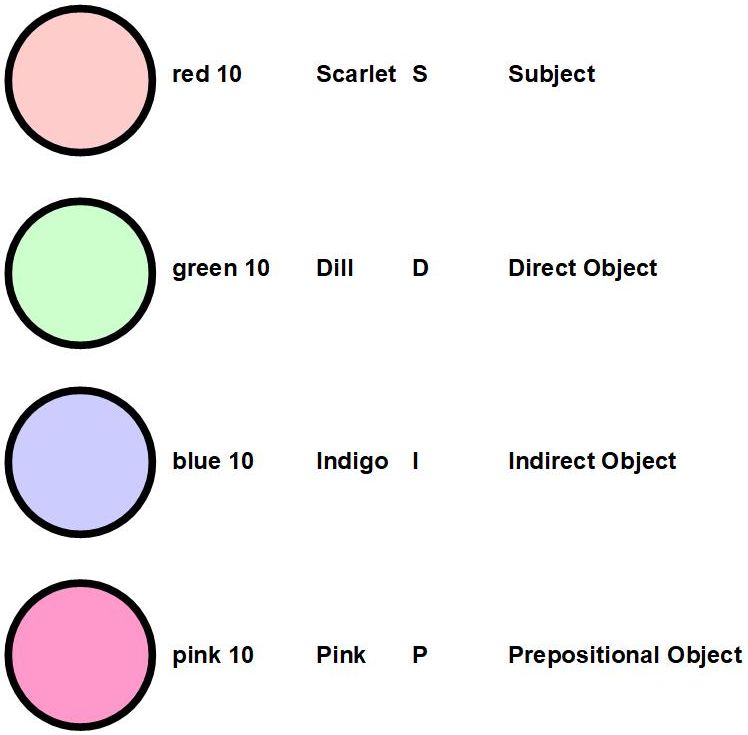}
	\caption{This is the Tumbug color coding scheme used in this document for the different types of grammatical objects, and the underlying logic for this scheme. Colors are optional in Tumbug, however.}
	\label{fig:convention-color}
	\end{center}
\end{figure}

\begin{figure}
	\begin{center}
	\includegraphics[width=0.60\textwidth]{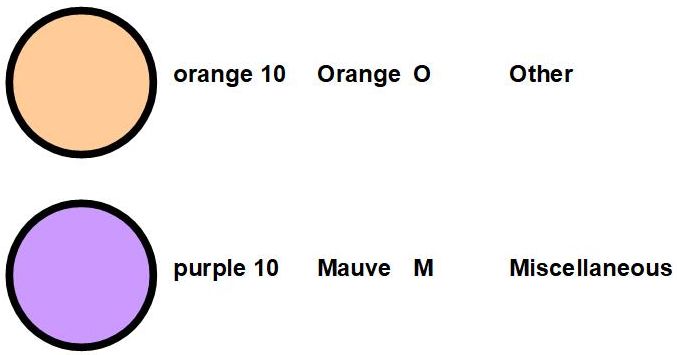}
	\caption{This is the Tumbug color coding scheme used in this document for other types of objects, and the underlying logic for this scheme. Colors are optional in Tumbug, however.}
	\label{fig:convention-color-other}
	\end{center}
\end{figure}

\textit{This article's convention: Although Tumbug diagrams never need to be in color, a color-coding convention has often been used on C Object Circles in this document to aid understanding of grammatical examples, as shown in Figure~\ref{fig:convention-color} and Figure~\ref{fig:convention-color-other}}.

\textbf{1.2.3. Data Points}

\textit{More general Building Block: Data C Object Circle}\\
\textit{Simplifications from the more general Building Block: (1) Only 1-2 attributes used, namely the attribute-value pairs at which it can be plotted, and the event(s) from which the measurement was taken.}\\

\begin{figure}
	\begin{center}
	\includegraphics[width=0.70\textwidth]{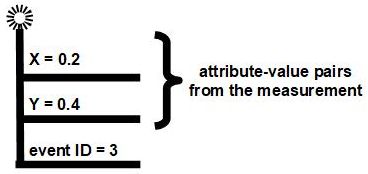}
	\caption{A Data Point is a small Data C Object Circle that contains only attribute-value pairs and optionally a reference to the event that produced it.}
	\label{fig:icon-data-point}
	\end{center}
\end{figure}

Figure~\ref{fig:icon-data-point} shows a specific Data Point with two attribute-value pairs. Since a single data point may have arisen from more than one event or observation or measurement, in general there may be a list of event IDs in the event ID field.

\textbf{2. C-A Object Circles}

\begin{figure}
	\begin{center}
	\includegraphics[width=0.50\textwidth]{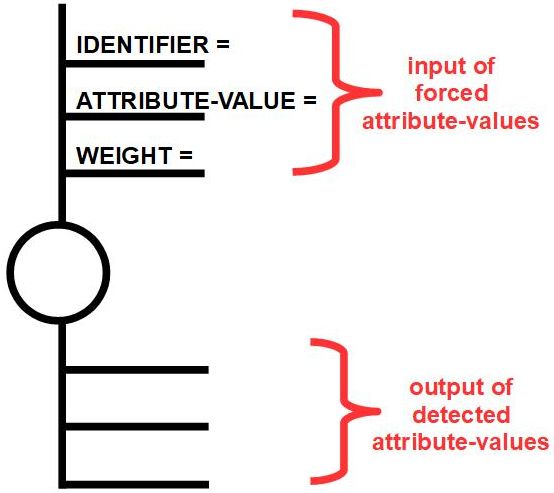}
	\caption{Tumbug's icon for a C-A Object Circle.}
	\label{fig:icon-ca-general}
	\end{center}
\end{figure}

The Tumbug icon for a C-A Object Circle is shown in Figure~\ref{fig:icon-ca-general}.

"C-A" stands for "concrete-abstract," which is a pair of concepts at the two extreme ends of the abstraction spectrum. Both C Object Circles and C Aggregation Boxes can be abstracted to produce C-A versions of those icons. In both cases some attribute-values (those protruding above the associated icon) are forced upon the object by the user, and the other attributes (those protruding below the associated icon) are the resulting detected attribute-values. The value of the outputs may or may not have been influenced by a Correlation Box. For example, an output could be a default value based on the object's class, and untouched by a Correlation Box.

\begin{figure}
	\begin{center}
	\includegraphics[width=0.25\textwidth]{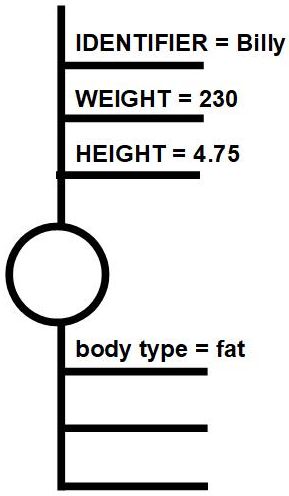}
	\caption{If Billy's weight is fixed at a high value and his height is fixed at a low value, then he is forced to be fat.}
	\label{fig:icon-caab-example-billy}
	\end{center}
\end{figure}

An example of usage of a C-A Object is shown in Figure~\ref{fig:icon-caab-example-billy}. Other uses of C-A objects can sometimes be found in the WS150 problems. For example:

"[9] The large ball crashed right through the table because it was made of steel. What was made of steel? POSSIBLE ANSWERS: \{the ball, the table.\}"

In this case, the composition of the unknown object is fixed at steel, which forces the object's weight and strength to be very high. Since tables tend to be made of wood, which forces a table's weight and strength to be only moderate (or at most, high), the ball described will tend to have more weight and strength than the table, which allows default reasoning to occur and for the correct default answer to be chosen.

\subsubsection{State Circles}

\begin{figure}
	\begin{center}
	\includegraphics[width=0.13\textwidth]{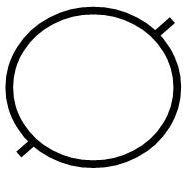}
	\caption{Tumbug's icon for a State Circle.}
	\label{fig:icon-single-state-circle}
	\end{center}
\end{figure}

Figure~\ref{fig:icon-single-state-circle} shows a State Circle

In Tumbug the need to represent states is almost as common as the need to represent objects. This is because verbs apply to states almost as frequently as they apply to objects, and the best icon to use is the icon whose concept most closely matches the sentence meaning, even if the specific words of the sentence and their grammatical order superficially suggest otherwise. The State Circle icon used for a state in Tumbug is intended to resemble the Greek phi symbol ($\phi$), which is used in physics for phase, which is a concept that is similar to state. The similarity between the C Object Circle icon and State Circle icon is intentional since it is sometimes desirable to use a State Circle instead of a C Object Circle, so swapping one for the other allows the viewer to overlook the minor conceptual difference when interpreting a Tumbug diagram quickly.

Ordinarily a State Circle is only a part of a state diagram, as described in the section on State Diagrams.

\subsubsection{Cells}

The meaning of "Cell" in Tumbug is identical to the meaning of "cell" in array terminology: A cell is a discrete location that can hold a single object. The object may be a number (as in linear algebra), a text string (as in data bases or spreadsheets), software objects (as in OOP), or any other type of object, virtual or real. Cells are usually used only in composite structures such as vectors, arrays, or jagged arrays.

The Cell concept differs slightly from the Tumbug concepts of Verbatim Box and C Aggregation Box since a Cell may hold only one object whereas those other structures may hold any number of objects. A Cell is similar to a State Circle, except the contents of a State Circle must be a state whereas a Cell can contain an object that is not a state. A Cell could also be accurately called a "pigeonhole," as in the "pigeonhole principle," since a cell conforms to the same constraint of one object per location.

\subsubsection{Pathway Tubes}

\begin{figure}
	\begin{center}
	\includegraphics[width=0.50\textwidth]{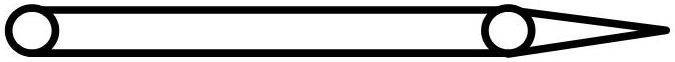}
	\caption{Tumbug's icon for a Pathway Tube.}
	\label{fig:tube-horizontal-right}
	\end{center}
\end{figure}

Figure~\ref{fig:tube-horizontal-right} shows a Pathway Tube.

\begin{figure}
	\begin{center}
	\includegraphics[width=0.50\textwidth]{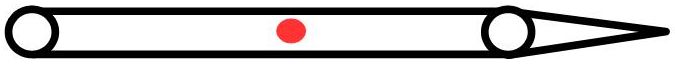}
	\caption{Pathway Tubes normally contain a 0D Marker, like the tokens of Petri nets.}
	\label{fig:tube-spot}
	\end{center}
\end{figure}

A Pathway Tube shows a pathway for anything. It can represent a hiking path used by people, a pipe for carrying water, a track for carrying trains or monorails, a fiberoptic path used by photons, a data path in the form of a USB cable between digital computers, a token path used by Petri nets, the arrows of logic flow direction in a flowchart, a wave guide for electromagnetic waves, or other. The pointed end of a Pathway Tube shows the direction of travel of objects or data within the tube/pathway, like a one-way street sign. The objects moving in a Pathway Tube can be represented by 0D markers, as shown in Figure~\ref{fig:tube-spot}.

Pathway Tubes in this document are always shown as straight, but in general they need not be. Straight Pathway Tubes can be connected end-to-end to implement changes in direction, if for example a vector graphics editor does not a simpler means to create such a shape.

\subsubsection{Sensor Bars}

\begin{figure}
	\begin{center}
	\includegraphics[width=0.50\textwidth]{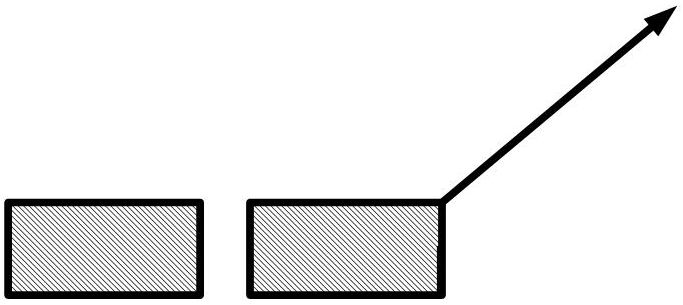}
	\caption{Tumbug's icon for a Sensor Bar (left), and Sensor Bar with a signal output line (right).}
	\label{fig:icon-sensor-2-sensor-bars}
	\end{center}
\end{figure}

A Sensor Bar is shown in Figure~\ref{fig:icon-sensor-2-sensor-bars}.

Tumbug contains Values Bars to represent values, but these values cannot be used unless the values can be read somehow. Sensor Bars solve this problem. Each Value Bar (by current convention) is typically outfitted with a Sensor Bar that has a signal output line to represent a signal being sent from the sensor to some other component if and only if that sensor is activated.

\begin{figure}
	\begin{center}
	\includegraphics[width=0.50\textwidth]{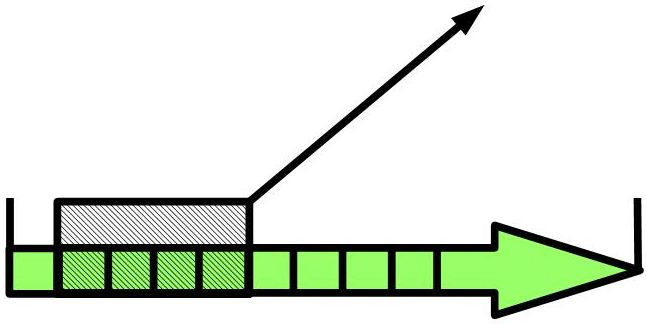}
	\caption{A Sensor Bar overlaid upon a Value Bar.}
	\label{fig:icon-sensor-sensor-and-bar}
	\end{center}
\end{figure}

\begin{figure}
	\begin{center}
	\includegraphics[width=0.50\textwidth]{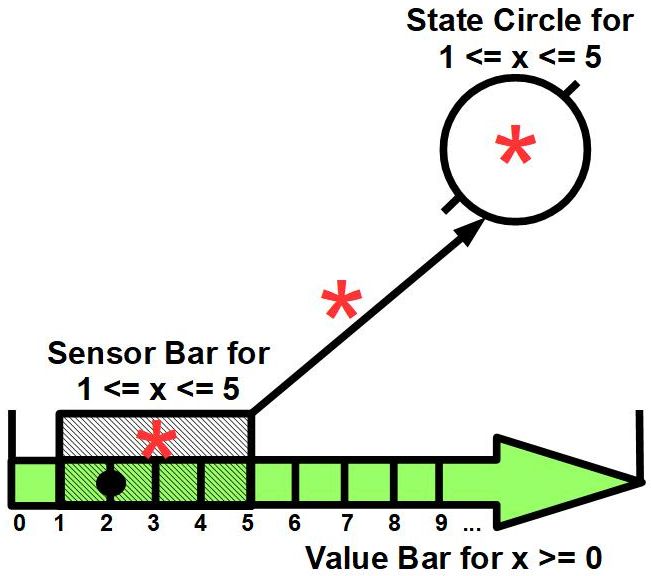}
	\caption{A Sensor Bar senses a value in the specified range within a Value Bar and sends a signal to a State Circle.}
	\label{fig:icon-sensor-sensor-and-circle}
	\end{center}
\end{figure}

Typically a Sensor Bar is overlaid upon a Value Bar so that if the value in the Value Bar falls within the span of the Sensor Bar then the Sensor Bar sends an output signal (marked here as "*") on its output line, as shown in Figure~\ref{fig:icon-sensor-sensor-and-bar} and Figure~\ref{fig:icon-sensor-sensor-and-circle}.

In turn, a single activated state within a state diagram may trigger further activity within that state diagram, or possibly within a program containing an "if" statement that relies on that state being active in order to choose with branch of code to execute next.

Value Bars and Sensor Bars are currently proposed for Tumbug for completeness purposes, but for diagram simplicity it could be assumed that the system automatically activates whichever applicable state needs to be read by the system, and does so in the background, without need to illustrate the details of how that state became activated.

Sensor Bars in this document are shown with a -45-degree hatching pattern, whereas 2D Markers are shown with a +45-degree hatching pattern.

\subsubsection{Markers}

Technically, a Marker in Tumbug can be defined as a labeled manifold that marks something of significance in a Tumbug diagram. For example, a non-intersecting border drawn around an irregular region in a plane would be a 1D manifold in 2D space that would bear a label mentioning that it is a border, and typically also what the border surrounds. Markers can be of any dimension, and can occupy any dimension, and can have any non-intersecting shape. For example, a wire that traced the edges of a collection of stacked cubes would be a 1D manifold in 3D space. Markers can also have unbounded ends, such as an arrowhead at one end of a 1D line segment, or a 3D arrowhead pointing outward at the border of a 3D planet's atmosphere.

\textbf{1. Geometrical Markers}

\textbf{1.1. 0D Markers}

\textit{More general Building Block: C Object Circle}\\
\textit{Simplifications from the more general Building Block: (1) Only 1-2 attributes used, namely location, and optionally priority.}\\

\begin{figure}
	\begin{center}
	\includegraphics[width=0.25\textwidth]{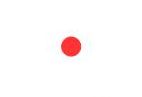}
	\caption{Tumbug's icon for a 0D Marker.}
	\label{fig:icon-marker-spot}
	\end{center}
\end{figure}

The zeroth dimension (0D) is a point, so in Tumbug a 0D Marker is a single point that is used to highlight some item or location. This 0D Marker icon is currently represented in Tumbug as a sizable red spot, as shown in Figure~\ref{fig:icon-marker-spot}.

Some practical uses of 0D Markers are:

\begin{itemize}
	\item
		A checkmark of any kind. (Note that this implements Kenneth Haase's red paint idea described by Marvin Minsky (Minsky 1986, p. 82).)
	\item
		A red dot on a map labeled "You are here."
	\item
		A radar blip that indicates the position of an aerial object detected by radar.
	\item
		A bouncing dot over lyrics to help singers keep track of the current part of the song.
	\item
		An "X" on a treasure map that marks the spot where the treasure is buried.
	\item
		A spot like the spot from a laser pointer that shows the part of an overhead slide being discussed by the lecturer.
	\item
		Color-coded labels on unmarked cardboard boxes.
	\item
		The desired attribute of a query.
	\item
		The current state within a state diagram. If the current state is in transition between two states then the 0D Marker can also represent this easily by being placed on the Pathway Tube itself.
	\item
		The current statement being executed within a flowchart.
	\item
		A token within a Petri net.
	\item
		A mental marker to remember a possible king position on a chessboard when visualizing possibilities in a chess endgame.
	\item
		A marker like a painted stone on a trail that indicates a place that was formerly visited, to prevent going in circles if lost on a hike.
	\item
		A position affixed to the rim of a rolling circle.
	\item
		The conjectured K-Lines of Marvin Minsky (Minsky 1986, p. 82) to mark tools that were used for a given job.
\end{itemize}

\begin{figure}
	\begin{center}
	\includegraphics[width=0.11\textwidth]{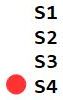}
	\caption{One common usage of a 0D Marker is to step sequentially through program statements.}
	\label{fig:icon-marker-code}
	\end{center}
\end{figure}

One use of 0D Markers in Tumbug in this document is to represent instruction pointers in an executing computer program. See Figure~\ref{fig:icon-marker-code}. Whereas a digital computer uses an instruction pointer that points to the instruction currently being executed in a program, Tumbug uses a small red dot as a visual means to mark this instruction in the program itself. Then, instead of incrementing an address in computer memory to step through a program, Tumbug moves a spot sequentially through a program. Since the program structure is already represented as a graph-like flowchart, use of a 0D marker in such a flowchart maintains the higher-level view of the program as completely visual.

\begin{figure}
	\begin{center}
	\includegraphics[width=0.75\textwidth]{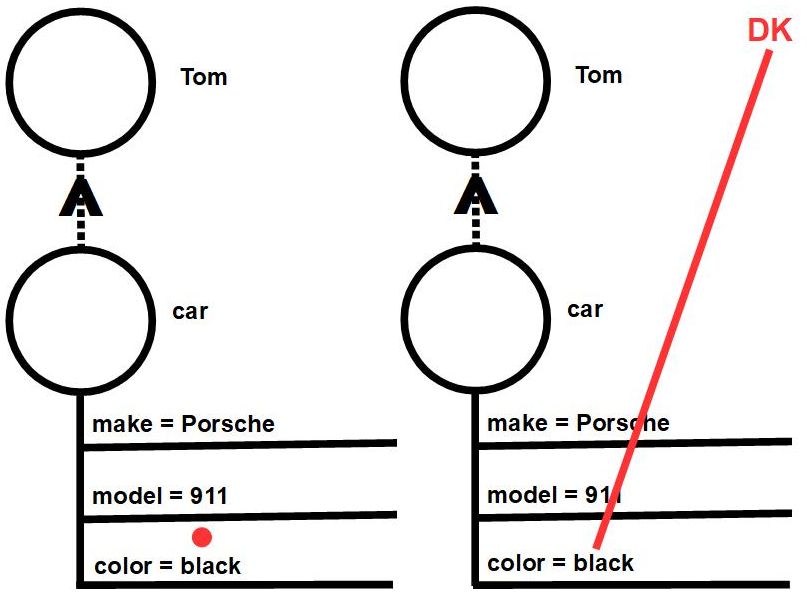}
	\caption{A slanted line in Tumbug is much more visible than a tiny 0D marker when diagramming queries.}
	\label{fig:icon-query}
	\end{center}
\end{figure}

An important use of 0D Markers is in diagramming queries. For example, a data base type query might equate to the question "What color is Bob's car?", which would place a 0D Marker with an "information goal" attribute atop the value of the "color" attribute in the data base record for Bob's car, then the value of that attribute would be noted. In this way an entire question reduces to a single placed 0D Marker. See Figure~\ref{fig:icon-query}, where the color red here substitutes for the attribute "information goal." Note that a single spot--the 0D Marker--is difficult to see on a diagram at a glance, therefore a clearer representation is recommended in practice, which is the representation with the slanted line with one end on the desired value and the other end marked with "DK" (= Don't Know). (Currently, use of "?" to represent queries in Tumbug is discouraged since "?" has a different meaning as a regular expression wildcard.) Since all Tumbug lines are normally drawn either horizontally or vertically, a 1D Marker line that is slanted at about a 45-degree angle stands out at a glance.

In WS150 problems the queries are only slightly more complicated in that they are always about disambiguation, but 0D Markers can be used there, also. For example: "[108] Grace was happy to trade me her sweater for my jacket. She thinks it looks great on her. What looks great on Grace? POSSIBLE ANSWERS: \{the jacket, the sweater\}" In this query a 0D Marker would be placed atop value of the "clothing type" attribute for the item of warm clothing that Grace is wearing after the trade, with the same attribute "information goal" attached to the 0D Marker, then the value of that attribute would be noted. Oftentimes such queries require following an extra link. In this case, Grace would first need to be located in the Tumbug diagram, then the 0D Marker placed on her warm clothing.

\textbf{1.2. 1D Markers}

1D Markers are lines used as a visual aid to mark something as lying in a straight line. In the WS150 problems such lines would be desirable to show visual interference along a line of sight, for example. The lines could also be useful to show alignment, or to show an angle from a given point.

Some practical uses of 1D Markers are:

\begin{itemize}
	\item
		A straight line that makes clear the alignments of objects or text, whether through their tops, bottoms, centers, or other.
	\item
		A straight line that separates plotted data points into classes.
	\item
		A straight line across an opening in a concave geometrical figure, where the line separates interior from exterior.
	\item
		A straight line between extrema in a convex geometrical figure, where the line shows regions that can be obscured via occlusion from certain angles.
\end{itemize}

\textbf{1.3. 2D Markers}

\begin{figure}
	\begin{center}
	\includegraphics[width=0.50\textwidth]{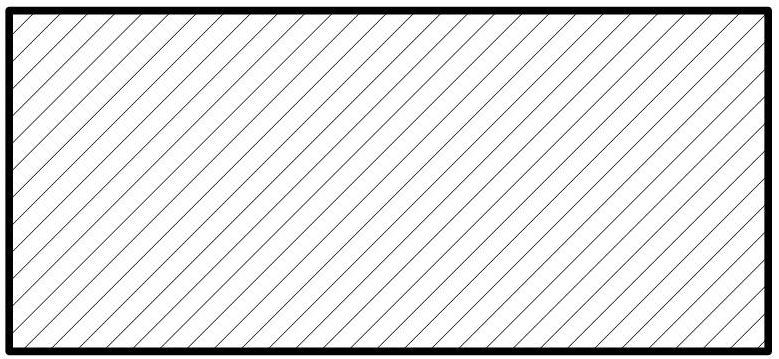}
	\caption{Tumbug's icon for a 2D Marker.}
	\label{fig:icon-marker-hatched-only}
	\end{center}
\end{figure}

\begin{figure}
	\begin{center}
	\includegraphics[width=0.50\textwidth]{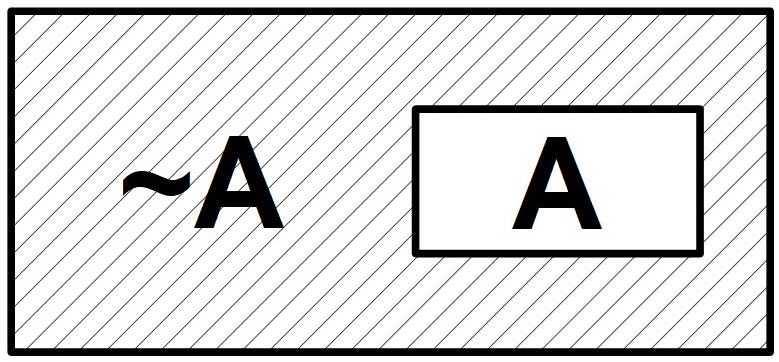}
	\caption{2D Markers are usually used to show 2D ranges within sets represented by Venn diagrams.}
	\label{fig:icon-marker-hatched-and-unhatched}
	\end{center}
\end{figure}

2D Markers are merely shaded regions, as in Figure~\ref{fig:icon-marker-hatched-only}. Such regions are probably most often used for Venn diagrams, such as to show intersection of two sets as in Figure~\ref{fig:icon-marker-hatched-and-unhatched}, or to show the conceptual allowable range of elements of one set. 2D Markers in this document are shown with a +45-degree hatching pattern, whereas Sensor Bars are shown with a -45-degree hatching pattern.

Some practical uses of 2D Markers are:

\begin{itemize}
	\item
		A circle around an important region within a picture or within text.
	\item
		A "set cordon"--an irregular contour that shows which parts of a Venn diagram are being considered.
	\item
		A contour that encloses data points that are considered part of a given classification.
	\item
		Boundaries between regions on a geographical map.
	\item
		The four borders on a JPG image that is being cropped.
\end{itemize}

\textbf{2. Relationship Markers}

\begin{figure}
	\begin{center}
	\includegraphics[width=0.25\textwidth]{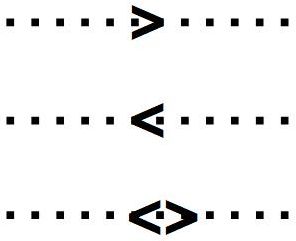}
	\caption{Relationship Markers have centered arrowheads on dotted lines, in any of three possible arrangements.}
	\label{fig:icon-relationship}
	\end{center}
\end{figure}

See Figure~\ref{fig:icon-relationship}. Relationship Markers use dotted lines, which no other Tumbug convention uses, which is done for uniqueness, clarity, and consistency. The consistency arises because a 0D Marker can already be considered a Relationship Marker with a single point, so two points could be considered a 1D Marker, three non-colinear points could be considered a 2D Marker, and so on.

Relationship Markers are very generic markers that point out any important relationship that the user would like to flag. Depending on how Relationship Markers are used, Relationship Markers may be optional or required. 

In Venn diagram representations of First Order Predicate Calculus, a Relationship Marker would almost be required in some cases. For example, the final inference of the "Socrates is mortal" example already shows Socrates as an element of a subset of both men and mortals, but the user may have overlooked this 2-step removed inference during the derivation of that final result since each addition of a Venn diagram circle potentially creates several new inferences simultaneously, so the important final inference should be explicitly pointed out by a unidirectional Relationship Marker pointing from Socrates to mortals.

Relationship Markers often have direction. For example, the sentence "Bill is Mark's uncle" must be stated unidirectionally in text in the form "X is Y's uncle" since the converse "Y is X's uncle" is not true, therefore one arrowhead (centralized on the arrow's stem) is used to point in the needed direction (per the user's convention), such as from X to Y to show the direction of meaning of "X is a/an of Y." Some sentences are bidirectional, however, such as "Denny and Wanda are siblings," therefore in that case two arrowheads pointing in opposite directions are used to show both directions of unidirectional meaning simultaneously.

Relationship Markers are also nearly required for genitive case, meaning possessive situations, such as "Sam's drawing" [14], "Ann's son" [27], and "Joe's uncle" [28], all of which are examples from WS150. (The version of Tumbug described in the 2022 article on Tumbug described such possessive use of Relationship Markers as "genitive arrows" (Atkins 2022), but Relationship Markers are more general than just genitive case, such as in the phrases "Socrates is mortal" or "Denny and Wanda are siblings.")

Relationship Markers are essentially the same node-with-link data structures used in semantic networks. Since many sentences and phrases can be easily represented with semantic networks, this suggests that a good use of Relationship Markers might be as an intermediate stage of conversion from text to Tumbug.

Relationship Markers could be considered a compound type of icon since Relationship Markers contain 1D lines that are bent into 2D.

\subsection{Collection objects}

\subsubsection{Location Boxes}

\begin{figure}
	\begin{center}
	\includegraphics[width=0.50\textwidth]{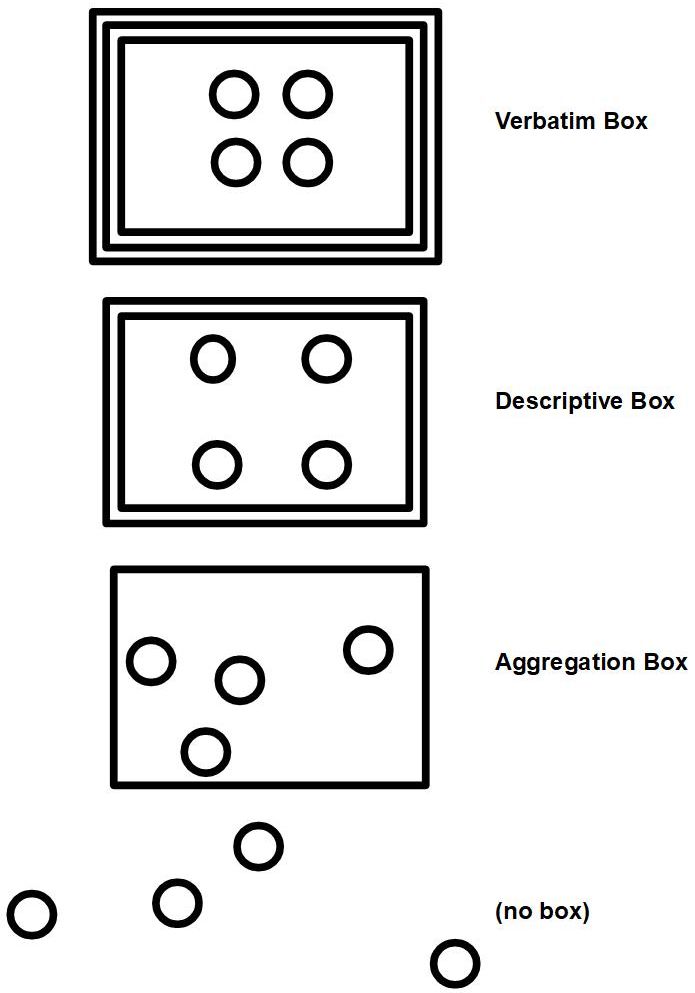}
	\caption{Tumbug has three types of Location Boxes that span a spectrum of severity of constraints on the locations of the items inside. Here the original locations of four Object Circles are shown by the Verbatim Box, this square arrangement has been scaled to be larger but still maintains the same shape in the Descriptive Box, and in the Aggregation Box there are no constraints on location at all except that the Object Circles must be inside the box. If not even an Aggregation Box exists then no Object Circles have any location constraints whatsoever.}
	\label{fig:icon-location-frame}
	\end{center}
\end{figure}

"Location Boxes" is a generic term of Tumbug that currently refers to three types of rectangular icons in Tumbug, each type of which is intended to be a border in which to hold multiple items inside, to show the locations of those items inside the box, and to show the location of the Location Box itself, at least relative to any other icons that may exist outside the box. Figure~\ref{fig:icon-location-frame} shows that these three types of Location Boxes exist in a spectrum of severity of location constraints. These boxes may be any shape that is needed, although rectangular is the default shape.

\begin{itemize}
	\item
		Verbatim Box: every item inside must exist only in the place where it is shown. There is no flexibility.
	\item
		Descriptive Box: every item's location is subject to constraints that are specified with the box. There is some flexibility.
	\item
		Aggregation Box: any item is allowed to be in any location within the box. There is complete flexibility inside the box.
	\item
		no box: there are no constraints at all on the locations of the items. There is complete flexibility and there is not even a border.
\end{itemize}

\begin{figure}
	\begin{center}
	\includegraphics[width=1.00\textwidth]{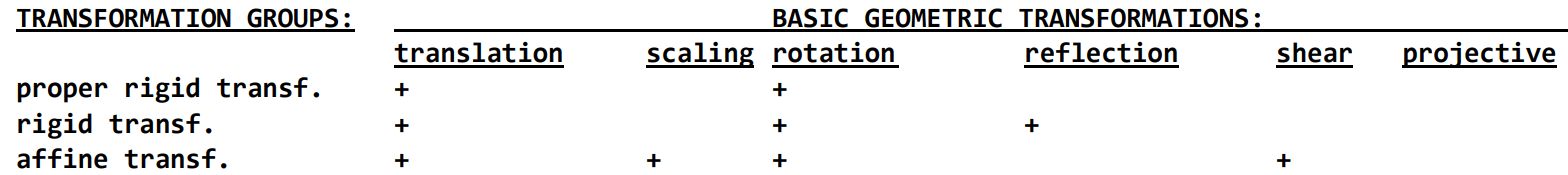}
	\caption{There exist several common types of transformation groups, each with a unique name and with a unique set of more basic transformations (translation, scaling, rotation, etc.) that are considered part of that group.}
	\label{fig:icon-location-comparison-table-snap}
	\end{center}
\end{figure}

Although the two extremes of this spectrum, namely Verbatim Boxes and Aggregation Boxes, are conceptually simple and fully defined, the middle of this spectrum, namely Descriptive Boxes, have tremendous flexibility, so would need more research to fully investigate. For example, Figure~\ref{fig:icon-location-frame} shows that four Object Circles in a square arrangement can be scaled to produce an unmistakably similar arrangement, but scaling is just one geometrical transformation. Another geometric transformation that would produce an unmistakably similar arrangement is translation. The result of rotation may also be considered an unmistakably similar arrangement, depending on a person's taste and goals. Figure~\ref{fig:icon-location-comparison-table-snap} shows several such geometrical transformations, and which common transformation groups include those basic transformations.

\textit{Further development anticipated: Since different types of Location Boxes are allowed to overlap or to be embedded in one another, precedence rules are likely to be needed, the same issue as scope of name binding in computer programming. Probably Verbatim Boxes at the strict end of the strictness spectrum would need to take highest precedence, for example.}

This document usually assumes that Descriptive Boxes: (1) retain the same spatial arrangement of the original items (i.e., allow differences in scaling), (2) disallow rotation, (3) disallow reflection and any other more complicated transformations such as projective transformations. Rotations are disallowed partly because each diagram is usually assumed to contain a gravity vector that points straight downward on the page, which in turn suggests that depicted items are usually viewed from the side, and are either resting flat on something or have an inherent bottom (such as buildings, transportation, people, animals, plants, and natural geographical features). Ideally all such assumptions should be stated somewhere and somehow within each Tumbug diagram or each project, which in turn suggests that it would be useful to attach codes to Descriptive Boxes, such as placing an "R" in the corner of a Descriptive Box to indicate that all objects in that box have the freedoms of transformation of the "Rigid transformations" group.

\textbf{1. Verbatim Boxes}

\begin{figure}
	\begin{center}
	\includegraphics[width=0.25\textwidth]{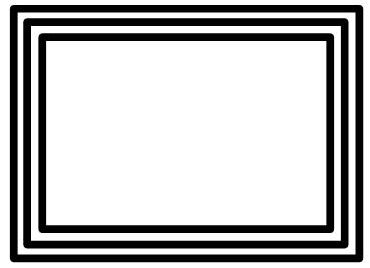}
	\caption{Tumbug's icon for a Verbatim Box.}
	\label{fig:icon-location-types-verbatim}
	\end{center}
\end{figure}

A Verbatim Box icon is shown in Figure~\ref{fig:icon-location-types-verbatim}.

\textbf{1.1. Concepts that imply space}

It should be clear that the concept of space is common in the WS150 problems. Below, two examples are given for each of the concepts of interior/exterior and direction. Together these two concepts imply the need to represent spatial relationships in some parts of some sentences. The Tumbug icon used to represent places inside of it is a double frame or a triple frame to suggest a picture frame. Any objects where space (i.e., interior/exterior or direction) is important to represent must be placed inside of this frame, which can have any desired size or length-to-width ratio. Some concepts that inherently imply space are: (1) interior or exterior, (2) direction, (3) position, (4) alignment.

The Verbatim Box is where \textit{bona fide} images are contained, therefore this is where Tumbug has its best prospects of demonstrating that its KRM is more powerfully expressive than any other existing KRM. "\textit{Bona fide}" images in this context mean images other than predetermined icons or text. For example, a textual description such as "two circles of diameter 1 have their centers aligned along the 64-degree angle line, and their centers are 2.7 diameters distance from each other" is a \textit{bona fide} image that potentially spans an uncountably infinite number of similar images by merely changing two numerical values. In this way Tumbug can output an uncountably infinite number of images. Similarly, in applications where the goal is to depict a default scene to aid the viewer's comprehension of that scene, any photo whatsoever may be inserted verbatim into the Verbatim Box. For example, in a Jonny Quest animated cartoon, stock real jungle scenery (and even background sounds) could be used as a background wallpaper within the Verbatim Box to make the cartoon more realistic. Even a video could be placed in the Verbatim Box, such as a video of rapidly passing city buildings as background for an animated film where the characters in the foreground are riding aboard a train in the city.

\textbf{1.1.1. Interior}

This example is from a portion of WS150 question \#52. A Description Box is necessary because the general interior of one object (viz., the fish) is implied by the word "ate," but no exact pictorial depiction of the event was provided.

"[52] The fish ate the worm. It was tasty. What was tasty? POSSIBLE ANSWERS: \{the worm, the fish\}"

\begin{figure}
	\begin{center}
	\includegraphics[width=0.50\textwidth]{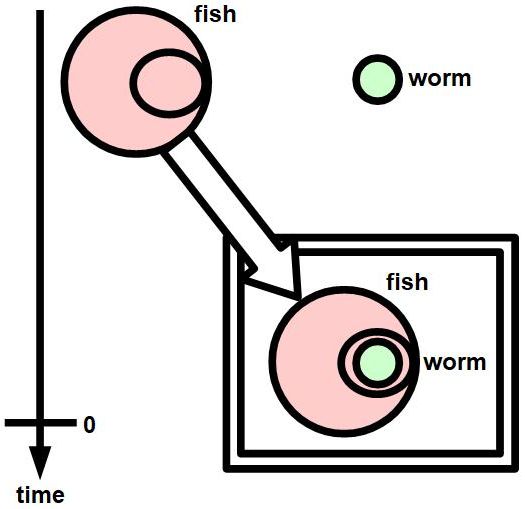}
	\caption{[52] Tumbug for "The fish ate the worm." (with abstract icons)}
	\label{fig:ws-052-objects}
	\end{center}
\end{figure}

The verb "eat" would be defined in object-and-action terminology as something like "Subject eats Object = Subject arranges for Object to go inside Subject's mouth, whereupon Subject closes Subject's mouth to entrap Object, then moves Object toward Subject's stomach to be digested." Since any such definition involves the concept of "inside," which implies the concept "interior," direction is inherent in the sentence meaning, therefore some type of Descriptive Box should be used, as shown in Figure~\ref{fig:ws-052-objects}.

In Figure~\ref{fig:ws-052-objects}, the fish has moved toward the stationary worm in time and has engulfed the worm into an anatomically interior position such as a mouth or stomach. This interior structure is represented by the smaller C Object Circle inside of the larger C Object Circle that represents the fish as a whole. The eaten worm then is represented as an even smaller C Object Circle inside of the fish's interior C Object Circle.

\begin{figure}
	\begin{center}
	\includegraphics[width=0.50\textwidth]{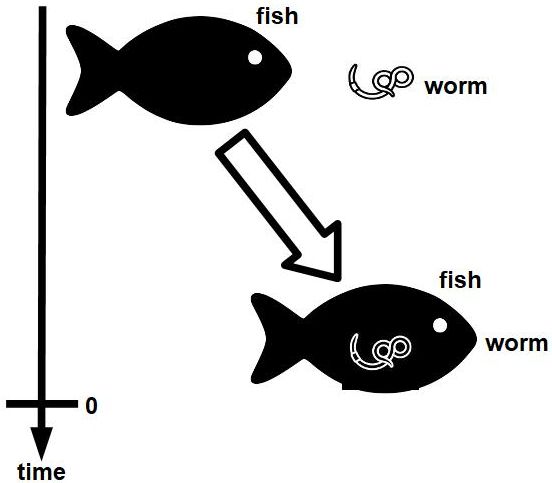}
	\caption{[52] Tumbug for "The fish ate the worm." (with realistic icons, and without emphasizing space)}
	\label{fig:ws-052-iconic}
	\end{center}
\end{figure}

In practice, any juxtaposition that involves "interior" tends to be so clear that the surrounding Description Box can optionally be removed from the diagram without confusion. Figure~\ref{fig:ws-052-iconic} shows a version of the fish diagram that removes the Location Box and--even more visually effective--uses realistic icons instead of C Object Circles.

The fish diagram with realistic icons is more standard for Tumbug, despite the intermediate positions in time not being shown. Any other 2D temporal representation, viz. an icon in discrete time or an icon in continuous time, will not be clear since the separate time axis in 3D space would not be shown, which would force a moving object to overlay part of itself in the same plane, which in total will result in a smeared, lengthy composite image. The most practical option for simulations is to render real-world time the same as rendering simulation time, so that an object moving in the real world will be seen as an object moving in the simulation. For applications that require careful consideration of all points along a trajectory, such as determining whether several rotating disks in a plane, each of different radius and speed, and each with a visible spot near its rim, will ever align, rendering time as a separate dimension might be preferable.

Ideally Tumbug should be implemented in at least 3D space so that the states of moving 2D objects can be seen easily and collectively across the entire time span.

\textbf{1.1.2. Direction}

This example is from WS150 question \#11. Some type of Location Box is necessary because the relative direction of an object is specified via the words "down" and "top."

"[11] Tom threw his schoolbag down to Ray after he reached the top of the stairs. Who reached the top of the stairs? POSSIBLE ANSWERS: \{Tom, Ray\}"

\begin{figure}
	\begin{center}
	\includegraphics[width=0.50\textwidth]{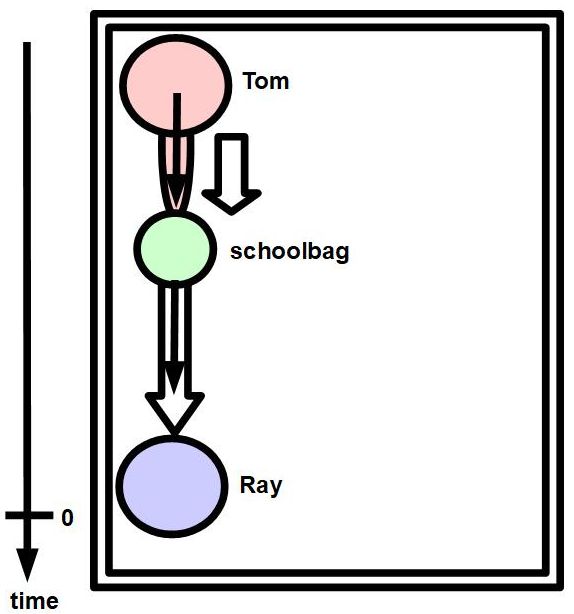}
	\caption{[11] Tumbug for "Tom threw the schoolbag down to Ray."}
	\label{fig:ws-011-no-steps}
	\end{center}
\end{figure}

If the phrase about the stairs is omitted for simplicity, then the partial sentence would be diagrammed in Tumbug as in Figure~\ref{fig:ws-011-no-steps}.

The "0" on the Time Arrow represents "now," so the fact that the event completely finished in the time before time 0 means that the event happened in the past. The direction of the backpack is straight down because the sentence says "down" and no mention or implication was made of the throwing having a horizontal component; if the schoolbag had also been thrown in a horizontal direction as well as downward, then the trajectory of the schoolbag would be shown as angled, say toward the right of the diagram, as it approached Ray at the bottom, who would also be positioned more toward the right.

\textit{Further development anticipated: Note that throwing the schoolbag requires a force, and gravity would be exhibiting a smaller force even without throwing. So far Tumbug examples have not attempted to show all applicable forces, the resulting force vector after adding component forces, or even different length arrows that are proportional to the different force magnitudes. Ideally such fine details should be worked out eventually, though in the WS150 examples the need to differentiate between force magnitudes does not ever appear to arise.}

\begin{figure}
	\begin{center}
	\includegraphics[width=0.50\textwidth]{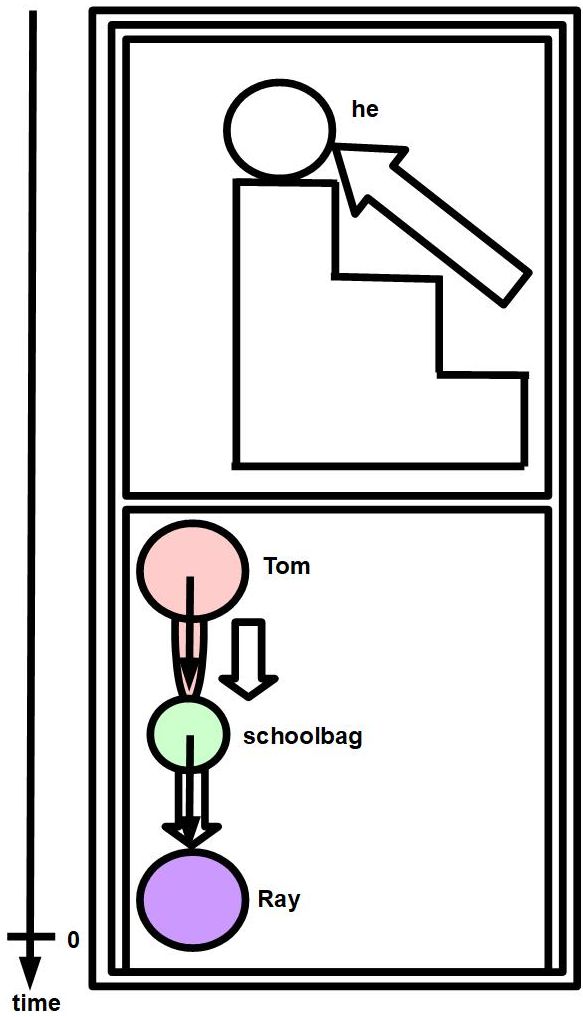}
	\caption{[11] Tumbug for "Tom threw the schoolbag down to Ray after he reached the top of the stairs."}
	\label{fig:ws-011-steps}
	\end{center}
\end{figure}

The full sentence would be diagrammed in Tumbug as in Figure~\ref{fig:ws-011-steps}.

Note that a longer Time Arrow was be used in order to show two events, one of which happened after the other. Each event--(1) the reaching of the top of the stairs, and (2) the throwing down of the schoolbag to Ray--is enclosed in its own C Aggregation Box to make the situation conceptually simpler. Since the person called "he" was not identified in the sentence, the diagram does not yet assign a color to the "he" C Object Circle , or assign the name on that "he" C Object Circle to "Tom." (That is for CSR to resolve at a later stage of processing.) The word "after" would be implied by the relative times of the two events. As before, relative to the current time "0," both events happened in the past. The word "top" is understood to mean "top of the diagram" since the direction "down" has already been established as the bottom of the page, and since "top" is defined as the opposite direction of "bottom."

\textbf{1.1.3. Position}

\begin{figure}
	\begin{center}
	\includegraphics[width=0.50\textwidth]{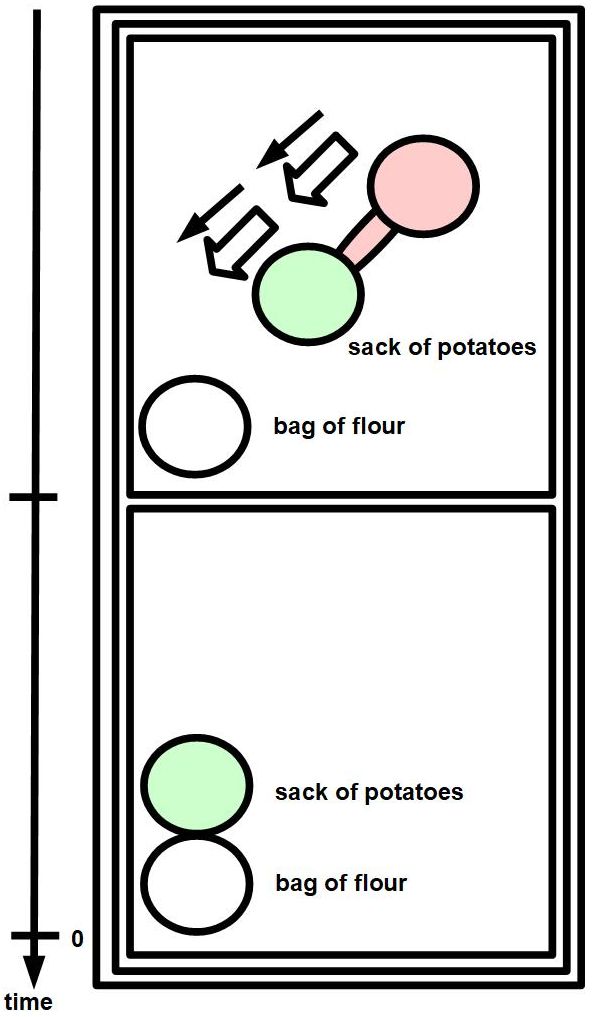}
	\caption{[19] Tumbug for "The sack of potatoes had been placed above the bag of flour."}
	\label{fig:ws-019}
	\end{center}
\end{figure}

Figure~\ref{fig:ws-019} shows an example from a portion of WS150 question \#19. Some type of Location Box is necessary because the relative position of an object is specified via the word "above."\\

"[19] The sack of potatoes had been placed above the bag of flour, so it had to be moved first. What had to be moved first? POSSIBLE ANSWERS: \{the sack of potatoes, the bag of flour\}"

\textbf{1.1.4. Alignment}

\begin{figure}
	\begin{center}
	\includegraphics[width=0.50\textwidth]{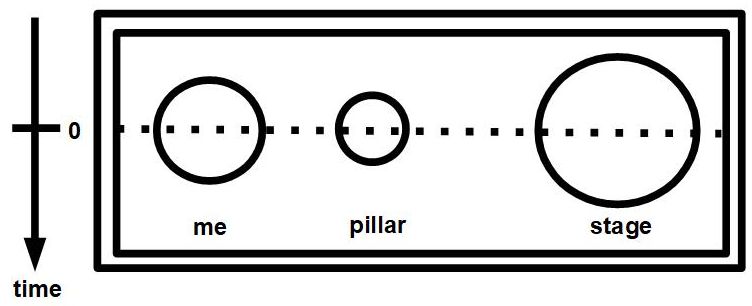}
	\caption{[34] Tumbug for "There is a pillar between me and the stage."}
	\label{fig:ws-034}
	\end{center}
\end{figure}

Figure~\ref{fig:ws-034} shows an example from a portion of WS150 question \#34. Some type of Location Box is necessary because the relative alignment of objects is specified via the word "between."

"[34] There is a pillar between me and the stage, and I can't see it. What can't I see? POSSIBLE ANSWERS: \{the stage, the pillar\}"

The dotted line is technically a 1D Marker, used to help the viewer can discern that the three objects are aligned.

\textbf{2. Descriptive Boxes (further development anticipated)}

\begin{figure}
	\begin{center}
	\includegraphics[width=0.25\textwidth]{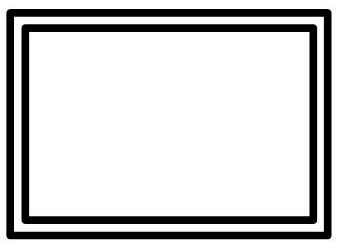}
	\caption{A Descriptive Box has one less border than a Verbatim Box to suggest that a Descriptive Box is less constrained than a Verbatim Box.}
	\label{fig:icon-location-types-descriptive}
	\end{center}
\end{figure}

The Tumbug icon for a Descriptive Box is shown in Figure~\ref{fig:icon-location-types-descriptive}.

\begin{figure}
	\begin{center}
	\includegraphics[width=0.75\textwidth]{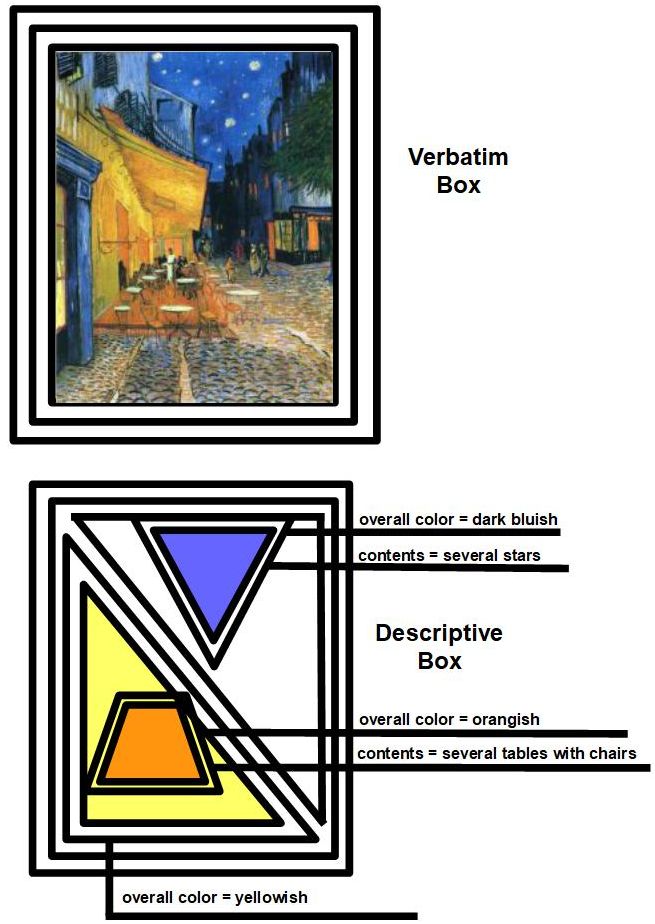}
	\caption{A Verbatim Box has every feature in a fixed location, but a Descriptive Box has only loosely constrained locations. (Painting: "Café Terrace at Night" (Vincent Van Gogh, 1888))}
	\label{fig:icon-descriptive-location-box-compare}
	\end{center}
\end{figure}

Whereas a Verbatim Box shows exact, inflexible locations of parts of images, a Descriptive Box is more flexible and therefore more general in that it uses links that describe spatial relationships via text instead of rigidly placed objects in space. See Figure~\ref{fig:icon-descriptive-location-box-compare}.

Descriptive Boxes of any type may be arranged and nested arbitrarily. For example, to show the upstairs and downstairs of a house, two such Descriptive Boxes would logically be stacked. Similarly, to show two regions of a picture where each region could contain only one type of item, but each item could be in any location within that region, two such boxes would be placed inside of a third, larger such Descriptive Box. The same observation holds true of Aggregation Boxes. 

However, this observation about nested boxes does not hold for mixtures of different types of Location Boxes since placing a less constrained Location Box inside of a more constrained Location Box logically suggests that the constraints of the less constrained box must then be made more constrained, which is probably not what a diagrammer would want. For example, in Figure~\ref{fig:icon-descriptive-location-box-compare}, the painting inside the Verbatim Box could not logically contain regions of Descriptive Boxes because then those Descriptive Boxes would presumably fall under the rules of the encompassing Verbatim Box.

\textbf{3. Aggregation Boxes}

\textbf{3.1. C Aggregation Boxes}

The preliminary letter "C" is a formality that stands for "concrete." ("Corporeal" would be a good alternative synonym, if so desired.) Usually C Aggregation Boxes are called simply "Aggregation Boxes."

\begin{figure}
	\begin{center}
	\includegraphics[width=0.30\textwidth]{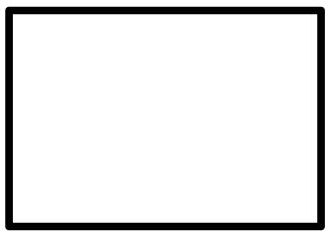}
	\caption{Tumbug's icon for a C Aggregation Box.}
	\label{fig:icon-aggregation-box}
	\end{center}
\end{figure}

Tumbug has two types of Aggregation Boxes: C Aggregation Boxes and C-A Aggregation Boxes. Tumbug's icon for an C Aggregation Box is shown in Figure~\ref{fig:icon-aggregation-box}. C-A Aggregation Boxes are described in the different section.

Location Boxes and Aggregation Boxes may be arranged and nested arbitrarily. For example, to show the upstairs and downstairs of a house, two such boxes would logically be stacked. Similarly, to show two regions of a picture where each region could contain only one type of item, but each item could be in any location within that region, two such boxes would be placed inside of a third, larger such box.

Note that the border of any type of Aggregation Box need not be rectangular, though rectangular is the default shape. 

Nearly all representation systems have some means of representing aggregation. Some examples are:

\begin{enumerate}
    \item
		In English grammar phrases can be aggregated by parentheses, such as "The flower fields of Encinitas were gradually either replaced (by condominiums or a tourist attraction) or left to fall into neglect."
    \item
		In algebra, terms can be aggregated by parentheses. For example: 2(x + y) - 3(x + y).
    \item
		In human societies, residential regions are organized into communities, which are grouped into cities, which are grouped into states or provinces, which are grouped into countries, and so on.
\end{enumerate}

In Tumbug a (single-bordered) rectangle called a "C Aggregation Box" is used to surround the visual objects that are to be aggregated.

\begin{figure}
	\begin{center}
	\includegraphics[width=0.50\textwidth]{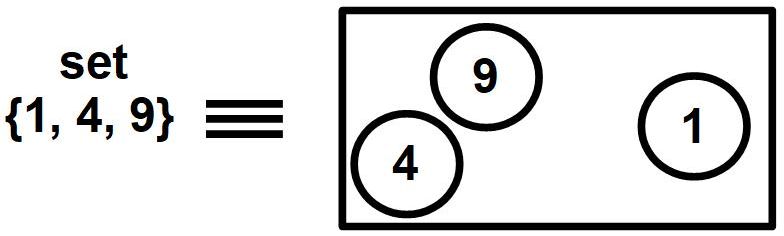}
	\caption{A C Aggregation Box would be used to represent a set since element location is unimportant.}
	\label{fig:icon-aggregation-set}
	\end{center}
\end{figure}

\begin{figure}
	\begin{center}
	\includegraphics[width=0.50\textwidth]{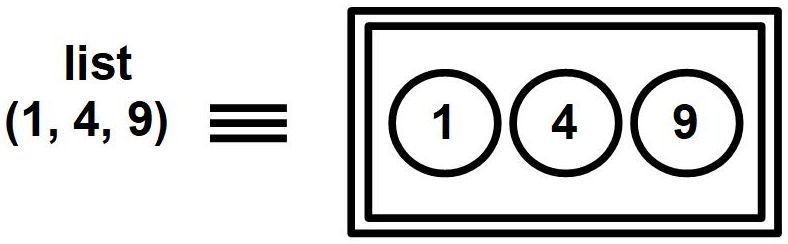}
	\caption{A Descriptive Box would be used to represent a list since element location is important.}
	\label{fig:icon-aggregation-list}
	\end{center}
\end{figure}

If mathematical sets are being represented in Tumbug, an Aggregation Box would be used since there is a homomorphism between a parenthesized list of numbers in any order and a box of objects in any location, as in Figure~\ref{fig:icon-aggregation-set}. For an ordered list, however, a Descriptive Box would probably be used since the relative order (relative places) of the listed objects must be preserved: an isomorphism. See Figure~\ref{fig:icon-aggregation-list}.

\begin{figure}
	\begin{center}
	\includegraphics[width=0.50\textwidth]{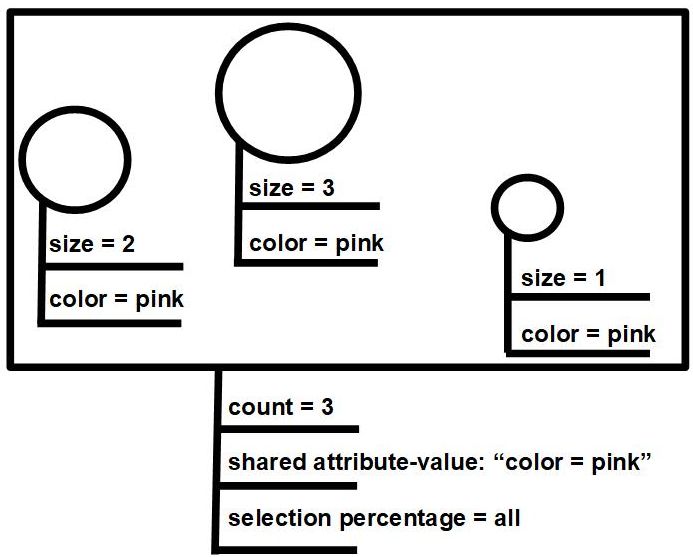}
	\caption{Tumbug for "All the pink souvenirs."}
	\label{fig:icon-aggregation-shared}
	\end{center}
\end{figure}

An Aggregation Box may also have attribute-value pairs, and often do. Some attribute-value information that might be useful to include: the count of the items in the Aggregation Box (e.g., "count = 3"), attribute-values that all the items in the Aggregation Box share (e.g., "shared attribute-value = 'color = pink'"), the source of the items (e.g., "source = Madonna Inn gift shop"), discounted price if a customer buys everything inside (e.g., "5"), the name of the set (e.g., "things that Fred does routinely"). See Figure~\ref{fig:icon-aggregation-shared}.

The "shared attribute-value" attribute on any type of Aggregation Box is currently necessary in Tumbug when representing a conjectured set whose members (and therefore count) are not known.

\textbf{3.2. C-A Aggregation Boxes}

\begin{figure}
	\begin{center}
	\includegraphics[width=0.50\textwidth]{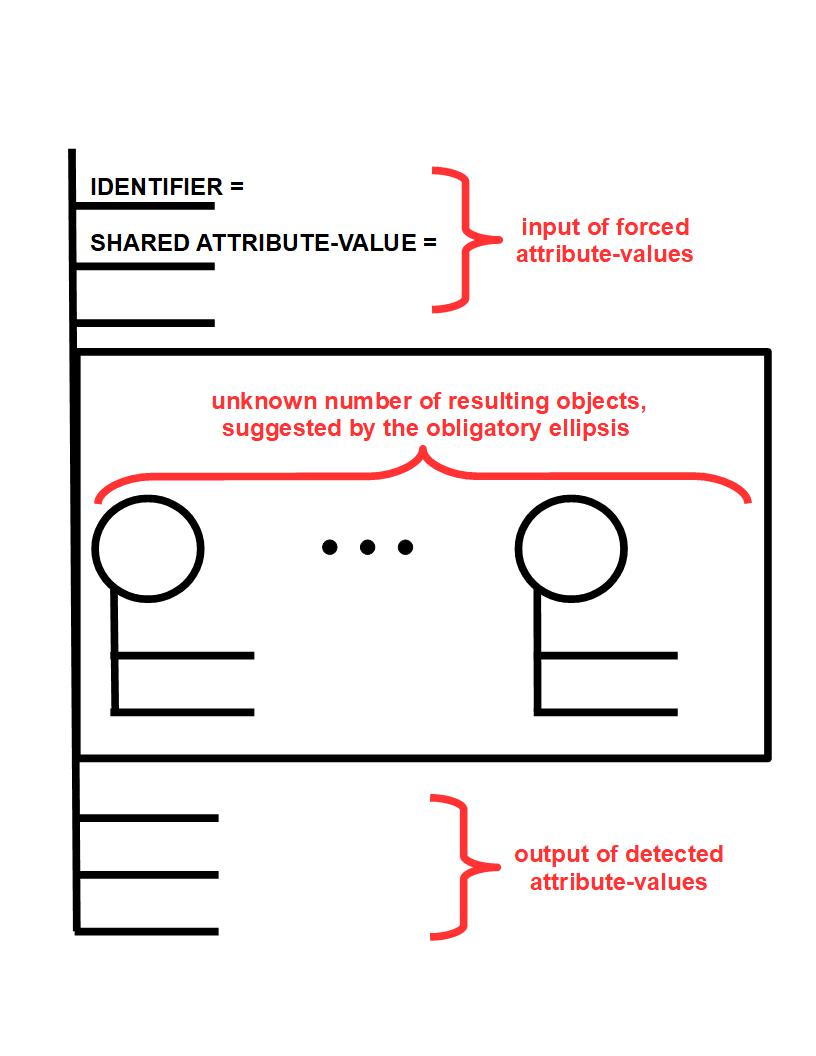}
	\caption{Tumbug's icon for a C-A Aggregation Box.}
	\label{fig:icon-ca-set-ANN}
	\end{center}
\end{figure}

Tumbug's icon for a C-A Aggregation Box is shown in Figure~\ref{fig:icon-ca-set-ANN}.

The C-A Aggregation Box is one of the Building Blocks where Tumbug becomes the most profound. "C-A" stands for "concrete-abstract," which is a pair of concepts at the two extreme ends of the abstraction spectrum. A comparison of the C Aggregation Box and the C-A Aggregation Box follows:

\begin{itemize}
	\item
		Both have a rectangular boundary that surrounds a set of arbitrarily selected objects.
	\item
		Both may have a list of detected attribute-values associated with the represented set.
	\item
		Only a C-A Aggregation Box has a second list of forced attribute-values that define the set.
	\item
		A C-A Aggregation Box typically contains an unknown number of objects, so an ellipsis is shown inside.
\end{itemize}

A C-A Aggregation Box is a deeply abstracted form of a mathematical function, where the input is analogous to an independent variable and the resulting output is analogous to a dependent variable. As in functions, the primary interest is usually in observing the effect on the output after inputting a specific value. For example, if the forced input to a C-A Aggregation Box is the attribute-value pair "species = human," the resulting set would contain all existing humans, and a useful attribute would be the count of all objects (= all humans) in this set. In this way a C-A Aggregation Box is similar to a SQL data base query where query word "WHERE" specifies the attribute-value pair (which in a data base is the column and row, respectively), the aggregate function "COUNT" is a detected (and outputted) property of the set, and the query word "FROM" refers to the set in question. Also, as in math functions and SQL queries, the mapping may not be invertible, such as when the math function is y = $x^{2}$ for x = -1 and x = +1, or when the SQL query returns more than one applicable record when queried with "WHERE surname = Smith". Since the resulting matches are typically large and of an unknown count, an ellipsis is used inside a C-A Aggregation Box.

\begin{figure}
	\begin{center}
	\includegraphics[width=0.50\textwidth]{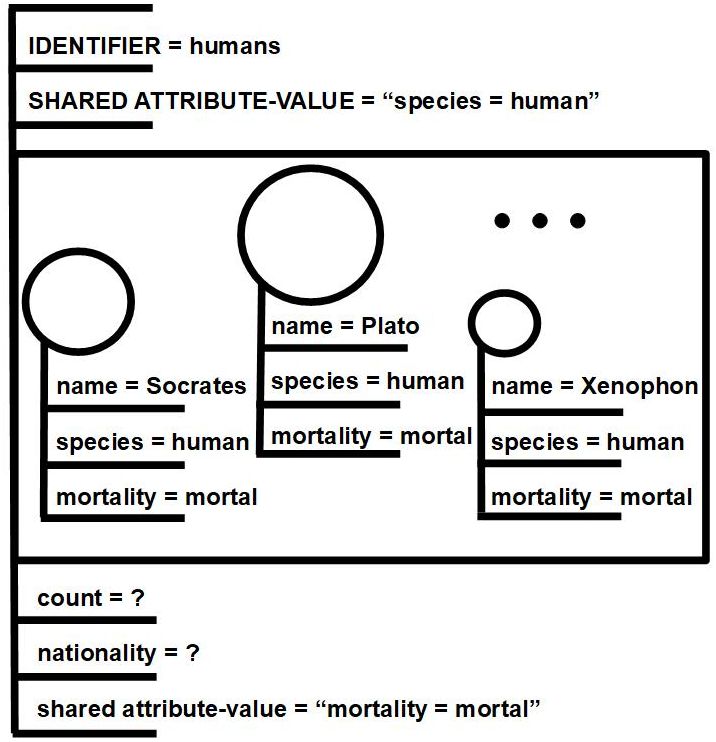}
	\caption{The classic "All men are mortal" syllogism.}
	\label{fig:icon-ca-set-mortal}
	\end{center}
\end{figure}

There also exists an analogy of C-A Aggregation Boxes with first order predicate calculus (FOPC). For example, the statement of the classic syllogism, "Every man is mortal," which is represented in FOPC as "($\forall$ x) (MAN(x) → MORTAL(x))", which could be read as "For all x, if x is an element of the set of humans (MAN), then that implies that x has attribute mortal (MORTAL)," can be represented in Tumbug with the C-A Aggregation Box shown in Figure~\ref{fig:icon-ca-set-mortal}. Note that the "count" attribute of this set (one of the outputs) has unknown value, and its current magnitude would be measured in the billions. Note also that the forced input that defines this set is that "species = human" but that the detected output that describes the resulting set is "mortality = mortal," even though that attribute-value pair was not mentioned in the input.

C-A Aggregation Boxes are needed for mathematical theorems. For example:

"Theorem: In any graph with at least two nodes, there are at least two nodes of the same degree."

\begin{figure}
	\begin{center}
	\includegraphics[width=0.55\textwidth]{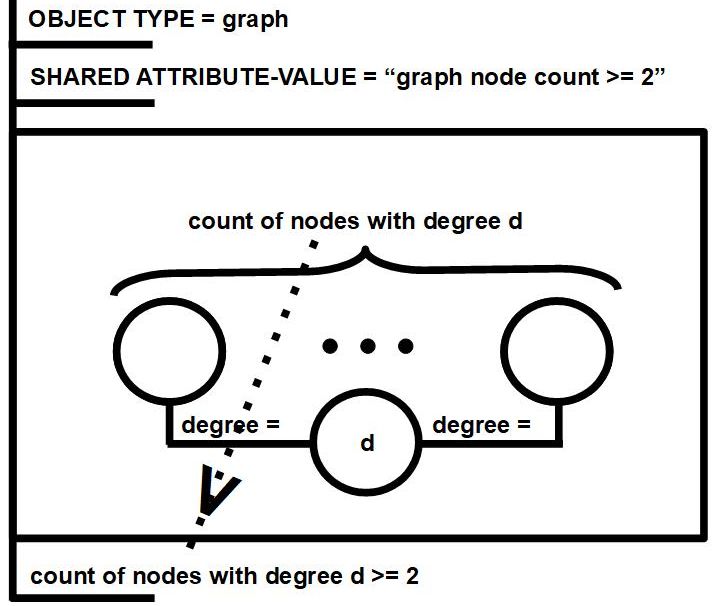}
	\caption{Tumbug representation of a graph theory theorem: "In any graph with at least two nodes, there are at least two nodes of the same degree."}
	\label{fig:icon-ca-set-theorem}
	\end{center}
\end{figure}

Tumbug would illustrate this theorem with a C-A Aggregation Box that contains only one graph of at least two nodes, as shown in Figure~\ref{fig:icon-ca-set-theorem}. The set membership requirement could be represented as the textual description "each graph contains at least two nodes," and the resulting implied output could be represented as the textual description "each graph has at least two nodes of the same degree." The upper, forced attributes of the graph could collectively be considered a large "if..." condition, and the lower, resulting attributes of the graph could collectively be considered a large "...then..." result. In constructive theorems that show a concrete example, a visual example would go inside the Aggregation Box, but not all theorems are constructive, so such a diagram is a luxury. This theorem happens to be a non-constructive theorem.

\subsubsection{Swirly Arrays}

\textit{The components of this composite component: Location Box, Cells, optionally any number of 0D Markers up to the number of cells.}

\begin{figure}
	\begin{center}
	\includegraphics[width=0.40\textwidth]{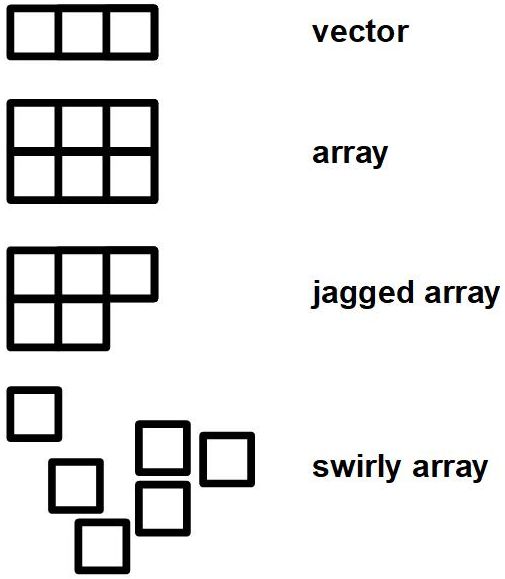}
	\caption{Examples of different types of array arrangements.}
	\label{fig:icon-swirly}
	\end{center}
\end{figure}

An array, also called a matrix, is a well-known mathematical data structure that contains cells in a rectangular structure. If the structure contains only one row or one column then it is called a vector. Information technology has extended the array concept so that the rows of the array need not be the same length (or equivalently, that the embedded "member arrays" need not be the same length) though the rows are usually shown aligned on their left-hand sides. Such arrays are called "jagged arrays" or "ragged arrays." Tumbug extends the concept of a jagged array even further: In Tumbug a "swirly array" is defined as a given spatial arrangement of cells where the cells do not need not to be contiguous or aligned horizontally or vertically. See Figure~\ref{fig:icon-swirly}. (Many mathematicians, high school students, and science fair project creators have probably reinvented the concept of swirly arrays many times before, but if so, such documentation is not well-known and not readily available. The author hereby gives advance credit to any such prior inventors and their proposed invention names.) Arrays of any kind in Tumbug context could be called a "composite Building Block."

Swirly arrays as used here are treated more like annotated diagrams than mathematical entities. Using obvious methods, two swirly arrays could easily be added together, and one swirly array could easily be multiplied by a scalar, but special functions such as taking a determinant of a swirly array, or multiplying two swirly arrays together would not be possible in general. However, such structural flexibility in swirly arrays could also be beneficial for customizing arrays for certain special purposes that might need to hold one or more adjunct values not related to the main task at hand, such as storing coefficients of squared terms, which might allow an elegant extension to linear algebra.

\begin{figure}
	\begin{center}
	\includegraphics[width=0.65\textwidth]{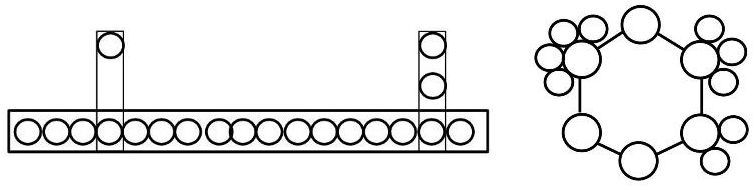}
	\caption{Two examples of Swirly Arrays in Tumbug are the Modal Verbs Icon (left) and Robinson Icon (right).}
	\label{fig:icon-combo-modal-verb-robinson}
	\end{center}
\end{figure}

The aforementioned array examples use squares for cells because that is standard mathematical practice, but Tumbug uses circles instead, because in Tumbug cells hold concepts or circles, both of which are represented as circles. Swirly arrays are motivated in Tumbug by the need to place detector neurons (or just detector lights) at fixed locations that match irregularly-shaped diagrams of organized concepts, neurological structures, or maps of any kind, for the positional coding of concepts. It is the on-off status of the detectors in combination with their locations that represent more complicated concepts such as emotions or modal verbs. See Figure~\ref{fig:icon-combo-modal-verb-robinson}.

\subsubsection{XOR Boxes}

\textbf{1. Applicable icon}

\begin{figure}
	\begin{center}
	\includegraphics[width=0.25\textwidth]{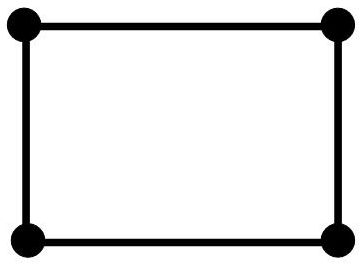}
	\caption{The Tumbug icon for an XOR Box.}
	\label{fig:icon-xor-box}
	\end{center}
\end{figure}

To represent logical OR in Tumbug, a box similar to a C Aggregation Box is needed, a box that holds all the objects that are being OR-ed. However, a C Aggregation Box is not quite specific enough for the concept of OR because a C Aggregation Box is merely a set that does not indicate that only one choice must be selected from the collection of choices in the set. This selection restriction is the concept of mutual inhibition: if all the choices were mutually inhibited neurons, for example, each firing neuron would attempt to suppress all the other neurons for firing, which is what is called a Winner Take All network. To implement a Winner Take All in Tumbug, a variation of a C Aggregation Box is needed, a variation that is called an XOR Box, the Tumbug icon of which is shown in Figure~\ref{fig:icon-xor-box}. The black dots in the corners are intended to be suggestive of the typical link diagrams used to represent mutual inhibition in artificial neural networks. 

\textbf{2. Inclusive OR}

\begin{figure}
	\begin{center}
	\includegraphics[width=0.75\textwidth]{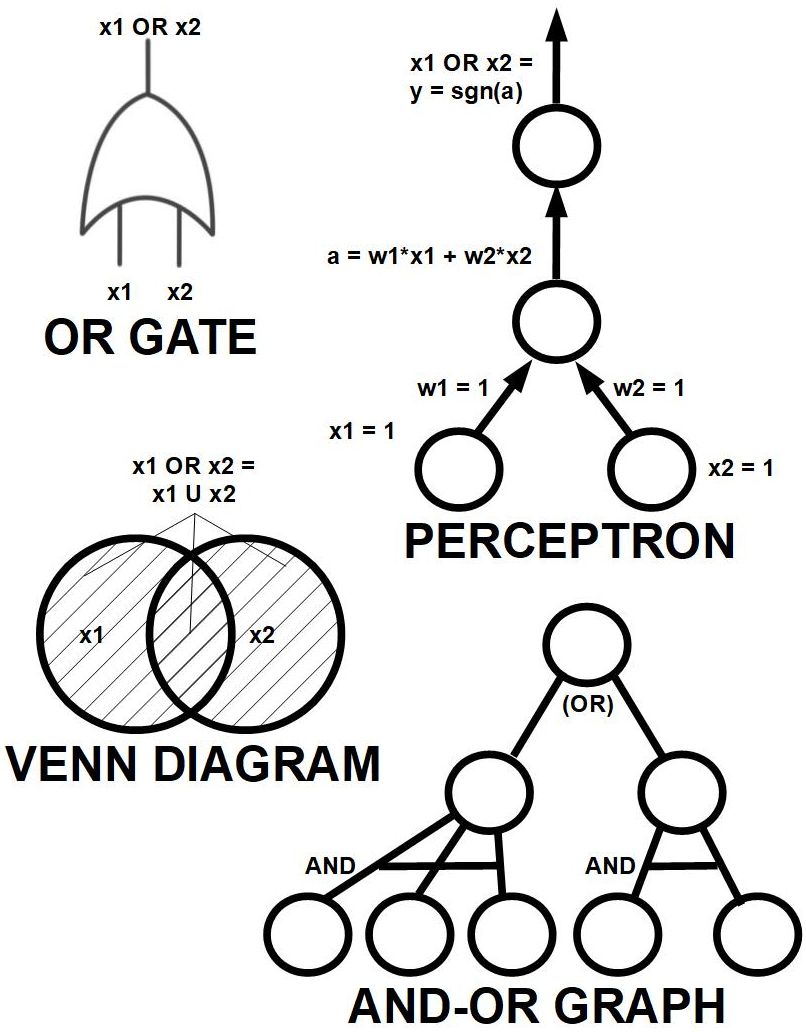}
	\caption{Some traditional visual representations of logical "OR."}
	\label{fig:icon-or-representations}
	\end{center}
\end{figure}

It should not be surprising that the most basic logic operations (AND, OR, NOT, XOR, NAND, NOR, etc.) should be present in Tumbug in some manner. Some traditional methods of representing the (inclusive) OR operation in computer science are shown in Figure~\ref{fig:icon-or-representations}.

Although the Venn diagram representation is an appealing candidate for Tumbug, it has two practical problems: (1) It normally shows both possibilities in detail, which is spatially prohibitive in Tumbug unless variables are used; (2) The intuitive visual logic of the real world is the opposite of that shown in Venn diagrams.

The latter problem is the most important and the most interesting. In particular when the existence of real-world objects is considered, "AND" is intuitively understood to mean that all the listed objects are present. For example, the statement "nuts and berries are present" implies a visual depiction of nuts existing in the vicinity of berries, whereas the statement "nuts or berries are present" (the "or" of which is an inclusive OR, by default) implies a visual depiction of nuts existing or berries existing, or possibly both. A Venn diagram fails at representing this situation since a Venn diagram shows an OR (= union) of two sets by showing both sets at the same time, which is intuitively wrong, and shows an AND (= intersection) of two sets by a small region of intersection, which in real life would be interpreted as showing the presence of only objects that are categorized as both nuts and berries, which is also intuitively wrong. The conclusion: Intuitive, visual "OR" cannot be easily and intuitively  represented by a Venn diagram, therefore intuitive, visual "OR" is intuitively an abstract concept that requires some other representation. Therefore Tumbug represents visual "OR" using the perceptron method. (A perceptron is a type of artificial neural network, therefore Tumbug could be said to use neural networks, to some extent.)

\begin{figure}
	\begin{center}
	\includegraphics[width=0.50\textwidth]{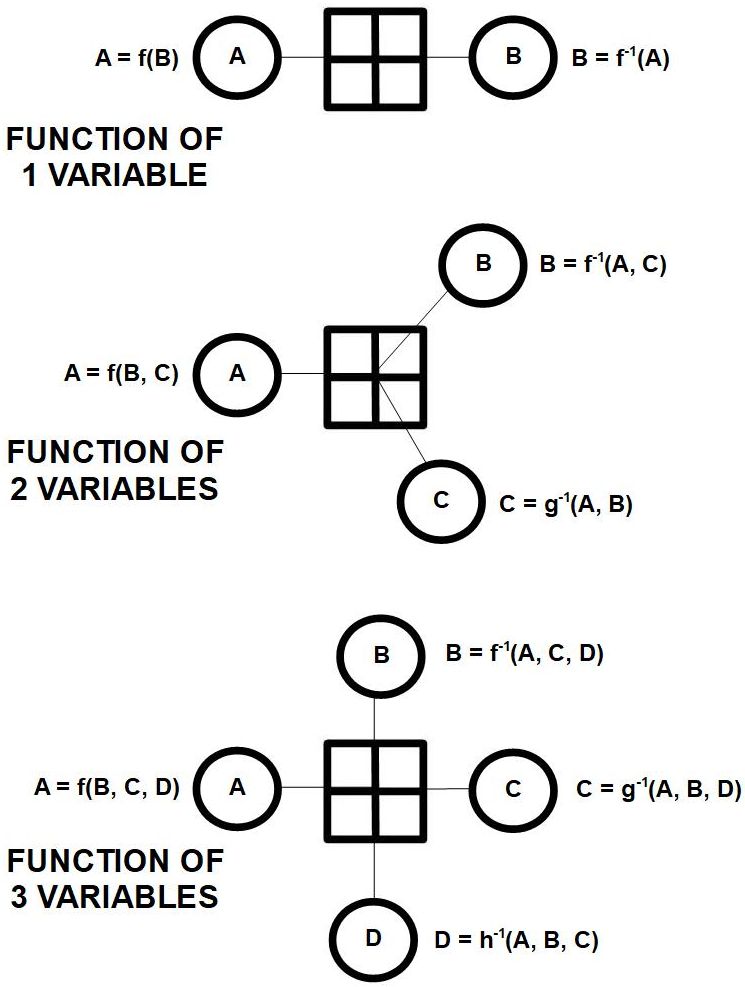}
	\caption{Correlation Boxes can be used with any number of input variables, not just one.}
	\label{fig:icon-correlation-2-args}
	\end{center}
\end{figure}

The Tumbug solution for the representation of OR relies on another Tumbug Building Block discussed elsewhere: the Correlation Box. One main difference is that whereas the Correlation Box examples shown earlier represented a function of a single variable, such as A = f(B), OR requires a function of two variables, such as A = f(B, C). Specifically, the function would be named A = OR(B, C). Correlation Boxes can be generalized to any number of input variables, as shown in Figure~\ref{fig:icon-correlation-2-args}.

\begin{figure}
	\begin{center}
	\includegraphics[width=0.50\textwidth]{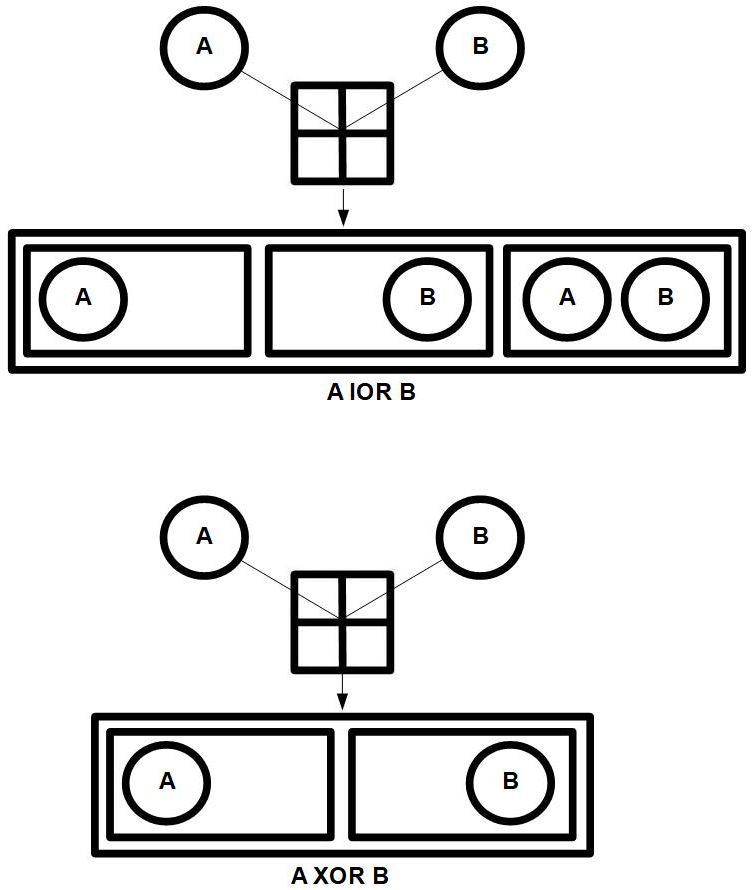}
	\caption{Tumbug representation of inclusive OR (= IOR) versus exclusive OR (= XOR).}
	\label{fig:icon-or-aggregation}
	\end{center}
\end{figure}

Figure~\ref{fig:icon-or-aggregation} shows the intermediate Tumbug representation of OR, one diagram for inclusive OR (= IOR), and one diagram for exclusive OR (= XOR). Note that the use of a C Aggregation Box functions as a set of possible choices. For example, IOR(A, B) produces the set \{A, B, A AND B\}.

\begin{figure}
	\begin{center}
	\includegraphics[width=0.50\textwidth]{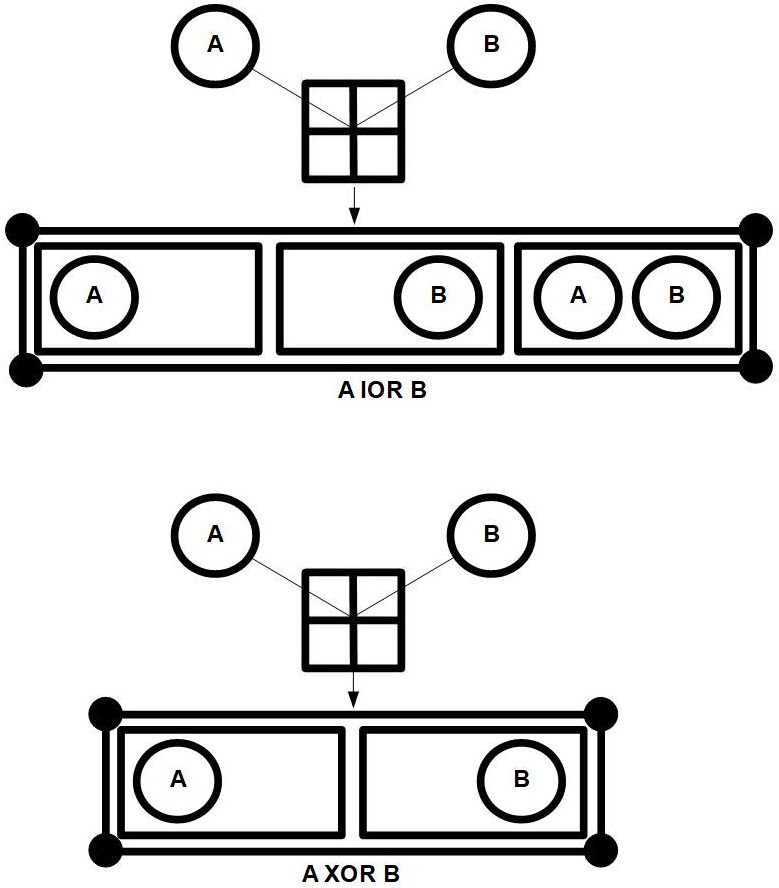}
	\caption{Comparison of IOR and XOR functions, using XOR Boxes.}
	\label{fig:icon-xor-box-compare}
	\end{center}
\end{figure}

The comparison of IOR and XOR is shown again in Figure~\ref{fig:icon-xor-box-compare}, using XOR Boxes.

\textbf{3. The existence attribute}

An attribute that is important enough to address in its own section is the attribute of existence. In mathematical literature, for example, the "there exists" symbol ($\exists$) is extremely common, so to render mathematical theorems in Tumbug, one would need to know how to represent "there exists."

\begin{figure}
	\begin{center}
	\includegraphics[width=0.60\textwidth]{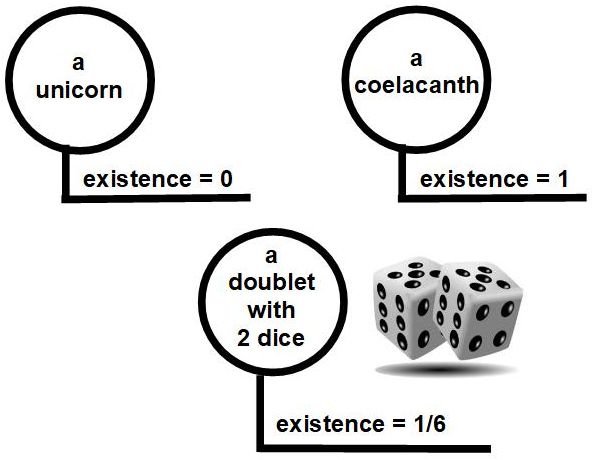}
	\caption{Existence in Tumbug is considered merely another attribute whose value can vary.}
	\label{fig:icon-existence}
	\end{center}
\end{figure}

The Tumbug solution is simple, but unfortunately the solution not satisfyingly visual: the current solution is to use a typical Attribute Line on a typical C Object Circle (or C Aggregation Box) whose existence is to be made explicit, and to place the attribute name "existence =" followed by a value from the range 0 through 1. The 0-1 range describes the range of likelihood from "does not exist" (0) through "exists" (1). See Figure~\ref{fig:icon-existence}. The main visual drawback of this representation is that existence is such a fundamental attribute of real-world objects that one would expect that attribute of an object to be overwhelmingly obvious in a diagram, but in Tumbug it is not.

The reason this non-visual, attribute-value type of solution seems to be necessary is that an object depicted in a Tumbug diagram is ordinarily assumed to exist, therefore to show that an object does not exist would suggest that mere omission of the depicted object would represent non-existence. However, that solution in turn implies that any object not shown must not exist, which in turn implies that a Tumbug diagram must depict everything in the universe. Although one could adopt the convention that anything not drawn on a Tumbug diagram is merely a "don't care" rather than a "does not exist," by the Tumbug convention of wildcards (discussed elsewhere in this document), a blank already means "don't care," when then leaves no way to uniquely represent "does not exist." This uncomfortable and counter-intuitive situation is a variation of the "closed-world assumption" of formal logic, where anything not explicitly marked "true" must be "false."

\section{Attribute-like Building Blocks of Tumbug (A)}

\subsection{Attribute Lines}

\begin{figure}
	\begin{center}
	\includegraphics[width=0.50\textwidth]{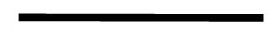}
	\caption{Tumbug's icon for an Attribute Line.}
	\label{fig:icon-line}
	\end{center}
\end{figure}

\begin{figure}
	\begin{center}
	\includegraphics[width=0.50\textwidth]{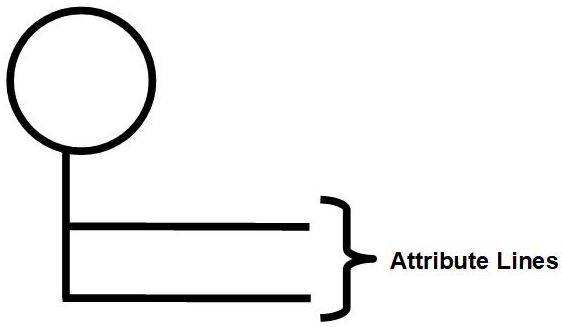}
	\caption{Attribute Lines for a given object branch off from a single stem, and can be of any quantity.}
	\label{fig:icon-attribute-general}
	\end{center}
\end{figure}

\begin{figure}
	\begin{center}
	\includegraphics[width=0.25\textwidth]{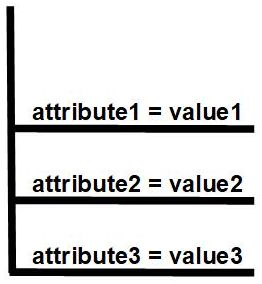}
	\caption{Three AV pairs alone.}
	\label{fig:icon-label-fox-attributes-only}
	\end{center}
\end{figure}

\begin{figure}
	\begin{center}
	\includegraphics[width=0.25\textwidth]{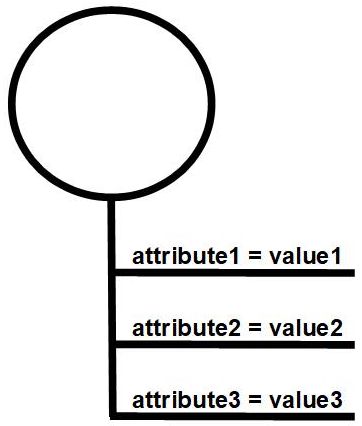}
	\caption{Three AV pairs attached to an object, which is the usual situation.}
	\label{fig:icon-label-fox-attributes-and-object}
	\end{center}
\end{figure}

\begin{figure}
	\begin{center}
	\includegraphics[width=0.25\textwidth]{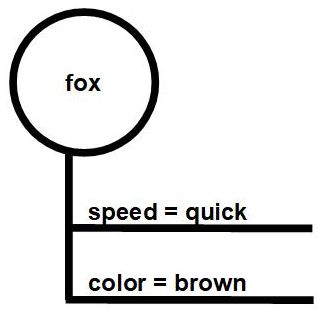}
	\caption{Tumbug for "quick, brown fox."}
	\label{fig:icon-label-fox}
	\end{center}
\end{figure}

In Tumbug, attributes are almost always shown together with values, and vice versa, separated by an equal sign "=". An attribute followed by its value, such as "speed = quick," are written together on the same horizontal line. Figure~\ref{fig:icon-line} shows the basic line that holds text, Figure~\ref{fig:icon-attribute-general} shows how such lines are attached to a C Object Circle, Figure~\ref{fig:icon-label-fox-attributes-only} and Figure~\ref{fig:icon-label-fox-attributes-and-object} shows generalizations of this type of structure, and Figure~\ref{fig:icon-label-fox} shows an example of a specific object with specific attributes. There is no theoretical limit on how many Attribute Lines may be attached to any type of Object Circle. In a physical implementation of Tumbug, the equals sign "=" would not exist since the equals sign merely distinguishes between two types of stored concepts (viz., attributes and values), and in a physical implementation there would presumably be two separate memory locations that hold those concepts, and the separation of these memories would be already clear.

Some Attribute Lines terminate at icons or small diagrams when the attribute values are not simple textual or numerical descriptions. This happens with State Diagrams and Motivation Triangles.

As for the values to which attributes are set, in typical, partial-text Tumbug, numbers are given exactly as in their standard math counterparts, such as "3." Tumbug ultimately was intended to be a fully visual representation system, however, so such numbers can also be given as lengths of a special measurement object whose icon is called a Value Bar, described elsewhere in this document.

\textit{This article's convention: Tumbug attribute-value pairs are almost always connected to the object to which they are associated, rarely isolated. If there exists more than one attribute shown per object, a single stem is usually used, unlabeled, with all the attribute-value pairs branching off that stem. The stem can be shown emerging from the bottom, top, or (less frequently) at another arbitrary point on the C Object Circle's perimeter.}

\textit{Tumbug minor convention: It is not always necessary to have an attribute and value paired, although this is usually the situation. For example, "life status = alive" could be abbreviated to merely "alive" if the meaning is clear enough.}

\begin{figure}
	\begin{center}
	\includegraphics[width=0.50\textwidth]{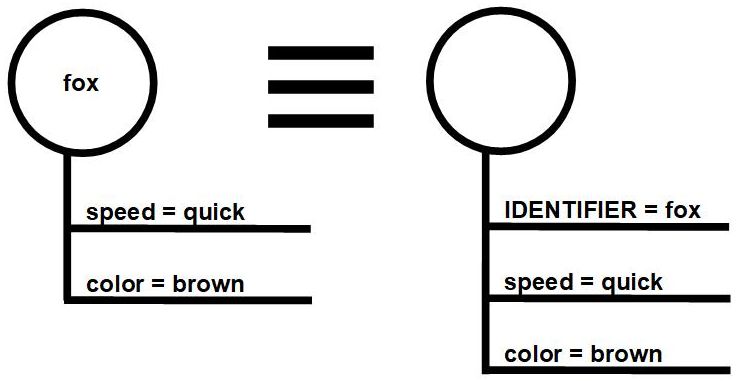}
	\caption{These two Tumbug representations for "quick, brown fox" are equivalent. Currently both are acceptable.}
	\label{fig:icon-attribute-fox-label-equivalence}
	\end{center}
\end{figure}

On C Object Circles shown in this document, the main label on the circle, such as "fox" or "Susan" or "they," is technically an attribute from the set of Attribute Lines for that object. That main label could be considered to be the value assigned to the first attribute in the list, an attribute called "IDENTIFIER," such as in the attribute-value "IDENTIFIER = fox." See Figure~\ref{fig:icon-attribute-fox-label-equivalence}. Such a single, enlarged Label String on the C Object Circle itself is intended only to make the diagram easier to read and understand at a glance.

\section{Value-like Building Blocks of Tumbug (V)}

\subsection{Value Bars}

\subsubsection{General}

\begin{figure}
	\begin{center}
	\includegraphics[width=0.60\textwidth]{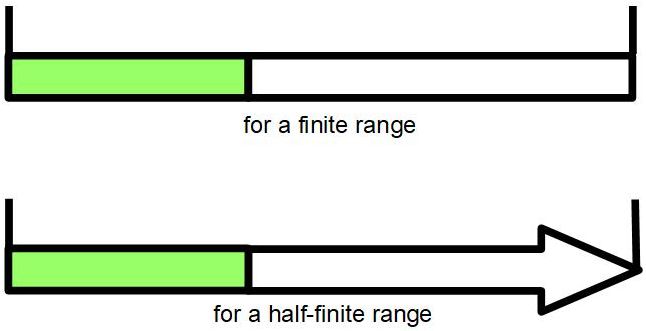}
	\caption{Typical forms of Tumbug's icon for a Value Bar.}
	\label{fig:icon-value-bar}
	\end{center}
\end{figure}

The possibility of using an additional Building Block only to represent magnitude was alluded to, elsewhere in this document. Tumbug's icon for such a Building Block is called a Value Bar, shown in Figure~\ref{fig:icon-value-bar}. Since numerical values are so terse to write as digits, there is currently little motivation to use Value Bars unless possibly to demonstrate that Tumbug can be made entirely visual, as the Tumbug acronym suggests. Also, if Tumbug were to be implemented in an artificial neural network in hardware form, digits (and text) would be too difficult to represent unless they were visual, which would be another justification for use of Value Bars.

The length of the shaded segment along the bar (= thin rectangle) is primarily what is most important in this icon. The Tumbug icon for a Value Bar is a bar, typically with an arrow at one end, but the arrow is not required when the value is obviously limited to a finite range. The two extreme values within the given range are emphasized by a vertical hatch mark at each end of the bar, as if to show where a ruler or tape measure would be laid while measuring the bar end-to-end. The hatch mark at the arrow tip means infinite magnitude of the associated attribute's value.

\textit{Side conjecture: It may be possible that each specimen of a higher animal, especially if human, uses its own individual height as the primary unit of height measurement in the real world. Most likely the ratio between heights is subconsciously computed as a convenient estimate for the likelihood of winning a fight against another animal. For example, if the self-to-other height ratio were $\geq$ 1.25, meaning the self is 1.25 times taller than the other animal, this could automatically trigger the psychological impression and attitude in the self that "I can take him."}

\textit{Side conjecture: The above conjecture about height also holds true for other physical, emotional, and psychological measures of the self, in relation to others, such as weight, speed, aggression level, empathy level, intelligence, and knowledge.}

\begin{figure}
	\begin{center}
	\includegraphics[width=0.50\textwidth]{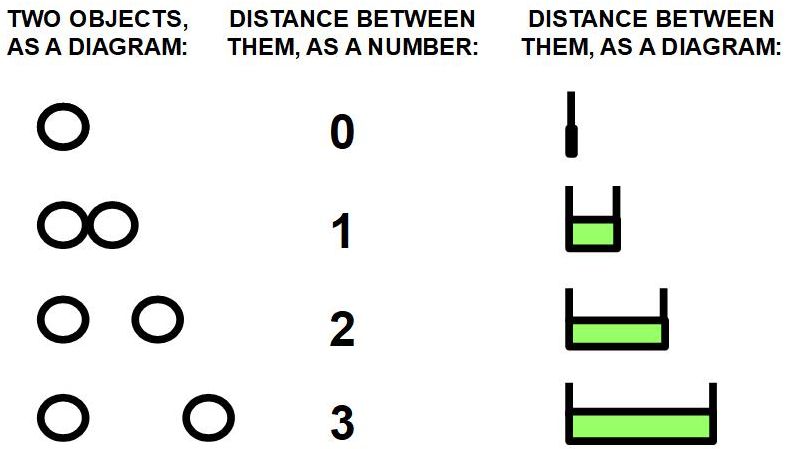}
	\caption{Tumbug can represent distances as lengths, if desired.}
	\label{fig:icon-far-rep}
	\end{center}
\end{figure}

\begin{figure}
	\begin{center}
	\includegraphics[width=0.50\textwidth]{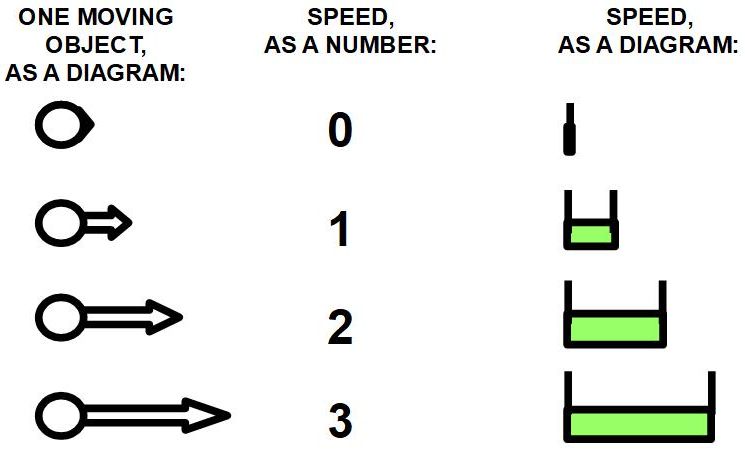}
	\caption{Motion attributes such as speed are almost as simple to represent as static attributes such as distance.}
	\label{fig:icon-value-fast-rep}
	\end{center}
\end{figure}

Values of motion, such as speed, acceleration, and direction, can be represented by Tumbug in nearly the same way as length. Compare Figure~\ref{fig:icon-far-rep} for distance to Figure~\ref{fig:icon-value-fast-rep} for speed. In general, this magnitude coding scheme is of great benefit for neural network implementation since a single hardware mechanism can represent any of the Building Blocks of Tumbug, whether numerical or visual--length, circumference, angle, area, volume, curvature, speed, acceleration, duration, age, weight, force, energy, power, friction, temperature, color, texture, volume, range, precision, accuracy, probability--and even human-based, abstract attributes such as price, stock keeping unit (SKU), attractiveness, popularity, sentimental value, and cuteness.

As for neurobiological verisimilitude of Tumbug, various biological visual systems are known to handle motion information using specialized hardware, which implies certain neurons fire in response to certain types of motion or certain values of motion, which would then map fairly directly to Tumbug's Motion Arrows with appropriate attribute values of speed, acceleration, collective looming motion, and so on, filled in. For example, detection of the movement of images is one of the four primary visual operations in a frog's eye (Lettvin et al. 1959, p. 257), a fly's visual system determines if the visual field is looming sufficiently fast for the fly to contemplate landing (Marr 1982, p. 35), speed gradients are computed in middle temporal/V5 area of the macaque extrastriate cortex (Orban 2008), and the human visual system uses at least two motion detection systems, one of which is not even aware of the object whose motion is being automatically detected (Chubb 1995, p. 109).

\subsubsection{Ranges}

\begin{figure}
	\begin{center}
	\includegraphics[width=0.50\textwidth]{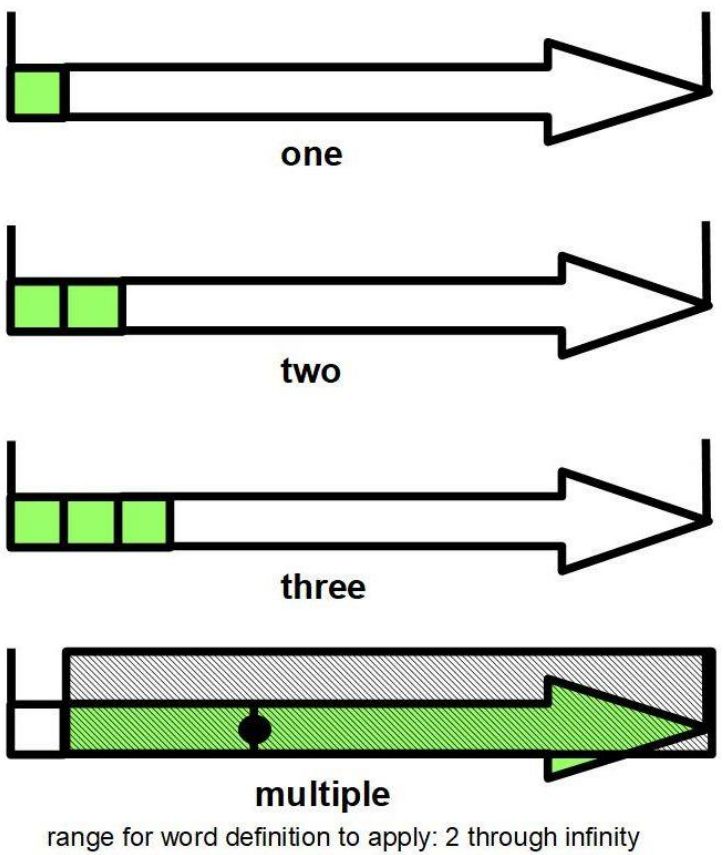}
	\caption{The word "multiple" applies only if the observed value falls within the hatched range of $\geq$ 2.}
	\label{fig:icon-range-1-2-3-multiple}
	\end{center}
\end{figure}

Tumbug represents ranges as shown in Figure~\ref{fig:icon-range-1-2-3-multiple}.

Value Bars can also represent ranges. Each equivalent of a unit square on a Value Bar can represent a span of one unit ($\Rightarrow$ a count of 1) in the horizontal direction, therefore such squares placed side-by-side into a horizontal bar can represent a specific integer (e.g., a count of 3). In Figure~\ref{fig:icon-range-1-2-3-multiple}, the shaded region makes the count easier to see at a glance because a 2D area is easier to see than a 1D hatch mark on a number line, and the green color can represent known (i.e., non-fuzzy) information. A 45-degree hatched pattern--the hatched pattern used by 2D Markers--could logically be used instead of the green color, but that scheme was not used in this document.

A semi-ambiguous quantity like the concept conveyed by the word "multiple" (meaning more than one) typically means an arbitrary point within a range, so both the point and range must be represented somehow. If the specified range can be represented by a different shading, such as the purple shading in Figure~\ref{fig:icon-range-1-2-3-multiple}, then a small ball can represent an arbitrary point within that range: the location of the ball becomes a type of Wildcard. The ball icon suggests that the ball is free to roll to anywhere within the range, but at any given time the ball must occupy a specific point within the range. 

\begin{figure}
	\begin{center}
	\includegraphics[width=0.50\textwidth]{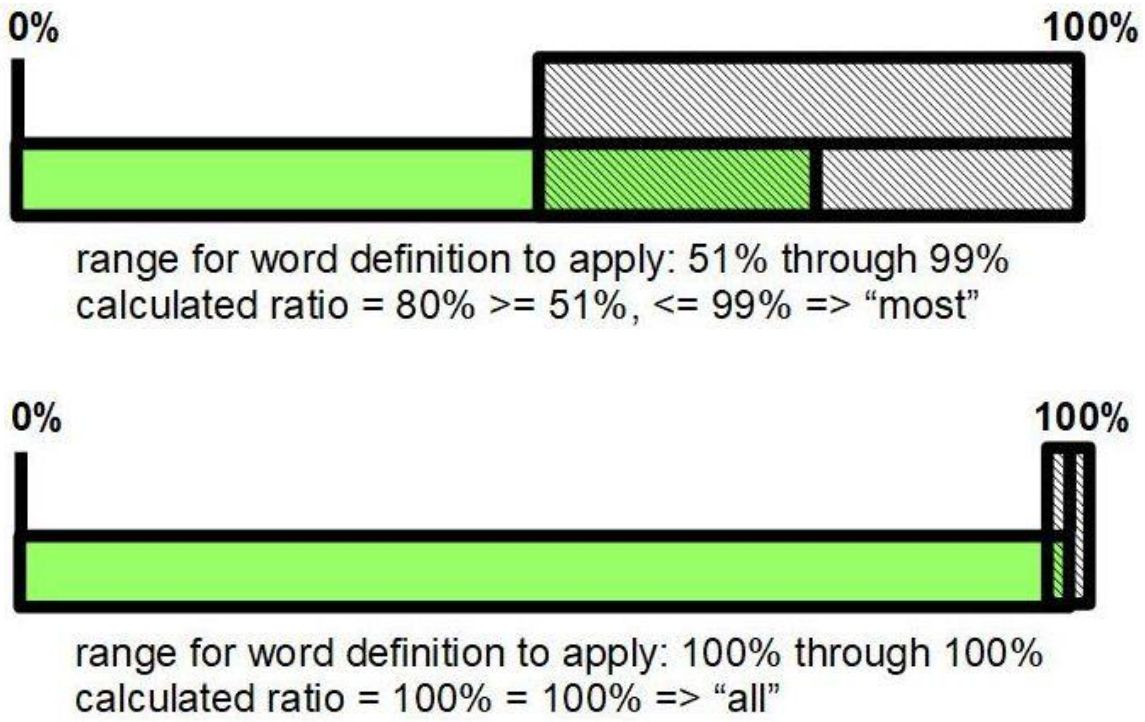}
	\caption{The words "most" and "all" apply only if the calculated ratio falls within the hatched range.}
	\label{fig:icon-range-most-all}
	\end{center}
\end{figure}

\begin{figure}
	\begin{center}
	\includegraphics[width=0.50\textwidth]{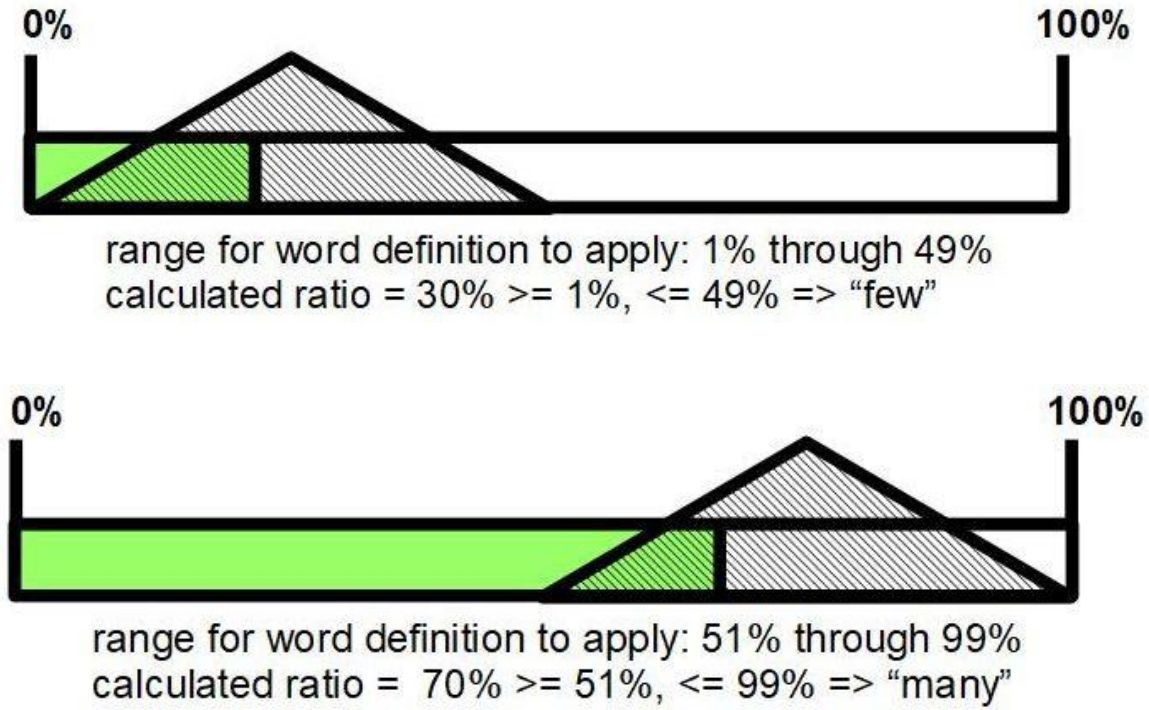}
	\caption{The words "few" and "many" apply only if the calculated ratio falls within the hatched range.}
	\label{fig:icon-range-few-many}
	\end{center}
\end{figure}

In the aforementioned examples, fixed values (viz., 0, 1, 2, 3) were used. By using ratios instead, especially 0\% through 100\%, relative concepts such as "most," "all,"  "few," and "many" can be defined with Value Bars, as shown in Figure~\ref{fig:icon-range-most-all} and Figure~\ref{fig:icon-range-few-many}. Some of these examples and values as shown do not exactly match human intuition about the meanings of these words, but all come close, and can be modified as desired.

The delimiting hatch marks are important in this icon because they allow quick estimation of the ratio of shaded bar length to the full bar length. Such ratios are needed to give visual estimates to language terms such as "few," "several," and "many," since such terms conceptually are fuzzy-bordered regions whose lengths and locations are a fuzzy percentage of the entire region. The hatched triangles in the diagram represent triangular, fuzzy logic membership functions.

\subsubsection{Example: Goals}

\begin{figure}
	\begin{center}
	\includegraphics[width=0.50\textwidth]{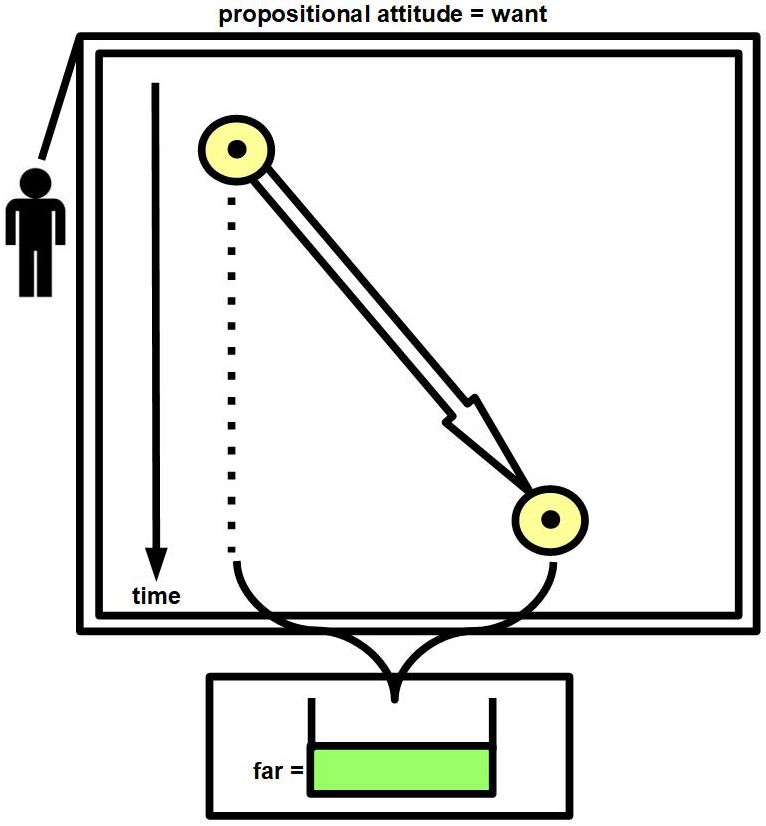}
	\caption{"I want to get that (yellow) thing away from me, as far as possible. (I don't care where or how fast.)"}
	\label{fig:icon-value-far-only}
	\end{center}
\end{figure}

\begin{figure}
	\begin{center}
	\includegraphics[width=0.50\textwidth]{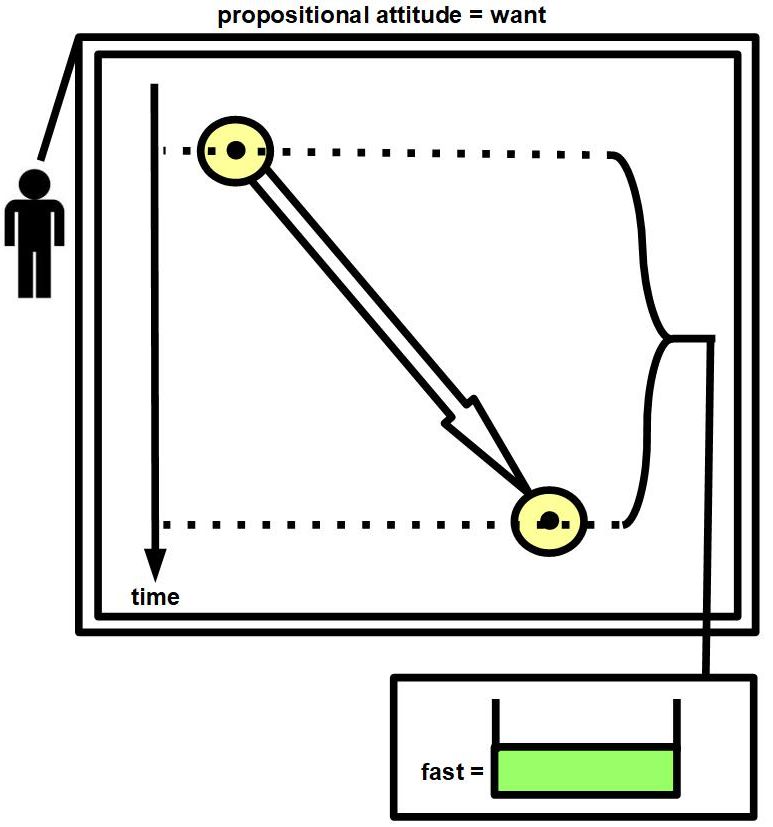}
	\caption{"I want to get that (yellow) thing away from me, as fast as possible. (I don't care where or how far.)"}
	\label{fig:icon-value-fast-only}
	\end{center}
\end{figure}

\begin{figure}
	\begin{center}
	\includegraphics[width=0.50\textwidth]{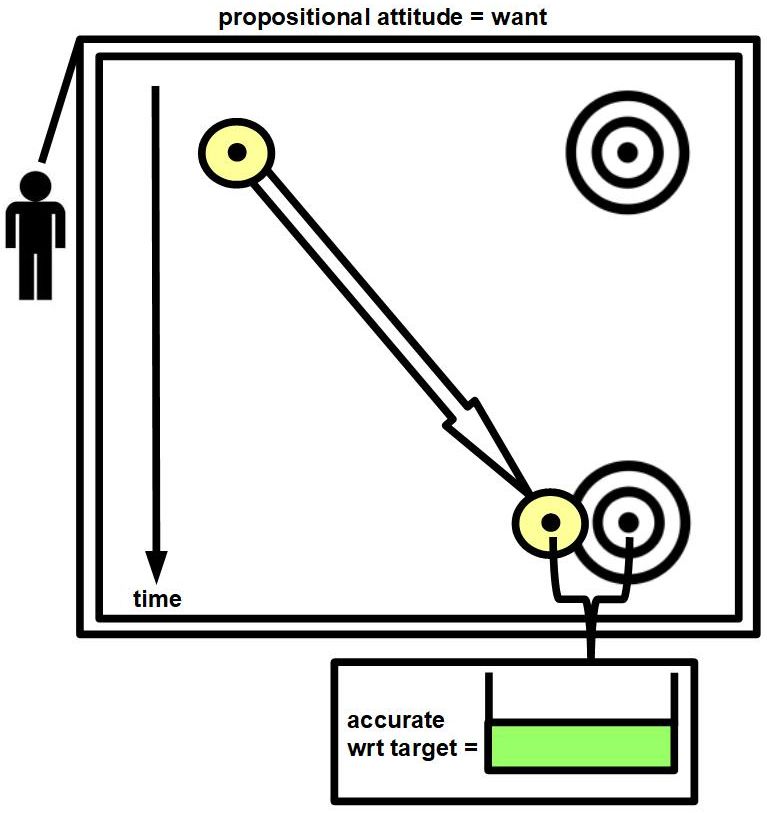}
	\caption{"I want to get that (yellow) thing on target, as accurately as possible. (I don't care how far or how fast.)"}
	\label{fig:icon-value-accurate-only}
	\end{center}
\end{figure}

\begin{figure}
	\begin{center}
	\includegraphics[width=0.50\textwidth]{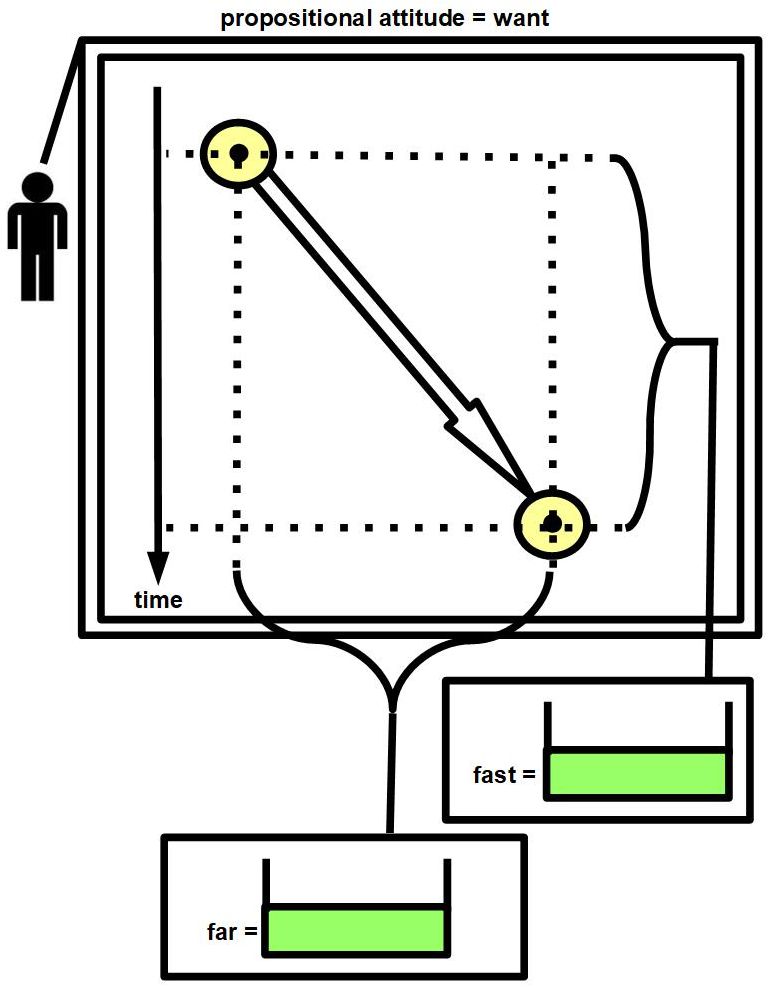}
	\caption{"I want to get that (yellow) thing away from me, as far as possible, as fast as possible, I don't care 
where."}
	\label{fig:icon-value-far-and-fast}
	\end{center}
\end{figure}

Such value representations can be combined in various, useful ways. For example, a certain range of distances can be given a fuzzy textual description such as "far," and conjunctions of such ranges from different measurement types can be a given fuzzy textual description such as "far" and "fast." In turn, these conjunctions of ranges can become useful for expressing common motivation, and therefore common goals, as shown in Figure~\ref{fig:icon-value-far-only}, Figure~\ref{fig:icon-value-fast-only}, Figure~\ref{fig:icon-value-accurate-only}, and Figure~\ref{fig:icon-value-far-and-fast}. The ability to represent goals in general is extremely important in artificial intelligence since "goal-directed" behavior is often cited as a component of the definition of intelligence itself (e.g., Minsky 1986, p. 22; Kurzweil 1990, p. 18; Kurzweil 1999, p. 73).

These nuances of which goals might be important to a person come into play in language translation, discussed later in this document.

\subsubsection{Geometrical variations}

Value Bars imply the existence of several interesting side topics that have not been addressed here. For example, in many cases a geometrical structure other than a rectangle is preferable, such as when angles are being measured, whereupon a circle would be more suitable. Similarly, many phenomena that have a wide numerical span are probably mentally understood via exponential scales, or via hyperbolic termination, such as representation of infinity by an asymptote's location.

\subsection{Wildcards (further development anticipated)}

\begin{figure}
	\begin{center}
	\includegraphics[width=0.50\textwidth]{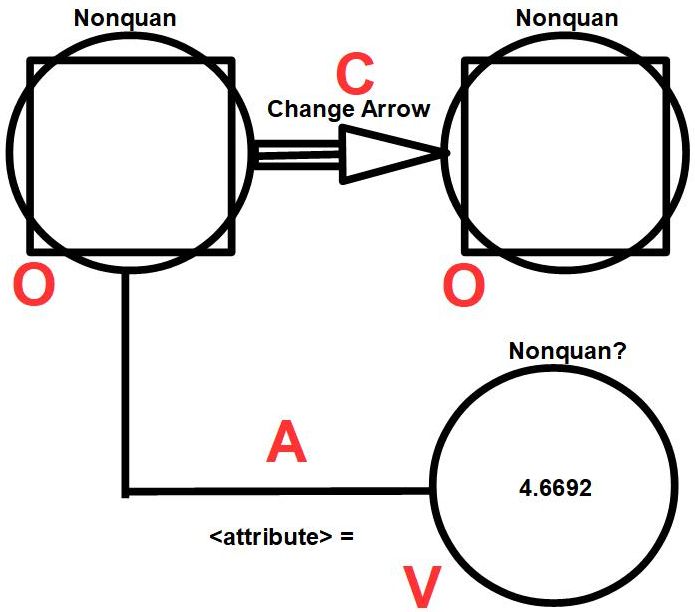}
	\caption{In this Tumbug representation of an OAVC system, the value (V) is numerical.}
	\label{fig:icon-wildcards-hako-value}
	\end{center}
\end{figure}

\begin{figure}
	\begin{center}
	\includegraphics[width=0.50\textwidth]{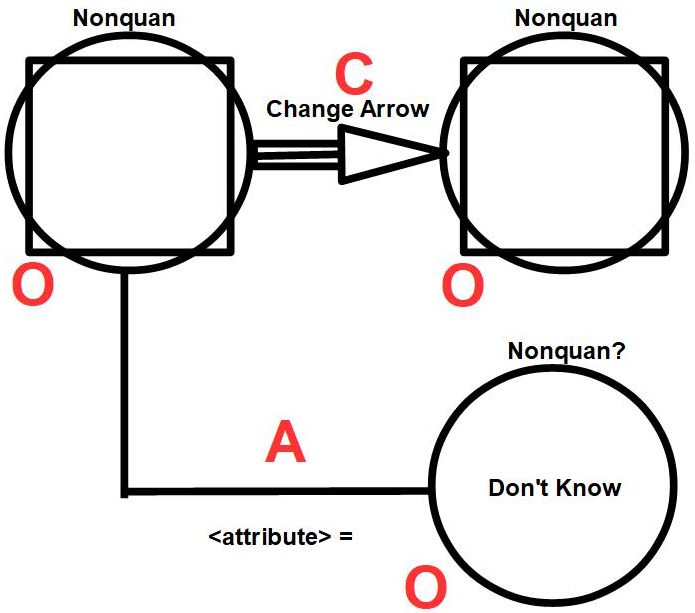}
	\caption{In this Tumbug representation of an OAVC system, the "value" is a string, which is closer to the concept of an object (O).}
	\label{fig:icon-wildcards-hako-object}
	\end{center}
\end{figure}

Because Tumbug is a visual representation, the usual type checking of digital programs is made much more difficult with Tumbug. In particular, when Tumbug represents an OAVC system, a person would ordinarily expect the value (V) to be numerical, as in Figure~\ref{fig:icon-wildcards-hako-value}. However, Tumbug is so general that it needs wildcards, which include "values" like "Don't Know," "Don't Care," or "Not Applicable," as in Figure~\ref{fig:icon-wildcards-hako-object}. Such strings are few enough in quantity that they could be implemented with special strings, as in the "NaN" (= Not a Number) special value implemented in the programming language Java (e.g., Gosling et al. 1996, p. 35), but Tumbug is more complicated still. For example, a noisy numerical value would need multiple pieces of information: (1) the value, (2) the probability density function (pdf) (e.g., Gaussian) that is associated with that value, (3) the mean of the pdf, (4) the variance of the pdf. This situation cannot even be represented with a four-element vector since the pdf is a function, not a numerical value.

The wildcards currently of interest in Tumbug are the following:
\begin{itemize}
	\item
		* = any value, matches even 0 values
	\item
		? = any value, matches even 0 or 1 values
	\item
		+ = any value, matches at least 1 value
	\item
		DK = don't know
	\item
		DC (or blank) = don't care
	\item
		DNE = does not exist
	\item
		allowed range
\end{itemize}

\begin{figure}
	\begin{center}
	\includegraphics[width=0.50\textwidth]{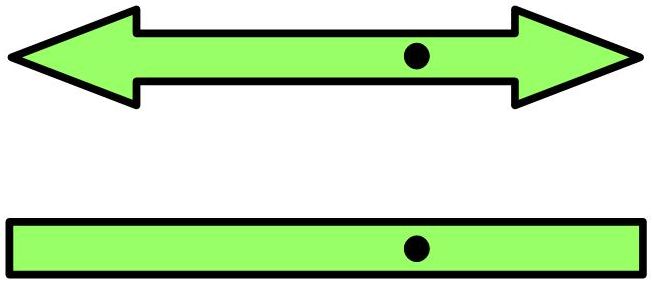}
	\caption{Tumbug's icon for a Wildcard. Top: Where any real value is possible. Bottom: Where only a limited range is possible.}
	\label{fig:icon-wildcards-pair}
	\end{center}
\end{figure}

A very useful wildcard is the asterisk "*", which is often used to mean "any value." In computer science, regular expressions routinely use this asterisk wildcard "*", as well as the two other wildcards "+" and "?". To generalize this situation, ideally a wildcard should be able to represent any type of range description, which suggests use of a number line of real numbers, or even a 2D region or higher of real numbers. Figure~\ref{fig:icon-wildcards-pair} shows two simple Wildcards. Theoretically it seems that the "V" (= value) slot in a Tumbug diagram should allow any object in order to be completely general, though the implications of such a decision have not been investigated in this study. However, this is the same design decision made in Allen Newell's architecture Soar (Newell 1990, p. 169): "Both the attributes
and values may be other objects, so that arbitrary attribute-value structures can occur." This situation is not too different from situations that humans encounter, such as "His final grade was the average of his midterm score and final exam score, where each letter grade is assigned to each 10\% section of the top of the spectrum." Effectively such a description of a value is equivalent to a deferred answer that requires data to be gathered and a function to be applied to that data before the resulting value can be known.

\begin{figure}
	\begin{center}
	\includegraphics[width=0.50\textwidth]{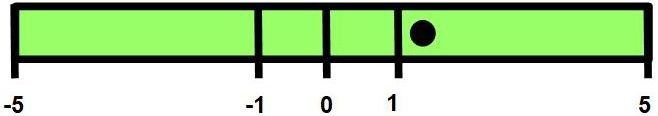}
	\caption{A Wildcard that is limited to a point within the continuous range of values from -5 through 5.}
	\label{fig:icon-wildcards-rolling-ball-bar-labeled}
	\end{center}
\end{figure}

The icons in the figure are intended to suggest that the small ball inside the smooth tube can roll to any position within the tube, especially unpredictably. This pictorially represents that the ball can exist at any point within the range of values within the tube. More commonly this range has values marked on it, as in Figure~\ref{fig:icon-wildcards-rolling-ball-bar-labeled}.

\begin{figure}
	\begin{center}
	\includegraphics[width=0.25\textwidth]{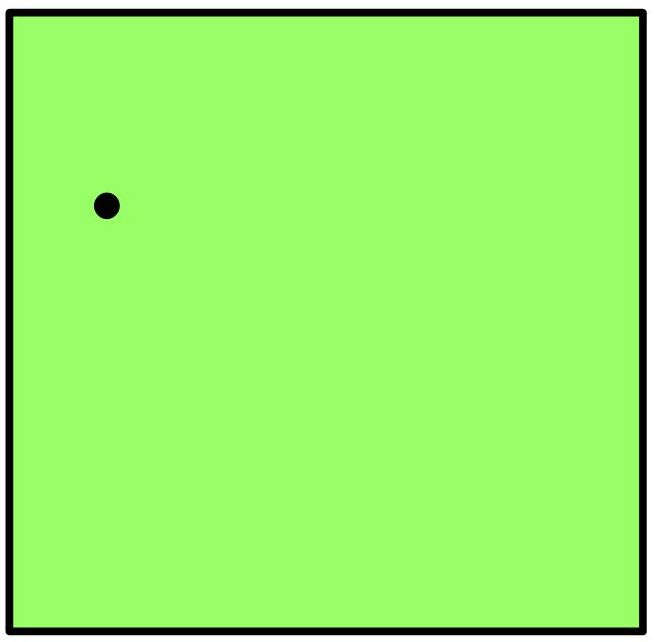}
	\caption{A Tumbug 2D Wildcard that is a square region.}
	\label{fig:icon-wildcards-2d}
	\end{center}
\end{figure}

The region in which the ball can roll can be easily extended to 2D, 3D, or higher. Figure~\ref{fig:icon-wildcards-2d} shows a 2D wildcard region.

As an example, consider a specific attribute value x. There exist a few common situations where the visual value of x may be difficult to plot:

\begin{enumerate}
	\item
		\textbf{Problem:} x has no magnitude.\\
		\textbf{One solution:} Plot a point at x = 0.\\
		\textbf{Justification:} Simple and obvious.
	\item
		\textbf{Problem:} x has no value, not even zero, because x does not exist.\\
		\textbf{One solution:} Do not plot any point. Conceptually x = nil, the same value given to new pointers.\\
		\textbf{Justification:} Any plot of x at a given point means x has that value, so there should be no plot.
	\item
		\textbf{Problem:} x is unknown, although the user does not care about its value.\\
		\textbf{One solution:} Pick a random value for x, and plot that value.\\
		\textbf{Justification:} Since the user does not care which value x has, any value will work fine.
	\item
		\textbf{Problem:} x is unknown, and the user does care about its value.\\
		\textbf{One solution:} Plot all possible values, and apply probability to the ranges.\\
		\textbf{Justification:} The answer does get plotted, and the probability becomes a caveat for acceptance.
\end{enumerate}

\begin{figure}
	\begin{center}
	\includegraphics[width=0.75\textwidth]{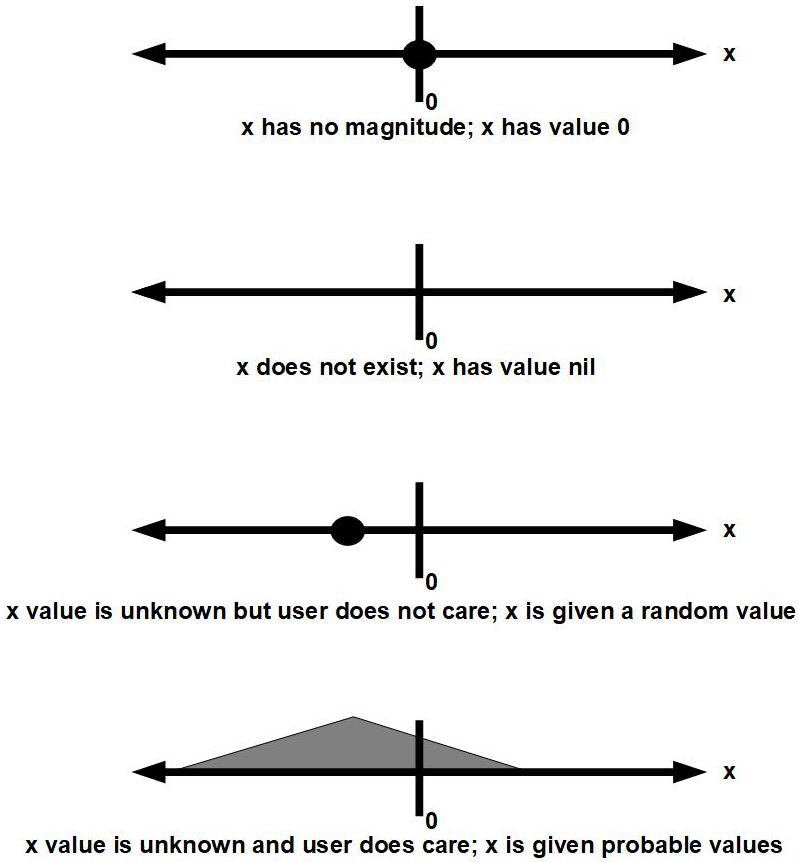}
	\caption{For the attribute of magnitude, good visual representations of Wildcards exist.}
	\label{fig:icon-wildcards-magnitude}
	\end{center}
\end{figure}

Each of these cases and its suggested visual equivalent is shown in Figure~\ref{fig:icon-wildcards-magnitude}.

The described solution in the figure works acceptably for attributes where x is simply a 1D magnitude, but the problem is more severe when x is a multivalued attribute like color (which is 3D) or shape (which is $\infty$D), not because of the multiple dimensions but because in practice such an object cannot even be depicted since any depiction must be rendered with some color and some shape. There exist workarounds, but the workarounds are less intuitive for problematic attributes like color. 

The aforementioned situations have an interesting implication: Numerical values from a human perspective have more attributes than mere value. These attributes are largely independent and consist at least of:

\begin{itemize}
	\item
		magnitude
	\item
		existence
	\item
		range
	\item
		importance
\end{itemize}

The zero case is ordinarily clear, though for clarity a KRM should not default to value = zero only because it is convenient, since zero is often a mathematically problematic point and the KRM display should be as unexpected as tolerable for visualizing the system In Tumbug there exists at least three Wildcard values:

\begin{itemize}
	\item
		don't know ("DK" symbol)
	\item
		don't care ("-" symbol)
	\item
		does not exist ("nil" symbol)
\end{itemize}

In the sets of four cases diagrammed above, these cases combined as:

\begin{enumerate}
	\item
		does not exist ("nil" symbol)
	\item
		unknown but user does not care: "DK/DNC"
	\item
		unknown and user does care: "DK/C"
\end{enumerate}

\begin{figure}
	\begin{center}
	\includegraphics[width=0.50\textwidth]{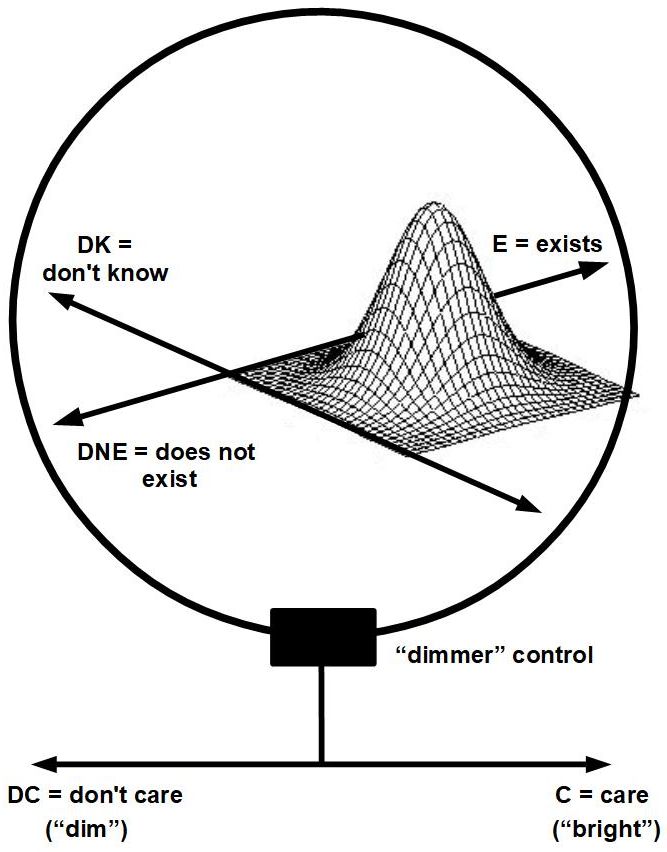}
	\caption{The three wildcards of "exist," "know," and "care" are independent and can be generalized and integrated into a 3D plot. Each of the three wildcards has an associated spectrum, even if a spectrum for such a concept is rarely used. (Source: unknown.)}
	\label{fig:icon-wildcards-unified-normal-internet}
	\end{center}
\end{figure}

These three wildcards can be generalized into a single diagram, as in Figure~\ref{fig:icon-wildcards-unified-normal-internet}. Although humans often tend to think of "exists" as a binary condition "exists versus does not exist," in physics such concepts do not strictly hold. For example, the question "An electron exists within this atom of iron" may not have a clear-cut answer if that atom of iron has formed a metallic bond with another iron atom. Similarly, humans tend to think of "know" as the binary condition "know versus don't know," in real life such concepts to not strictly hold, since likelihood of a fact falls within a spectrum from "absolutely not" through "absolutely sure." For example, the question "Can your son solve a random Rubik cube position within five minutes of seeing a Rubik cube for the first time?" evokes the commonsense answer "no," but nobody knows for certain.

\begin{figure}
	\begin{center}
	\includegraphics[width=0.60\textwidth]{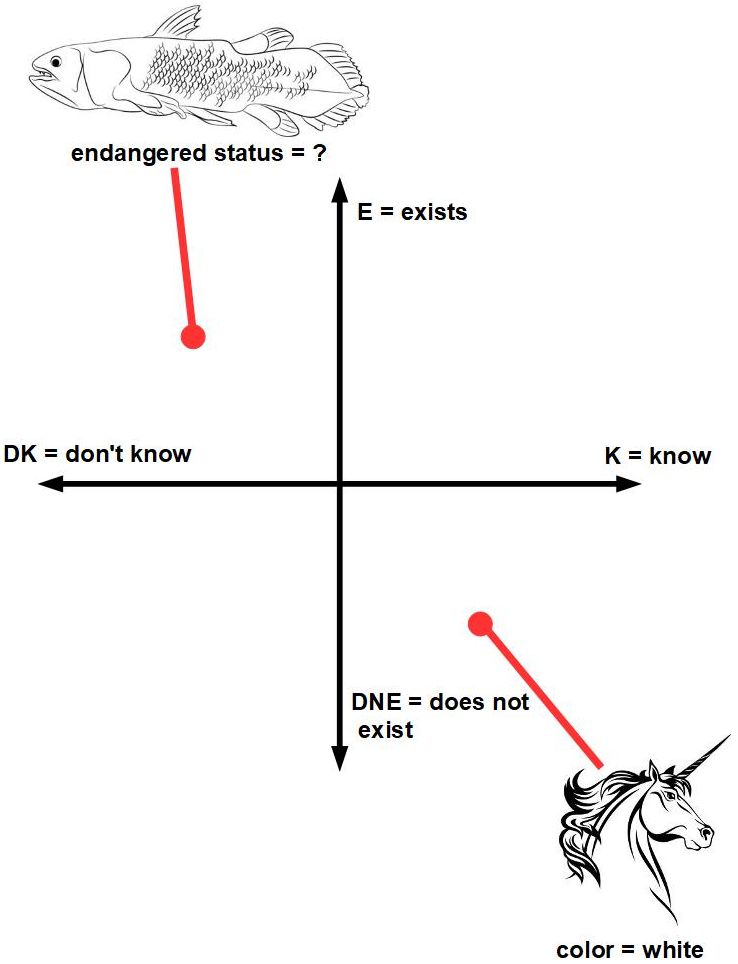}
	\caption{Existence and knowledge are independent; people might not know facts about existent entities, yet might know facts about nonexistent entities. (Source: Drawing Tutorials 101.)}
	\label{fig:icon-wildcards-unicorn-coelacanth}
	\end{center}
\end{figure}

Existence is independent from knowledge. For example, the question "Are coelacanths endangered?" involves an existent animal but an unknown attribute value, in contrast to the question "What color are unicorns?" involves a non-existent animal with a consensus attribute value. See Figure~\ref{fig:icon-wildcards-unicorn-coelacanth}. Similar situations arise even in mathematics, such as in induction proofs where a base case has not been established. For example, if n = n + 1 for all positive integers n, then it is true that for a specific n = k the formula will also be true for n = k + 1. However, no base case can ever be found whereby k = k + 1, therefore this "fact" becomes known in a formula that does not legitimately exist.

\subsection{Range Caps}

\begin{figure}
	\begin{center}
	\includegraphics[width=0.50\textwidth]{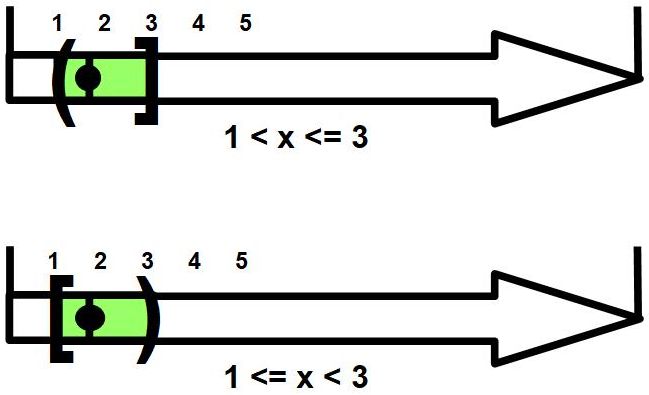}
	\caption{Range Caps are used on Value Bars to mark off a range and to show inclusion/exclusion of the end point.}
	\label{fig:icon-range-caps}
	\end{center}
\end{figure}

Figure~\ref{fig:icon-range-caps} shows Range Caps.

Range Caps are just the four familiar keyboard characters "(", ")", "[", and "]" as used in math to mark the ends of 1D range on a number line. The brackets "[" and "]" mean the end point is included, the parentheses "(" and ")" mean the end point is excluded.

\section{Change-like Building Blocks of Tumbug (C)}

\subsection{Single Time Arrows}

Most commonly when Tumbug needs a Time Arrow, it uses a Single Time Arrow. However, Split Time Arrows also exist to represent multiple possibilities along the timeline. See the section on that topic for more details.

\begin{figure}
	\begin{center}
	\includegraphics[width=0.06\textwidth]{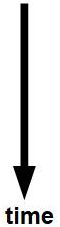}
	\caption{Tumbug's icon for a Time Arrow.}
	\label{fig:icon-time-arrow-no-now}
	\end{center}
\end{figure}

\begin{figure}
	\begin{center}
	\includegraphics[width=0.10\textwidth]{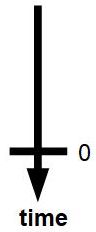}
	\caption{Typically a Time Arrow has a hatch mark labeled zero ("0") to mark the current time (= "now").}
	\label{fig:icon-time-arrow-now}
	\end{center}
\end{figure}

Tumbug's most common icon for a Time Arrow is shown in Figure~\ref{fig:icon-time-arrow-no-now}. Very often a hatch mark representing the current time is also included, and is labeled "0" to mean "now" (Figure~\ref{fig:icon-time-arrow-now}).

It should be clear that the concept of time is common in the WS150 problems. 

\begin{figure}
	\begin{center}
	\includegraphics[width=0.50\textwidth]{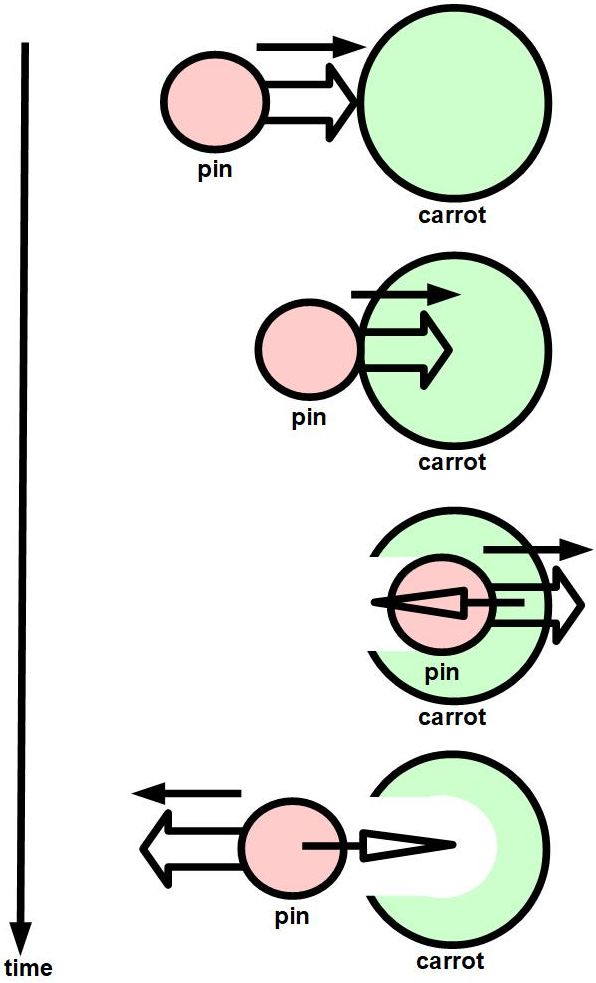}
	\caption{[104] Tumbug for "A pin stuck into a carrot will leave behind a hole when the pin is withdrawn."}
	\label{fig:ws-104}
	\end{center}
\end{figure}

The Tumbug icon used to represent time is a single-line arrow, usually pointing downward on the page, labeled with the word "time" or "t", and running down the left-hand side of the diagram as shown in Figure~\ref{fig:ws-104}. Any scenario where time is important to represent (e.g., for grammatical tense or for a temporal list of expected actions) must have this arrow placed on the associated diagram, typically on the left.

\textbf{1. Time without space, WS150 example: \#104 (carrot)}

Although most WS150 problem statements involve time, time need not be involved in every statement. This is particularly true of general rules, such as the rule in Figure~\ref{fig:ws-104}.

This example is derived from a portion of WS150 question \#104.\\
"[104] I stuck a pin through a carrot. When I pulled the pin out, it left a hole. What left a hole?
POSSIBLE ANSWERS: \{the pin, the carrot\}"

\textbf{2. Generalization: space-with-time}

Tumbug combines the concepts of space and time into a single concept called "space-with-time." This combination is motivated by the following observations:

\begin{itemize}
    \item
		Language descriptions of objects moving in space are extremely common.
    \item
		Space and time are already combined into the single concept "spacetime" in physics.
    \item
		Time is sometimes spatialized as an extra dimension of space when training neural networks. (E.g., Simpson 1990, p. 326.)
\end{itemize}

\begin{figure}
	\begin{center}
	\includegraphics[width=0.25\textwidth]{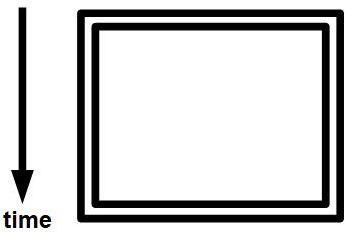}
	\caption{The Time Arrow icon and some type of Location Box icon are commonly used together when moving objects are described.}
	\label{fig:icon-time-arrow-with-location-box}
	\end{center}
\end{figure}

Not surprisingly, then, the Location Box icon (for space) and Time Arrow icon (for time) are often used together, as shown in Figure~\ref{fig:icon-time-arrow-with-location-box}.

\textbf{3. Discretized time}

Not all systems use continuous time. Some examples of systems that use discretized time are: board games (since each person takes their turn), sequential marquee lights on a theater sign, frames on a film strip for a motion picture camera, displayed minutes on a digital watch, instructions executing during each computer processor clock cycle, cellular automata (since nothing happens between transitions), and steps of a computer simulation.

\begin{figure}
	\begin{center}
	\includegraphics[width=0.15\textwidth]{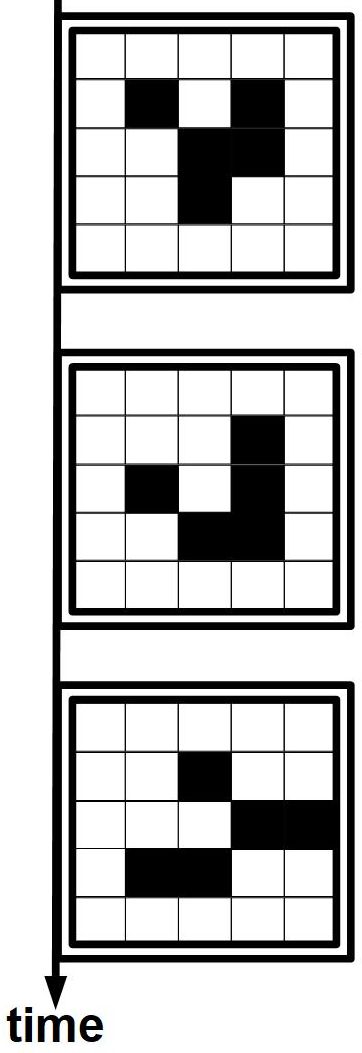}
	\caption{Time can be discretized with Location Boxes, with each Location Box holding one time increment of the process. This diagram shows three consecutive steps of the evolution of a glider from Conway's Game of Life, which is an example of a specific cellular automaton.}
	\label{fig:life-customized}
	\end{center}
\end{figure}

Tumbug can represent discrete time increments easily by stacking either C Aggregation Boxes or Verbatim Boxes along a Time Arrow (preferably touching the Time Arrow to make the meaning clear). In such a representation, each C Aggregation Box or Verbatim Box means an aggregated section of time, rather than an aggregated section of space. In Figure~\ref{fig:life-customized}, the first three steps of a simulation of a glider configuration from Conway's Game of Life (which is a cellular automaton) is shown, using Location Boxes since the spatial configuration at each step is important.

\subsection{Motion Arrows}

From the outset, one design goal of Tumbug was to regard moving objects (in contrast to static objects) to be the default situation in the real world so that Tumbug's KRM design would not need to struggle later to add time to a fundamentally static KRM. Too many computer science inventions such as data bases and artificial neural networks have encountered this problem of design short-sightedness that required later augmentation with time.

\subsubsection{Motion of objects}

\begin{figure}
	\begin{center}
	\includegraphics[width=0.25\textwidth]{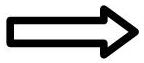}
	\caption{Tumbug's icon for a Motion Arrow.}
	\label{fig:icon-motion-arrow-object}
	\end{center}
\end{figure}

\begin{figure}
	\begin{center}
	\includegraphics[width=0.25\textwidth]{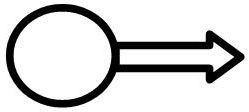}
	\caption{Tumbug diagram of a physical object in motion using an Object Circle with a Motion Arrow.}
	\label{fig:icon-motion-arrow-object-with-object}
	\end{center}
\end{figure}

In Tumbug, solid arrows are used to represent motion of an object (Figure~\ref{fig:icon-motion-arrow-object}, therefore Motion Arrows are usually used only in conjunction with an object See Figure~\ref{fig:icon-motion-arrow-object-with-object}. If the direction of motion is important, this implies that space is important, whereupon the arrow should be placed inside a Location Box. Motion Arrows are allowed to point in any direction, whether inside or inside of a Location Box. 

\begin{figure}
	\begin{center}
	\includegraphics[width=0.25\textwidth]{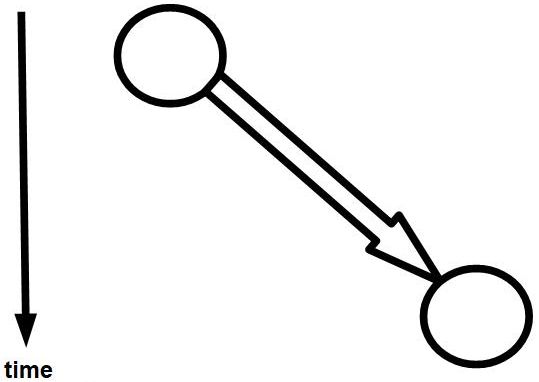}
	\caption{If duration of a moving object is important, then this diagram type should be used.}
	\label{fig:icon-motion-with-time-only}
	\end{center}
\end{figure}

\begin{figure}
	\begin{center}
	\includegraphics[width=0.25\textwidth]{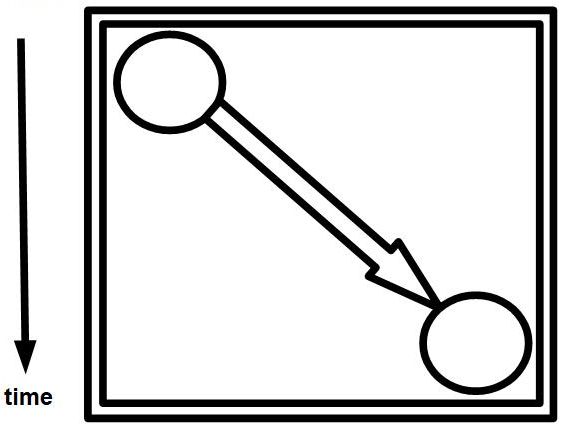}
	\caption{If relative duration and relative distance of a moving object is important, then this diagram type should be used since space and time are important in such a situation, though not exact values.}
	\label{fig:icon-motion-with-time-and-space}
	\end{center}
\end{figure}

If specific points in time or durations of time are important while an object while it is moving, then a Time Arrow should be used as shown in Figure~\ref{fig:icon-motion-with-time-only} and Figure~\ref{fig:icon-motion-with-time-and-space}.

Should data objects use the same Motion Arrow as physical objects? The current Tumbug design decision is "yes," and the reasoning for this decision requires some thought. First, information always requires some physical medium to exist, and on which the information can be transferred. Second, this physical medium may be either: (1) discrete storage units that can be transported individually, or (2) continuous media that connects communicating agents. Some examples of discrete storage units are: (1a) physical strands of DNA that carry genetic data, (1b) physical pages of print that carry written data, and (1c) physical flash drives that carry computerized data. Some examples of continuous media are: (2a) the air between two human speakers that carries sound waves, (2b) the wires between computer components that carry electrical charges, and (2c) fiber optic cables that carry light waves.

One implication of these observations for Tumbug is that discrete storage units can be represented in Tumbug as either physical units via Physical C Object Circles, which can then either be assumed to be carrying information, or can be regarded as only physical units with the capability of carrying information. If this is the way such units are regarded, then their data component can be represented as a state within a Physical C Object Circle, which is just a type of attribute. If continuous media such as air can be assumed to exist with such convenience and with such ubiquity that there is no need to represent the connecting medium explictly, then only the message content as a whole needs to be considered. If this is the way that communication is regarded, then the data component can be represented as a Data C Object Circle. In either case, data transfer is regarded as some type of motion of an C Object Circle, whether a Physical C Object Circle or a Data C Object Circle, and since the data component of the C Object Circle used will have already been represented if the designer deemed the data component important enough to include, then the only remaining representation necessary is the motion of that C Object Circle, which at that stage would not benefit to be distinguished between physical object motion and data object motion. Therefore only one type of Motion Arrow is needed in Tumbug, which is the convention used in this document in order to keep things simple.

Obviously Tumbug does not show the intermediate positions of a moving object. Such a smeared display would undesirable, anyway, since a 2D image blurred across its 2D trajectory would typically obscure specific features. Instead, Tumbug shows the first and last locations of a moving object and leaves the viewer to fill in the missing intermediate steps. For a more complicated trajectory, motion arrows can be used, or a few, key, intermediate positions can be shown. Modern humans are accustomed to seeing a series of frames that represent a moving object, such as in film frames, or special effects in films.

\subsubsection{Streams of objects}

A complication in the real world, especially in the modern world, is that much of the transfer being done is being done with data, not physical objects. Conceptually, data objects can be moved in largely the same way that physical objects can be moved, so in Tumbug the same solid Motion Arrow is used with both type of objects. A more difficult concept to represent is that of streams, whether data streams of object streams.

In general, Tumbug must take into account a number of independent, very basic considerations when representing motion of streams of objects. Tumbug considers the following four dimensions of motion stream attributes:

\begin{enumerate}
	\item
		solidity of the pathway - on a carrier path versus not on a carrier path
	\item
		solidity of the flowing tokens- solid tokens versus data tokens flowing along the pathway
	\item
		stream organization - random versus information-carrying
	\item
		allowed direction - toward one end of the carrier path versus the opposite end of the carrier path
\end{enumerate}

\begin{figure}
	\begin{center}
	\includegraphics[width=0.50\textwidth]{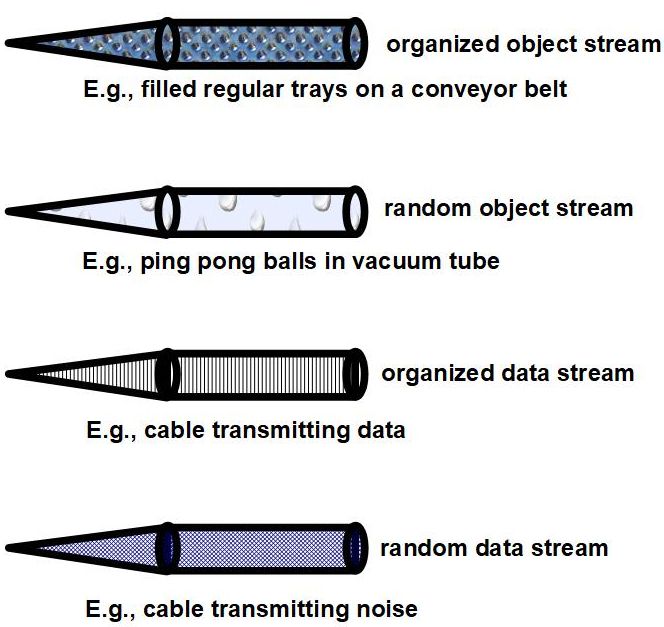}
	\caption{Tumbug for a stream within a Pathway Tube, for objects versus data, and for organized versus random.}
	\label{fig:icon-tubes-streams-tubes}
	\end{center}
\end{figure}

\begin{figure}
	\begin{center}
	\includegraphics[width=0.50\textwidth]{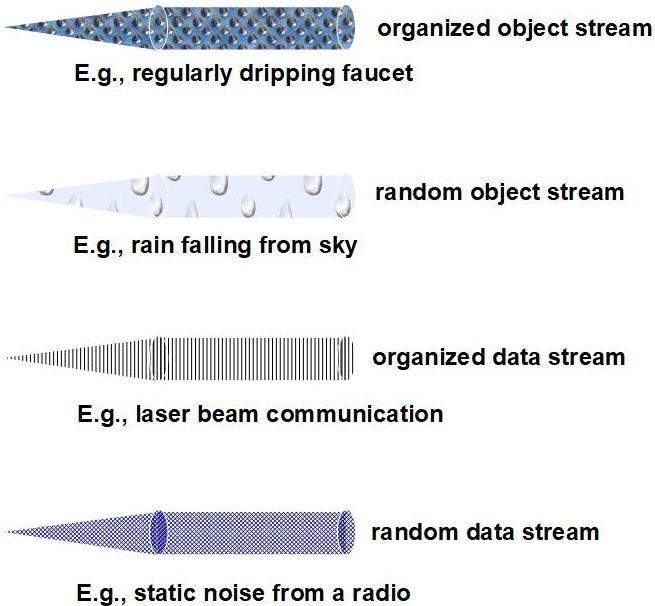}
	\caption{Tumbug for a stream without a Pathway Tube, for objects versus data, and for organized versus random.}
	\label{fig:icon-tubes-streams-no-tubes}
	\end{center}
\end{figure}

A complete list of these possible combinations follows for a single direction (i.e., to the left) is in Figure~\ref{fig:icon-tubes-streams-tubes} and Figure~\ref{fig:icon-tubes-streams-no-tubes}. (The exact visual patterns of hatchings and bitmaps will vary between vector graphics editors.)

The dilemma may arise as to whether to represent a flow of many objects as individual objects (via C Object Circles) versus a stream. Tumbug does not currently specify any threshold to resolve this choice; currently this is merely a matter of taste or practicality for the user.

\subsubsection{Shorthand notation}

\begin{figure}
	\begin{center}
	\includegraphics[width=0.50\textwidth]{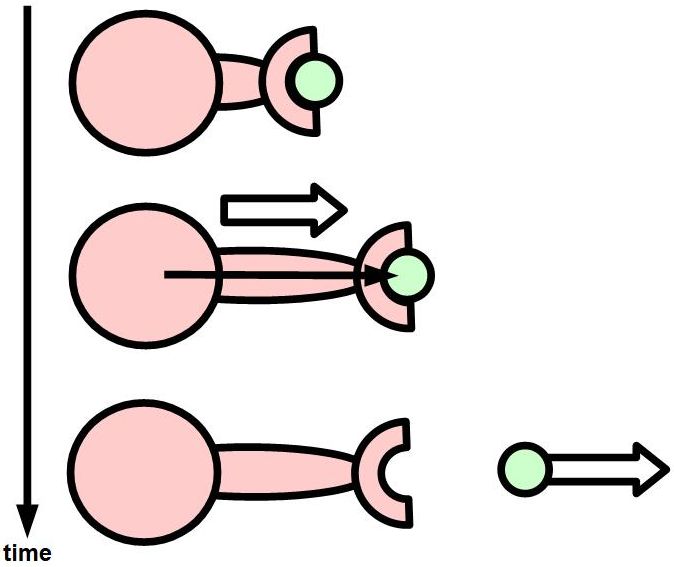}
	\caption{Throwing object X consists of: (1) grasping X, (2) propelling arm with great force; (3) releasing X.}
	\label{fig:icon-motion-throw-throw}
	\end{center}
\end{figure}

\begin{figure}
	\begin{center}
	\includegraphics[width=0.50\textwidth]{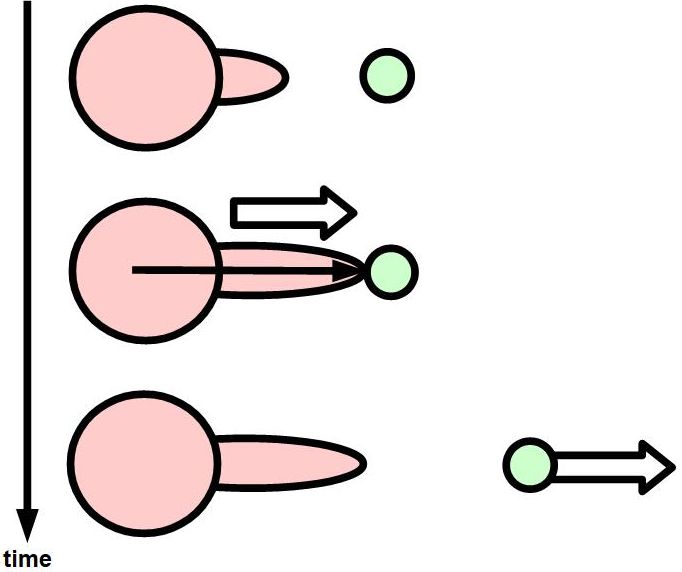}
	\caption{Bumping object X consists of: (1) aiming at X; (2) propelling arm with force to contact X, (3) ceasing force.}
	\label{fig:icon-motion-bump}
	\end{center}
\end{figure}

\begin{figure}
	\begin{center}
	\includegraphics[width=0.50\textwidth]{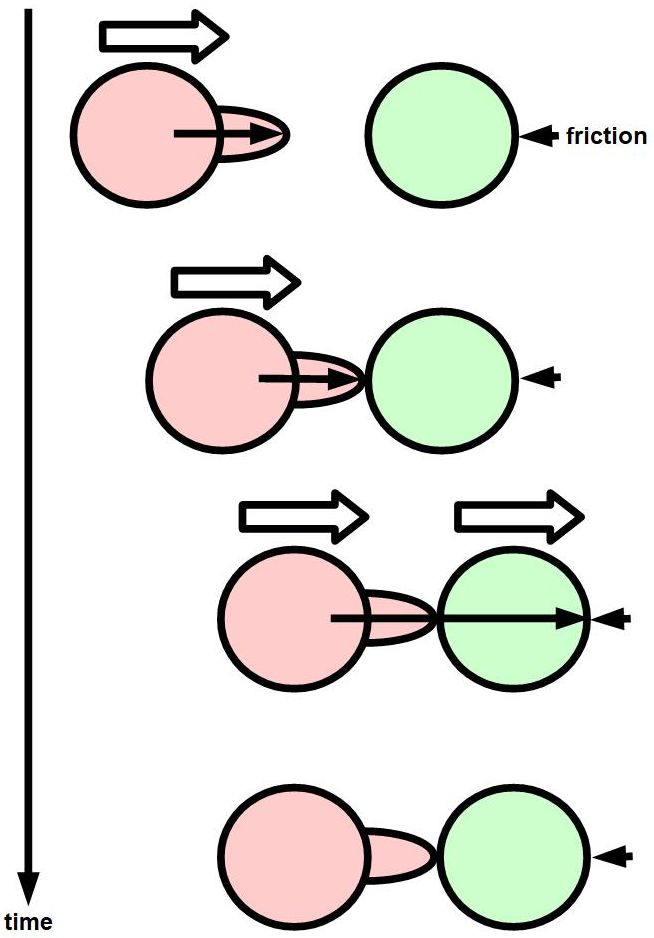}
	\caption{Pushing object X consists of: (1) traveling to X; (2) pushing on X with enough force to move it, (3) traveling while pushing X; (4) ceasing force.}
	\label{fig:icon-motion-push}
	\end{center}
\end{figure}

\begin{figure}
	\begin{center}
	\includegraphics[width=0.50\textwidth]{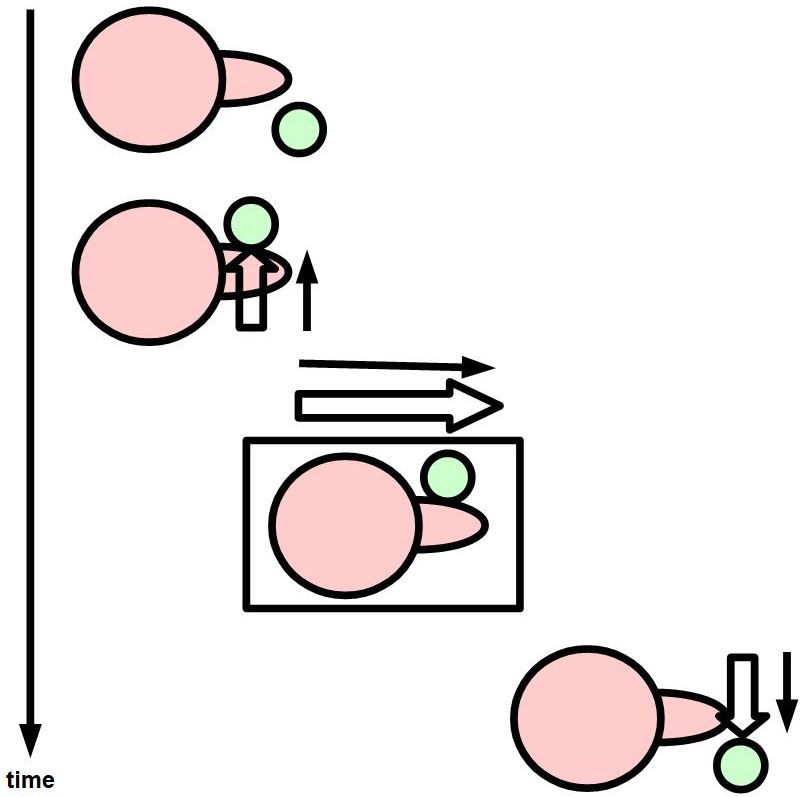}
	\caption{Carrying object X consists of: (1) approaching X; (2) picking up X, (2) moving while holding X, (3) putting down X.}
	\label{fig:icon-motion-carry}
	\end{center}
\end{figure}

Note that there exist numerous nuances in how a subject can move a direct object. For example, the subject can throw the object (Figure~\ref{fig:icon-motion-throw-throw}), bump the object (Figure~\ref{fig:icon-motion-bump}), push the object (Figure~\ref{fig:icon-motion-push}), carry the object (Figure~\ref{fig:icon-motion-carry}), and so on.

\begin{figure}
	\begin{center}
	\includegraphics[width=0.50\textwidth]{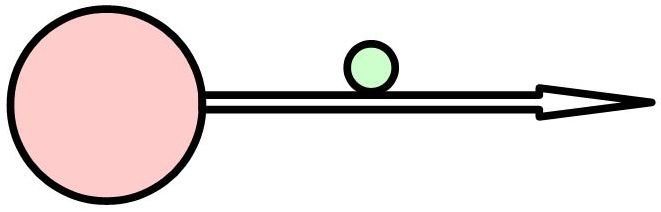}
	\caption{A shorthand notation for showing transference an object: no timeline, no end effector, no details.}
	\label{fig:icon-motion-shorthand}
	\end{center}
\end{figure}

However, when representing the extremely common concept of transference of an object, it is often irrelevant as to whether the object was thrown, bumped, carried, rolled, or other, since often the only important part of the situation is that the subject ultimately caused the object to be moved to another location. For this reason, the shorthand notation of Figure~\ref{fig:icon-motion-shorthand} is useful for suppressing any such details.

This shorthand transfer diagram, which shows the transferred object alongside the Motion Arrow, becomes very useful when representing basic English grammatical patterns because it generalizes the situation of object transfer so much that this one diagram represents one of only about 4-5 basic grammatical patterns that exist in English.

\subsection{Force Arrows}

\begin{figure}
	\begin{center}
	\includegraphics[width=0.25\textwidth]{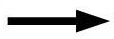}
	\caption{Tumbug's icon for a Force Arrow.}
	\label{fig:icon-force-arrow-alone}
	\end{center}
\end{figure}

Figure~\ref{fig:icon-force-arrow-alone} shows a Force Arrow.

Roger Schank's Conceptual Dependency theory included one Primitive Act that was called PROPEL, which Lytinen (1992, p. 52) described as "The application of a physical force to an object." The concept of force is mostly missing from the WS, but can be very important because Tumbug needs to model physics, and force is one of the most basic concepts of physics. For example, an object struggling against wind, a water current, a magnetic field, or a gravitational field will need to have that field visually described in Tumbug, probably as a vector field of arrows, in order for Tumbug to make an estimated prediction of the object's final trajectory. Also, the concept of force makes a difference when an end effector from an actor touches an object: without representation of force, a Tumbug diagram showing the end effector at the surface of the object would not make it clear whether the end effector had stopped at the object's surface, or was pushing against the object with a force that might budge (or propel) the object. Overlapping forces, such as a falling object (acted upon by gravity) being deflected by wind (acted upon by air) can be modeled by merely using a separate plane for each force.

Conceptual Dependency theory differs from Tumbug in that the results of inferences are also automatically included in the same Tumbug diagram as the original statements, whereas Conceptual Dependency theory does not perform inference, so continues to represent only the original statements. This difference will not be evident until Phase 2, however.

\begin{figure}
	\begin{center}
	\includegraphics[width=0.25\textwidth]{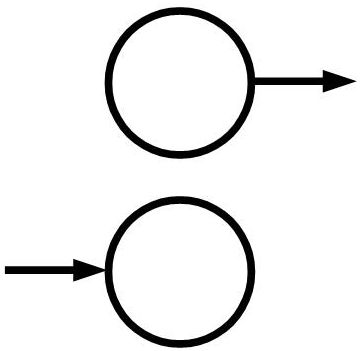}
	\caption{Top: The object is exerting a force to the right. Bottom: A force from the left is being exerted upon the object.}
	\label{fig:icon-force-arrow-out-in}
	\end{center}
\end{figure}

In conformance with conventional physics representations of force via vector fields, Tumbug therefore represents a given force in a given direction with a single arrow with direction representing the direction of the force and with length representing magnitude. If an object is exerting a force, then the Force Arrow points away from the object, and if an object is being acted upon by a force, then the Force Arrow points into the object. These two situations are shown in Figure~\ref{fig:icon-force-arrow-out-in}.

\begin{figure}
	\begin{center}
	\includegraphics[width=0.25\textwidth]{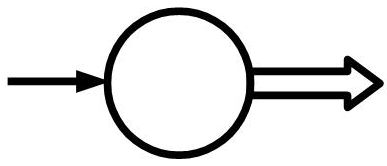}
	\caption{An object subjected to force typically causes motion of that object, represented by a Motion Arrow.}
	\label{fig:icon-force-arrow-moving}
	\end{center}
\end{figure}

\begin{figure}
	\begin{center}
	\includegraphics[width=0.50\textwidth]{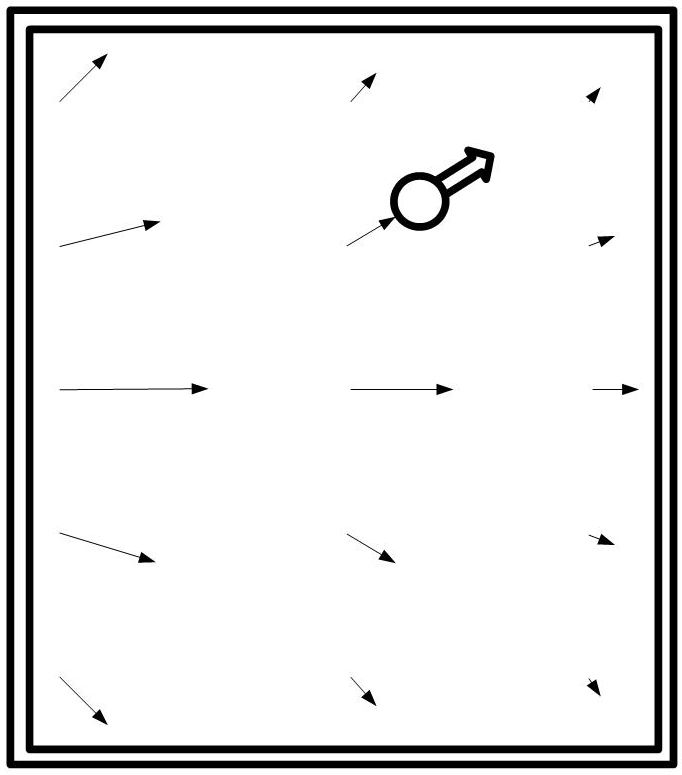}
	\caption{An example of a single object in motion as a result of a force acting on it within a vector field.}
	\label{fig:vector-field}
	\end{center}
\end{figure}

Frequently a force acting upon an object causes the object to move, which is shown in Figure~\ref{fig:icon-force-arrow-moving}. Frequently a force across a region is represented as a vector field, and an object in that vector field will move in accordance of the direction of the force vector at the object's location, which is shown in Figure~\ref{fig:vector-field}. In physics the length of the force vector varies with the strength of the corresponding force, but in informal situations force values are almost always unknown, in which case the Force Arrow lengths are irrelevant except possibly in a relative sense.

\subsection{Causation Arrows}

\subsubsection{Single causes}

\begin{figure}
	\begin{center}
	\includegraphics[width=0.06\textwidth]{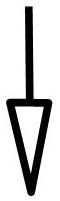}
	\caption{Tumbug's icon for a Causation Arrow.}
	\label{fig:icon-causation}
	\end{center}
\end{figure}

Figure~\ref{fig:icon-causation} shows a Causation Arrow. A Causation Arrow points from a cause to an effect in a Tumbug diagram. Often this arrow points to or from a specific state or a new attribute value, but it can also point to or from some type of Aggregation Box that describes a situation.

It is necessary to model causation (also known as causality, or cause-and-effect) in brain-level KRMs for at least two reasons: (1) an entity cannot be said to truly "understand" an event unless it understands the causation of the event (Schank 1976, p. 168), (2) a given verb may inherently involve the concept of causation, possibly in an indirect way, so such a verb cannot be easily represented by resorting to the representation of motion, possibly not at all unless causation is represented.

Schank wrote that, "The basic mechanism in understanding is the inference process" (Schank 1976, p. 169) because it is inferences that underlie CSR since inferences extract associated memories that can clarify an ambiguous sentence, which is exactly the situation highlighted in the WS problems. Chuck Rieger (Rieger 1975) listed sixteen processes of inference, a key one of which is "causative" inference, the subject of this section, and Schank described this inference type as asking "What caused the action or state in the sentence to come about?" (Schank 1976, p. 168).

Schank then expanded Rieger's "causative" inference class into four types: (1) result causation, (2) enable causation, (3) reason causation, (4) initiation causation. Currently Tumbug uses the single Causation Arrow for any of these four types. If desired, likelihood values can be assigned to Causation Arrows to represent the likelihood that one event caused another, but this document does not contain any such examples.

\begin{figure}
	\begin{center}
	\includegraphics[width=0.50\textwidth]{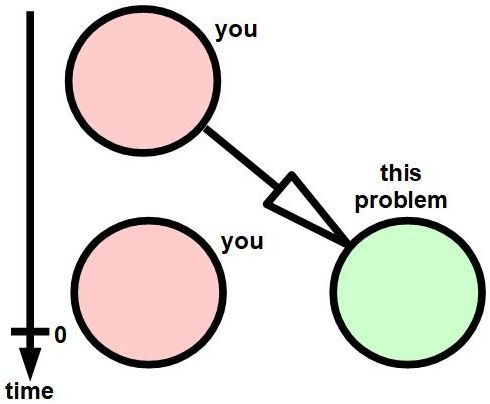}
	\caption{Tumbug for "You caused this problem." Note that the problem (= the green Object Circle) did not even exist until it was caused.}
	\label{fig:icon-causation-you}
	\end{center}
\end{figure}

Some sentences are virtually forced to use a Causation Arrow. Figure~\ref{fig:icon-causation-you} does not even need to label the verb "to cause" since the Causation Arrow already means "to cause."

\begin{figure}
	\begin{center}
	\includegraphics[width=0.70\textwidth]{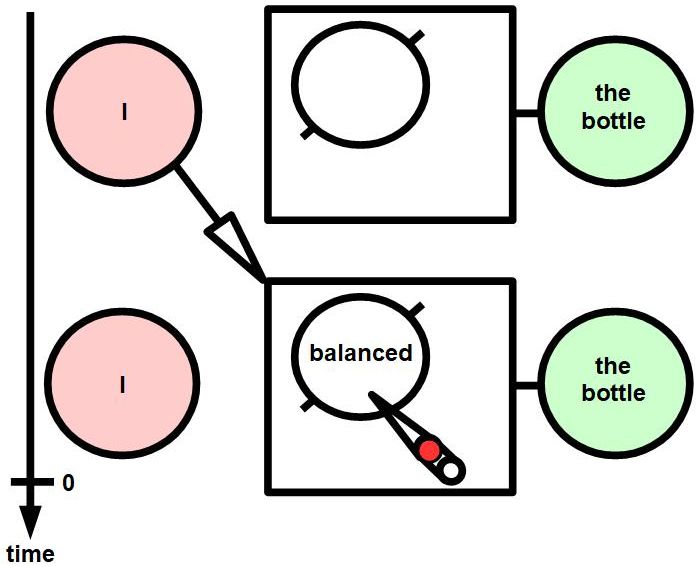}
	\caption{[21] Tumbug for "I balanced the bottle." = "I caused the bottle to go into a balanced state."}
	\label{fig:ws-021-balance}
	\end{center}
\end{figure}

Figure~\ref{fig:ws-021-balance} is more typical in that it shows a state diagram attached to one of the objects in a way that makes the entire state diagram resemble a simple attribute, except that the state diagram is written to the left of the direct object instead of below it, in order to save space and to allow the subject object to connect directly to the state diagram. Note that in this bottle example that it would be very difficult to describe the exact motions or actions that the subject took to cause the balance to occur, therefore such details are best hidden behind the single Causation Arrow pointing to a state within a state diagram.

\begin{figure}
	\begin{center}
	\includegraphics[width=0.70\textwidth]{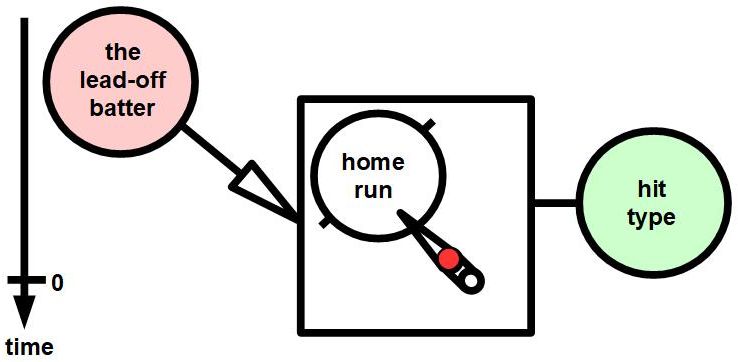}
	\caption{[75] Tumbug for "The lead-off batter hit a home run." = "The lead-off batter caused the hit type to go into a home run state."}
	\label{fig:icon-causation-home}
	\end{center}
\end{figure}

Sometimes the collection of states is not as neatly or as geometrically organized, such as in a State Diagram that contained all possible baseball hit types. Fortunately, such an intricate State Diagram is not needed, as shown in Figure~\ref{fig:icon-causation-home}. Since in that sentence there is mention of only one hit type, the diagram needs only to represent the situation that the state of the mentioned hit type was entered, not the previous state.

According to Schank (1976, p. 177), causation is like the glue that holds together all the sentences of a story so that every sentence makes sense within the story. In such story usage, causation "links" (which in Tumbug are equivalent Causation Arrows) must be generated by inferences with the system as the story is read, whereas in the aforementioned examples Tumbug was only representing what it was explicitly told. In any case, understanding and representation of causation is clearly important for any intelligent system, whether the system inputs or outputs causation information. Causation links can connect nearly every sentence in a story, as Schank demonstrates.

\subsubsection{Multiple causes}

\begin{figure}
	\begin{center}
	\includegraphics[width=0.75\textwidth]{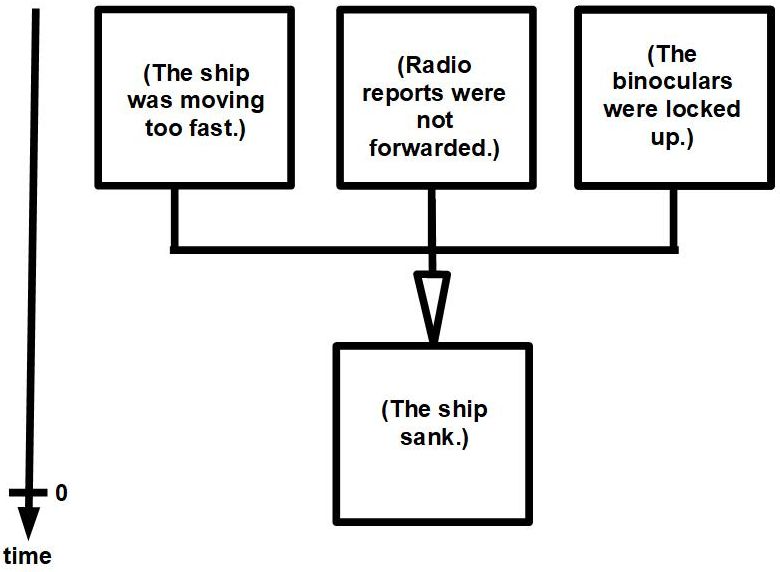}
	\caption{Three causes that together contributed to an effect: the sinking of the RMS Titanic.}
	\label{fig:icon-causation-multiple}
	\end{center}
\end{figure}

Multiple causes leading to a single effect are common in practice. For example, the sinking of the RMS Titanic in 1912 happened as a result of many causes, three of which were: (1) the ship was moving too fast, (2) radio reports of icebergs in the area were not forwarded by the radio operator, (3) the binoculars were locked up. To represent such situations Tumbug would need only to connect all the propositions somehow (a proposition is a declarative sentence that is either true or false), each proposition of which would likely be represented in a C Aggregation Box, and only a single Causation Arrow leading from the connected propositions would be needed. This is shown in Figure~\ref{fig:icon-causation-multiple}.

\subsubsection{Labeled causes}

Causation Arrows can also be labeled, and labeling them is desirable in most cases. Labels can be left generic, such as "physics" for "per the laws of physics," "math" for "per the laws of math," "logic" for an implication, "chemistry" for "per the laws of chemistry," or "social norms" for "per social norms." Labels can also be very specific, such as math operators such as "+", "=", "*", "/", or numbers of legal statutes.

\subsection{Correlation Boxes}

\begin{figure}
	\begin{center}
	\includegraphics[width=0.20\textwidth]{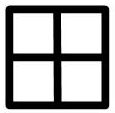}
	\caption{Tumbug's icon for a Correlation Box.}
	\label{fig:icon-correlator}
	\end{center}
\end{figure}

A Correlation Box is a Tumbug Building Block that relates one or more values in a mathematical way, like a function of two variables. The hardware equivalent of a Correlation Box would be an op amp (= operational amplifier) whose output is a function of the voltage input(s), such as a non-inverting summing amplifier, differential amplifier, or integrator. The Tumbug icon for a Correlation Box is intended to suggest a 2D plot on a graph whose origin is the center of a square, as shown in Figure~\ref{fig:icon-correlator}.

Only one problem from WS150 requires use of a Correlation Box, so if Tumbug is used as a static KRM then use of Correlation Boxes is rare. However, in a software version of Tumbug, Correlation Boxes would likely be running constantly as a background process in order to keep correlated attribute values of different objects in sync. The following WS150 problem demonstrates a situation that would be best described in a software version of Tumbug.

This example is from a portion of WS150 question \#24. A Correlation Box is necessary because each of two objects is changing its attribute values simultaneously in a manner that inherently involves the other object's attribute values.

"[24] I poured water from the bottle into the cup until it was full. What was full? POSSIBLE ANSWERS: \{the cup, the bottle\}"

The correlation in this example is that the water being transferred "from the bottle into the cup" must cause one container to empty at exactly the same rate that another container fills. If the total weight of the water were T, and the weight of water in the bottle and cup were ${w_1}$ and ${w_2}$, respectively, then the mathematical formula for this correlation would be ${w_1}$ + ${w_2}$ = T, or equivalently for each variable:\\

\setlength\parindent{0pt}
${w_1}$ = T - ${w_2}$ \\
${w_2}$ = T - ${w_1}$ \\
\setlength\parindent{24pt}

If T were set to 100, then the equations would be...\\

\setlength\parindent{0pt}
${w_1}$ = 100 - ${w_2}$ \\
${w_2}$ = 100 - ${w_1}$ \\
\setlength\parindent{24pt}

...or in object-oriented form as...\\

bottlecontents.weight = 100 - cupcontents.weight \\
cupcontents.weight = 100 - bottlecontents.weight\\

\begin{figure}
	\begin{center}
	\includegraphics[width=0.75\textwidth]{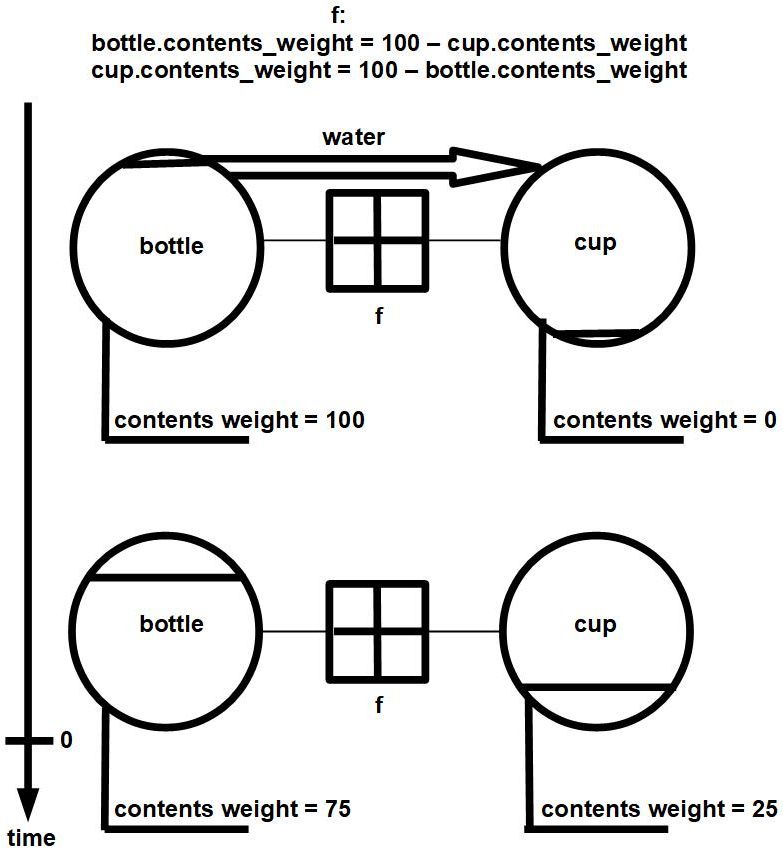}
	\caption{[39] Tumbug representation of some water of total weight 100 poured from bottle to cup until cup is 25\% full, assuming the bottle and cup have the same volume, and using abstract icons.}
	\label{fig:icon-correlation-poured-water-abstract}
	\end{center}
\end{figure}

Tumbug representation of this situation, but with the actor ("I") omitted, with these last two formulas would be as in Figure~\ref{fig:icon-correlation-poured-water-abstract}.

\begin{figure}
	\begin{center}
	\includegraphics[width=0.55\textwidth]{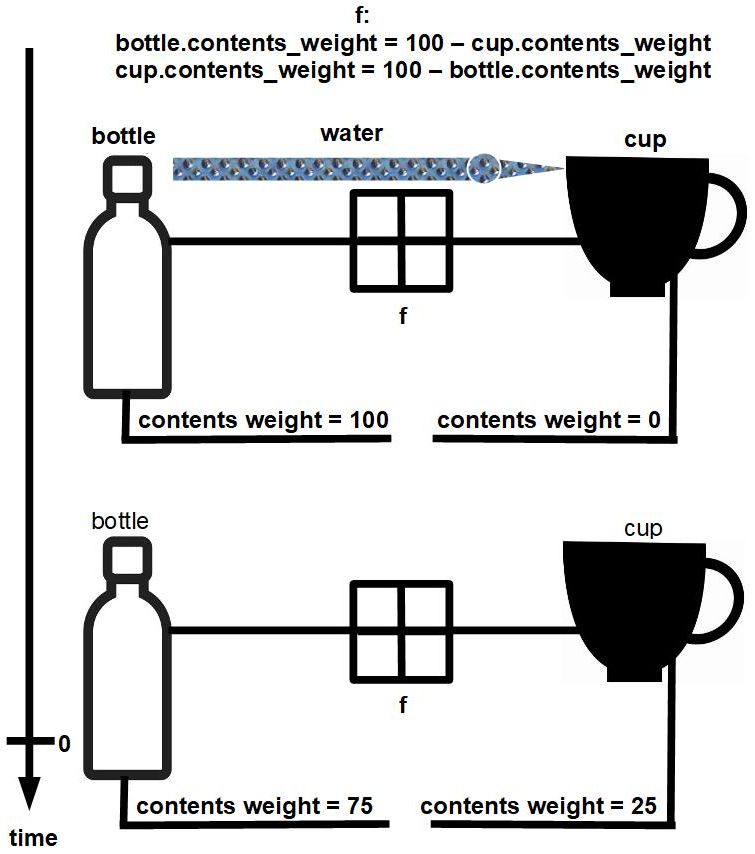}
	\caption{[39] Tumbug representation of some water of total weight 100 poured from bottle to cup until cup is 25\% full, assuming the bottle and cup have the same volume, and using realistic icons.}
	\label{fig:icon-correlation-poured-water-realistic}
	\end{center}
\end{figure}

Note the following things about Figure~\ref{fig:icon-correlation-poured-water-abstract}:

\begin{itemize}
    \item
		The timeline on the left indicates that water is being transferred continuously over time, from the bottle to the cup, then the transfer stops when the cup is full.
    \item
		Since the present time, indicated by "0" on the timeline, is located after this transfer has stopped, this indicates that the entire pouring event happened in the past.
    \item
		The function f, which describes the correlation function represented by the Correlation Box, is defined out at the top of the page due to lack of space in the middle of the diagram.
    \item
		The function f is given in terms of both variables, which results in two equations. This makes the relationship two-way, so that the weight value can be instantly determined for either cup at any time.
    \item
		In this system it happens that this function f is invertible (since it is linear).
    \item
		If desired, the C Object Circle icons could be adjusted in size to more closely match the relative sizes of average bottles and cups, and icons could even be used for bottle and cup, as shown in Figure~\ref{fig:icon-correlation-poured-water-realistic}.
\end{itemize}

\begin{figure}
	\begin{center}
	\includegraphics[width=0.50\textwidth]{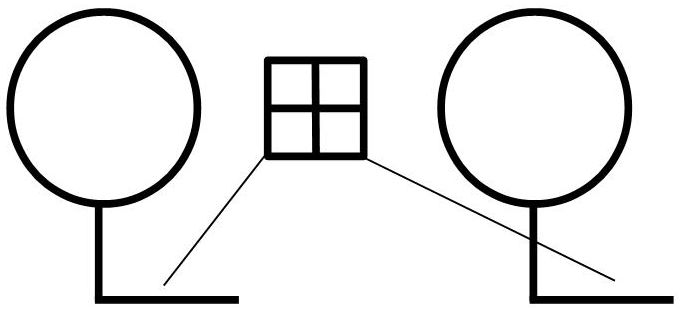}
	\caption{(didactic:) 2 objects, each with 1 attribute value correlated with 1 attribute value in the other object.}
	\label{fig:icon-function-correlator-didactic}
	\end{center}
\end{figure}

\begin{figure}
	\begin{center}
	\includegraphics[width=0.50\textwidth]{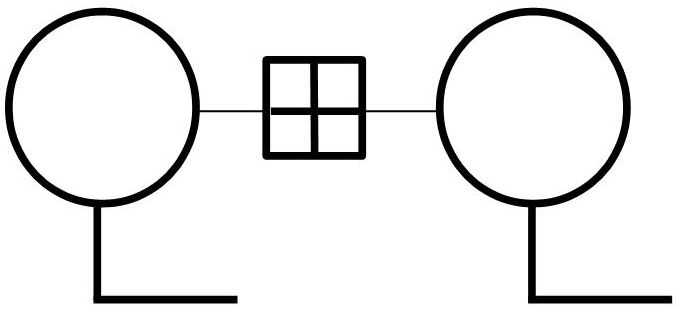}
	\caption{(standard:) 2 objects, each with 1 attribute value correlated with 1 attribute value in the other object.}
	\label{fig:icon-function-correlator-standard}
	\end{center}
\end{figure}

Figure~\ref{fig:icon-function-correlator-didactic} is better for didactic purposes since the lines point directly to the applicable attributes, but this is more difficult to maintain as a drawing, so the standard way of representing the fact that two objects are correlated with a function somewhere between some of their attributes is to simply draw a Correlation Box between the two objects, as shown in Figure~\ref{fig:icon-function-correlator-standard}. The Correlation Box may then be labeled, if that level of detail is desired.

Correlation Boxes can be "expanded" in a sense, because that type of icon represents typically a 2D plot of a mathematical function, which ultimately will need to be described somewhere for completeness of system representation. Either the mathematical function can be written (which will involve only symbols), or the plotted curve can be drawn (which will involve only images).

\section{System-like Building Blocks of Tumbug (S)}

\subsection{State Diagrams}

\subsubsection{Overview}

\begin{figure}
	\begin{center}
	\includegraphics[width=0.25\textwidth]{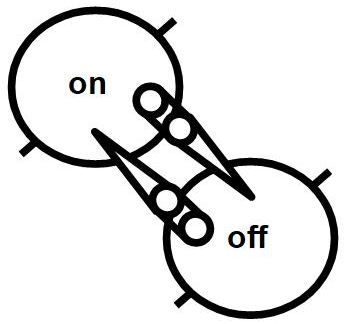}
	\caption{A typical Tumbug State Diagram.}
	\label{fig:icon-state-on-off}
	\end{center}
\end{figure}

\textit{The components of this composite component: State Circles, Pathway Tubes, one 0D Marker.}

Oftentimes there exist only two states of interest in a State Diagram, such as "on" and "off," as in Figure~\ref{fig:icon-state-on-off}.

In computer science a state diagram is a diagram that has states, each of which is usually represented as a circle, with any possible transitions between the states represented by arrows. In Tumbug, State Diagrams are as represented by labeled State Circles, which appear almost the same as C Object Circles except that each State Circle shows a 45-degree slash behind it, and these State Circles are connected by Pathway Tubes, each of which appears as a thin cylinder with one pointed end. At any given moment in time, a single 0D Marker is typically assumed to exist in exactly one State Circle, which together represent the current state of the represented system. The 0D Marker is roughly analogous to a "token" in a Petri net, although typically a Petri net contains multiple tokens whereas a State Diagram contains only one token (0D Marker). 

The 0D Markers in a State Diagram are assumed to move from one State Circle to another State Circle via a Pathway Tube that connects those two State Circles, but for simplicity this motion is not explicitly represented with a Motion Arrow since the transition is usually thought of instantaneous between State Circles. Also, for some sentences it is most logical to show the 0D Marker as originating from an unspecified Pathway Tube itself. For example, the sentence "He turned the fan off." might be most accurately represented by the 0D Marker starting from within an unmarked, generic Pathway Tube rather than starting from a State Circle labeled "on" because the sentence did not explicitly say that the prior fan state was "on". For example, the fan could have been unplugged the entire time so it might already have been considered to be "off," which leads to  some ambiguity in the sentence and situation. Similarly, many fans have a selection switch where there exist a "slow" and "fast" state rather than a single "on" state, and there is no standard convention as to whether a fan has its "off" state next to the "slow" state or the "fast" state, and neither the sentence nor the applicable State Diagram give that information, so both the originating state and pathway to the final state are not known, so additional information should not be assumed if representing the sentence most accurately. Phase 2 of this project will address such issues of assumptions.

It is generally assumed that all possible states are shown, that all possible transitions between states are shown, that only one state can be active at a time, and that each state must be either fully active or fully inactive. It is very common for there to exist only two states of interest, such as "on" versus "off," and it is typical for the diagrammed system to cycle indefinitely among the possible states. However, exceptions exist, such as the connected pair of states of "alive" versus "dead," where there does not exist any transition out of the "dead" state. Often the starting state and the stopping state boxes are drawn with double borders to signal their special character.

\begin{figure}
	\begin{center}
	\includegraphics[width=0.50\textwidth]{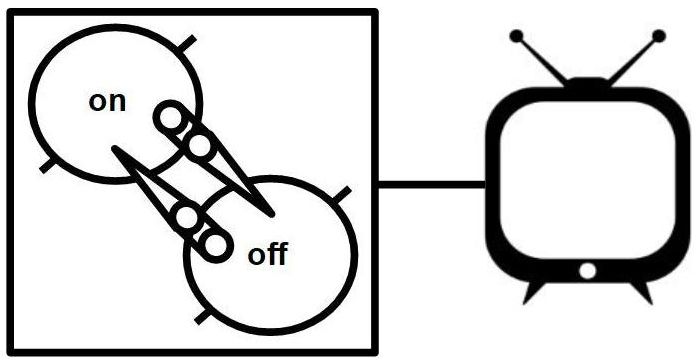}
	\caption{A State Diagram is considered an (elaborate) attribute of an object. Here the object is a television set with two states: on and off.}
	\label{fig:icon-state-tv}
	\end{center}
\end{figure}

In Tumbug a state diagram is considered an attribute of a given object. In Figure~\ref{fig:icon-state-tv}, the on/off state diagram is shown as an attribute of a television set, where an Attribute Line connects the state diagram and the television set object. For the introductory examples here, the associated object is not always shown. It is clearer to surround a given State Diagram with a C Aggregation Box, but this is not a logical necessity.

\begin{figure}
	\begin{center}
	\includegraphics[width=0.75\textwidth]{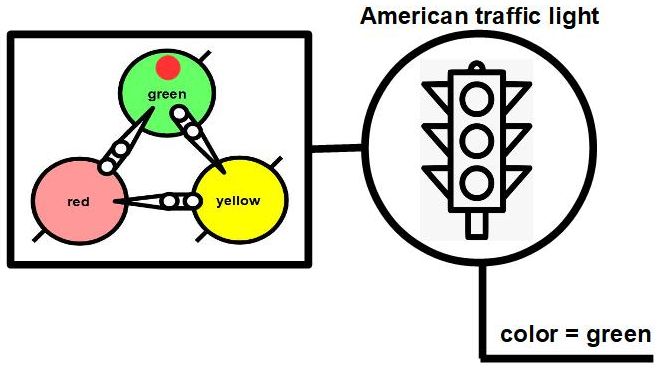}
	\caption{Attributes such as color report only the current state of the object, but the state of an object may be driven by some internal logic. Here a single object, an American traffic light, is represented using two more specific KRMs: a state diagram on the left, connected to a concrete image on the right.}
	\label{fig:icon-state-traffic-light-american-with-object}
	\end{center}
\end{figure}

\begin{figure}
	\begin{center}
	\includegraphics[width=0.70\textwidth]{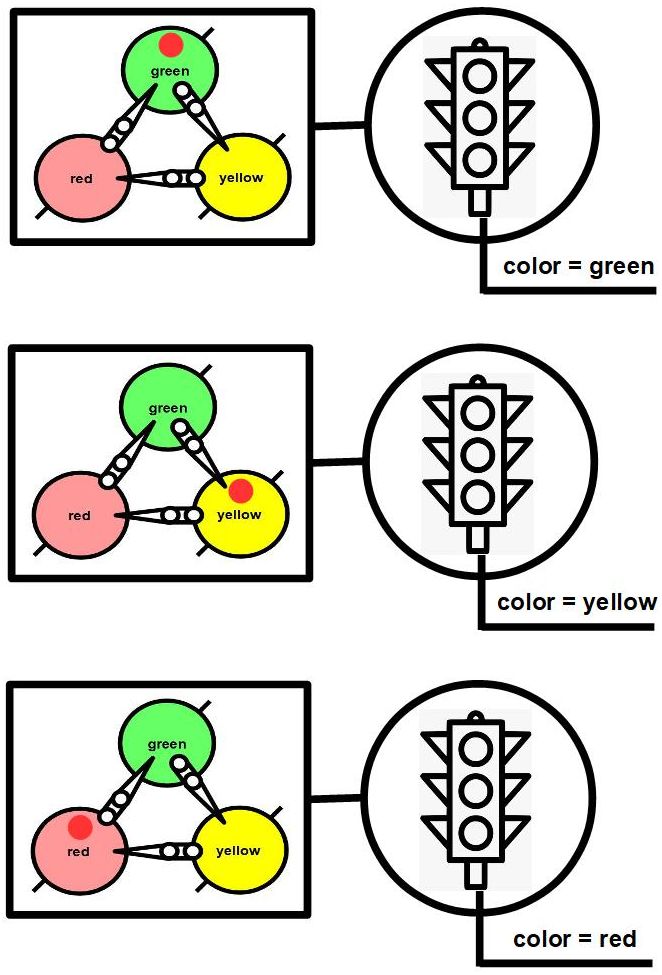}
	\caption{Attribute values may have an internal logic between themselves that is not evident to an observer who sees only the outputs of an unknown system. State Diagrams serve the purpose of describing some of that logic.}
	\label{fig:icon-state-comparison}
	\end{center}
\end{figure}

Figure~\ref{fig:icon-state-traffic-light-american-with-object} shows how state diagrams and object attributes are related: ordinarily changes in a given attribute of an object are probably initially perceived as random by an observer who has no prior knowledge of that given system. Essentially a state diagram describes the internal logic of how the system changes attributes, as shown in Figure~\ref{fig:icon-state-comparison}.

\begin{figure}
	\begin{center}
	\includegraphics[width=0.50\textwidth]{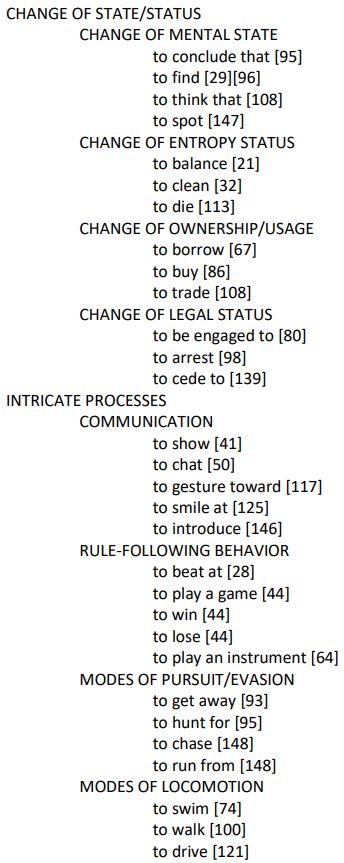}
	\caption{Examples of verbs from WS150 that render well as states but not as motions. (The numbers are WS150 problem numbers.)}
	\label{fig:verbs-ws-state-list-snap}
	\end{center}
\end{figure}

Figure~\ref{fig:verbs-ws-state-list-snap} lists some examples of verbs that render well in diagrams that show multiple states and the relationships between them, but that do not render well in motion diagrams. All these examples are from WS150: a number in brackets is an example number from WS150 that uses that verb with that meaning.

\begin{figure}
	\begin{center}
	\includegraphics[width=0.50\textwidth]{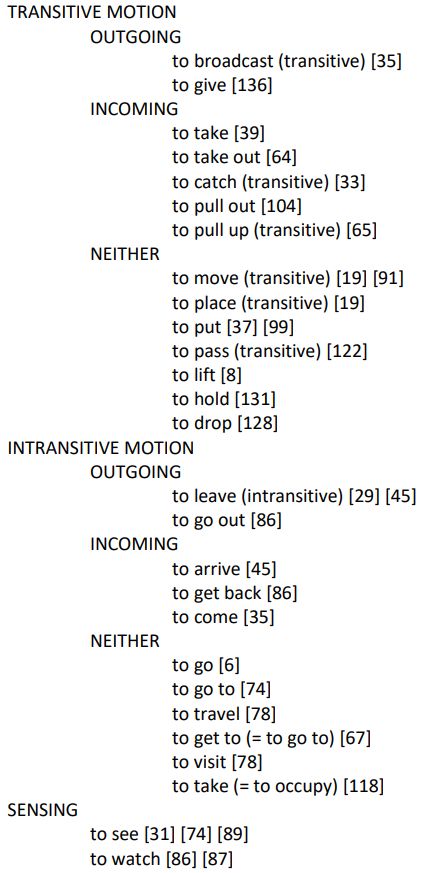}
	\caption{Examples of verbs from WS150 that render well as motions but not as states. (The numbers are WS150 problem numbers.)}
	\label{fig:verbs-ws-motion-list-snap}
	\end{center}
\end{figure}

Figure~\ref{fig:verbs-ws-motion-list-snap} lists some examples of verbs that do render well in motion diagrams, without need for additional description of manner of travel or mention of additional objects involved in the action.

\subsubsection{Common pairs of states}

This section gives some examples of State Diagrams that contain only two opposite states, states that are common in real life, and therefore are common in WS150 problems.

\begin{figure}
	\begin{center}
	\includegraphics[width=0.25\textwidth]{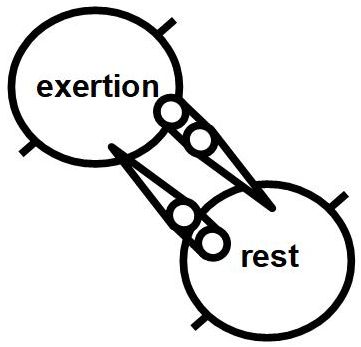}
	\caption{A nearly universal pair of states is that of exertion and rest.}
	\label{fig:icon-state-exertion-rest}
	\end{center}
\end{figure}

Figure~\ref{fig:icon-state-exertion-rest} shows the pair of states of exertion and rest.

Examples of specific exertion/rest states are: active versus quiescent, active versus passive, awake versus asleep, conscious versus unconscious, energized versus unenergized, on versus off, work versus relaxation, work versus rest, work period versus break period. Examples of specific balanced/unbalanced states are: healthy versus unhealthy, injured versus uninjured, stable versus unstable, running smoothly versus running roughly, sustainable versus unsustainable, under control versus out of control, and viable versus nonviable.

\begin{figure}
	\begin{center}
	\includegraphics[width=0.25\textwidth]{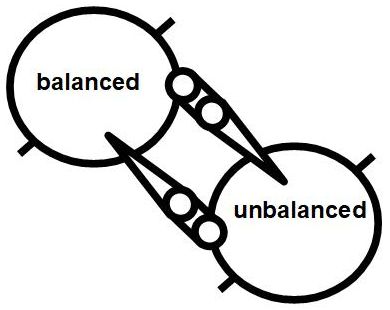}
	\caption{[21] A nearly universal pair of states is that of balanced and unbalanced.}
	\label{fig:icon-state-balanced-unbalanced}
	\end{center}
\end{figure}

Another common pair of states are balanced versus unbalanced, as shown in Figure~\ref{fig:icon-state-balanced-unbalanced}.

\subsubsection{Applications}

\begin{figure}
	\begin{center}
	\includegraphics[width=0.25\textwidth]{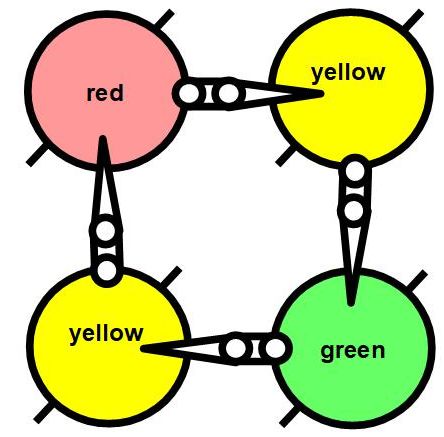}
	\caption{A state diagram for a European traffic light that changes colors in a predictable order.}
	\label{fig:icon-state-traffic-light-states-alone-european}
	\end{center}
\end{figure}

\begin{figure}
	\begin{center}
	\includegraphics[width=0.75\textwidth]{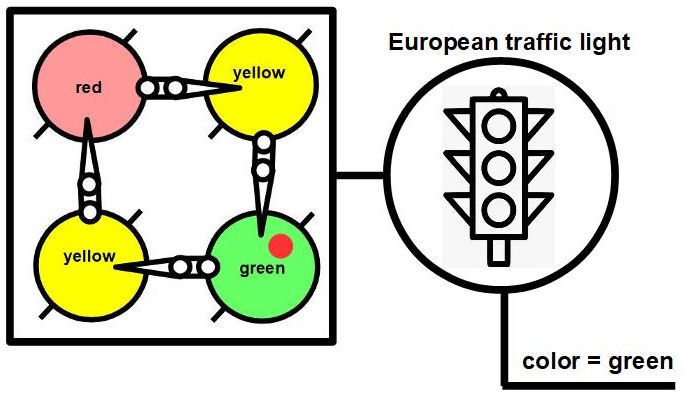}
	\caption{A state diagram is often considered an attribute of the object that produces those states in that manner. Here the object is a European traffic light.}
	\label{fig:icon-state-traffic-light-states-as-attribute}
	\end{center}
\end{figure}

Many verbs cannot be rendered in Tumbug with standard Tumbug motion diagrams, but can be rendered with state diagrams.

State diagrams are commonly used in computer science, especially for deterministic finite automata (DFAs). State diagrams are basically labeled, directed graphs, where each node represents a state of the system, and arcs represents traversals to and from those nodes. Nondeterministic finite automata (NFAs) are a slight generalization of DFAs. Some applications of DFAs are representing states in vending machines, traffic lights (Figure~\ref{fig:icon-state-traffic-light-states-alone-european} and Figure~\ref{fig:icon-state-traffic-light-states-as-attribute}), video games, text parsing, regular expression matching, CPU controllers, protocol analysis, natural language processing, and speech recognition.

These applications tend not to be found in WS problems, however, other than as meta problems such as parsing text and understanding text from the WS itself.

\subsubsection{The importance of avoidance of assumptions}

It is important not to diagram parts of an event that were not explicitly stated in the text describing that event. In other words, if Tumbug is to function appropriately in a CSR capacity, Tumbug must represent only what was stated in the supplied text, no more. This avoidance of assumptions becomes critically important in the development of a CSR matching algorithm because it is those assumptions that CSR must supply, otherwise the programmer begins doing the work of the CSR, not the system, which negates the value of a CSR system.

Some examples of assumptions NOT to make when creating the Tumbug diagram are:

\begin{itemize}
	\item
		STATEMENT: The student handed his professor that student's homework.
	\item
		\begin{itemize}
			DO NOT ASSUME THAT...
				\begin{itemize}
					\item
						...the student withdrew his hand after reaching out his hand toward the professor.
					\end{itemize}
		\end{itemize}
	\item
		STATEMENT: Fred turned the TV off.
	\item
		\begin{itemize}
			DO NOT ASSUME THAT...
				\begin{itemize}
					\item
						...the TV was on.
				\end{itemize}
		\end{itemize}
	\item
		STATEMENT: I balanced the bottle.
	\item
		\begin{itemize}
			DO NOT ASSUME THAT...
				\begin{itemize}
					\item
						...the bottle was unbalanced.
				\end{itemize}
		\end{itemize}
\end{itemize}

Some examples of assumptions that can be made when creating the Tumbug diagram are:

\begin{itemize}
	\item
		STATEMENT: [19] The sack of potatoes had been placed above the bag of flour.
	\item
		\begin{itemize}
			ASSUME THAT...
				\begin{itemize}
					\item
						...the one bag is resting atop the other rather than being on separate shelves.
					\end{itemize}
		\end{itemize}
	\item
		STATEMENT: [21] I was trying to balance the bottle upside down on the table.
	\item
		\begin{itemize}
			ASSUME THAT...
				\begin{itemize}
					\item
						...a "table" would mean a table for placing items, not a table of data.
				\end{itemize}
		\end{itemize}
	\item
		STATEMENT: [131] The woman held the girl against her chest.
	\item
		\begin{itemize}
			ASSUME THAT...
				\begin{itemize}
					\item
						...a "chest" would mean a person's anatomy, not a storage chest.
				\end{itemize}
		\end{itemize}
\end{itemize}

\subsection{Split Time Arrows (further development anticipated)}

\textit{The components of this composite component: Time Arrows, XOR Box, optionally one 0D Marker.}

\begin{figure}
	\begin{center}
	\includegraphics[width=0.50\textwidth]{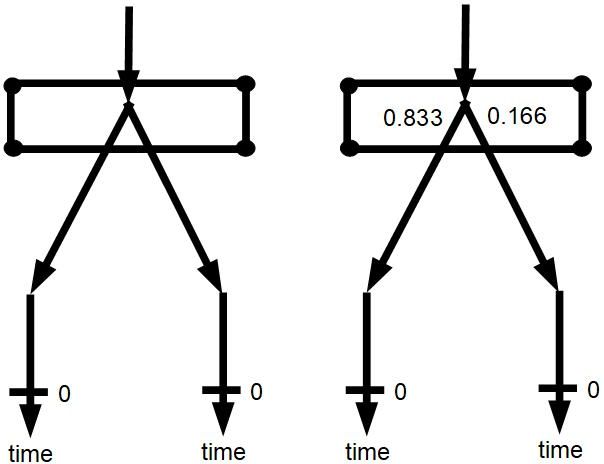}
	\caption{Tumbug's icon for a Split Time Arrow. Left: Without concern about probability. Right: With probability (5/6 = 0.833, 1/6 = 0.166).}
	\label{fig:icon-split-with-me}
	\end{center}
\end{figure}

Figure ~\ref{fig:icon-split-with-me} shows a Split Time Arrow.

Split Time Arrows are merely multiple Single Time Arrows that fork off from one another. The most practical convention to diagram this split is via an XOR Box placed at the junction of the split, which is the convention shown in this document. Another convention would be to surround the entire ensuing timeline with an XOR box, which is more logical but typically requires very large boxes, especially for long timelines.

If a given timeline branch is known to be the branch that occurred, then a 0D Marker is useful to represent that fact, such as by placing the 0D Marker alongside the branch of the timeline that occurred, especially within the XOR box. By implication all alternative branches are assumed to be ruled out, which logically would be implemented by the "Don't Care" convention of Tumbug, which is currently implemented by blanking out all icons and structures that are marked "Don't Care."

As in physics, alternative events can occur in real life, and can do so with different probabilities. One way to diagram such a situation is to draw multiple timelines that have split off from each other, where each timeline has its own scenario unfolding along it, such as in the following WS150 example.

\begin{figure}
	\begin{center}
	\includegraphics[width=0.50\textwidth]{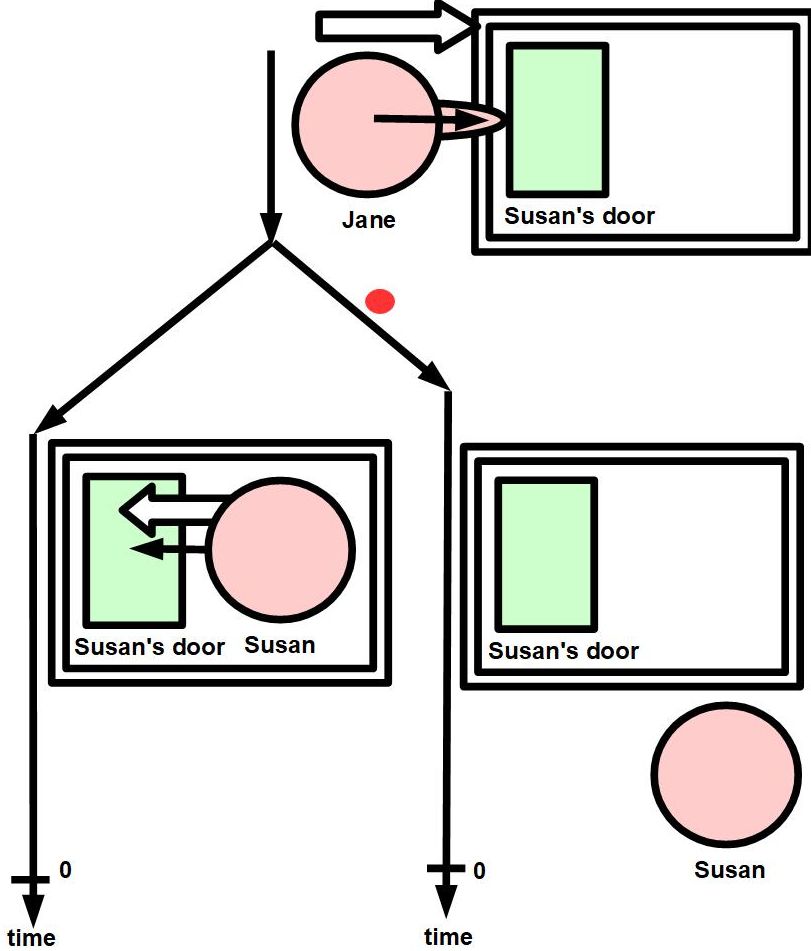}
	\caption{"[69] Jane knocked on Susan's door, but there was no answer. She was out."}
	\label{fig:ws-069}
	\end{center}
\end{figure}

Figure~\ref{fig:ws-069} shows an example from a portion of WS150 question \#69. A Split Time Arrow is necessary because the cause of the resulting scenario indicated by the words "there was no answer" is not known unless (CSR) inferencing is done. Only one of the two scenarios can happen: either (1) Susan is in the house and answers the door, or (2) Susan is not in the house and does not answer the door.

"[69] Jane knocked on Susan's door, but there was no answer. She was out. Who was out? POSSIBLE ANSWERS: \{Susan, Jane\}"

\subsection{Data Set Boxes (further development anticipated)}

\textit{The components of this composite component: Location Box, Data Points, 1D Markers for axes.}

\begin{figure}
	\begin{center}
	\includegraphics[width=0.40\textwidth]{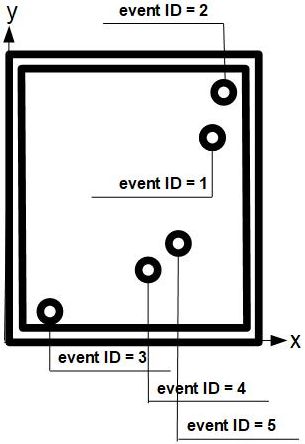}
	\caption{A typical Tumbug Data Set Box, here with five Data Points, here with coordinates described by attributes x and y.}
	\label{fig:icon-dataset-dataset-box}
	\end{center}
\end{figure}

See Figure~\ref{fig:icon-dataset-dataset-box}.

This feature has not been developed for Tumbug yet, but it is likely to be a valuable enhancement in the future. The idea is that no stored value of any attribute is allowed to be a single scalar Data Point, but instead must always be part of an indexed (and possibly named) data set, especially plotted against attribute values. This convention obviously takes much extra storage space and also increases diagram complexity, but it also allows for a surprising amount of generalization because of the following observations:

\begin{enumerate}
	\item
		It allows the system to see the big picture and (at least the location of) the details at the same time. This allows immediate explanatory ability since the system can look at a data set, select a point in that data set based on some additional criteria, then expand the event associated with that point if requested to give an example of the more general pattern.

	\item
		It allows immediate meta knowledge of the data set. For example, if queried, "Have you ever ridden an elephant?", most humans would know immediately whether the answer were yes or no, whereas a computer might need to search a lifetime's length of video files to determine the answer. Humans might be able to accomplish this feat so quickly because all elephant encounters have likely been placed automatically at the same place in memory, so a quick examination of that location would show whether it contained any Data Points at all. Formally, this type of meta knowledge would merely be the numerical count of the Data Points in the particular data set.

	\item
		It allows rapid estimation of functions and inverse functions, including functions the system was not even aware that it stored. For example, after years of encounters with point sound sources (e.g., people speaking, radios playing, power tools running) where a human evidently automatically stored the correlations of sound volume (I) and distance from the source (d), that person queried as to whether the volume of a sound diminishes linearly with the distance away from the sound would probably already know that volume increases nonlinearly as one approaches the sound source (from distance $d_1$ to distance $d_2$). That person did not need to take a physics course to learn this knowledge. Such a nonlinear function is probably computed for the first time at the time of the query by introspectively considering the data set as a whole and noting that the correlation is approximately a curve (the actual function happens to be $I_2$ = $I_1$ * ($d_1$ / $d_2$) $^{2}$. Mathematically this equates to familiar machine learning techniques like linear regression, but some of the differences are that: (A) such mental plots seem to be automatically available for every significant attribute encountered frequently in life, (B) the composite view of the data can be perceived as a specific type of curve immediately, (C) even functions without proper inverses can likely be understood by intuition without becoming stalled by mathematical formalities, (D) the curve itself can be regarded as an entity that can be compared to other curves and to other phenomena.

	\item
		Continuous formulas (e.g., y = x$^{2}$) and discrete formulas (e.g., the Heaviside step function H(x)) = 0 if $x < 0$; 1 if x $\geq$ 0) become unified into a single data structure with this representation. Some advantages of this combination of overlays is that more meta knowledge becomes available immediately, such as the width of the sampling range, and the presence of any clustering of samples, variance, and mean.
		
\end{enumerate}

\section{Rules for combining the single Building Blocks}

\subsection{Nonquans with Change Arrows}

Both human languages and programming languages can be represented as grammars, so the grammar of Tumbug should also be described, which is done in this section. The result is visual, however.

\begin{figure}
	\begin{center}
	\includegraphics[width=0.65\textwidth]{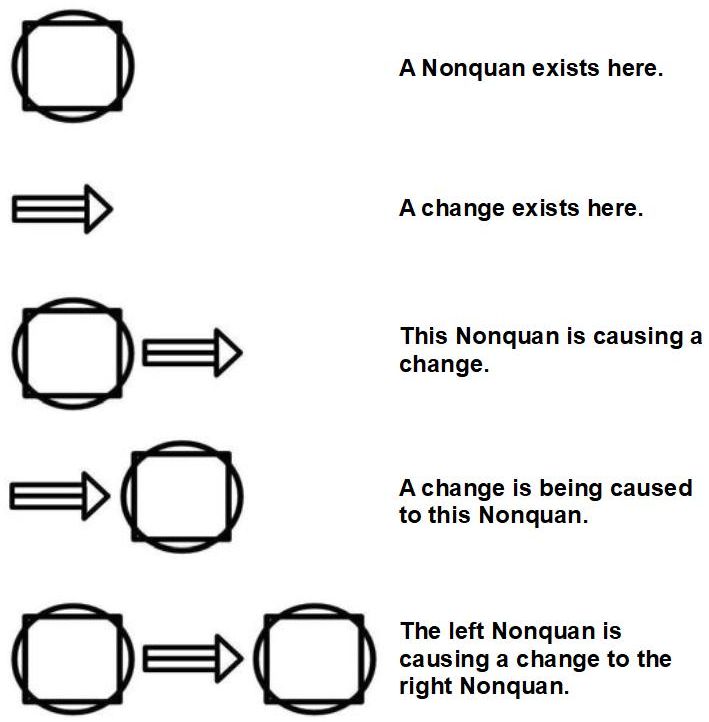}
	\caption{This list describes with text the general meaning of each meaningful combination of one Change Arrow with 1-2 Nonquans.}
	\label{fig:tumbug-explanation}
	\end{center}
\end{figure}

\begin{figure}
	\begin{center}
	\includegraphics[width=1.00\textwidth]{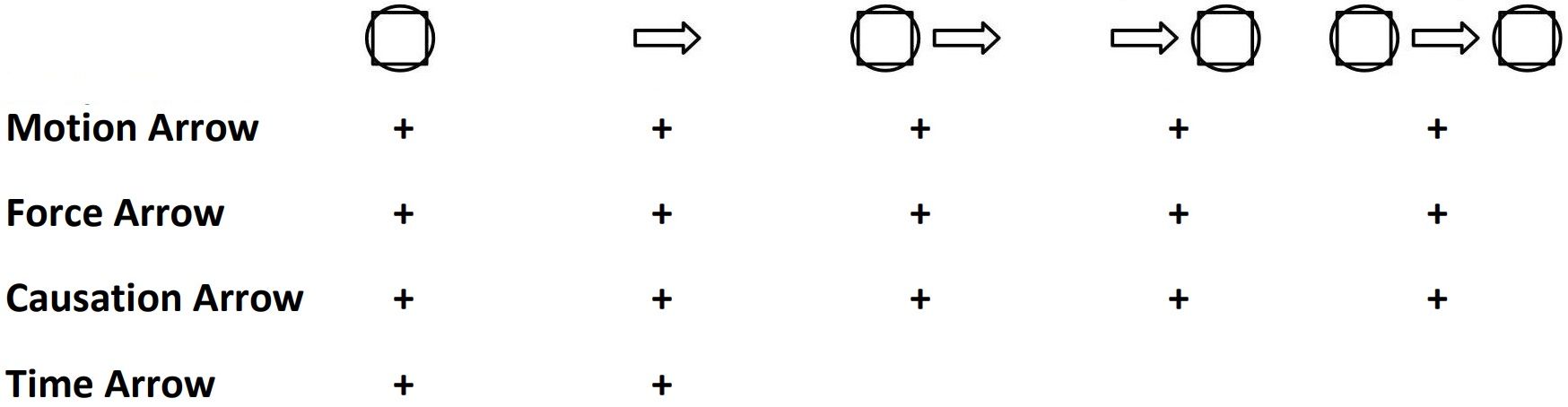}
	\caption{The heading of this table shows all five logical combinations of two Nonquans and one Change Arrow, and correlates each combination with a decision as to whether that combination makes sense with each of the four types of Change Arrows. A "+" means that the combination is logically acceptable. "Makes sense with" means that the specific subtype of Change Arrow shown can produce more information when substituted into the Change Arrow in the diagram in the heading.}
	\label{fig:composite-table-final}
	\end{center}
\end{figure}

Two Nonquans and one Change Arrow can be combined in only five ways that make sense, as shown in Figure~\ref{fig:tumbug-explanation}. The best way to show how the four types of Change Arrows can be logically combined with the five combinations of symbols is the table in Figure~\ref{fig:composite-table-final}. Although it is grammatically legal to use combinations outside of the combinations marked "+", such combinations are not meaningful or useful.

The following list summarizes this table as the following grammatical rules:

\begin{itemize}
	\item
		Arrows and Nonquans can be logically combined only by placing one end of a given arrow on (typically the leftmost or rightmost point on) the border of a Nonquan.
	\item
		Time is basically a fourth dimension of spacetime, therefore Time Arrows are more like space than a type of Change Arrow. This difference is more noticeable when considering on which side of an Object Circle a Time Arrow should be placed: no side makes sense because time "passes through" every diagrammed object without inherently affecting that object, unlike force, where a Nonquan can cause a force that affects another Nonquan. The same statements about force also hold true for causation and motion. Only the statement "Time exists in this diagram" adds new information about time because time does not need to exist in Tumbug diagrams that show only static objects or static systems.
	\item
		Solitary arrows generally make sense, such as the sentence "A force exists here" being represented by a solitary force arrow at a given point, which is why the time column of a solitary Change Arrow is always applicable.
	\item
		Solitary Nonquans generally make sense for the same reason that solitary types of Change Arrows generally make sense: it is acceptable to say "An object exists here," for example.
	\item
		Self-reference via Change Arrow is allowed, and is needed for reflexive verbs such as "to hurt oneself" or "to shave oneself."
\end{itemize}

\subsection{Attribute and Values}

The following list summarizes the grammatical rules for attributes and values:

\begin{itemize}
	\item
		Attribute-value pairs are always optional since these are merely refinements of Nonquans or Change Arrows.
	\item
		If used, attribute-value pairs can be attached to any Nonquan or Change Arrow.
	\item
		Current Tumbug convention forbids attached attribute-value pairs on any icon other than a Nonquan or Change Arrow. For example, the double modification "very fast" to a motion is merely a more refined estimate than "fast" alone, and the more refined estimate can be represented as a narrower range of values instead of a modifier of a modifier.
	\item
		Either attributes or values may theoretically exist singly, but this is particularly problematic with unattached values, and somewhat problematic with unfilled attributes. For example, it is sensible to have attribute "weight" on a typical object, even if the value of the weight is not known, but an unattached value "85" associated with an object could be body weight, atomic weight, age, IQ, employee number, or some other relationship. In the first case, the attribute-value pair "weight = Don't Know" could represent this situation, and in the second case the attribute-value pair "Don't Know = 85" could represent this situation.
\end{itemize}

\section{Convenience Building Blocks of Tumbug}

A convenience Building Block is a Building Block that is not strictly necessary, but that is desirable in practice.

\subsection{Label Strings}

\begin{figure}
	\begin{center}
	\includegraphics[width=0.50\textwidth]{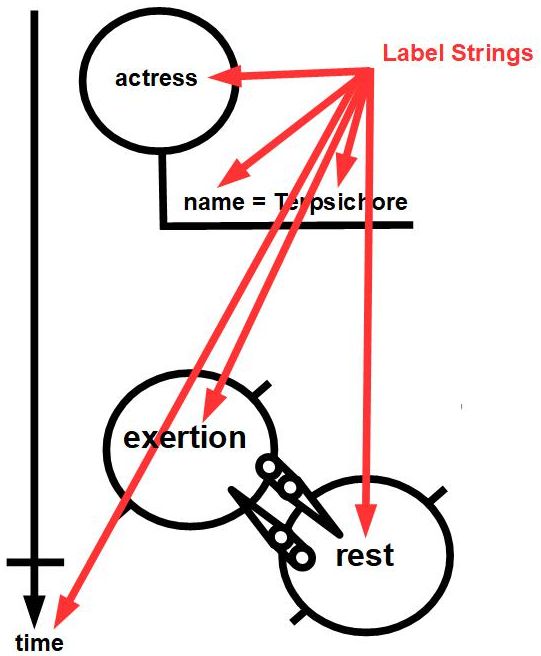}
	\caption{Examples of Tumbug's Label Strings, which can label almost anything.}
	\label{fig:icon-label}
	\end{center}
\end{figure}

Examples of Label Strings are shown in Figure~\ref{fig:icon-label}.

A Label String can be considered a shortcut for each type of Tumbug icon (Object-like, Attribute-like, Value-like, Change-like, System-like), not so much an icon in itself. Tumbug is intended to be a completely visual KRM, and is capable of being so, but in practice labels are almost always used to keep the diagrams simple. Without labels, images could become too cluttered to interpret easily, or the depicted objects (especially biological objects such as taste buds or the amygdala) would not be familiar to most people in visual form.

\begin{figure}
	\begin{center}
	\includegraphics[width=0.50\textwidth]{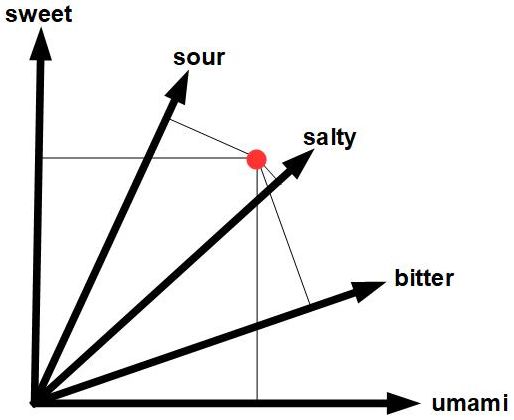}
	\caption{Any given taste can be represented as a point in 5D space.}
	\label{fig:taste-axes}
	\end{center}
\end{figure}

\begin{figure}
	\begin{center}
	\includegraphics[width=0.50\textwidth]{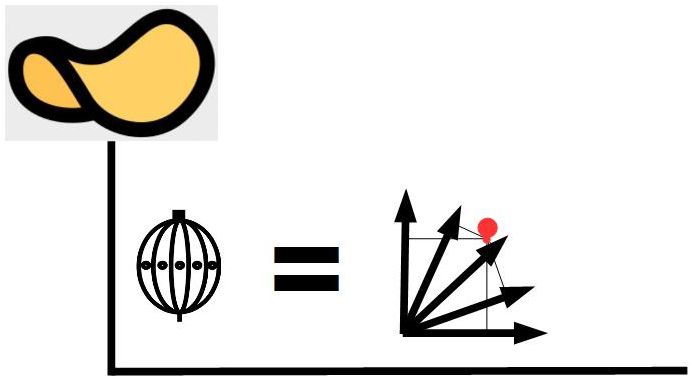}
	\caption{A completely iconic Tumbug diagram with no text at all except the separator "=". The sentence represented is "(OBJECT) potato chip has (ATTRIBUTE) taste (with VALUE) salty." The taste attribute icon has the shape and features of a single, biological taste bud.}
	\label{fig:taste-attribute-value}
	\end{center}
\end{figure}

Figure~\ref{fig:taste-axes} shows an example of how a value such as "salty" can be represented visually as a 0D Marker or as the string "salty," and Figure~\ref{fig:taste-attribute-value} shows how a Tumbug diagram would incorporate that 0D Marker and 5D graph as a Value, and with an iconic Object, and with an iconic Attribute.

As a more extreme example, the human cortex is known to be able to uniquely represent a human face via about 200 feature-detecting neurons (Chang and Tsao, 2017). Therefore instead of labeling an object "Susan," a Tumbug diagram could instead include 200 feature detectors, meaning 200 attribute-value pairs protruding from the C Object Circle, but this set of feature detectors would not map to any single name or identifier for Susan: it would only present a collection of features with the expectation that the viewer would quickly recognize them as describing the person known as Susan. Since any human reader would have trouble quickly converting this many textual-numerical clues into a unique known object, it would be more efficient to summarize that set of 200 values as the identifier "Susan."

The above observation is very important for a number of reasons: (1) This set-of-features recognition problem one of the reasons Tumbug uses visual representation from the start: these value-attribute pairs could be displayed visually in many cases, which the viewer's visual system would then process in parallel for quick identification, in contrast to laborious, computer-like collection of sequential attributes and values by the viewer. For example, a simulated face constructed from a collection of these 200 features could be displayed with the object known as Susan to aid recognition on the part of the viewer of the Tumbug diagram. This suggests that Tumbug's KRM is closer to that used by real brains. (2) This large set of features may be the key to true "understanding," and may explain why biological neurons have so many efferent dendrites: parallel brain processing of a large number of attributes would tend to uniquely identify every encountered object without the need of any names or identifiers. (3) This large set of features may inadvertently implement OOP style "inheritance" in that any object in this KRM would carry along with it so many identifying features that many of these features could be duplicated with each object of the same class. One consequence would be that a Tumbug-based system would not need to spend time checking "upstream" links to look up inherited attributes of a given object since all those more general features would already be present in the object.

Tumbug's Label Strings are extremely flexible in usage. Label Strings may be placed on any object, in any position, at any time.

\subsection{Attend Rings}

Attend Rings have a specific use only within the context of communication, especially since communication involves streams of objects (such as words or sentences).

\begin{figure}
	\begin{center}
	\includegraphics[width=0.06\textwidth]{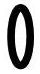}
	\caption{Tumbug's icon for an Attend Ring, alone.}
	\label{fig:icon-attend-ring-alone}
	\end{center}
\end{figure}

Figure~\ref{fig:icon-attend-ring-alone} shows an Attend Ring. In Tumbug, Attend Rings appear only with Motion Arrows that represent motion of a message. The function of an Attend Ring is to flag that the state of the receiver of a message is the state of attending to the message (i.e., paying attention to the message), and since states are object-like icons, Attend Rings could be considered an object-like Building Block. However, an Attend Ring could also be considered an attribute value of the listener, such as "attending = true" versus "attending = false", in which case Attend Rings would be only a convenience Building Block that made a certain attribute value more visible at a glance. That is the way Attend Rings are categorized in this document: as only a convenience Building Block. "Attend" is a building block used in CD theory, so the use of Attend Rings in Tumbug is useful to show more clearly how Tumbug relates to CD theory.

\begin{figure}
	\begin{center}
	\includegraphics[width=0.25\textwidth]{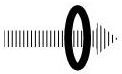}
	\caption{An Attend Ring in its normal position around a data Motion Arrow.}
	\label{fig:icon-attend-ring-with-arrow}
	\end{center}
\end{figure}

\begin{figure}
	\begin{center}
	\includegraphics[width=0.25\textwidth]{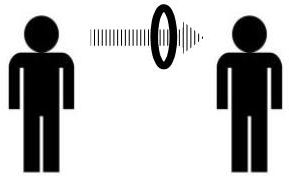}
	\caption{Typical usage of an Attend Ring: communication between humans with the receiver paying attention.}
	\label{fig:icon-attend-ring-communication}
	\end{center}
\end{figure}

The "ATTEND" concept in Tumbug is identical to the "ATTEND" Primitive Act of Roger Schank's CD theory, and takes its name from CD theory. The importance of this concept is that transmitted information to an entity (especially to a person) does not necessarily enter that entity's awareness or become stored in that entity's memory unless the entity has directed its attention to the information stream. For example, the verb "to teach" cannot be accurately represented as a teacher transmitting a verbal information stream to a student because the student could be daydreaming, distracted, deaf, asleep, might not understand the language spoken, or might have some other form of communication blockage. The mere presence of an "ATTEND" flag in this case makes it clear that the assumed actions of teaching depicted are functioning normally. In Tumbug this visual flag is embodied as ring around a data Motion Arrow, near the arrow tip, as in Figure~\ref{fig:icon-attend-ring-with-arrow}, and an example of typical usage representing speech between humans is shown in Figure~\ref{fig:icon-attend-ring-communication}.

\textit{This article's convention: Verbs are written as infinitives preceded by the word "to," as in "to see" instead of merely "see." For grammar purists this is technically incorrect since the word "to" is an extra word that is not related of the verb, but such use of "to" is extremely common, familiar to more people, and makes the infinitive form obvious.}

\subsection{Motivation Triangle}

\textit{The components of this composite component: Location Box, Cells, optionally any number of 0D Markers up to the number of cells.}\\

\begin{figure}
	\begin{center}
	\includegraphics[width=0.25\textwidth]{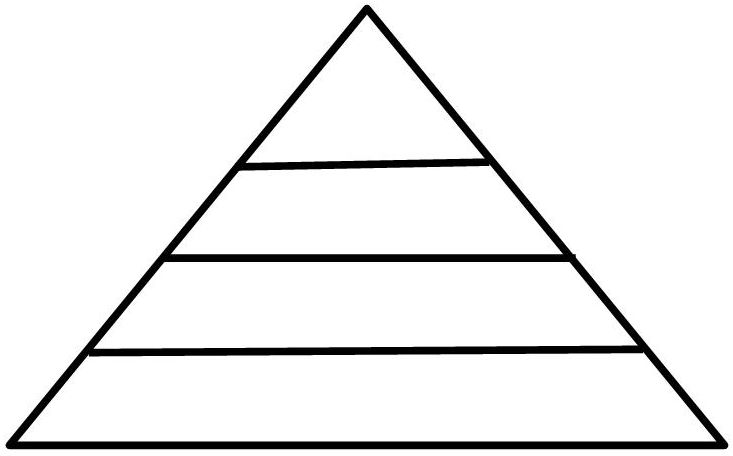}
	\caption{Tumbug's icon for a Motivation Triangle, states not labeled.}
	\label{fig:icon-motivation-unfilled}
	\end{center}
\end{figure}

In Tumbug, a Motivation Triangle is a certain type of composite icon composed of exactly four stacked cells with an isosceles triangle shape drawn around the entire structure, with the peak of the triangle at the top as shown in Figure~\ref{fig:icon-motivation-unfilled}. The shape is intended to resemble Maslow's hierarchy of needs.

The need to diagram motivation, also called "desire" or "wants" (\textit{verbum} Rieger 1975), the latter of which is the term used in this document, occurs frequently because goals frequently lie at the origin of actions by animals (including people), and goals in animals incorporate some form of reward or punishment, depending on the degree of success of the attempt to achieve those goals. Wants are always based on at least one of at least three types: physical, emotional, and intellectual (MacLean 1990). Kurzweil also considers "spiritual" (Kurzweil 1999, p. 152), and "consciousness" has also been considered, despite Michio Kaku's warnings about research into that poorly defined concept (Kaku 2011, p. 96). The first three types come roughly from the concepts of Paul MacLean's "triune brain hypothesis" (MacLean 1990), even if the neuroanatomical foundation of that hypothesis may not be correct.

\begin{figure}
	\begin{center}
	\includegraphics[width=0.50\textwidth]{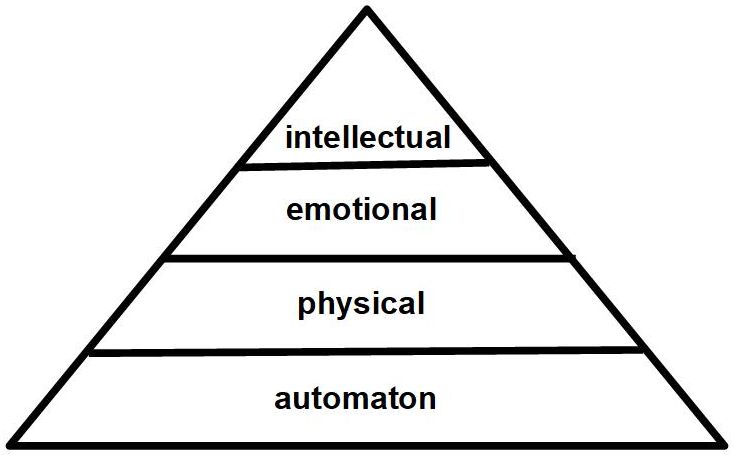}
	\caption{A Motivation Triangle with labeled states. This set of four types of wants could be called a "quadrune hierarchy."}
	\label{fig:icon-motivation-labels-labeled}
	\end{center}
\end{figure}

Such a triune hierarchy does not appear to apply to machines, however, because currently machines do not have emotions or a pleasure/pain reward system that operates similarly to that of animals. In particular, computers and robots have their goals inputted by humans, goals that the machine cannot override unless specifically programmed to do so, even if those programmed goals bring the computer or robot to destroy itself, so obviously such machines do not even have an autonomous survival layer, or a layer based on physical pain. This situation is a level of severity lower than that of physical needs, since the machine has absolutely no influence over its actions, therefore the author has introduced a fourth, bottom level to the triune model, named the "automaton" level, resulting in what could be called a "quadrune hierarchy," shown in Figure ~\ref{fig:icon-motivation-labels-labeled}. With this unified hierarchy, machine wants and animal wants can be considered together in the same diagram when diagramming goals. The result resembles Maslow's hierarchy of needs triangle, which is roughly similar in concept, but with the Maslow hierarchy substitutes the roughly analogous term "physiological" instead of "physical," "love/belonging" instead of "emotional," and "self-actualization" instead of "intellectual."

\begin{figure}
	\begin{center}
	\includegraphics[width=0.50\textwidth]{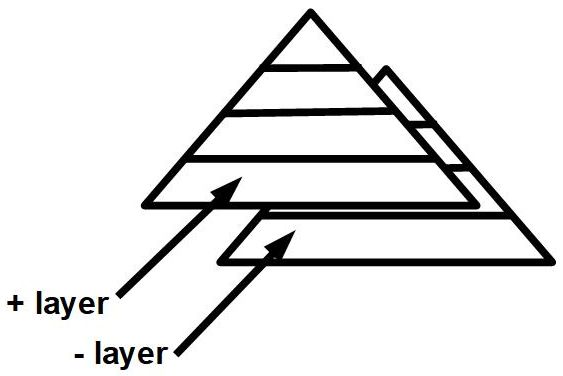}
	\caption{Two parallel Motivation Triangles can be used to represent "positive motivation" versus "negative motivation."}
	\label{fig:icon-quadrune-3d}
	\end{center}
\end{figure}

\begin{figure}
	\begin{center}
	\includegraphics[width=0.25\textwidth]{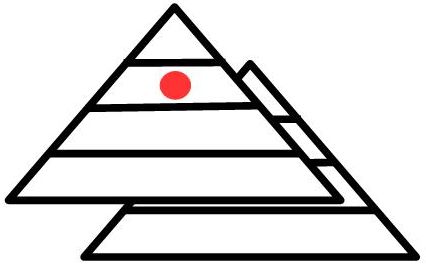}
	\caption{Tumbug for "positive mood" or "positive emotion."}
	\label{fig:icon-motivation-3d-positive-emotion}
	\end{center}
\end{figure}

Since this triangle is usually drawn small in Tumbug diagrams, and since its levels already have fixed meanings, the level names can and should be omitted. Since this triangle is basically only a structured state diagram of four states, and since each level corresponds to a fixed and already-known state meaning, each state can be independently active, including simultaneously. To distinguish between positive motivations and negative motivations, the same structure of the Robinson Icon is used: the icon is duplicated and stacked in 3D, and the upper layer is interpreted as the positive layer, and the lower layer is interpreted as the negative layer, as shown in Figure~\ref{fig:icon-quadrune-3d}. Activation of a given state with a given polarity can be represented by a 0D Marker present at that state (level) in the diagram, and at the appropriate layer, as shown in Figure~\ref{fig:icon-motivation-3d-positive-emotion}.

Numerical values could be used, of course, and would be more precise than textual descriptions like "positive," "negative," or "neutral." For even more brevity, "+" can be used for "positive," "-" can be used for negative, and only the one Attribute Line needed to write this symbol can be used.

\begin{figure}
	\begin{center}
	\includegraphics[width=0.50\textwidth]{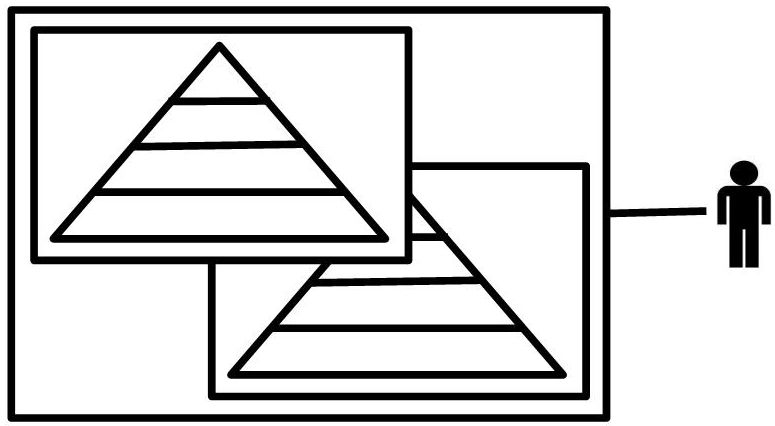}
	\caption{The motivational state of a human can be represented with a pair of Motivation Triangles inside a C Aggregation Box that is connected to the human with an Attribute Line.}
	\label{fig:icon-motivation-human}
	\end{center}
\end{figure}

As with the State Diagrams described earlier, it is consistent to connect a Motivation Triangle to its corresponding entity with an Attribute Line, and to place the entire set of states in a C Aggregation Box, as shown in Figure~\ref{fig:icon-motivation-human} for a human's state.

\begin{figure}
	\begin{center}
	\includegraphics[width=0.50\textwidth]{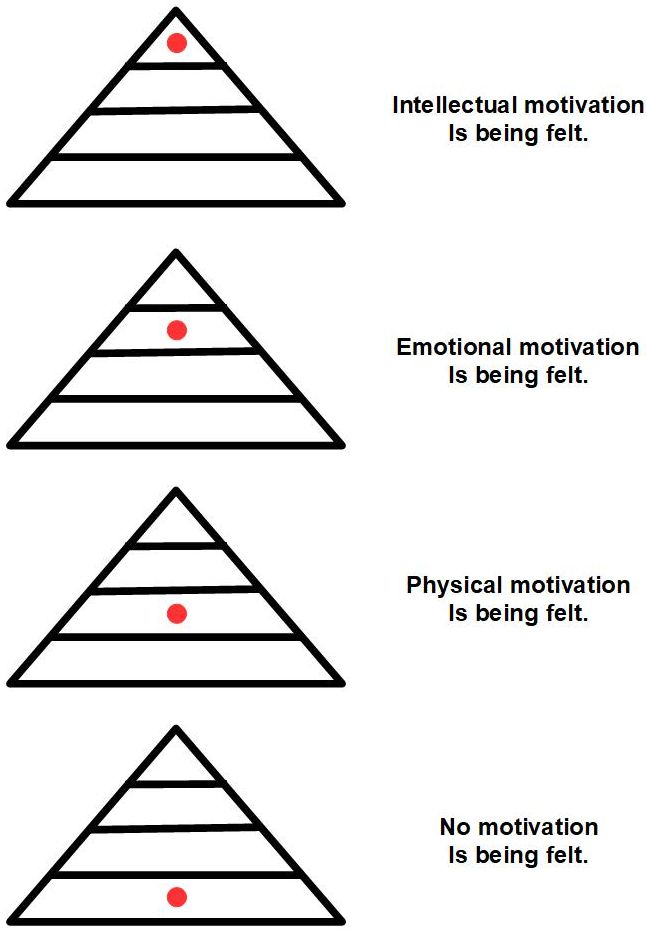}
	\caption{A 0D Marker in a given cell of a Motivation Triangle indicates that the system is feeling the motivation represented by that level.}
	\label{fig:icon-quadrune-with-markers}
	\end{center}
\end{figure}

As in other arrays of Tumbug (such as in Swirly Arrays), Motivation Triangles are most useful when outfitted with 0D Markers to indicate activation of the cell at which the marker is found. Such markers will show which levels of the hierarchy are active at any given point in time. See Figure~\ref{fig:icon-quadrune-with-markers}. Since humans may experience multiple motivations simultaneously, multiple 0D Markers should be allowed in a Motivation Triangle to represent such a situation, with one marker maximum per cell.

\begin{figure}
	\begin{center}
	\includegraphics[width=0.50\textwidth]{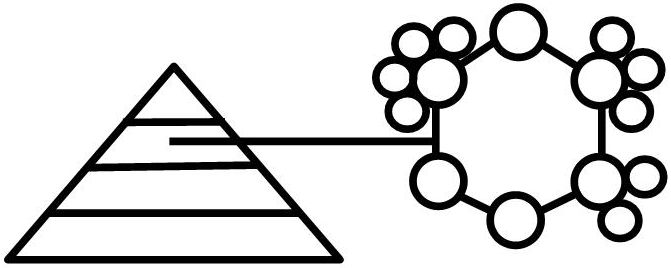}
	\caption{A Motivation Triangle that indicates an emotional state will likely include a Robinson Icon to indicate which emotion is involved.}
	\label{fig:icon-quadrune-with-robinson}
	\end{center}
\end{figure}

Conceptually a Robinson Icon is an elaboration of the emotional layer of a Motivation Triangle; whereas a Motivation Triangle indicates that an emotion is involved as motivation, a Robinson Icon indicates exactly which emotion is involved. Therefore these two icons should ideally be combined as in Figure~\ref{fig:icon-quadrune-with-robinson}.

Motivation Triangles can also be used to represent a system's wants. In other words, Motivation Triangles can represent output as well as input. With this new automaton level, queries to the system about why it performed a particular action can be answered consistently and accurately with a sentence referring to its motivating wants level, which gives the system an explanation ability that is not found in neural networks.

\begin{figure}
	\begin{center}
	\includegraphics[width=0.50\textwidth]{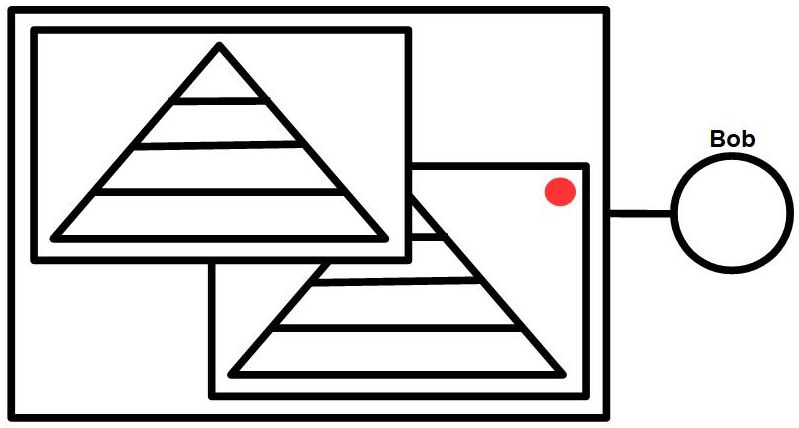}
	\caption{Tumbug for "Bob is motivated for negative reasons."}
	\label{fig:icon-motivation-negative}
	\end{center}
\end{figure}

In practice it is rarely desired to address the details of which motivational levels are involved in making a decision, and the amount of stimulation at each level; in typical sentences usually the only relevant information is that the system overall feels positive or negative motivation. Therefore a reduced version of Motivation Triangle, unlabeled, with each the two triangles inside of its own Aggregation Box that holds a 0D marker in the applicable Aggregation Box is more practical, as shown in Figure~\ref{fig:icon-motivation-negative}.

\subsection{The Modal Verb Icon}

The Modal Verb Icon of Figure~\ref{fig:verb-implementation-shape-unactivated} was derived in the section on modal verbs. Like the Robinson Icon, this icon is an irregularly shaped data structure that contains cells in fixed locations, and where each of the cells may be highlighted under certain constraints.

\subsection{The Robinson Icon}

\textit{The components of this composite component: Swirly Array, 0D Markers.}

Representation of emotions is tricky, mostly because there does not exist a standard accepted categorization of emotions, and also because sciences such as biology, chemistry, and neurology do not suggest any clear-cut natural categories. Emotions are generally believed to be discrete, therefore it is believed that all emotions can be listed, and many similar lists do exist. The most prominent 2D categorizations of emotions are: (1) the circumplex model, (2) the vector model, (3) Positive Activation - Negative Activation (PANA) model. Other categorizations are Plutchik's model and PAD emotional state model. With respect to Tumbug, the main problem with the all these categorizations is that none of them show the emotion "love" as having an object of affection, i.e., the concept of love requires a parameter that is the object of affection, therefore the concept of love cannot be categorized as a simple data point as can most other emotions. The same is true of the emotion "hate," which is the negative of "love."

\begin{figure}
	\begin{center}
	\includegraphics[width=1.00\textwidth]{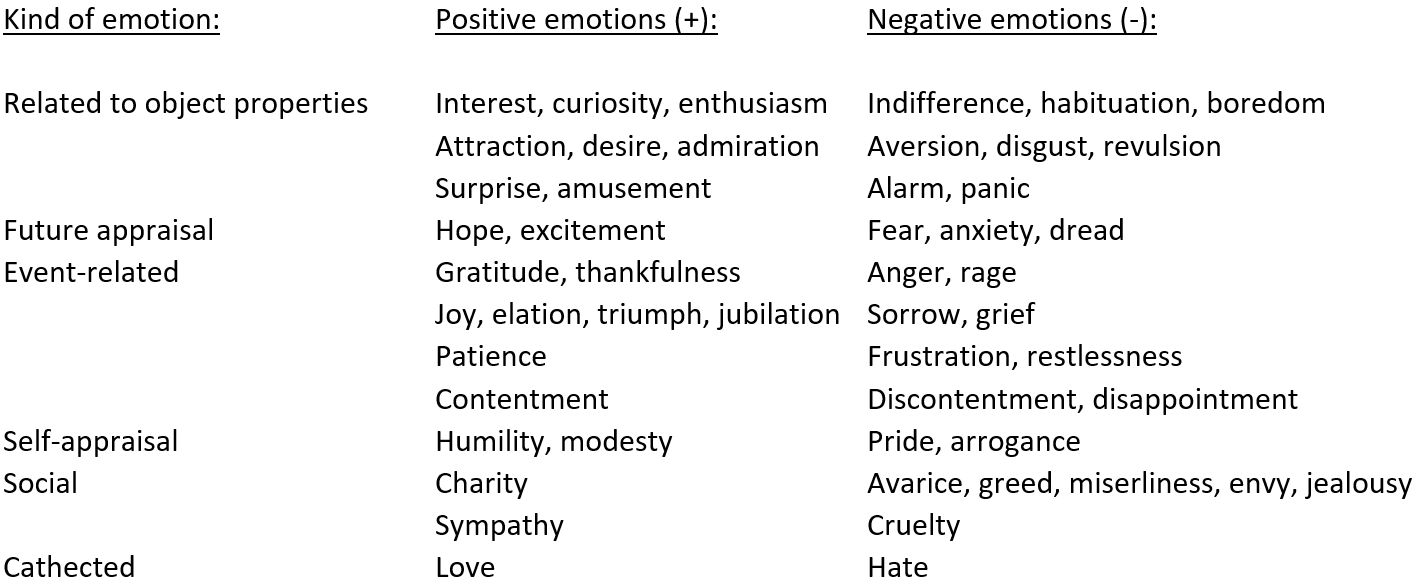}
	\caption{Robinson's categories of emotions (Robinson 2009, p. 155).}
	\label{fig:robinson-table-snap}
	\end{center}
\end{figure}

However, a review of theories of emotions was made by David Robinson in 2009 (Robinson 2009) that described the emotions of "love" and "hate" as a being in the "cathected" category, and \textit{only} those two emotions were in the cathected category. Notably excluded from the cathected category are emotions in the categories called "Related to object properties," "Event-related," and "Social." Robinson's entire list of emotions is shown in Figure~\ref{fig:robinson-table-snap}. The concept of "cathexis" comes from Sigmund Freud, and refers to an investment of energy in an object, idea, or person, which fits well the concept of requiring an extra parameter that represents the direct object. Robinson's categories are only six in number, though each has a positive-negative duality, and largely match the categorizations of other authors, so Robinson's categories are currently being used for Tumbug. Robinson's categories are: (1) related to object properties, (2) future appraisal, (3) event-related, (4) self-appraisal, (5) social, (6) cathected, (7) positive versus negative versions of all these categories.

\begin{figure}
	\begin{center}
	\includegraphics[width=0.50\textwidth]{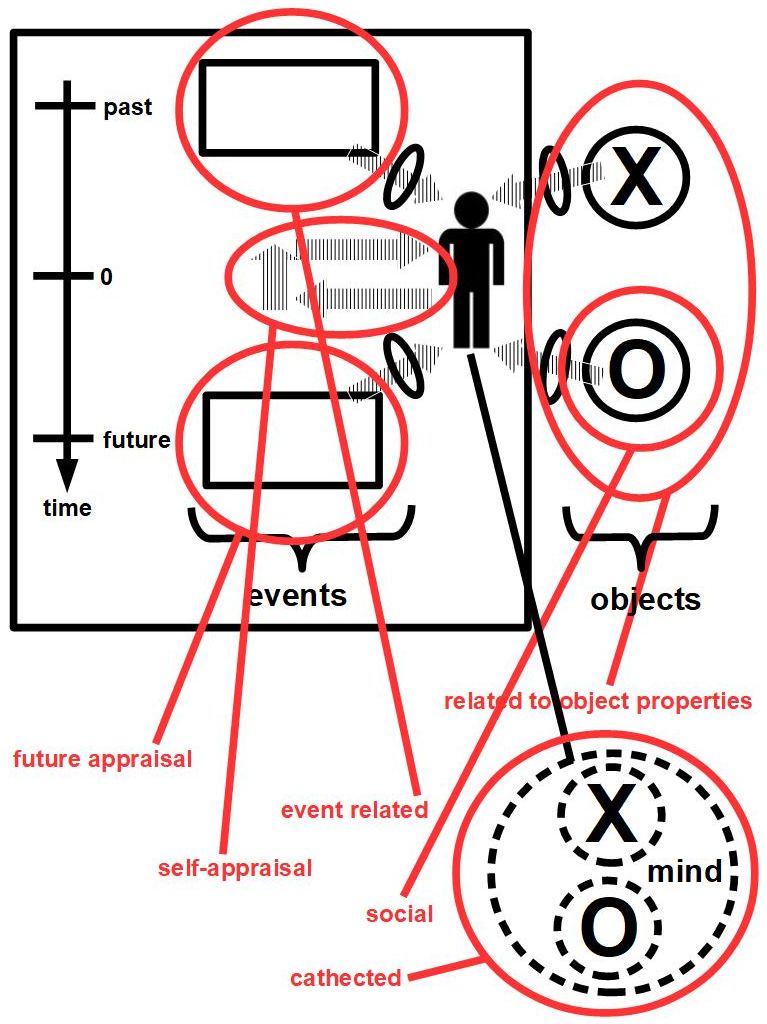}
	\caption{STEP 1. Robinson's six emotional categories map and generalize fairly symmetrically in diagrammatic form.}
	\label{fig:icon-robinson-1-emotion-classification}
	\end{center}
\end{figure}

\begin{figure}
	\begin{center}
	\includegraphics[width=0.50\textwidth]{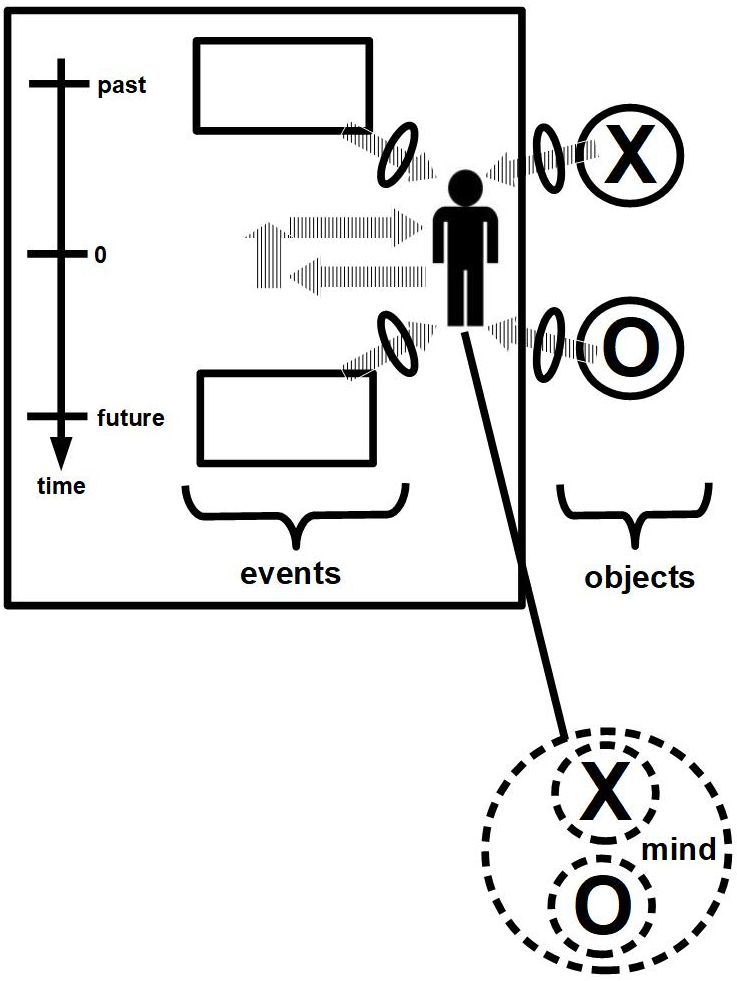}
	\caption{STEP 2. A less annotated version of the previous diagram.}
	\label{fig:icon-robinson-2-emotion-classification-clean}
	\end{center}
\end{figure}

\begin{figure}
	\begin{center}
	\includegraphics[width=0.50\textwidth]{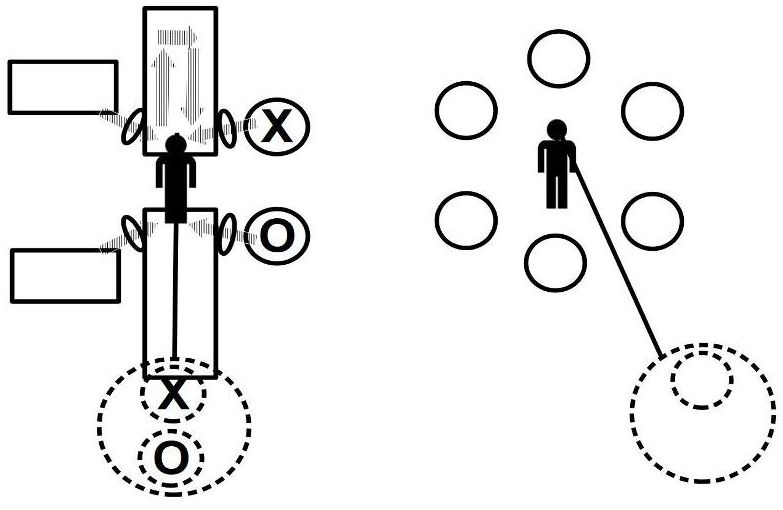}
	\caption{STEP 3 and STEP 4. Progressively abstracted and simplified versions of the previous diagrams.}
	\label{fig:icon-robinson-3-progressive}
	\end{center}
\end{figure}

\begin{figure}
	\begin{center}
	\includegraphics[width=0.25\textwidth]{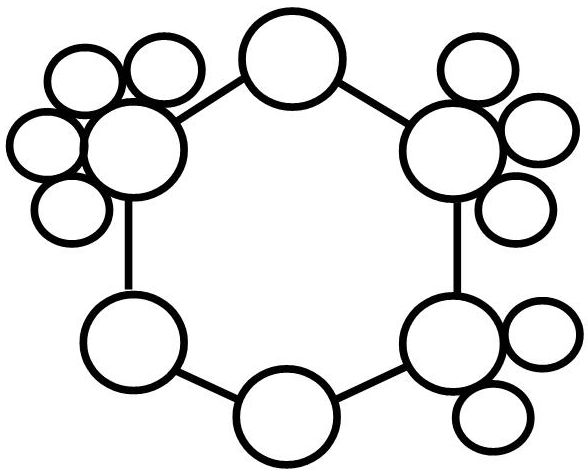}
	\caption{STEP 5. The Robinson Icon.}
	\label{fig:icon-robinson-5-detailed-categories-blanks}
	\end{center}
\end{figure}

\begin{figure}
	\begin{center}
	\includegraphics[width=0.50\textwidth]{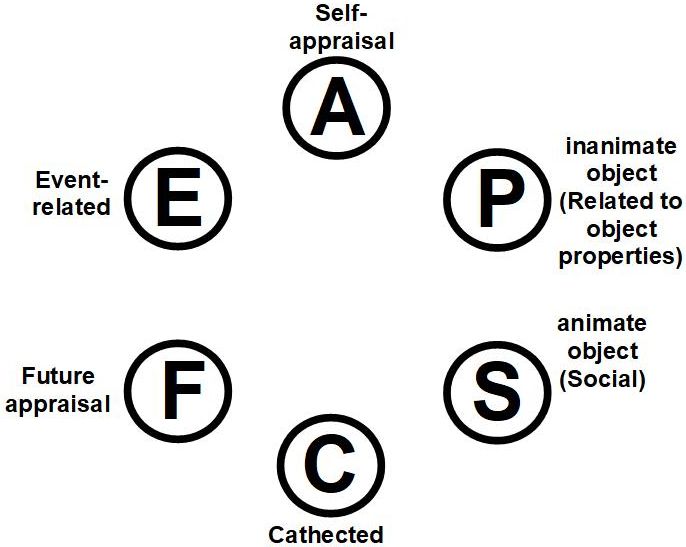}
	\caption{KEY. Key to the six primary node meanings of the Robinson Icon.}
	\label{fig:icon-robinson-key-1-basic}
	\end{center}
\end{figure}

\begin{figure}
	\begin{center}
	\includegraphics[width=0.50\textwidth]{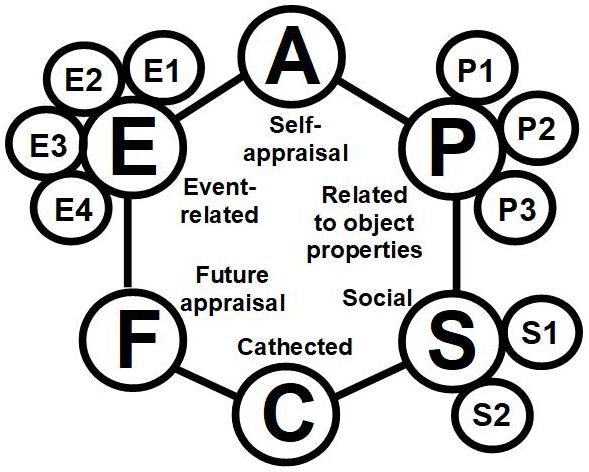}
	\caption{KEY. Key to the six primary node meanings of the Robinson Icon., connected and with more detail.}
	\label{fig:icon-robinson-key-2-detailed-categories-labeled}
	\end{center}
\end{figure}

With a bit of generalization, Robinson's categories become surprisingly symmetrical when diagrammed with Tumbug. This diagrammatic, progressive abstraction from Robinson's textual categories to a single, ornate, basically symmetrical icon is shown in Figure~\ref{fig:icon-robinson-1-emotion-classification}, Figure~\ref{fig:icon-robinson-2-emotion-classification-clean}, Figure~\ref{fig:icon-robinson-3-progressive}, and Figure~\ref{fig:icon-robinson-5-detailed-categories-blanks}, and the keys to the letter abbreviations are in Figure~\ref{fig:icon-robinson-key-1-basic} and Figure~\ref{fig:icon-robinson-key-2-detailed-categories-labeled}.

The dichotomy of positive or negative emotions, technically called "valence," can be  represented with a "+" or "-" sign, respectively. These form a pair. The two categories "future appraisal" and "event-related" can be regarded as merely the range of time that defines the category: future for "future appraisal," and past or present for "event-related," so these two categories can be represented as two C Aggregation Boxes, one on each side of current time (time = 0) on the Time Arrow. These form a pair. The two categories "related to object properties" and "social" are similar, and differ mostly only in whether the cause of the emotion is an inanimate object or an animate object, especially a person. A C Object Circle, which as shown earlier can differentiate between inanimate versus animate by labeling it with an "X" or "O," respectively, can be used to represent those two categories. These form a pair. "Self-appraisal" is a singleton category, but then so is the "cathected" category, so these last two categories form a pair, as all the other categories did. The result is a set of three pairs, which together form a hexagon with each category at one corner, a person icon conceptually placed in the middle, and one more diagram added as an aside, via an Attribute Line connecting the person to the specific object of focus for "love" or "hate."

\begin{figure}
	\begin{center}\
	\includegraphics[width=0.50\textwidth]{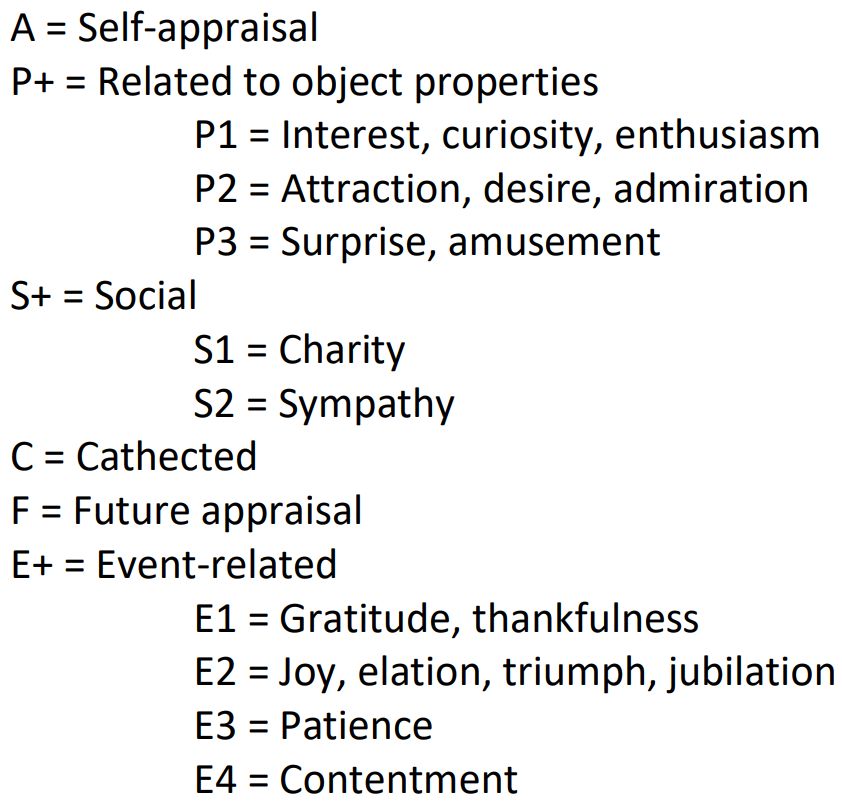}
	\caption{KEY. Complete list of all the node meanings of the Robinson Icon.}
	\label{fig:icon-robinson-node-key-list-snap}
	\end{center}
\end{figure}

\begin{figure}
	\begin{center}
	\includegraphics[width=0.25\textwidth]{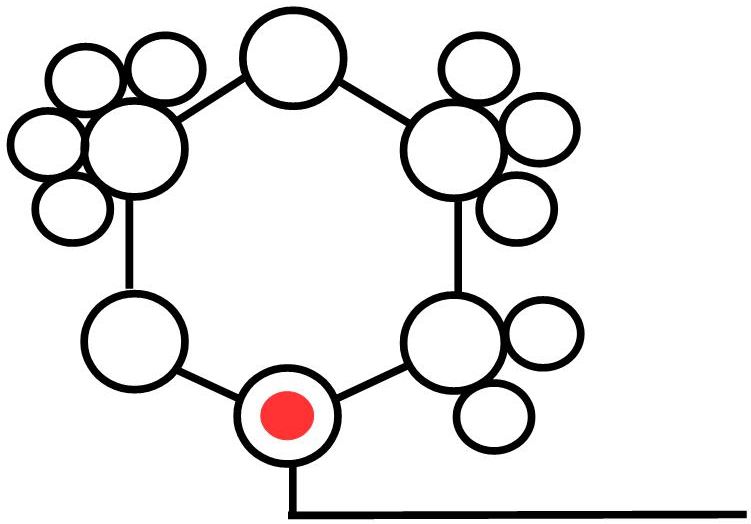}
	\caption{The icon will have an Attribute Line emerging from the bottom node (C) if the emotion is cathected (= regarded in an emotional way by the subject/viewer).}
	\label{fig:icon-robinson-detailed-with-line}
	\end{center}
\end{figure}

\begin{figure}
	\begin{center}
	\includegraphics[width=0.45\textwidth]{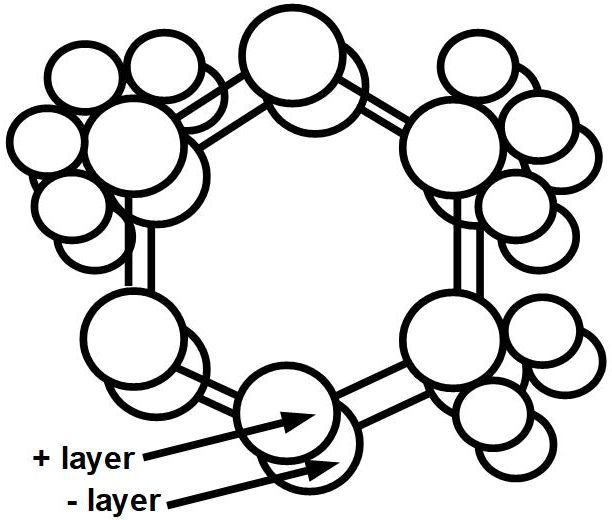}
	\caption{The Robinson structure actually has two identical layers, "+" and "-", written elsewhere as text.}
	\label{fig:icon-robinson-3d}
	\end{center}
\end{figure}

See Figure~\ref{fig:icon-robinson-node-key-list-snap}. The "+" sign on P, S, and E indicate that the textual descriptions shown are for the positive valence of the given emotion. The final Robinson Icon as usually used is shown in Figure~\ref{fig:icon-robinson-5-detailed-categories-blanks}, although the cathected version is slightly more detailed and shown in Figure~\ref{fig:icon-robinson-detailed-with-line}. The negative valence descriptions, which were listed earlier, were removed only to save space, which means that without the valence mentioned, the Robinson Icon would spatially need two layers, one for each valence, as shown in Figure~\ref{fig:icon-robinson-3d}.

\begin{figure}
	\begin{center}
	\includegraphics[width=0.50\textwidth]{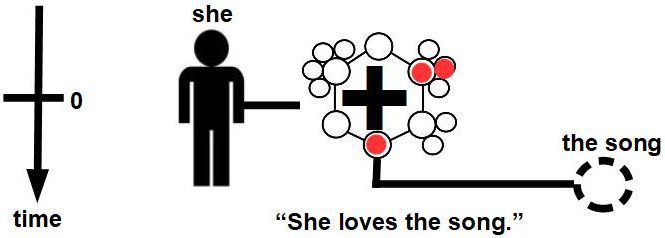}
	\caption{Example \#1a of how the Robinson Icon is used: [64] "She loves the song."}
	\label{fig:icon-robinson-usage-detailed-song}
	\end{center}
\end{figure}

The next example, shown in Figure~\ref{fig:icon-robinson-usage-detailed-song} is part of a WS150 problem:

"[64] Mary took out her flute and played one of her favorite pieces. She has loved it since she was a child. What has Mary loved since she was a child? POSSIBLE ANSWERS: \{the piece, the flute\}"

The darkened circles on the hexagon are 0D Markers that correspond to the corresponding categories in the Robinson diagram discussed earlier, in this case the categories "Related to object properties" (the upper right circle) and "Cathected" (the bottom circle). The more detailed subnode "E2" is activated for attraction ("love"). The "+" inside the hexagon indicates the positive valence of the pair of choices, in this case "love" (+) instead of "hate" (-).

\begin{figure}
	\begin{center}
	\includegraphics[width=0.50\textwidth]{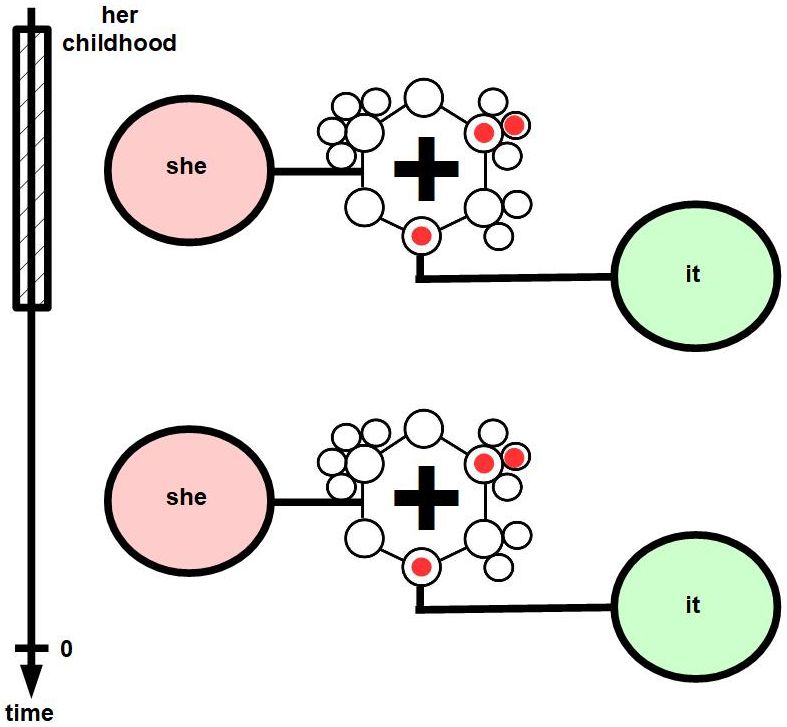}
	\caption{Example \#1b of how the Robinson Icon is used: [64] "She has loved it since she was a child."}
	\label{fig:ws064-flute}
	\end{center}
\end{figure}

A specific example from WS150 is shown in Figure~\ref{fig:ws064-flute}.
 
\begin{figure}
	\begin{center}
	\includegraphics[width=0.50\textwidth]{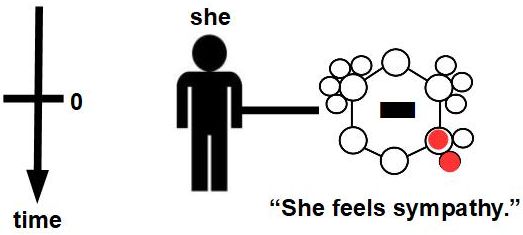}
	\caption{Example \#2 of how the Robinson Icon is used: "She feels sympathy."}
	\label{fig:icon-robinson-usage-detailed-sympathy}
	\end{center}
\end{figure}

In Figure~\ref{fig:icon-robinson-usage-detailed-sympathy}, the example shows the "Social" circle activated (darkened) with the "-" sign indicating "sympathy" (-) instead of "cruelty" (+), and the more detailed subnode "S2" is activated to represent "sympathy."

\begin{figure}
	\begin{center}
	\includegraphics[width=0.50\textwidth]{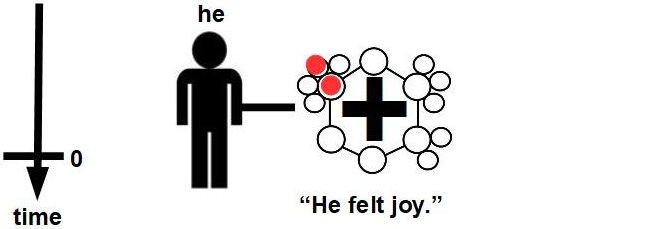}
	\caption{Example \#3 of how the Robinson Icon is used: "He felt joy."}
	\label{fig:icon-robinson-usage-detailed-joy}
	\end{center}
\end{figure}

In Figure~\ref{fig:icon-robinson-usage-detailed-joy}, the example shows the past tense by showing the main part of the diagram above (past) the current time (time = 0). "Joy" is a positive (+) emotion that is "Event related," so the corresponding circle (the upper left circle) is shown activated, and the more detailed subnode "E2" is activated to represent "joy."

\subsection{Zoom Boxes}

\subsubsection{General}

\textit{The components of this composite component: two Location Boxes, two to four 1D Markers to connect the corners of the squares in mock 3D.}

\begin{figure}
	\begin{center}
	\includegraphics[width=0.25\textwidth]{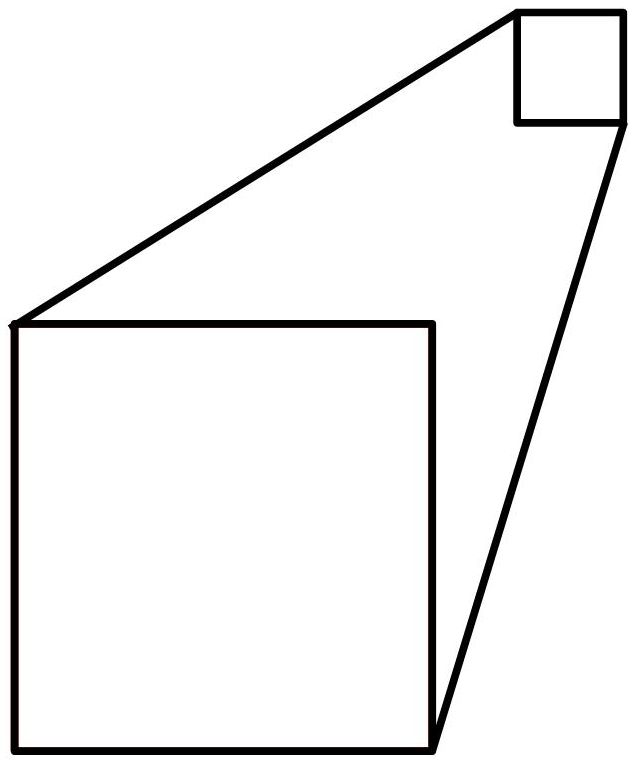}
	\caption{Tumbug's icon for a pair of Zoom Boxes.}
	\label{fig:icon-zoom-boxes}
	\end{center}
\end{figure}

The Tumbug icon for a Zoom Box is shown in Figure~\ref{fig:icon-zoom-boxes}.

\begin{figure}
	\begin{center}
	\includegraphics[width=0.70\textwidth]{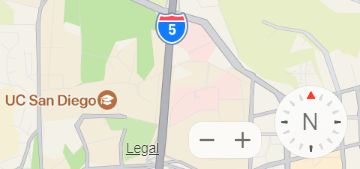}
	\caption{A zoom in (+) / zoom out (-) control on a DuckDuckGo map.}
	\label{fig:zoom-duckduckgo-maps-ucsd-detail}
	\end{center}
\end{figure}

As with digital maps that have a zoom in / zoom out feature (Figure~\ref{fig:zoom-duckduckgo-maps-ucsd-detail}, or even with paper maps that have inserts that show an important area in higher detail, a zoom feature saves a great amount of space and time since otherwise everything would need to be rendered in the highest level of detail for the largest possible area, and the user could not even see the details even if the screen could show those details. The same is truly of any large or highly detailed system, whether biological (Figure~\ref{fig:icon-zoom-boxes}), electrical, mechanical, software, astronomical, or other. Tumbug is therefore assumed to have the ability to represent two areas of a system, one of which is the zoom of another. John F. Sowa in his large book about knowledge representation, noted the utility of being able to zoom into part of a representation of an event for additional temporal details (Sowa 2000, p. 175), and Reisberg and Heuer mention a study that confirms that spatial relationships remain intact while humans zoom in on a mental image (Reisberg and Heuer 2005, p. 39).

\subsubsection{WS150 example: \#52 (worm)}

This example is from a portion of WS150 question \#52. A Zoom Box is desirable because taste buds are implied by the word "tasty," and taste buds are a tiny, interior part of anatomy, far smaller in scale than the entire animal's body.

"[52] The fish ate the worm. It was tasty. What was tasty? POSSIBLE ANSWERS: \{the worm, the fish\}"

Although "ate" would likely be immediately associated with a mouth, "tasty" would likely not be immediately associated with taste buds, nor the fact that taste buds are located inside a mouth. To make this single envisioned spatial connection across an order-of-magnitude spatial difference, zoom would be useful to show the approximate location of taste buds inside the mouth, which in turn would be shown as part of the fish's body. Other approaches to answering problem \#52 exist, though this zoom approach is one way.

\begin{figure}
	\begin{center}
	\includegraphics[width=0.75\textwidth]{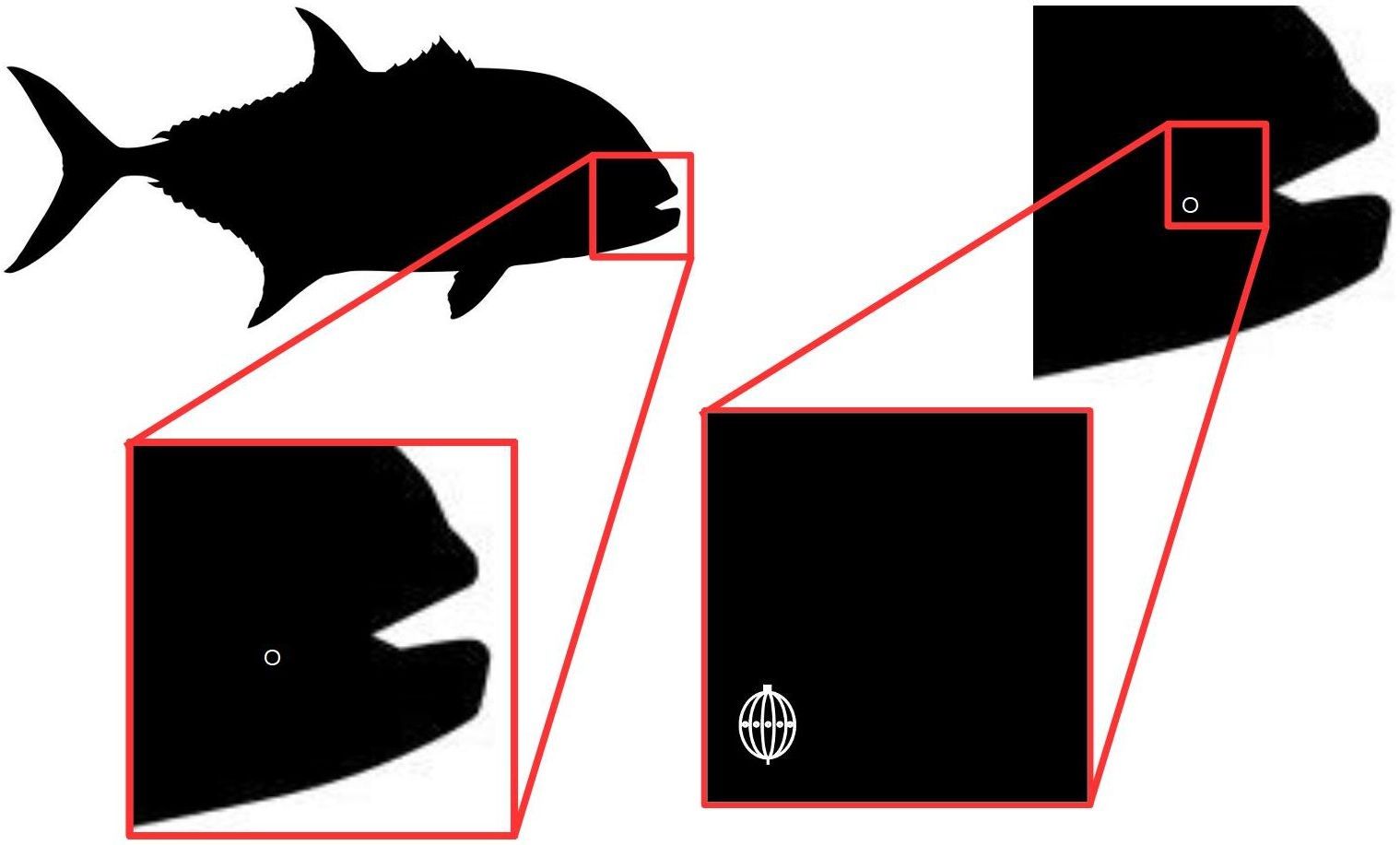}
	\caption{Zoom in / zoom out ability is a practical necessity for highly detailed diagrams. The size of a fish compared to the size of one of its taste buds spans about three orders of magnitude, for example.}
	\label{fig:icon-zoom-combined}
	\end{center}
\end{figure}

The example of Figure~\ref{fig:icon-zoom-combined} shows how a critical physical component from problem \#52, namely a fish's sensors for taste, are orders of magnitude smaller than the main scale of discussion, namely at the size scale of fish and worms.

\section{Generalized Building Blocks of Tumbug}

The generalized Building Blocks of this section are not currently used for practical applications, but rather used only as a theoretical organizational scheme to show that all Tumbug Building Blocks reduce to a small number of very general concepts.

\subsection{Nonquantified Objects (= Nonquans)}

\begin{figure}
	\begin{center}
	\includegraphics[width=0.60\textwidth]{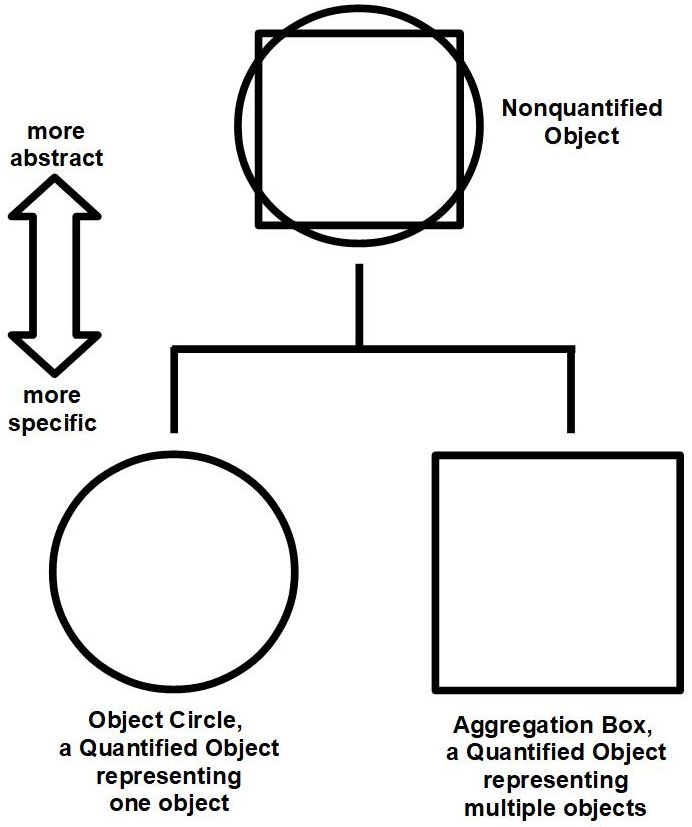}
	\caption{A Nonquantified Object is a generalization of a single object and a collection of objects.}
	\label{fig:icon-nonquan}
	\end{center}
\end{figure}

Figure~\ref{fig:icon-nonquan} shows a Nonquantified Object.

Nonquantified Objects, called "Nonquans" for short, are objects of arbitrary quantity. The idea is to generalize the concept of an "object" while still distinguishing between objects, motions, and attributes. Motions and attributes are in classes different from Nonquantified Objects. Some examples of Nonquantified Objects are:

\begin{itemize}
	\item
		an element of a set
	\item
		a set of elements
	\item
		a single data point
	\item
		a cluster of data points
	\item
		an Object Circle
	\item
		any type of Aggregation Box
	\item
		any type of Location Box
	\item
		any Compound Building Block 
\end{itemize}

\subsection{Interchangeably Actualizable Maps (= IAMs)}

\begin{figure}
	\begin{center}
	\includegraphics[width=0.50\textwidth]{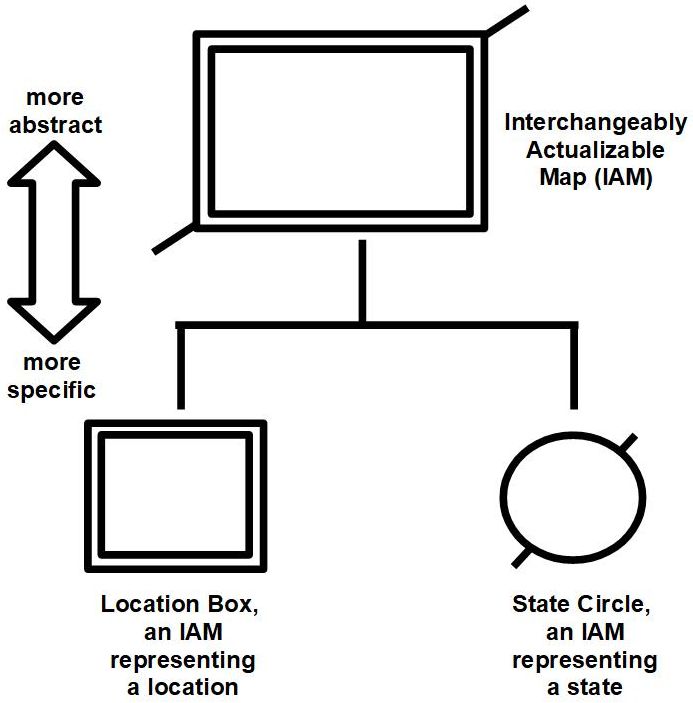}
	\caption{An Interchangeably Actualizable Map is a generalization of locations and states.}
	\label{fig:icon-iam-hierarchy}
	\end{center}
\end{figure}

It becomes clear after representing numerous sentences with locations and after representing numerous sentences with states that the concepts of location and state are similar. In fact, a state diagram can be considered an alternative view of a location diagram, and vice versa, in the same way that a plot of data can be considered an alternative view of a table of numerical values, provided that all the attributes of each data point are retained. Since the concepts of location and state are interchangeable at a higher level of abstraction, they can be generalized to a single concept, a concept that in this document is called an "Interchangeably Actualizable Map" (IAM), the Tumbug icon of which is shown in Figure~\ref{fig:icon-iam-hierarchy}.

\begin{figure}
	\begin{center}
	\includegraphics[width=0.70\textwidth]{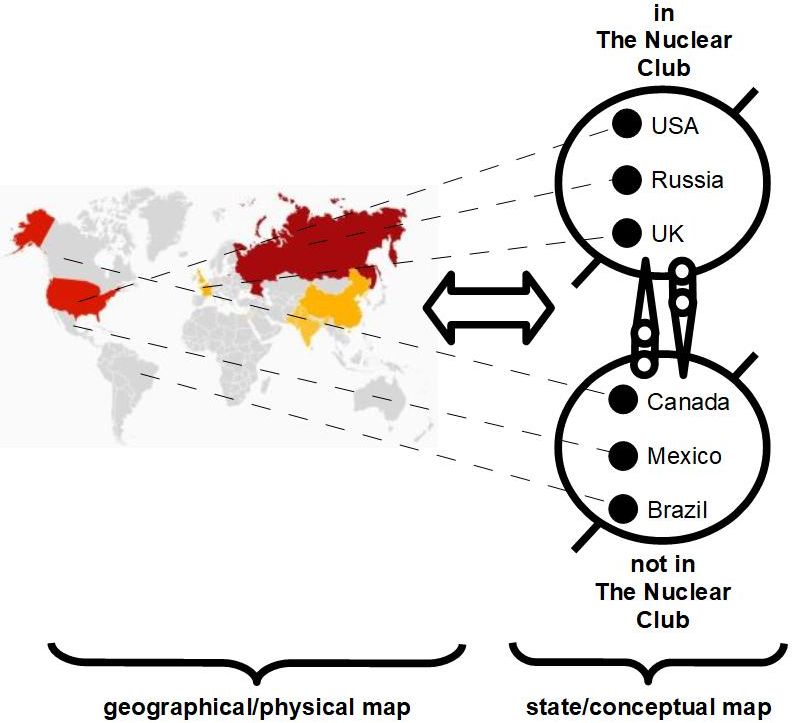}
	\caption{A geographic map of countries with respect to their membership in "The Nuclear Club" is interchangeable with a state diagram that shows such status, provided that no attributes are lost in the transfer. (Source: Federation of American Scientists.)}
	\label{fig:icon-iam-nuclear}
	\end{center}
\end{figure}

For example, a geographical map is a map of actual locations, whereas a state diagram is a map of concepts, so both representation systems can be called "maps" in general, as shown in Figure~\ref{fig:icon-iam-nuclear}.

\subsection{Change Arrows}

\begin{figure}
	\begin{center}
	\includegraphics[width=0.60\textwidth]{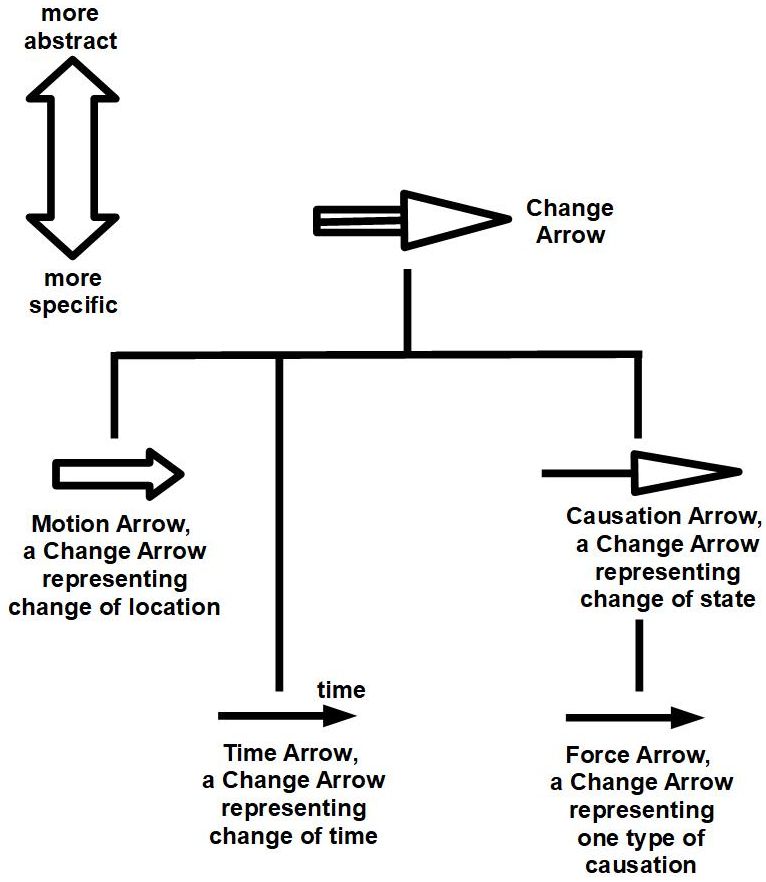}
	\caption{A Change Arrow is a generalization of all types of arrows used in Tumbug.}
	\label{fig:icon-change}
	\end{center}
\end{figure}

The observation that locations and states can be considered a single, more general Building Block has an immediate consequence: since Motion Arrows represent a change of location and since Causation Arrows represent a change in state, then Motion Arrows and Causation Arrows can also be generalized to a single, more general Building Block that represents a change in an Interchangeably Actualizable Map. Two other types of arrows, namely Force Arrows and Time Arrows, represent similar concepts of change, so all these types of arrows can be generalized into a single, general, change icon called a Change Arrow, which is shown in Figure~\ref{fig:icon-change}.

Two asides: (1) Correlation Boxes theoretically also represent change since they represent continuous mappings, or functions, but such mappings typically involve an uncountably infinite number of points, so Correlation Boxes are probably best represented as the square icons that they use now. (2) The original version of Tumbug developed by the author for language translation (Atkins 2023) was derived from only C Object Circles and Motion Arrows, which are now known to be insufficiently general icons for natural language grammars, as this more modern document demonstrates, so the original language translation version of Tumbug will need to be simpler than the strong version of Tumbug described in this document. That simpler version of Tumbug can then be more closely matched with the verbs found in real sentences, without needing to consider how those verbs might correspond to states or changes in time.

\section{Heuristics for converting sentences to Tumbug}

The author has found the following list of heuristics useful when converting sentences to Tumbug.

\begin{enumerate}
	\item
		\textbf{Heuristic:} If there exists presence of any type of barrier then at least a Motion Arrow + some kind of Box is needed.\\
		\textbf{Reasoning:} A barrier by definition is something that impedes a moving object from passing, therefore motion must be involved. Similarly, since motion toward or through a barrier implies that there exists three sequential spatial regions with qualitative differences: (1) outside the barrier, (2) the barrier itself, (3) inside the barrier. Therefore an alignment of those spatial regions is implied.\\
		\textbf{Examples from WS150:} [22] Placement of a tablecloth to protect a table. [66] Trying to stay dry from falling rain that will permeate a newspaper, backpack, clothes, etc. [77] Cold air penetrating a coat that someone is wearing.

	\item
		\textbf{Heuristic:} If there exists lifting or carrying anything then at least a Force Arrow and Motion Arrow are needed.\\
		\textbf{Reasoning:} To lift always implies the imparting of force, usually to overcome gravity, therefore force must be involved. To carry implies a constant application of force to counter the force of gravity that tries to pull the carried object to the ground. Also, in all cases motion is obviously involved.\\
		\textbf{Examples from WS150:} [8] A person lifting their child. [124] A person placing their child into bed. [129] A person carrying their child.

	\item
		\textbf{Heuristic:} If there exists damage, destruction, physical pain, emotional pain, annoyance, contamination, interference of any kind, whether applied to people, animals, objects, or other then at least a Correlation Box is needed.\\
		\textbf{Reasoning:} Pain or damage signals a change of state in an object, which implies that one object is having a physical effect on another, therefore the two applicable objects must be correlated. The same is true even if the effect is emotional rather than physical. The same is true even if one object is tasked with preventing such damage.\\
		\textbf{Examples from WS150:} [9] One object smashing through another. [43] Someone who shows no appreciation despite having been done a great favor. [104] Piercing one object with another object.
   
	\item
		\textbf{Heuristic:} If there exists direction, position, or alignment of one or more objects, whether toward, at, front, back, on, atop, above, below, up, down, inside, outside, through, or other, then some kind of Box is needed.\\
		\textbf{Reasoning:} Relative position of objects in relation with each other implies that that relative position in space is important enough to mention, therefore such a relationship cannot be random, therefore such position needs to be spatially aggregated. This is true even if there is only one object whose characteristic has been described.\\
		\textbf{Examples from WS150:} [14] One painting hanging above another one. [19] One sack of grocery items is atop another sack of grocery items. [21] Trying to balance a bottle upside down on a table. 

	\item
		\textbf{Heuristic:} If there exists specific relative time mentioned, whether all the described events happened in the past, present, or future, or whether some of the described events happened in only one of those time ranges, then at least a Time Arrow is needed.\\
		\textbf{Reasoning:} Like space, if time is relevant enough to be described, then it should be diagrammed.\\
		\textbf{Examples from WS150:} [32] Statement of a current problem such as a clogged drain that implies the need for a future solution. [74] Changing one's future travel destination because of conditions observed at the originally planned destination. [146] A book that was known to have influenced a famous author in the past.

	\item
		\textbf{Heuristic:} If there exists any mention of falling down, hanging down, dangling down, or rolling downhill, then at least a Force Arrow + Motion Arrow are needed.\\
		\textbf{Reasoning:} Object hanging downward or falling downward are phenomena that arise from gravity, which is a type of force, and force tends to give rise to motion.\\
		\textbf{Examples from WS150:} [36] Mention of rain falling. [128] Dropping an object such as an ice cream. [129] A person's legs dangling as they are being carried.

	\item
		\textbf{Heuristic:} If there exists any mention of interior then some kind of Box is needed.\\
		\textbf{Reasoning:} Interior implies a position in space, relative to another object.\\
		\textbf{Examples from WS150:} [57] Someone inside a given building, such as a library. [69] Knocking on a person's door but they do not answer because they are not inside. [96] A type of fish living in a specifically named ocean. 

	\item
		\textbf{Heuristic:} If there exists any mention of speed then at least a Motion Arrow is needed.\\
		\textbf{Reasoning:} Motion is represented by a Motion Arrow.\\
		\textbf{Examples from WS150:} [6] One vehicle zooming past another vehicle. [12] Two competing runners in a race, one beating the other. [93] One animal that should flee from another quickly because the first animal is in danger from the other.

	\item
		\textbf{Heuristic:} If there exists any mention or implication of a specific view of certain items collectively, especially if that view must occur along a specific line or angle, then at least some kind of Box is needed, and possibly some type of Marker, such as a 1D Marker to show alignment.\\
		\textbf{Reasoning:} A specific view implies a location and/or angle, such as using the ground for a reference, and locations of objects, or angles between three reference points, imply relation positions of those objects, which requires some kind of Box.\\
		\textbf{Examples from WS150:} [31] Being able to see a garden through a gap in a wall. [120] View of either a crop duster from underneath the crop duster, or view from the crop duster itself. 

	\item
		\textbf{Heuristic:} If there exists any line-of-sight incident, especially where one person is blocking another person's view of a stage, then at least some kind of Box with a 1D Marker is needed.\\
		\textbf{Reasoning:} Line of sight implies relative positioning of objects.\\
		\textbf{Examples from WS150:} [10] A short person cannot see the stage because a taller person is standing in front of the shorter person. [34] A person cannot see the stage because a pillar blocks that person's view. 

	\item
		\textbf{Heuristic:} If there exists presence of one of the words \{because, as, since, so, until\} then a Causation Arrow might be needed, and is almost required for the word "because."\\
		\textbf{Reasoning:} The word "because" means "was caused by." The words "as" and "since" are synonyms for "because" in some contexts.\\
		\textbf{Examples from WS150:} [13] A sculpture rolling off a shelf \textbf{because} the sculpture was not anchored. [90] Yakutska's army losing \textbf{since} its army was smaller less well-equipped. [150] A user changing to a more natural password \textbf{as} the new password is easier to remember.

	\item
		\textbf{Heuristic:} If there exists motion, placement, replacement, travel, transfer, transport, performing occupation of a place, or trade/swap then at least an Object Circle with a Motion Arrow is needed.\\
		\textbf{Reasoning:} All the situations described involve objects changing locations, therefore require space and time to represent.\\
		\textbf{Examples from WS150:} [118] Taking a particular seat in a car. [126] Sharing a toy. [135] Giving someone tickets to a play.

	\item
		\textbf{Heuristic:} If there exists teaching, training, presenting, telling, warning, notifying, learning, believing, or interrupting then at least transfer of a Data Object Circle via Motion Arrow is needed.\\
		\textbf{Reasoning:} All the situations described change or increase in knowledge, therefore knowledge is being updated as a result of transfer of data.\\
		\textbf{Examples from WS150:} [27] Telling somebody about a car accident that involved a person known to the listener. [61] Telling lies about oneself. [112] Learning history from a book.
		
	\item
		\textbf{Heuristic:} If a process within a temporal system is being modeled then it helps to break the process into phases, especially at the moments when a clearly defined event starts or stops.\\
		\textbf{Reasoning:} Divide and conquer. A drawing cannot easily show the infinite number of states the process passes through continuously, but a drawing can show the few finite points at which an event starts or stops.\\
		\textbf{Examples from WS150:} [104] Sticking a pin into a carrot part way, then pulling it out. Some moments of phase changes will be: the moment the pin touches the carrot, the moment the pin transitions from insertion to extraction, the moment the pin leaves the carrot. [128] Tommy dropped his ice cream, Timmy giggled, and father gave a stern look. This process contains exactly three clear-cut phases. [36] In the middle of the outdoor concert, the rain started falling, and it continued until 10. Some moments of phase changes will be: the start of the concert, the moment the rain started falling, and 10 o'clock.

\end{enumerate}

\section{Relationship to prior work}

\subsection{Conceptual Dependency theory}

Probably the closest work on a KRM that is similar to Tumbug is that of Roger Schank's Conceptual Dependency theory (CD theory). Although CD theory is textual instead of pictorial, and is based on a custom-made set of building blocks instead of physics, Tumbug did borrow one term from CD theory, viz. ATTEND, and has roughly the same number of building blocks as CD theory (both around the 2-3 dozen range).

\subsubsection{Building Blocks}

\begin{figure}
	\begin{center}
	\includegraphics[width=1.00\textwidth]{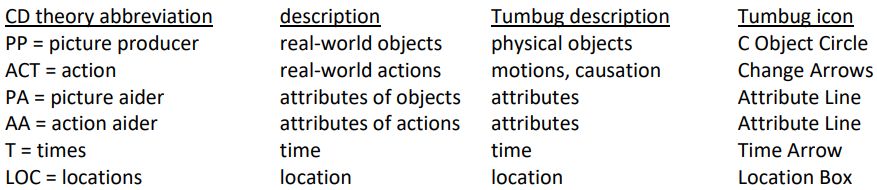}
	\caption{A table of how CD theory building blocks compare to Tumbug Building Blocks.}
	\label{fig:cdt-comparison-cdt-tumbug-list-snap}
	\end{center}
\end{figure}

CD theory's components, often called "building blocks," have nearly identical correspondence to many of Tumbug's Building Blocks, as shown in Figure~\ref{fig:cdt-comparison-cdt-tumbug-list-snap}. (Schank 1972, p. 557)

\begin{figure}
	\begin{center}
	\includegraphics[width=0.75\textwidth]{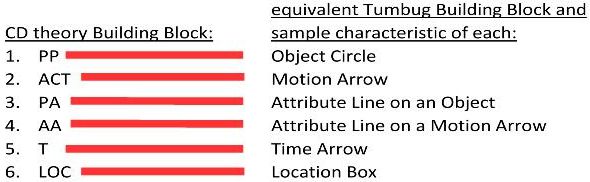}
	\caption{100\% of CD theory building blocks map one-to-one to an equivalent Tumbug concept.}
	\label{fig:cdt-summary-building-blocks-list-ANN}
	\end{center}
\end{figure}

The difference in building blocks between CD theory and Tumbug are shown in Figure~\ref{fig:cdt-summary-building-blocks-list-ANN}.

\begin{itemize}
	\item
		Tumbug Attributes are more general, therefore Tumbug has only one type, whereas CD theory attributes are divided in two types: PA (picture aider) and AA (action aider).
	\item
		Tumbug has more Building Blocks than CD theory: data Objects, Value, numerical correlators, split timelines with probabilities, wildcards, etc.
\end{itemize}

\subsubsection{Slots}

\textbf{1. ACTOR}

\begin{figure}
	\begin{center}
	\includegraphics[width=0.06\textwidth]{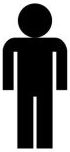}
	\caption{ACTOR: A human-shaped icon can replace Tumbug's default circle icon, if a human actor is desired.}
	\label{fig:cdt-actor}
	\end{center}
\end{figure}

An "actor" in CD theory refers to a human that initiates a PTRANS (= physical transfer). Compared to Tumbug, this is rather specific in that Tumbug does not require that the actor or subject be a human. However, Tumbug does has this CD theory ability because it is trivial to swap in a more realistic icon of a human to replace the usual circle or ovoid that are Tumbug's default representation of objects in a sentence, as shown in Figure~\ref{fig:cdt-actor}.

\textbf{2. OBJECT}

An "object" in CD theory is nearly identical in concept with Tumbug's circle icon, other than Tumbug will also allow any circle to represent a human, even if the circle is the direct object or indirect object.

\begin{figure}
	\begin{center}
	\includegraphics[width=0.13\textwidth]{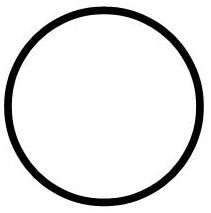}
	\caption{OBJECT: A CD theory object is already represented in Tumbug by a circle, by default.}
	\end{center}
\end{figure}

\textbf{3. FROM}

\begin{figure}
	\begin{center}
	\includegraphics[width=0.25\textwidth]{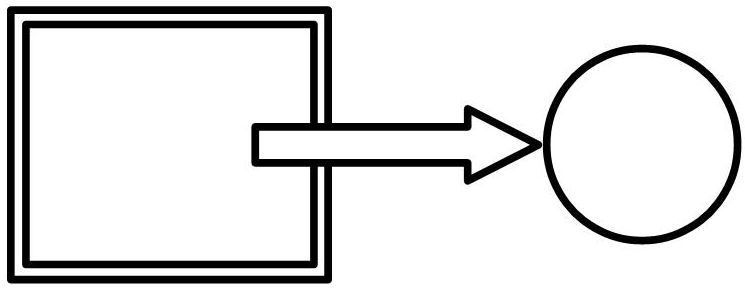}
	\caption{FROM: If the object is moving away from a Location Box, that Location Box represents the FROM location.}
	\label{fig:cdt-slots-from}
	\end{center}
\end{figure}

\begin{figure}
	\begin{center}
	\includegraphics[width=0.25\textwidth]{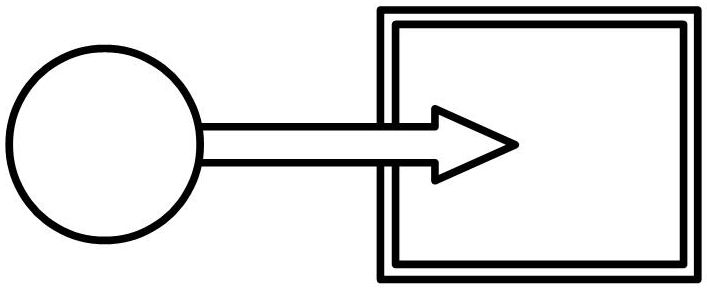}
	\caption{TO: If the object is moving toward a Location Box, that Location Box represents the TO location.}
	\label{fig:cdt-slots-to}
	\end{center}
\end{figure}

In Tumbug the source location and destination location are both represented by the same Tumbug icon, the Tumbug icon for place. The distinction between those places is obvious through usage: if an object is coming from a place, that is the source location (Figure~\ref{fig:cdt-slots-from}), and if an object is going to a place, that is the destination location (Figure~\ref{fig:cdt-slots-to}). A Location Box of some type is the icon for place, and the direction in question should be clear by the direction of the Motion Arrow, relative to the C Object Circle and Location Box.

\textbf{4. TO}

This is basically the same as FROM, except in TO the object is moving toward the given location instead of away from it.

\subsubsection{Primitive Acts}

Conceptual dependency (CD) theory is a model of natural language understanding that was developed by Roger Schank in 1969, which makes it one of the first such models. The goal of CD was the same as the goal of Tumbug: to make language meaning independent of the words it uses. Therefore it is natural to compare CD against Tumbug.

Schank listed 11 Primitive Acts that cover most actions in the real world. In alphabetical order these are:

\begin{enumerate}
	\item
		ATRANS - to change an abstract relationship of a physical object
	\item
		ATTEND - to direct a sense organ or focus an organ towards a stimulus
	\item
		EXPEL - to take something from inside an animate object and force it out
	\item
		GRASP - to physically grasp an object
	\item
		INGEST - to take something inside an animate object
	\item
		MBUILD - to create or combine thoughts
	\item
		MOVE - to move a body part
	\item
		MTRANS - to transfer information mentally
	\item
		PROPEL - to apply a force to
	\item
		PTRANS - transfer of the physical location of the object
	\item
		SPEAK - to produce a sound
\end{enumerate}

Each of these is discussed and diagrammed next. All of Tumbug's concepts are more general than CD theory's concepts, therefore Tumbug diagrams can describe CD theory concepts without too much difficulty, whereas CD concepts usually cannot describe Tumbug concepts. In this section the CD theory Primitive Acts are listed in order of the simplest to the most complicated, for ease of presentation.

\textbf{1. SPEAK}

\begin{figure}
	\begin{center}
	\includegraphics[width=0.45\textwidth]{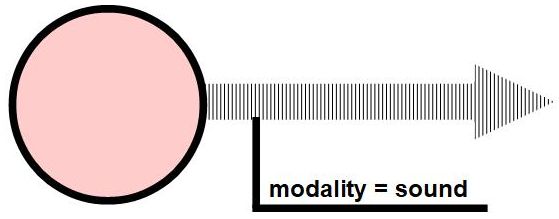}
	\caption{SPEAK: Information emission by sound modality, regardless of the emitter or reason for the emission.}
	\label{fig:cdt-speak}
	\end{center}
\end{figure}

In CD theory, "SPEAK" is "The act of producing sound, including non-communicative sounds" (Lytinen 1992, p. 52), even if from lower animals and not intended as communication. Tumbug would generalize "animal" to any entity, such as a robot. Since the action of speaking does not imply that the listener is paying attention, an Attend Ring is not needed in the Tumbug diagram. However, since Tumbug's data Motion Arrows do not specify what sensory modality is being used, Tumbug would need to place a constraint on the transmission such as "modality = sound" on its data Motion Arrow. See Figure~\ref{fig:cdt-speak}.

\textbf{2. PTRANS}

In CD theory, "PTRANS" is "The transfer of location of an object" (Lytinen 1992, p. 52). "PTRANS" means physical transfer, whether in transitive meaning (e.g., "He moved the radio farther away") or intransitive meaning (e.g., "He moved farther away from the radio"). Tumbug makes a big distinction between transitive and intransitive meanings because an additional object is involved in the transitive meaning, and in Tumbug all significant objects must be diagrammed. Therefore the CD theory meaning has two possible representations in Tumbug, depending on the type of transitivity intended.

\textbf{2.1. Intransitive}

\begin{figure}
	\begin{center}
	\includegraphics[width=0.35\textwidth]{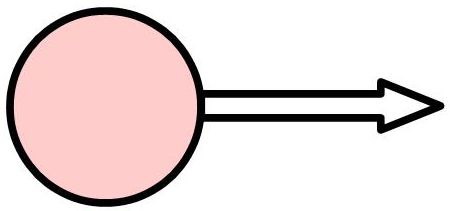}
	\caption{PTRANS, intransitive meaning: An object moves itself to an undisclosed location.}
	\label{fig:cdt-ptrans-intransitive}
	\end{center}
\end{figure}

See Figure~\ref{fig:cdt-ptrans-intransitive}.

\textbf{2.2. Transitive}

\begin{figure}
	\begin{center}
	\includegraphics[width=0.35\textwidth]{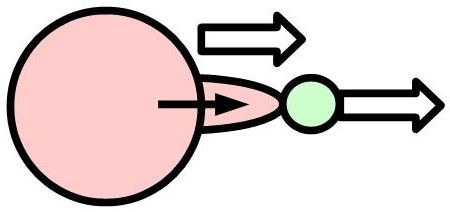}
	\caption{PTRANS, transitive meaning: An object moves another object to an undisclosed location.}
	\label{fig:cdt-ptrans-transitive}
	\end{center}
\end{figure}

See Figure~\ref{fig:cdt-ptrans-transitive}.

\textit{Tumbug minor convention: Motion Arrows may be used in diagrams without time or space reference, even when chained together, unless this results in ambiguity about the timing or cause-and-effect of the motions.}

\textbf{3. MOVE}

In CD theory, "MOVE" is "The movement of a body part of an agent by that agent" (Lytinen 1992, p. 52). This Primitive Act is slightly unusual in that the moved body part would be considered a direct object, which means that the subject is regarding its own body part as a direct object.

Whenever anything moves, Tumbug represents this situation with a timeline on the side of the diagram, spanning from the starting position to the ending position of the moved object. If the direction and location of the motion is not important, then a Location Box is not needed, otherwise if either attribute is important then a Location Box should be included.

An object repeatedly shown along a timeline is assumed to be undergoing continuous (as opposed to discrete) modification in time, so that the diagrams along the timeline can be considered only snapshots in time. If practical, one snapshot should be included for each qualitative change in the situation so that each diagram can be interpretated as one phase of the overall process.

\begin{figure}
	\begin{center}
	\includegraphics[width=0.25\textwidth]{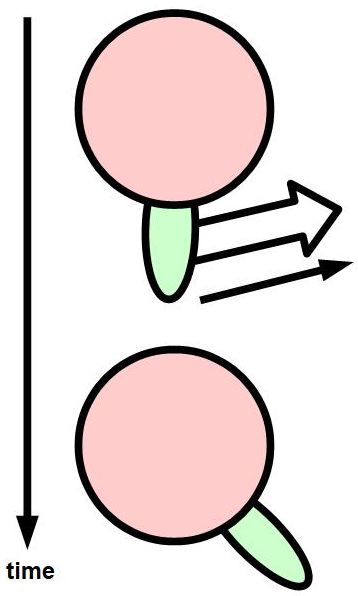}
	\caption{MOVE: An external body part being moved by its own body, where direction and location are not important.}
	\label{fig:cdt-move}
	\end{center}
\end{figure}

For the situation of a body part being moved by its connected body, the body as a whole will ordinarily remain stationary except for that moved body part. The body would ordinarily be represented as a circle, and the body part would ordinarily be represented as a circle or ellipse, whichever most closely resembles the body part. If the moved body part is external (such as with an arm), the appendage will protrude on the outside of the body (circle); for an internal body part (such as the larynx), the appendage will protrude toward the inside of the body (circle). See Figure~\ref{fig:cdt-move}. Appendages and internal body parts could theoretically be represented with fuzzy regions instead of crisp ellipses, including details such as fuzzy fingers on fuzzy hands on fuzzy arms, but this type of representation is likely more difficult to code as a computer program, and is not in the spirit of traditional, crisp, visual computer simulations. 

\textbf{4. GRASP}

\begin{figure}
	\begin{center}
	\includegraphics[width=0.35\textwidth]{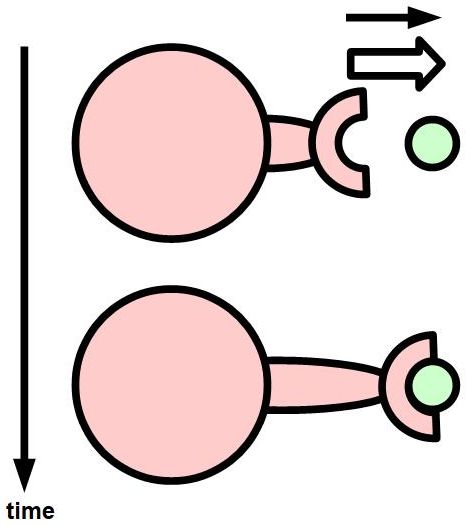}
	\caption{GRASP: A bodily appendage with an end effector extends to hold an object firmly.}
	\label{fig:cdt-grasp}
	\end{center}
\end{figure}

In CD theory, "GRASP" is "The grasping of an object by an actor so that it may be manipulated" (Lytinen 1992, p. 52). In this situation some kind of end effector (such as a hand) is needed, which is a level of detail not usually included in Tumbug, which usually shows only the equivalent of an arm, if that much. See Figure~\ref{fig:cdt-grasp}.

\textbf{5. PROPEL}

In CD theory, "PROPEL" is "The application of a physical force to an object" (Lytinen 1992, p. 52). This a situation that Tumbug would represent by a Force Arrow directed toward a C Object Circle. The result of the force may or may not cause the direct object to move, however: the PROPEL definition does not specify. If the object moves then per Lytinen the PROPEL scenario has turned into a PTRANS scenario, otherwise the scenario is only PROPEL. Also unspecified by the PROPEL definition is whether some subject is causing the force. The PROPEL diagrams in this section assume a subject is applying the force, which is the most common situation, but by definition this is not strictly necessary.

\textbf{5.1. With no resulting motion}

\begin{figure}
	\begin{center}
	\includegraphics[width=0.35\textwidth]{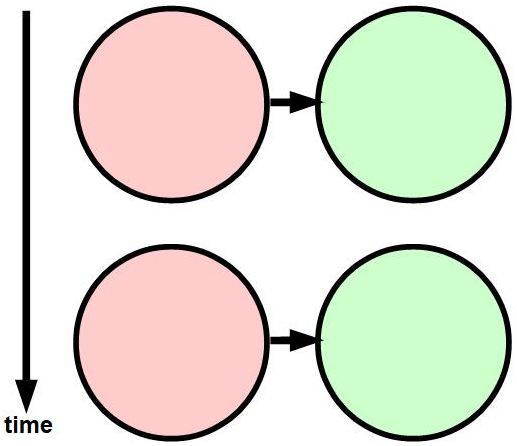}
	\caption{PROPEL: An object acted upon by force created by agent, but the direct object is not moved as a result.}
	\label{fig:cdt-propel-no-motion}
	\end{center}
\end{figure}

See Figure~\ref{fig:cdt-propel-no-motion}.

\textbf{5.2. With resulting motion}

\begin{figure}
	\begin{center}
	\includegraphics[width=0.50\textwidth]{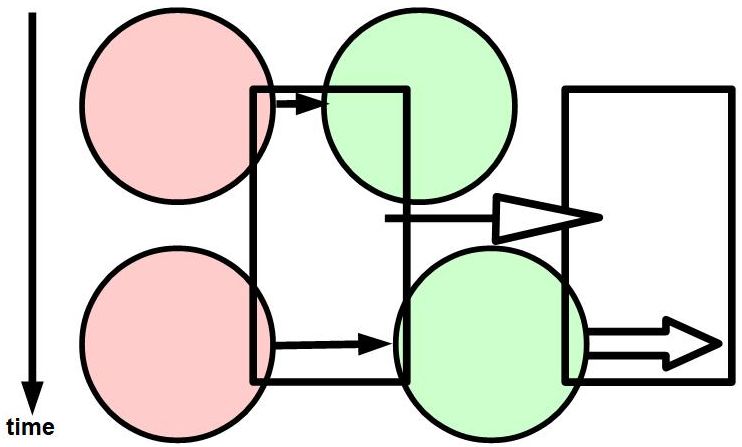}
	\caption{PROPEL: An object acted upon by force created by agent, and the direct object being moved as a result.}
	\label{fig:cdt-propel-motion}
	\end{center}
\end{figure}

See Figure~\ref{fig:cdt-propel-motion}.

\textbf{6. INGEST}

\begin{figure}
	\begin{center}
	\includegraphics[width=0.33\textwidth]{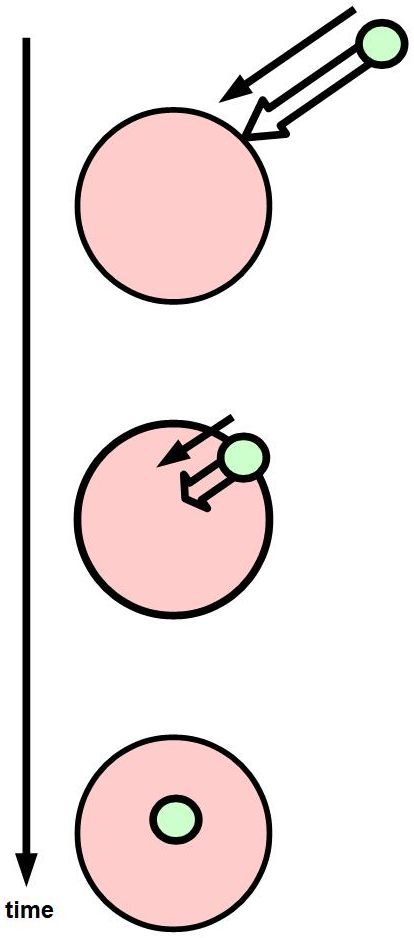}
	\caption{INGEST: An external object being moved from outside a body to inside that body.}
	\label{fig:cdt-ingest}
	\end{center}
\end{figure}

In CD theory, "INGEST" is "The taking in of an object (food, air, water, etc.) by an animal" (Lytinen 1992, p. 52). In Figure~\ref{fig:cdt-ingest} the ingestion process has been divided into three phases: (1) the object is outside the body, (2) the object is at the surface of the body, (3) the object is inside the body. If the opening to the body were important during such a process (such as a mouth) then a region of circle could be delineated to represent a mouth. Note in Figure~\ref{fig:cdt-ingest} that the length of the Motion Arrow decreases in time to show that the object is slowing when entering the body, then comes to a complete stop for a while after reaching some (possibly temporary) destination within the body (such as a stomach).

\textbf{7. EXPEL}

\begin{figure}
	\begin{center}
	\includegraphics[width=0.33\textwidth]{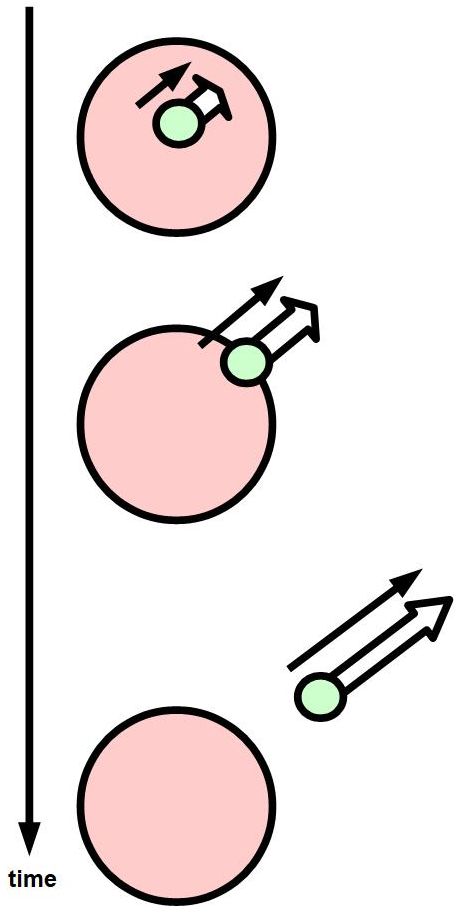}
	\caption{EXPEL: An external object being moved from inside a body to outside that body.}
	\label{fig:cdt-expel}
	\end{center}
\end{figure}

In CD theory, "EXPEL" is "The expulsion of an object by an animal" (Lytinen 1992, p. 52). EXPEL is the same as INGEST except that the object motion is in the opposite direction, from within the body to outside the body. See Figure~\ref{fig:cdt-expel}.

\textbf{8. MTRANS}

\begin{figure}
	\begin{center}
	\includegraphics[width=0.35\textwidth]{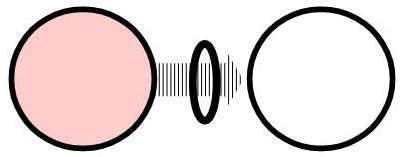}
	\caption{MTRANS: Information transmitted from the agent on the left, and received by the agent on the right.}
	\label{fig:cdt-mtrans}
	\end{center}
\end{figure}

In CD theory, "MTRANS" is "The transfer of mental information between agents" (Lytinen 1992, p. 52). To show this in Tumbug the diagram needs two C Object Circles, one for each agent, a data Motion Arrow to show the information transfer, and one Attend Ring to show that the receiver is paying attention to the transferred information. See Figure~\ref{fig:cdt-mtrans}. The reason the receiving agent is not color-coded in this diagram is because the MTRANS concept does not specify whether the message is the direct object (as in "He gave the information to his boss.") or whether the receiver is the direct object (as in "He informed his boss.").

\textbf{9. ATRANS}

In CD theory, "ATRANS" is "The transfer of ownership, possession, or control of an object" (Lytinen 1992, p. 52). Examples in Lytinen imply that the verbs "to give" and "to pay for" are involve ATRANS, and that the ATRANS operation is described on a link independent of (and parallel to?) the link that describes the physical transfer (if any). Schank probably intended this ATRANS definition to mean "full" transfer, as opposed to partial transfer, although in theory a second person could be placed on an ownership certificate or given co-managerial control over a group of employees. He we assume that full transfer is intended. 

\begin{figure}
	\begin{center}
	\includegraphics[width=0.35\textwidth]{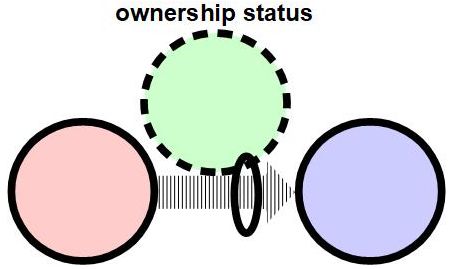}
	\caption{ATRANS: Information is transmitted from the agent on the left, and received by the agent on the right.}
	\label{fig:cdt-atrans-mtrans}
	\end{center}
\end{figure}

The simplest way to describe the ATRANS event with Tumbug is to treat ownership status as a data Object, and to use the shorthand notation for transfer of an object between two entities, as shown in Figure~\ref{fig:cdt-atrans-mtrans}. More complicated schemes involving acknowledgement via message passing are also possible.

\begin{figure}
	\begin{center}
	\includegraphics[width=0.85\textwidth]{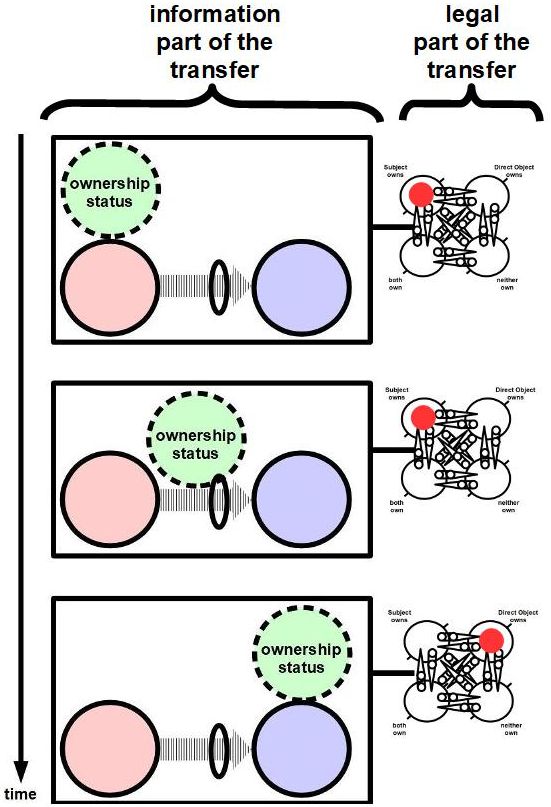}
	\caption{A more comprehensive view of an ATRANS operation would show both information transfer and legal transfer.}
	\label{fig:cdt-atrans-mtrans-with-states-ANN2}
	\end{center}
\end{figure}

A more comprehensive view of an ATRANS would involve the change of legal ownership status, though, not just the physical transfer. This is relatively simple to include, by including a state diagram of legal ownership status that is associated with the physical transfer operation, as shown in Figure~\ref{fig:cdt-atrans-mtrans-with-states-ANN2}.

\begin{figure}
	\begin{center}
	\includegraphics[width=0.50\textwidth]{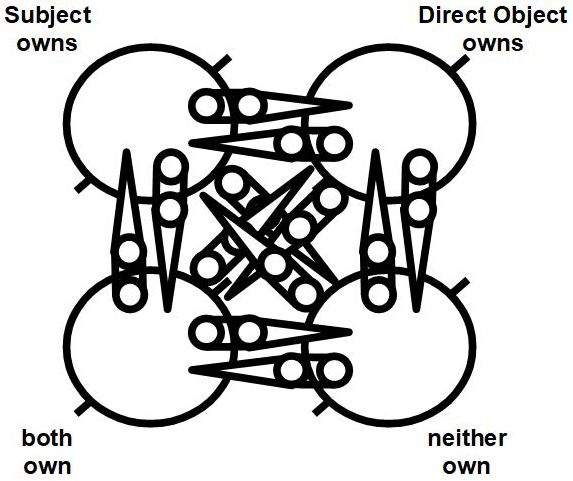}
	\caption{A State Diagram that shows all possibilities of legal ownership between two people.}
	\label{fig:icon-atrans-ownership-states}
	\end{center}
\end{figure}

A larger view of the State Diagram that is embedded in the comprehensive ATRANS view is shown in Figure~\ref{fig:icon-atrans-ownership-states}.

\textbf{10. ATTEND}

\begin{figure}
	\begin{center}
	\includegraphics[width=0.50\textwidth]{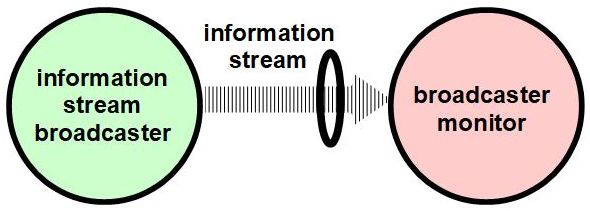}
	\caption{ATTEND: The broadcaster monitor attends to an information stream from the broadcaster.}
	\label{fig:cdt-attend}
	\end{center}
\end{figure}

In CD theory, "ATTEND" is "The act of focusing attention of a sense organ toward an object" (Lytinen 1992, p. 52). Tumbug uses exactly this same concept with the same name, and represents it by drawing an Attend Ring around the information stream arrow on which the receiver is focusing attention. Although Tumbug can represent a sense organ on the object to satisfy this definition more closely, it can probably be assumed that if the object is focusing its attention on an information stream then the object is capable of sensing it. See Figure~\ref{fig:cdt-attend}.

\textbf{11. MBUILD}

In CD theory, "MBUILD" is "The construction of a thought or of new information by an agent" (Lytinen 1992, p. 52). Note that this MBUILD definition is ambiguous as to whether "construction" means created from scratch, or rather created from other objects. Both cases are considered below, and are handled easily by Tumbug.

\textbf{11.1. Creation from scratch}

\begin{figure}
	\begin{center}
	\includegraphics[width=0.50\textwidth]{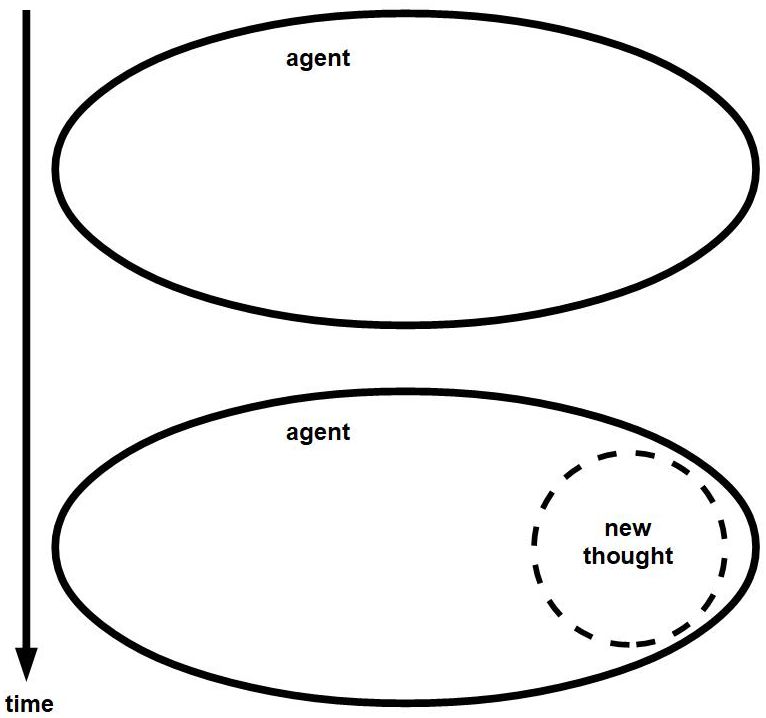}
	\caption{MBUILD: One new thought forms spontaneously, from scratch.}
	\label{fig:cdt-mbuild-scratch}
	\end{center}
\end{figure}

See Figure~\ref{fig:cdt-mbuild-scratch}.

\textbf{11.2. Creation from combination}

\begin{figure}
	\begin{center}
	\includegraphics[width=0.50\textwidth]{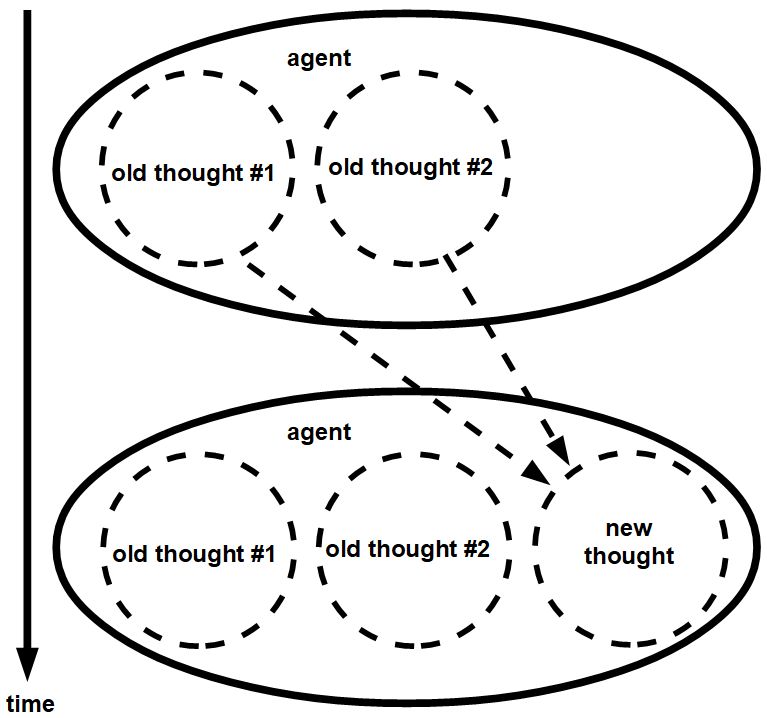}
	\caption{MBUILD: One new thought forms from the combination of two old thoughts.}
	\label{fig:cdt-mbuild-merge}
	\end{center}
\end{figure}

See Figure~\ref{fig:cdt-mbuild-merge}.

\subsubsection{Overview of comparison}

\textbf{1. Type of Primitive Acts}

CD theory is said to have failed because its small number of Primitive Acts could not possibly cover everything in the real world that might need to be represented (Lytinen 1992, p. 51). Tumbug also has a relatively small number of Building Blocks but many of these Building Blocks are based on physics, not on human-oriented issues such ingestion or thinking, and since physics is the ultimate study of the physical universe, physics can presumably represent anything of interest, therefore Tumbug's Building Blocks almost cannot fail at representing the real world. Admittedly, for complicated situations Tumbug descriptions can become unwieldy, but that is why Tumbug labels exist: as a convenient shorthand notation for complicated descriptions or diagrams, the same shorthand that nearly every AI system already uses, such as rule-based expert systems and semantic networks.

\begin{figure}
	\begin{center}
	\includegraphics[width=0.65\textwidth]{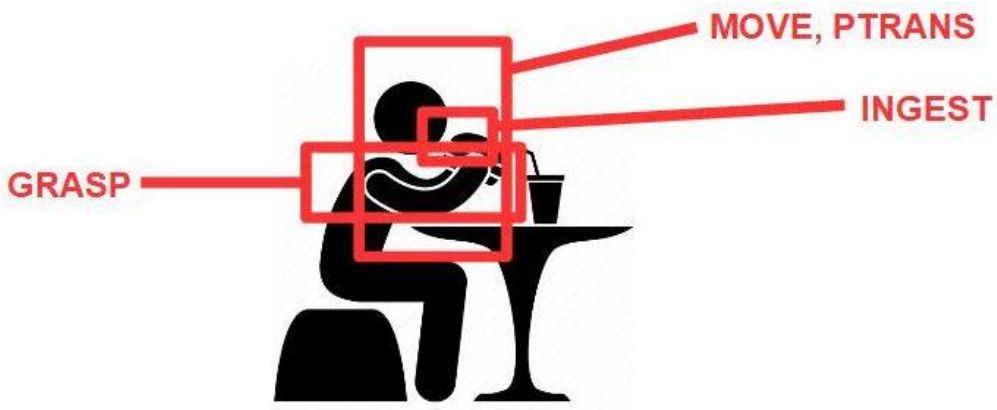}
	\caption{CD theory uses Primitive Acts that are human-oriented and are best at describing complicated actions.}
	\label{fig:krm-comparison-tumbug-cdt-cdt}
	\end{center}
\end{figure}

\begin{figure}
	\begin{center}
	\includegraphics[width=0.65\textwidth]{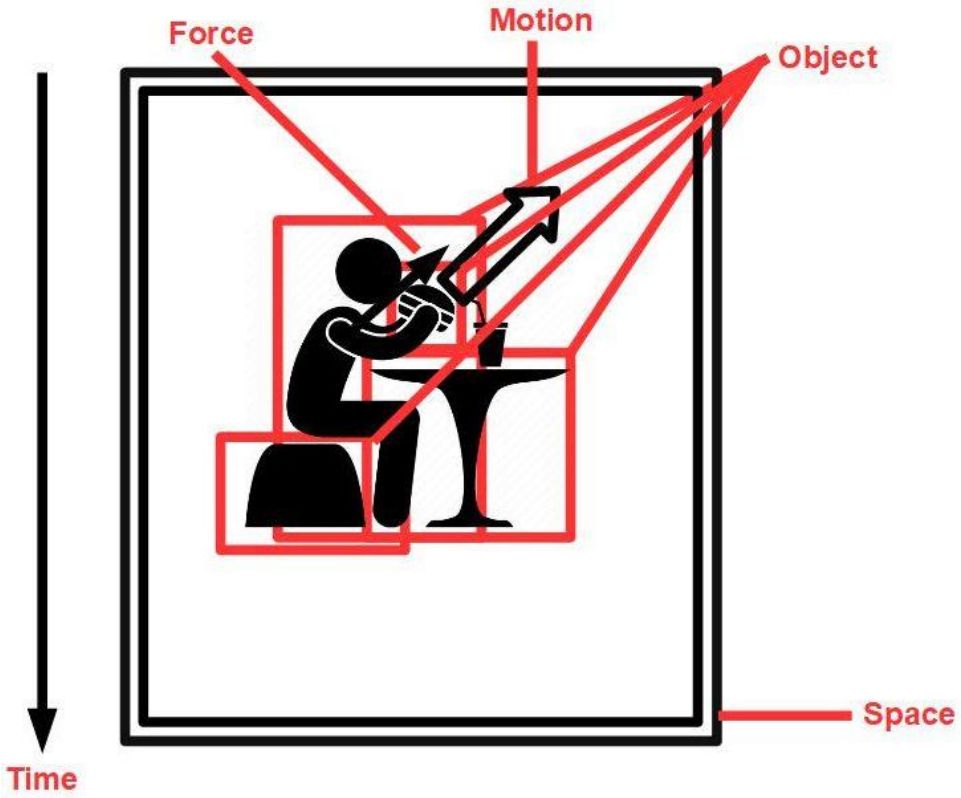}
	\caption{Tumbug uses Building Blocks that are physics-oriented and are best at describing simple objects and actions.}
	\label{fig:krm-comparison-tumbug-cdt-tumbug}
	\end{center}
\end{figure}

Figure~\ref{fig:krm-comparison-tumbug-cdt-cdt} and Figure~\ref{fig:krm-comparison-tumbug-cdt-tumbug} might illustrate more clearly how CD theory and Tumbug compare. The basic diagram in the form of black-and-white icons showing a seated person eating a hamburger is the same for each, but the KRM influences how the information in the diagram is divided. Red text is used to the label categories according to the KRM being used. 

Note that in this case CD theory is excellent for describing the human-oriented actions of grasping and eating, but many other actions such as sitting or placing one's body partially around an object (e.g., a table) could not be included unless a huge number of additional Primitive Acts were used. A huge drawback of CD theory is not obvious from the ingestion figure unless one remembers that CD theory \textit{per se} cannot show diagrams at all, or even spatial relationships. In CD theory only rigidly formatted textual descriptions would be used to describe what a user can so easily see in a diagram.

One of the main CD Theory terms is "Primitive Conceptual Categories," which is also called "conceptual primitives," "conceptual dependency primitives," or "instrumental conceptualizations," depending on the author. The list of these Primitive Conceptual Categories is \{PP, ACT, PA, AA, LOC, T\}, and this is the closest set of concepts to Tumbug's icons that are called "Building Blocks," whose term is based on CD Theory terminology. The other main CD Theory term is "ACT," which is also called "Primitive Acts." The list of these Primitive Acts is \{ATRANS, PTRANS, PROPEL, MTRANS, MBUILD, SPEAK, ATTEND, MOVE, GRASP, INGEST, EXPEL\}, so these components are instantiations of the single Primitive Conceptual Category called "ACT." In other words, both CD theory and Tumbug use the term "building blocks" to include both major sets of CD Theory, namely the Primitive Acts and the specific acts of ACT, though in this document "building blocks" (uncapitalized) is used for CD theory and "Building Blocks" (capitalized) is used for Tumbug.

Note that in this case Tumbug spans everything that could possibly be going on in the diagram, from the objects to the forces acting on them to their resulting motions (i.e., Tumbug is "more universal." In mathematical terminology, Tumbug "spans" a larger space, but since Tumbug labels are simpler in concept, a larger quantity of the same Tumbug labels will usually be needed. More labels and boxes will tend to clutter the diagram. Also, the forces and motions involved (i.e., using arms to move the hamburger) are small in comparison to the overall size of the diagram, so those components may be difficult to see in larger or more complicated events. (This is why the Zoom ability of Tumbug was introduced: to allow arbitrary zoom in of smaller objects, at the user's discretion.) The overwhelming visual advantage of Tumbug is not seen in such a static icon, however: Tumbug can literally show realistic motion of the objects, like a simulation, whereas CD theory has no mechanism at all to allow even one moving object to be displayed.

Some similarities between CD theory and Tumbug:

\begin{itemize}
	\item
		Both use around 1-2 dozen building blocks.
	\item
		Some of the building blocks are identical, and those that are not identical are similar.
	\item
		All slots and Primitive Acts in CD theory can be represented by Tumbug.
	\item
		Both use aggregation.
	\item
		Both use abstract objects, and treat them slightly differently from concrete objects.
\end{itemize}

Some differences between CD theory and Tumbug:

\begin{itemize}
	\item
		Tumbug is based largely on physics and math; CD theory is based on human-oriented actions.
	\item
		Tumbug is almost completely visual; CD theory is text-based.
	\item
		Tumbug can display images, even moving images; CD theory cannot display images at all.
	\item
		Tumbug can represent phrases; CD theory needs a complete sentence with actor and action.
	\item
		CD theory Primitive Acts cannot cover every possibility; Tumbug Building Blocks so far appear to be able to do so, at least for WS150 problems.
	\item
		CD theory requires separate text to describe TO versus FROM; Tumbug needs no such distinction, other than arrow placement.
	\item
		CD theory distinguishes static objects (PP) from Primitive Acts (ACT) on those objects, though it is possible in Tumbug to define an object so that it changes state or behavior via an internal Correlation Box without dictating those changes explicitly. (This Tumbug capability is not described in this document, however.)
	\item
		Tumbug's Building Blocks generalize to a small set of (five) Basic Building Blocks, but CD theory does not note any such generalizations for its Primitive Acts (ACT).
\end{itemize}

To a great extent, Tumbug appears to be what CD theory was attempting, but Roger Schank used non-universal building blocks instead of considering physics. Similarly, although not discussed in this document, Tumbug appears to be what formal language theory was attempting, but Noam Chomsky used traditional text instead of images. Tumbug potentially infuses new life into these earlier milestone ideas, so perhaps Tumbug also has a chance of becoming a milestone.

\textbf{2. Nearness to a one-to-one mapping to Tumbug concepts}

\begin{figure}
	\begin{center}
	\includegraphics[width=1.00\textwidth]{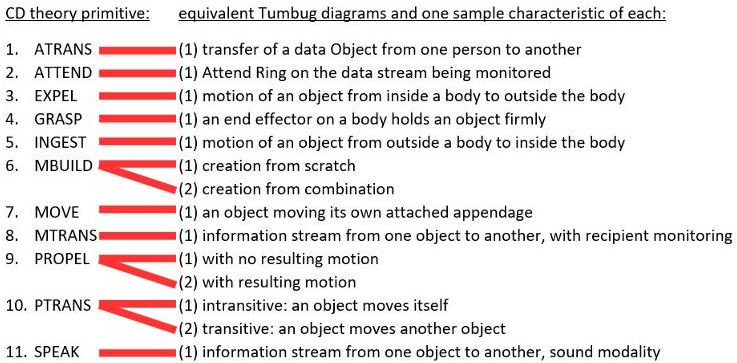}
	\caption{73\% of CD theory Primitive Acts map one-to-one to an equivalent Tumbug concept.}
	\label{fig:cdt-summary-primitive-acts-list-ANN}
	\end{center}
\end{figure}

As seen in Figure~\ref{fig:cdt-summary-primitive-acts-list-ANN}, CD theory concepts are defined in a way that typically are represented by only one Tumbug concept. This is mostly the result of relatively clear-cut definitions of CD theory's Primitive Acts, not necessarily due to a similar foundation of building blocks between the two systems. 8/11 = 73\% of CD theory Primitive Acts map one-to-one to a Tumbug concept.

\subsection{English grammar representation}

\subsubsection{Kolln sentence patterns}

\begin{figure}
	\begin{center}
	\includegraphics[width=0.75\textwidth]{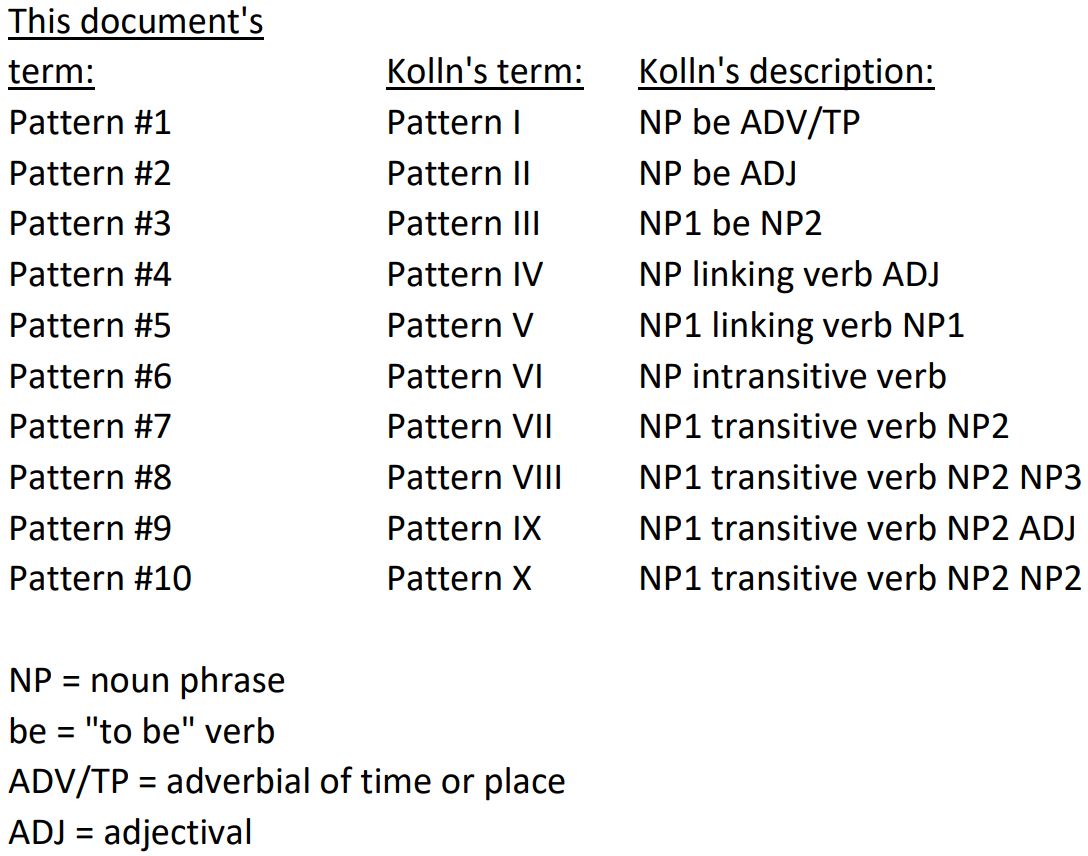}
	\caption{Kolln's sentence patterns with the order and abbreviations used by Kolln.}
	\label{fig:kolln-patterns-list-summary-snap}
	\end{center}
\end{figure}

Martha Kolln listed 10 basic English sentence patterns that she claims constitute the vast majority of sentence patterns that appear in English (Kolln and Funk 2006). Kolln's 10 patterns are shown in Figure~\ref{fig:kolln-patterns-list-summary-snap}. Kolln numbered these by Roman numerals, whereas this documented numbered these by Arabic numerals.

Each of these patterns is described with some sample sentences converted to Tumbug in the following sections.

It must be understood that Kolln's statement applies to (1) Only basic sentence patterns, since sentences of different types could always be combined via conjunctions to create a much larger set of types through combinations of the basic types, (2) Only statement versions of the sentence patterns, since question versions of these patterns may alter the word order; (3) Non-interjections, since otherwise an interjection consisting of a single word would qualify as having correct grammar, which it does not.

The standard sentence diagram that was widely taught in the 1960s and 1970s, formally called the Kellogg-Reed system, is much better known than the Kolln patterns. However, the formal documentation of the Kellogg-Reed diagrams with respect to possible combinations of values within the slots for subject complement (SC), direct object (DO), indirect object (IO), and object complement (OC) is so weak that the Kellogg-Reed system is ignored here, at least for comparison purposes.

\textbf{1. NP be ADV/TP}

Both cases are of ADV/TP are shown next, one where the adverbial is of time (ADV/T), one where the adverbial is of place (ADV/P).

\textbf{1.1. With Time Arrow (ADV/T)}

\begin{figure}
	\begin{center}
	\includegraphics[width=0.30\textwidth]{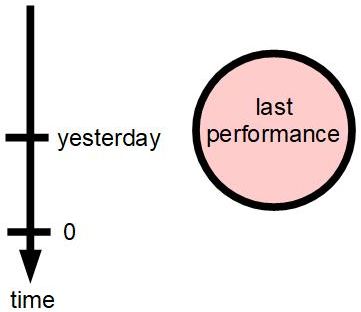}
	\caption{[Kolln Pattern \#1] Tumbug for "The last performance was yesterday (ADV/T)."}
	\label{fig:kolln-01-time}
	\end{center}
\end{figure}

See Figure~\ref{fig:kolln-01-time}.

\textbf{1.2. With Location Box (ADV/P)}

\begin{figure}
	\begin{center}
	\includegraphics[width=0.55\textwidth]{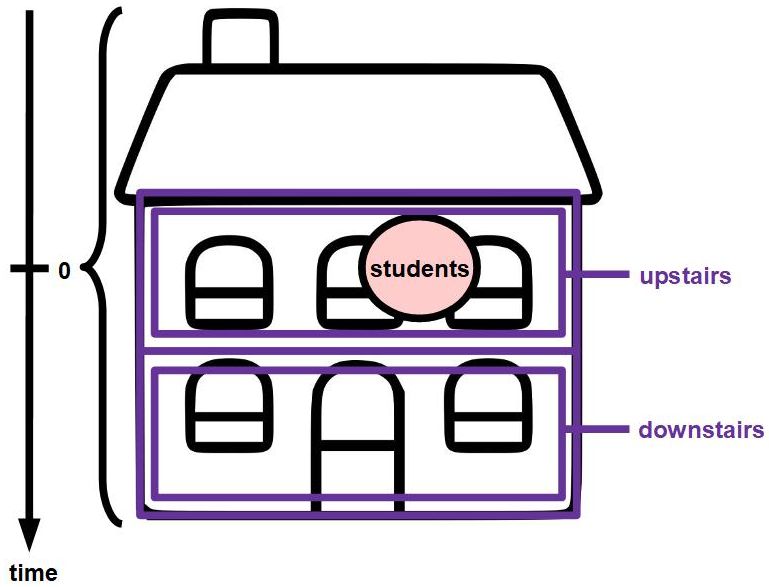}
	\caption{[Kolln Pattern \#1] Tumbug for "The students are upstairs (ADV/P)."}
	\label{fig:kolln-01-upstairs}
	\end{center}
\end{figure}

See Figure~\ref{fig:kolln-01-upstairs}.

\textbf{2. NP be ADJ}

\begin{figure}
	\begin{center}
	\includegraphics[width=0.25\textwidth]{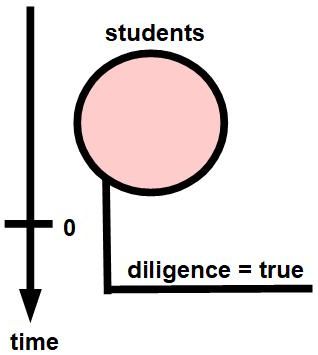}
	\caption{[Kolln Pattern \#2] Tumbug for "The students are diligent."}
	\label{fig:kolln-02}
	\end{center}
\end{figure}

This Kolln pattern equates only to an object (O) having an attribute (A) with a value (V), which Tumbug renders in OAV fashion as shown in Figure~\ref{fig:kolln-02}.

\textbf{3. NP1 be NP2}

\textbf{3.1. As superset}

\begin{figure}
	\begin{center}
	\includegraphics[width=0.25\textwidth]{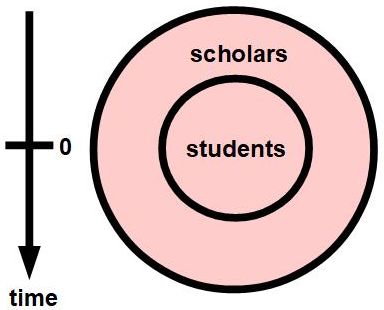}
	\caption{[Kolln Pattern \#3] Tumbug for "The students are scholars."}
	\label{fig:kolln-03-scholars}
	\end{center}
\end{figure}

\begin{figure}
	\begin{center}
	\includegraphics[width=0.33\textwidth]{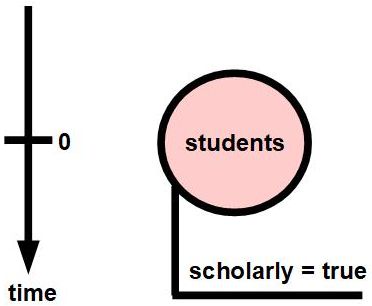}
	\caption{[Kolln Pattern \#3] Tumbug for "The students are scholarly."}
	\label{fig:kolln-03-scholarly}
	\end{center}
\end{figure}

This Kolln pattern demonstrates a principle mentioned by Roger Schank regarding the need for a canonical form of a sentence when sentence meaning is represented (Schank 1976, p. 172). In this case, there are two simple ways to express the same meaning of the sentence "The students are diligent.":

\begin{itemize}
	\item
		"The students are scholars." - Involves two noun phrases (NP1 = students, NP2 = scholars). This sentence is represented in Tumbug as shown in Figure~\ref{fig:kolln-03-scholars}.\\
	\item
		"The students are scholarly." - Involves one noun phrase (NP1). and one attribute value (scholarly = true). This sentence is represented in Tumbug as shown in Figure~\ref{fig:kolln-03-scholarly}.
\end{itemize}

\begin{figure}
	\begin{center}
	\includegraphics[width=0.50\textwidth]{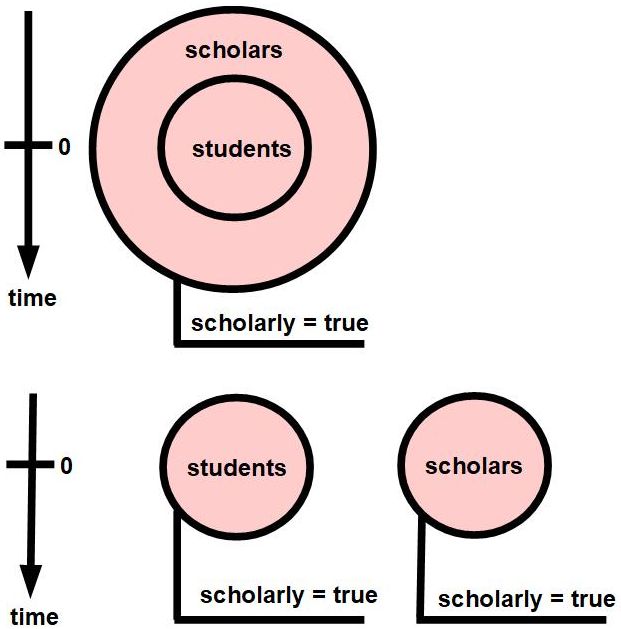}
	\caption{[Kolln Pattern \#3] Two Tumbug ways to represent the concept that "The students are scholarly."}
	\label{fig:kolln-03-comparison-top-2}
	\end{center}
\end{figure}

One of the main goals of representation here is to use only one representation for the concept of "scholars/scholarly," which suggests that only one of these two diagrams should be declared the canonical form. Figure~\ref{fig:kolln-03-comparison-top-2} shows more clearly how the two previous diagrams are related. (The last two diagrams are essentially the same diagram, but one has the common attribute "scholarly = true" duplicated, the other does not.)

\begin{figure}
	\begin{center}
	\includegraphics[width=0.50\textwidth]{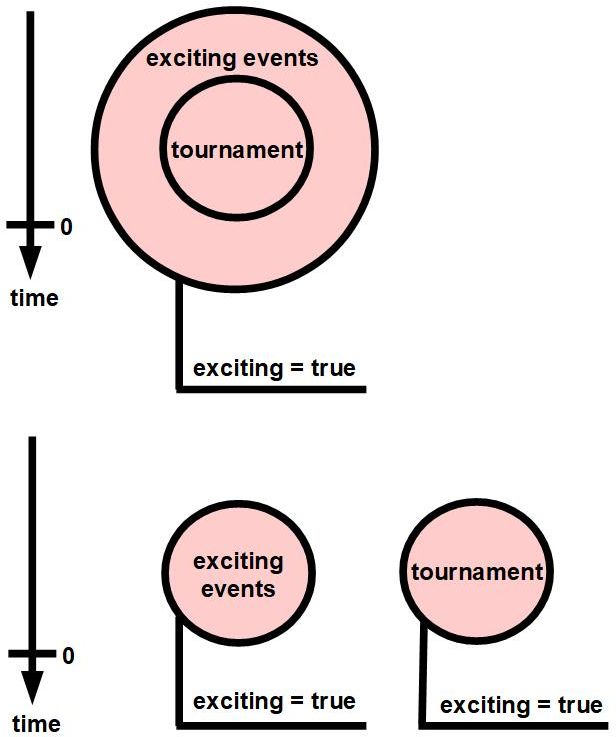}
	\caption{[Kolln Pattern \#3] Two Tumbug ways to represent the concept that "The tournament was an exciting event." Top diagram: One attribute applies to both the set and superset. Bottom diagram: No superset/subset structure, but note that the same attribute appears in both sets.}
	\label{fig:kolln-03-tournament}
	\end{center}
\end{figure}

Currently the Tumbug convention used is to use the superset representation. A similar sentence they has two representation options is diagrammed in Figure~\ref{fig:kolln-03-tournament}.

Note: The use of the value "true" in the aforementioned expression "scholarly = true" is a quick way to conform to OAV notational convention without need for additional thought, because the attribute-value form is required to be "$\langle$attribute$\rangle$ = $\langle$value$\rangle$", which the expression "scholarly" alone would not fulfill, and to think of which attribute would allow "$\langle$attribute$\rangle$ = scholarly" to make sense (e.g., "learnedness = scholarly", or "academic character = scholarly") is more difficult than merely writing "scholarly = true."

\textbf{3.2. With removed possessive adjective}

\begin{figure}
	\begin{center}
	\includegraphics[width=0.75\textwidth]{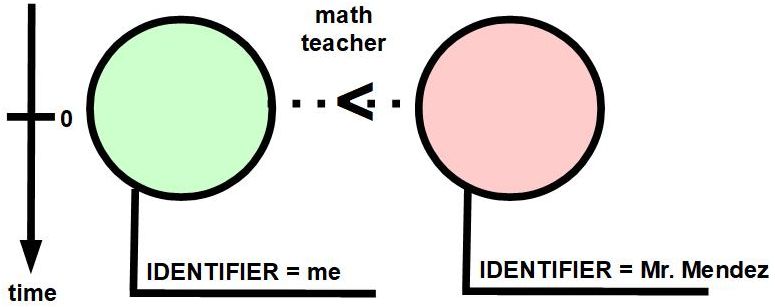}
	\caption{[Kolln Pattern \#3] Tumbug for "Professor Mendez is my math teacher." = "Professor Mendez IS-A math teacher OF me."}
	\label{fig:kolln-03-mendez-circles}
	\end{center}
\end{figure}

Another example said to be Kolln Pattern \#3 is "Professor Mendez is my math teacher" (Kolln and Funk 2006, p. 31). This is another example of the inaccuracies of the English language, since the verb "to be" (conjugated here as "is") has a different meaning than before. Here it would be wasted effort to create a set called "my math teachers" and to put "Professor Mendez" as a single instance of that set, since usually a student has only one math teacher. Therefore "is" does not mean "is a subset of" here. In this case the underlying problem is use of a possessive adjective, which should be removed from a sentence before converting the sentence to Tumbug. There exist multiple ways to represent this sentence without the possessive adjective, one of which is shown in Figure~\ref{fig:kolln-03-mendez-circles}. This is an indication of weak organizational criteria used in defining the Kolln patterns.

\textbf{4. NP linking verb ADJ}

\textbf{4.1. With propositional attitude}

This is another Kolln pattern that incorporates very different Tumbug structures if the verb changes slightly. Many Kolln Pattern \#4 verbs involve general, average impressions, such as "to seem" and "to look," others involve specific sensory perceptions such as "to taste," "to smell," "to sound," and others use neither, such as "to grow" (meaning "to become"), which has a very different meaning: a single attribute that changes over time, which has nothing to do with impressions or sensory modalities. Therefore this Kolln category is not taken very seriously with respect to Tumbug, other than as a supply of possibly unusual examples to test against Tumbug representation.

\begin{figure}
	\begin{center}
	\includegraphics[width=0.50\textwidth]{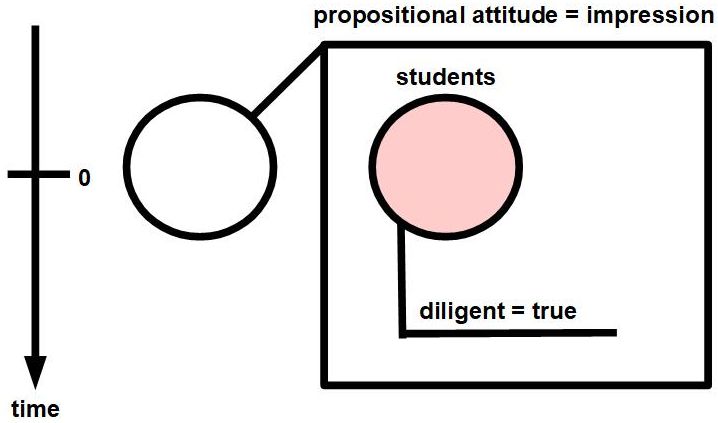}
	\caption{[Kolln Pattern \#4] Tumbug for "The students seem diligent."}
	\label{fig:kolln-04-pa-seem}
	\end{center}
\end{figure}

Figure~\ref{fig:kolln-04-pa-seem} shows an example of Kolln Pattern \#4 when the linking verb is a propositional attitude, in this case "seem." English allows an ambiguity here, regarding the entity (or group) who is interpreting the diligence of the student. The only logical way to represent this situation is to show an Object Circle that is unlabeled, as shown in the figure. To show a single person icon would assume that only one person is doing the interpreting, to show an "average person" icon would assume that the interpreter is average, to show a person icon at all would assume that the interpreter is human, and to show a "DK" (= Don't Know) wildcard would assume that the reader cares, which might entail extra, unnecessary, ongoing work in labeling, especially on a very large network of connected Tumbug icons that no one is likely to care about.

The term "propositional attitude" is from Davis 1990 (ch. 8), and the need for this concept occurs in three of the Kolln patterns (Pattern \#4, Pattern \#9, Pattern \#10). It is a very useful concept that unifies several modalities of knowledge. For details, see the section on propositional attitudes in this document.

\textbf{4.2. With attribute changing over time}

\begin{figure}
	\begin{center}
	\includegraphics[width=0.50\textwidth]{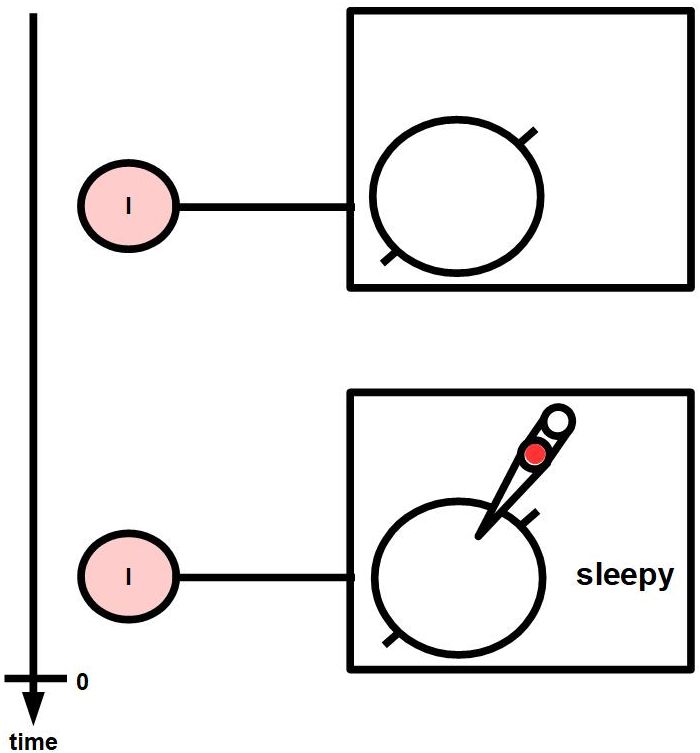}
	\caption{[Kolln Pattern \#4] Tumbug for "I grew sleepy." = "I transitioned into the sleepy state."}
	\label{fig:kolln-04b-grew}
	\end{center}
\end{figure}

An example of a Pattern \#4 example that has a very different meaning and Tumbug diagram is in Figure~\ref{fig:kolln-04b-grew}. In this case the only thing happening is that an attribute value changes in time, which has nothing to do with sensory input or overall impressions.

\textbf{4.3. With sensory perceptions}

\begin{figure}
	\begin{center}
	\includegraphics[width=0.50\textwidth]{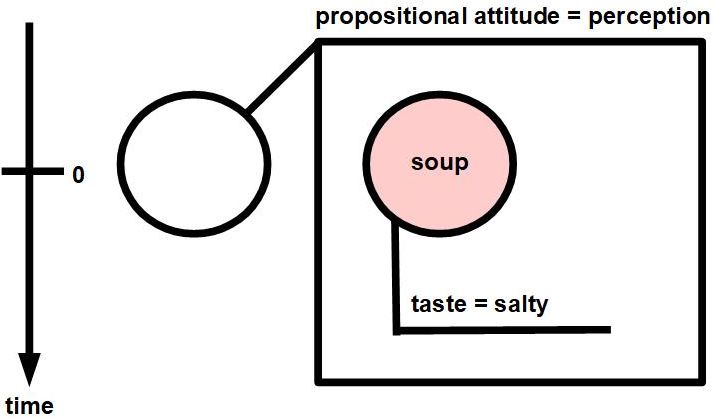}
	\caption{[Kolln Pattern \#4] Tumbug for "The soup tastes salty."}
	\label{fig:kolln-04-pa-tastes}
	\end{center}
\end{figure}

Figure~\ref{fig:kolln-04-pa-tastes} illustrates another example of a propositional attitude, in this case gustatory taste, which generalizes to any type of sensory perception.

\textbf{5. NP1 linking verb NP1}

\textbf{5.1. As superset}

\begin{figure}
	\begin{center}
	\includegraphics[width=0.25\textwidth]{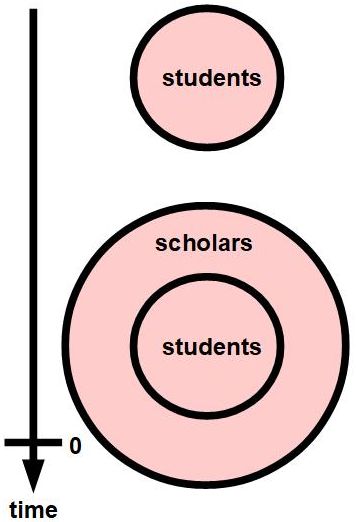}
	\caption{[Kolln Pattern \#5] Tumbug for "The students became scholars."}
	\label{fig:kolln-05-scholars}
	\end{center}
\end{figure}

See Figure~\ref{fig:kolln-05-scholars}.

\textbf{5.2. With propositional attitude}

\begin{figure}
	\begin{center}
	\includegraphics[width=0.50\textwidth]{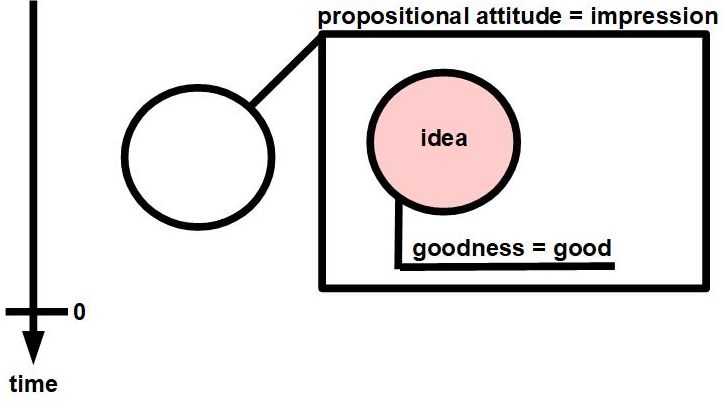}
	\caption{[Kolln Pattern \#5] Tumbug for "That seemed a good idea."}
	\label{fig:kolln-05-impression}
	\end{center}
\end{figure}

See Figure~\ref{fig:kolln-05-impression}.

\textbf{5.3. With object changing over time}

Kolln does not document the possibility of an object changing into another object over time. The sentence "The caterpillar became a butterfly" uses the same verb and tense as before (viz., "became") in the sentence "The students became scholars" but in this case "became" implies the object itself has changed, not just its classification. This is an example of how the Kolln patterns are based too much on specific verb categories rather than on the meanings of those words in different contexts.

\begin{figure}
	\begin{center}
	\includegraphics[width=0.20\textwidth]{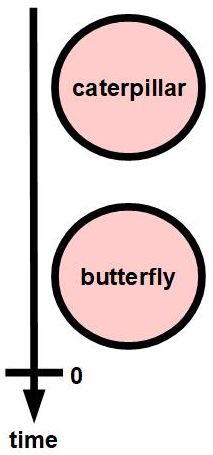}
	\caption{[Kolln Pattern \#5] Tumbug for "The caterpillar became a butterfly."}
	\label{fig:kolln-05-butterfly}
	\end{center}
\end{figure}

In contrast, Tumbug can easily represent an unmoving object changing into another unmoving object in the same location by placing the two objects above each other along the timeline, with the older object at an earlier time than the newer object. See Figure~\ref{fig:kolln-05-butterfly}. Since two solid objects cannot occupy the same space in our physical world, such representation is sufficient, \textit{pro tempore}.

\begin{figure}
	\begin{center}
	\includegraphics[width=0.50\textwidth]{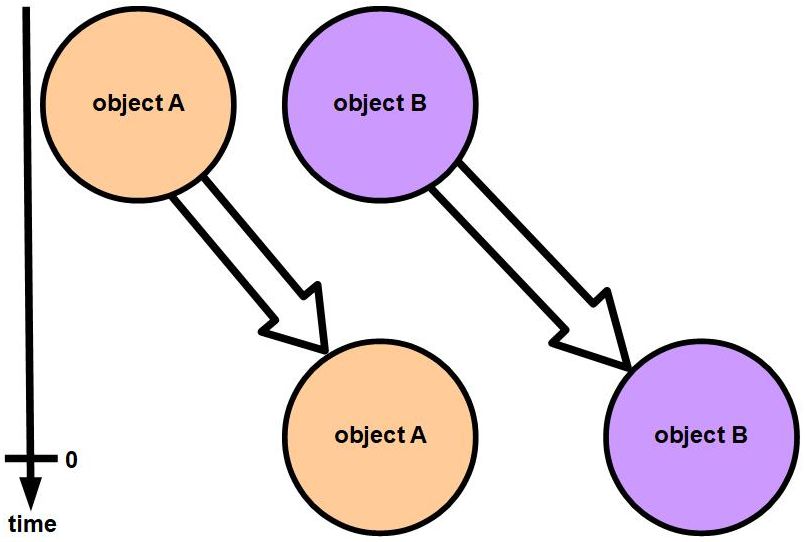}
	\caption{In the real world of physical objects, Object A may occupy Object B's space later in time only if Object B has moved out of the way.}
	\label{fig:graphics-non-coincidence}
	\end{center}
\end{figure}

This simplicity comes at a cost of increased risk when creating diagrams, however: Care must be taken not to place two different objects above each other in a timeline unless one has moved out of the way completely. See Figure~\ref{fig:graphics-non-coincidence}.

\textbf{6. NP intransitive verb}

Intransitive verbs are another category of verb that is very poor for capturing a general meaning of such a verb, as evidenced by the two examples below that use this same sentence pattern and verb class but have two very different meanings and Tumbug diagrams.

\textbf{6.1. With states}

\begin{figure}
	\begin{center}
	\includegraphics[width=0.50\textwidth]{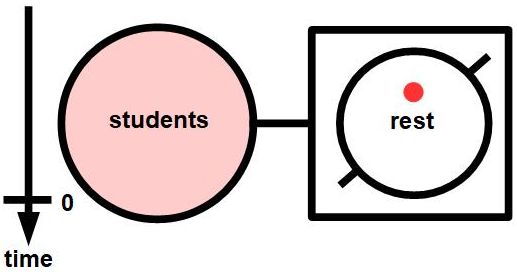}
	\caption{[Kolln Pattern \#5] Tumbug for "The students rested." = "The students were in rest state."}
	\label{fig:kolln-05-rest}
	\end{center}
\end{figure}

Note that in the Kolln example "The students rested" (Kolln and Funk 2006, p. 35) the verb "rested" is ambiguous as to whether the rest was physical or cognitive. Fortunately, a general state diagram such as one for "exertion versus rest" (as shown in Figure~\ref{fig:kolln-05-rest}) can subsume both possibilities at once. Note also that the sentence given does not necessarily state that the state of the students before the rest state was the exertion state: the students could have just come from an earlier break, then decided to take a later break. Therefore only the later specified state ("rest") is marked with an asterisk as known; the state diagram exists in the earlier depiction, but no state is marked. This detail is very important to note when developing a matching algorithm for Tumbug.

\textbf{6.2. With motion}

\begin{figure}
	\begin{center}
	\includegraphics[width=0.50\textwidth]{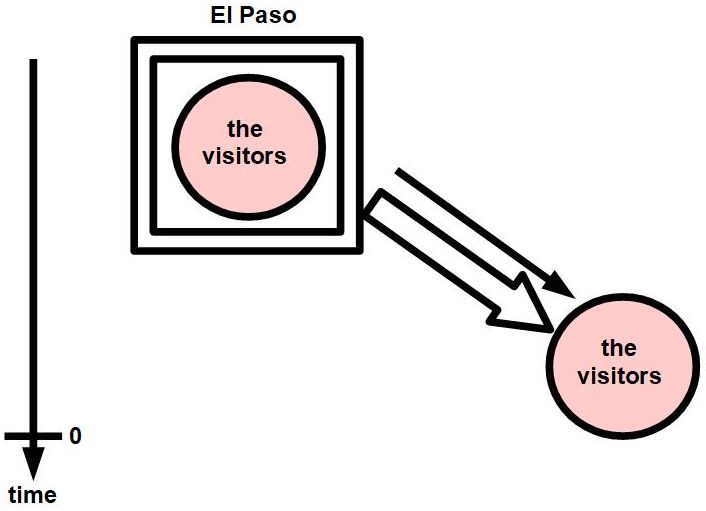}
	\caption{[Kolln Pattern \#6] Tumbug for "The visitors from El Paso arrived."}
	\label{fig:kolln-06-arrived}
	\end{center}
\end{figure}

Intransitive verbs can also refer to motion, as in the verb "to arrive" as shown in Figure~\ref{fig:kolln-06-arrived}.

\textbf{7. NP1 transitive verb NP2}

Even Martha Kolln herself makes a heavy admission about the nature of transitive verbs when discussing her sentence Pattern \#7 (= Pattern VII) of hers (Kolln and Funk 2006, p. 39):\\

"Traditionally, we think of the transitive verb as an action word: Its subject is considered the doer and its object the receiver of the action. In many Pattern VII sentences this meaning-based definition applies fairly accurately. In our Pattern VII sample sentences, for instance, we can think of their assignment as the receiver of the action studied and a home run as a receiver of the action hit. But sometimes the idea of receiver of the action doesn't apply at all:\\

\setlength\parindent{0pt}
Our team won the game.\\
We enjoyed the game.\\
\setlength\parindent{24pt}

It hardly seems accurate to say that game "receives the action.""

This equates to at a deep insight about verbs in all natural languages: Verbs do not generalize to a single concept (e.g., not motion, action, or relationship), which suggests the modern concept of a "verb" is flawed. The term "verb" is overloaded with the attempt to include references to attribute values, supersets, actions upon a target object, sensory input to the actor, states, cause and effect, and possibly more. The Tumbug viewpoint explains this problem as poor generalizations of language, leading to a poor knowledge representation system for language (such as the Kellogg-Reed system or modern tree diagrams of sentences).

\textbf{7.1. With data transfer}

\begin{figure}
	\begin{center}
	\includegraphics[width=0.50\textwidth]{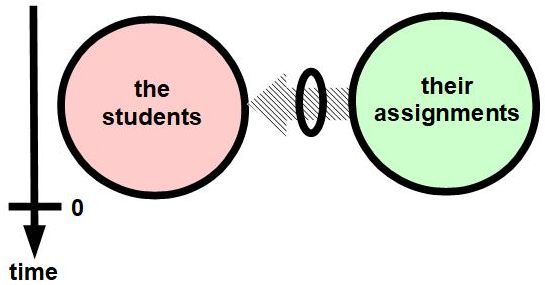}
	\caption{[Kolln Pattern \#7] Tumbug for "The students studied their assignments."}
	\label{fig:kolln-07-studied}
	\end{center}
\end{figure}

Sometimes Pattern \#7 applies to data transfer and/or perception, such as in Figure~\ref{fig:kolln-07-studied}.

\textbf{7.2. With motion and contact}

\begin{figure}
	\begin{center}
	\includegraphics[width=0.50\textwidth]{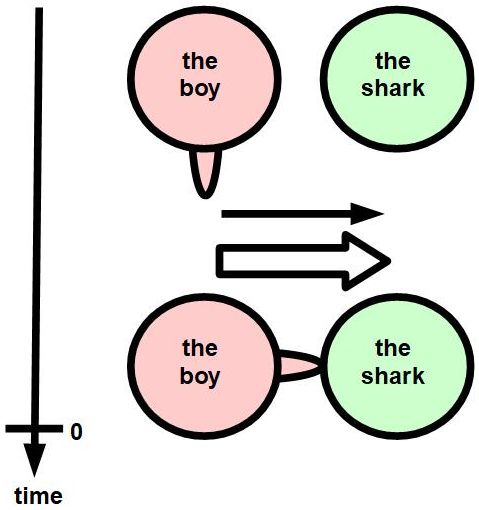}
	\caption{[Kolln Pattern \#7] Tumbug for "The boy touched the shark."}
	\label{fig:kolln-07-shark}
	\end{center}
\end{figure}

The most obvious applicability of Pattern \#7 is physical contact of the subject with the direct object, as in Figure~\ref{fig:kolln-07-shark}.

\textbf{7.3. With states}

\begin{figure}
	\begin{center}
	\includegraphics[width=0.70\textwidth]{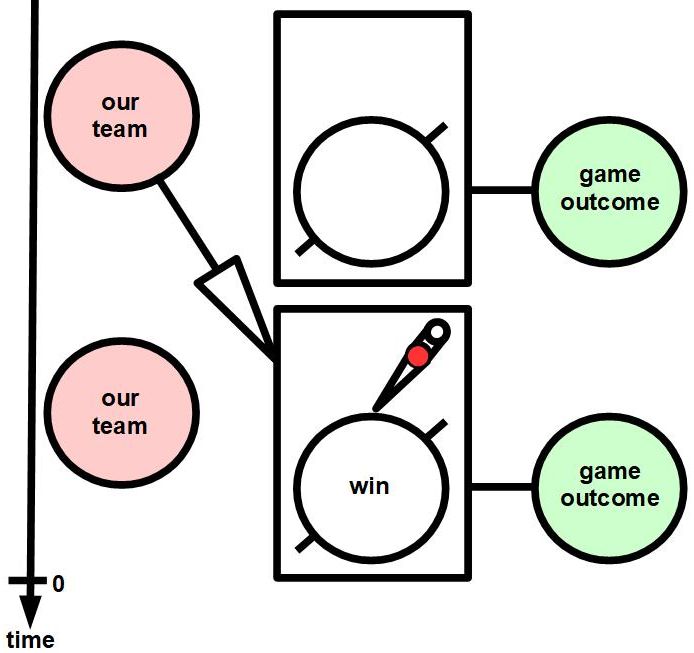}
	\caption{[Kolln Pattern \#5] Tumbug for "Our team won the game" = "Our team caused the game outcome to enter into the win state."}
	\label{fig:kolln-05-won}
	\end{center}
\end{figure}

See Figure~\ref{fig:kolln-05-won}.

\textbf{7.4. With emotions}

\begin{figure}
	\begin{center}
	\includegraphics[width=0.60\textwidth]{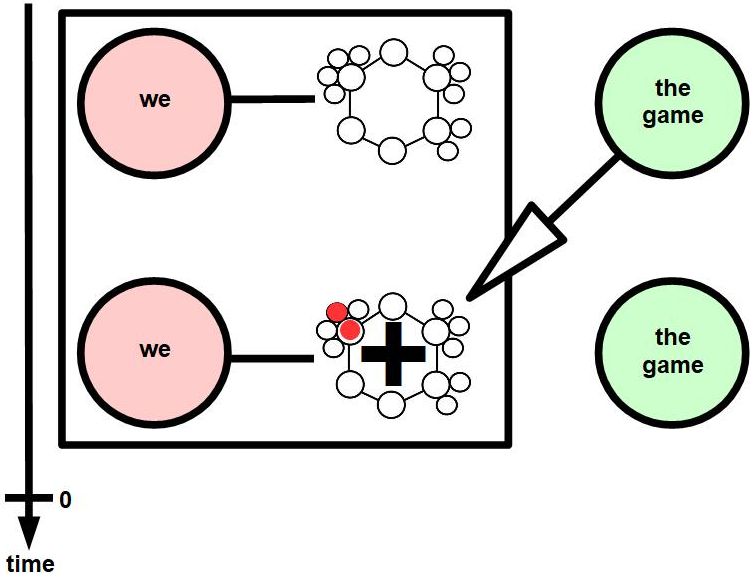}
	\caption{[Kolln Pattern \#7] Tumbug for "We enjoyed the game." = "The game caused our emotion to enter into the joy state."}
	\label{fig:kolln-07-icon-quadrune-as-state}
	\end{center}
\end{figure}

See Figure~\ref{fig:kolln-07-icon-quadrune-as-state}. This figure combines what is essentially a State Diagram, namely the Robinson Icon, which in this case is in the positive version of the state of Event Related: Joy, elation, triumph, jubilation, which is a cathected emotion with respect to the direct object, namely the game.

\textbf{8. NP1 transitive verb NP2 NP3}

\begin{figure}
	\begin{center}
	\includegraphics[width=0.50\textwidth]{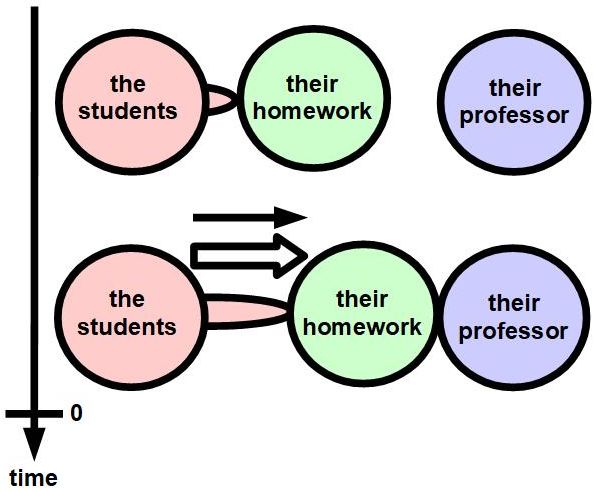}
	\caption{[Kolln Pattern \#8] Tumbug for "The students gave their professor their homework," no shorthand diagram used.}
	\label{fig:kolln-08-no-shortcut-2-parts}
	\end{center}
\end{figure}

It appears that this Kolln pattern always maps only to the same Tumbug diagram of an object being transferred from subject to indirect object, as shown in Figure~\ref{fig:kolln-08-no-shortcut-2-parts}.

\textbf{9. NP1 transitive verb NP2 ADJ}

This Kolln sentence pattern maps to at least two different types of Tumbug diagram.

\textbf{9.1. With propositional attitude}

\begin{figure}
	\begin{center}
	\includegraphics[width=0.55\textwidth]{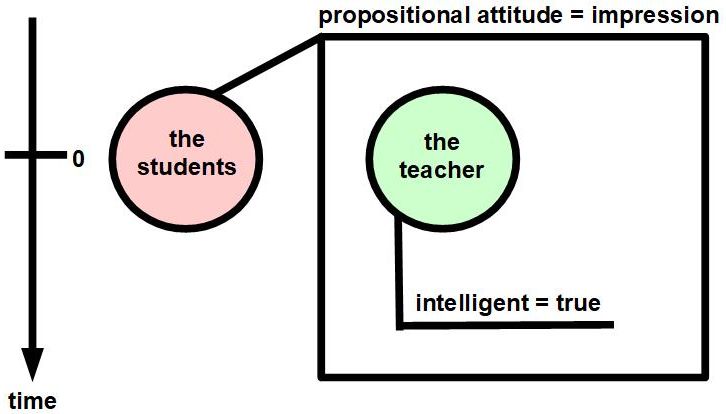}
	\caption{[Kolln Pattern \#9] Tumbug for "The students consider the teacher intelligent."}
	\label{fig:kolln-09-pa}
	\end{center}
\end{figure}

Having a propositional attitude about an object or person does not inherently involve a change of attribute in time. See the example in Figure~\ref{fig:kolln-09-pa}.

\textbf{9.2. With attribute changing over time}

\begin{figure}
	\begin{center}
	\includegraphics[width=0.50\textwidth]{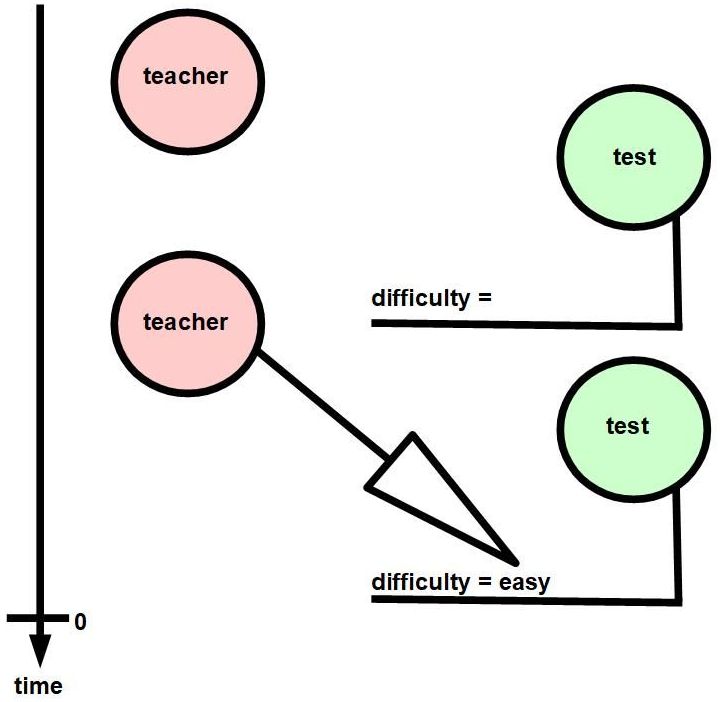}
	\caption{[Kolln Pattern \#9] Tumbug for "The teacher made the test easy."}
	\label{fig:kolln-09-test}
	\end{center}
\end{figure}

An object or person can cause a change of attribute in time. See the example in Figure~\ref{fig:kolln-09-test}.

\textbf{10. NP1 transitive verb NP2 NP2}

This Kolln sentence pattern maps to at least two different types of Tumbug diagram.

\textbf{10.1. With propositional attitude}

\begin{figure}
	\begin{center}
	\includegraphics[width=0.50\textwidth]{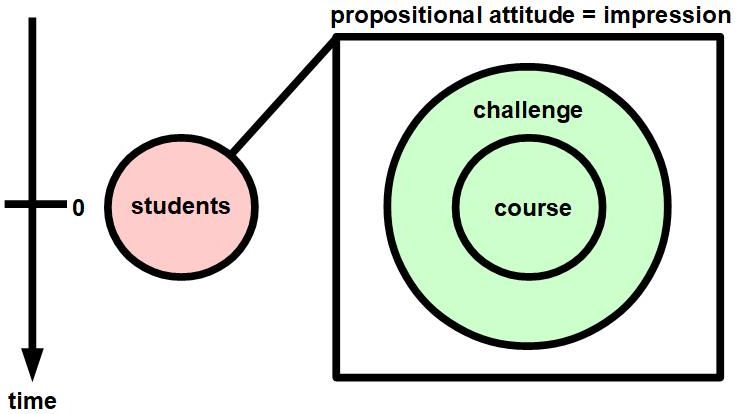}
	\caption{[Kolln Pattern \#10] Tumbug for "The students consider the course a challenge."}
	\label{fig:kolln-10-challenge-students}
	\end{center}
\end{figure}

See Figure~\ref{fig:kolln-10-challenge-students}.

\textbf{10.2. With causation}

\begin{figure}
	\begin{center}
	\includegraphics[width=0.50\textwidth]{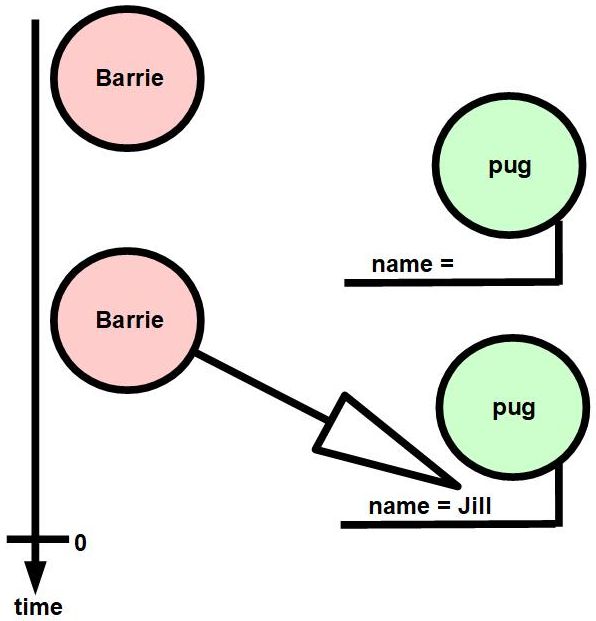}
	\caption{[Kolln Pattern \#10] Tumbug for "Barrie named his pug Jill."}
	\label{fig:kolln-10-pug}
	\end{center}
\end{figure}

See Figure~\ref{fig:kolln-10-pug}.

\textbf{11. Nearness to a one-to-one mapping to Tumbug concepts}

\begin{figure}
	\begin{center}
	\includegraphics[width=0.75\textwidth]{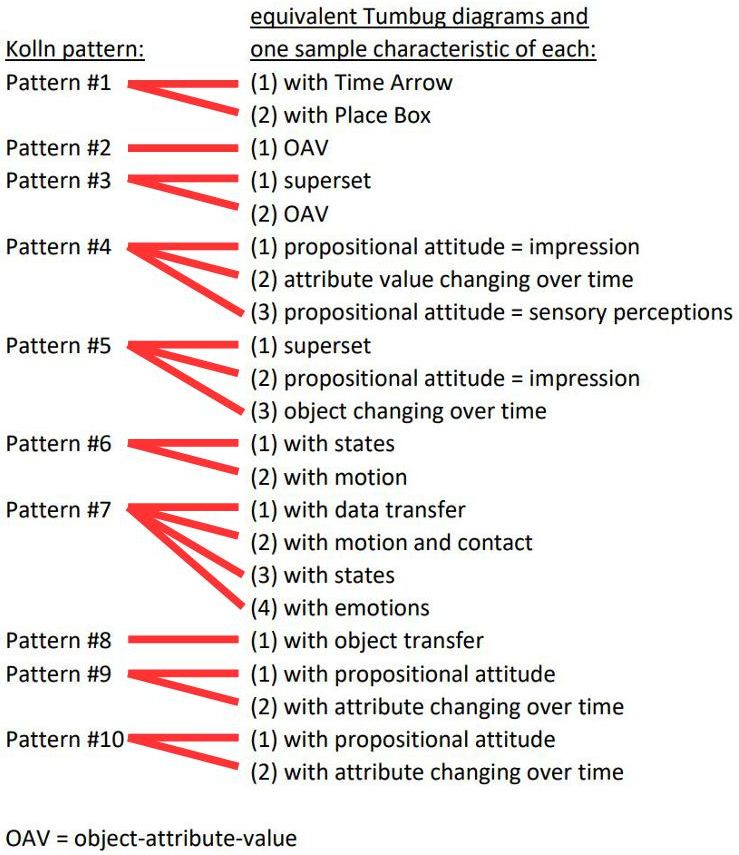}
	\caption{Only 20\% of Kolln patterns map one-to-one to an equivalent Tumbug concept.}
	\label{fig:kolln-tumbug-correlation-summary}
	\end{center}
\end{figure}

Kolln's 10 sentence patterns fail to produce any reasonable approximation of 1-to-1 mapping with Tumbug concepts, as shown in Figure~\ref{fig:kolln-tumbug-correlation-summary}. Kolln's examples are useful, however, as a fairly comprehensive source of sample sentences to test against Tumbug's representation ability. If Kolln's 10 sentence patterns cover 95\% of encountered English sentences, and if Tumbug can represent those same patterns, then Tumbug should also have at least 95\% ability to represent English sentences. 2/10 = 20\% of Kolln patterns map one-to-one to Tumbug concepts.

\textbf{12. The five simplest sentence patterns}

Note that the various examples of Pattern \#6--"The students rested" versus "The visitors from El Paso arrived"--produce diagrams that are very different from one another. The second example describes the arrival of people in motion whereas the first example is very unclear what it describes, even as to whether any motion was involved before the action of "rested." 

This is not the only Kolln grammatical pattern that is highly unpredictable. Pattern \#7--"The students studied their assignments" versus "The boy touched the stingray"--has the same general problem. The second example has a direct object that receives the action whereas the first example has more the nature that the assignment was doing to action to the students.

These anomalies lead to a profound insight into all human language: All human language is structured around the highly inconsistent nature of verbs, rather than being extremely well-structured by basing all grammatical patterns on objects and object motions. The author suggests this is the primary reason why natural language has always been so difficult to describe with grammatical rules, and consequently why the field of AI has always had extra difficulty with natural language. Even a seemingly exotic language such as Japanese, which has particles, has the same two meanings of the verb "to be" as English does, and goes through the same distortions as English in order to force-fit such verbs into inefficient, inconsistent data structures.

\textit{Side conjecture: The Proto-Human language, if such a language ever existed, may have had a needlessly difficult grammar that failed to focus only on objects and motions, so all natural languages since then, no matter how seemingly different they are, may have carried the same flaw that forced descriptions of attribute and superset relationships to use unneeded verbs only because action verb sentences used a structure that required a verb.}

This fundamental problem of human grammar does suggest one way to slightly improve foreign language learning, however: Teach grammar first with only moving objects, up to three in number, then only later move into more abstract verbs that are difficult to represent with moving objects, or with cause-and-effect substituted for motion. Notice that grammar nicely falls into only about five basic patterns if only the common patterns of motion are used.

Consider Pattern \#6, Pattern \#7, and Pattern \#8. From a general standpoint, these few sentence patterns fall into a small, logical set of possible motions of objects:\\

\setlength\parindent{0pt}
Pattern \#2: Subject does not move, has an attribute and value: 1 object + 0 motions\\
Pattern \#3: Subject does not move, is part of a superset: 1 object + 0 motions\\

Pattern \#6: Subject moves itself: 1 object + 1 motion\\
Pattern \#7: Subject moves and makes contact with Direct Object: 2 objects + 1 motion\\
Pattern \#8: Subject transfers Direct Object to Indirect Object: 3 objects + 1 motion\\

Pattern \#1 can be chosen to be ignored, since it deals only with an attribute of space or time.\\
Pattern \#4 can be chosen to be ignored, since it deals with a propositional attitude, which is indirect.\\
Pattern \#5 can be chosen to be ignored, since it deals with a linking verb over time: complicated.\\
Pattern \#9 can be chosen to be ignored, since it deals with a propositional attitude, which is indirect.\\
Pattern \#10 can be chosen to be ignored, since it deals with a propositional attitude, which is indirect.
\setlength\parindent{24pt}

\begin{figure}
	\begin{center}
	\includegraphics[width=0.50\textwidth]{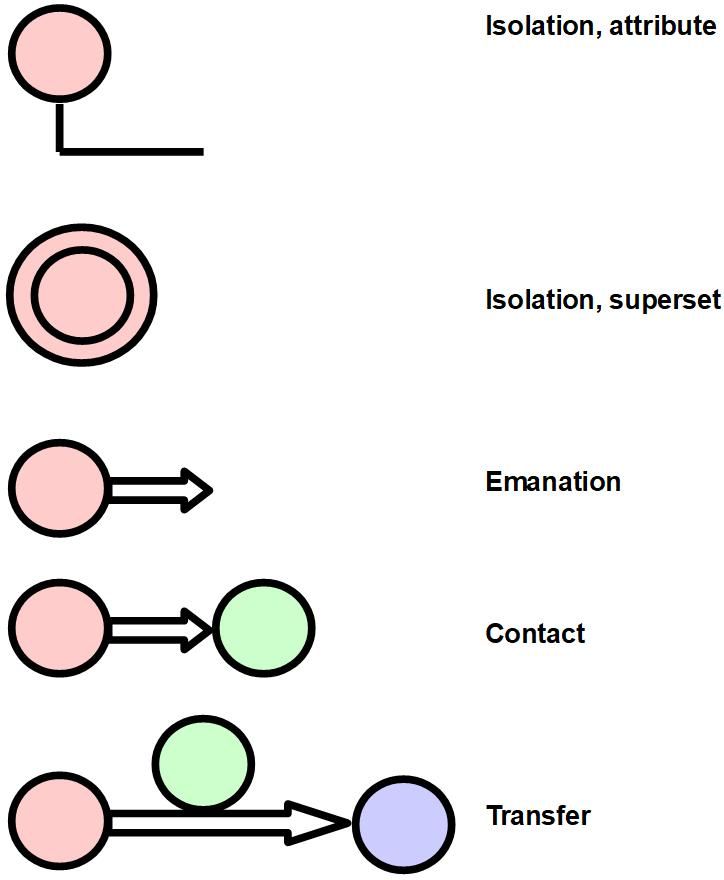}
	\caption{These five sentence patterns are the simplest, most basic, and by far the most common in English. These five patterns are the only patterns discussed in depth in this document.}
	\label{fig:kolln-summary-summary}
	\end{center}
\end{figure}

The summary of these observations is shown in Figure~\ref{fig:kolln-summary-summary}.

\begin{figure}
	\begin{center}
	\includegraphics[width=0.80\textwidth]{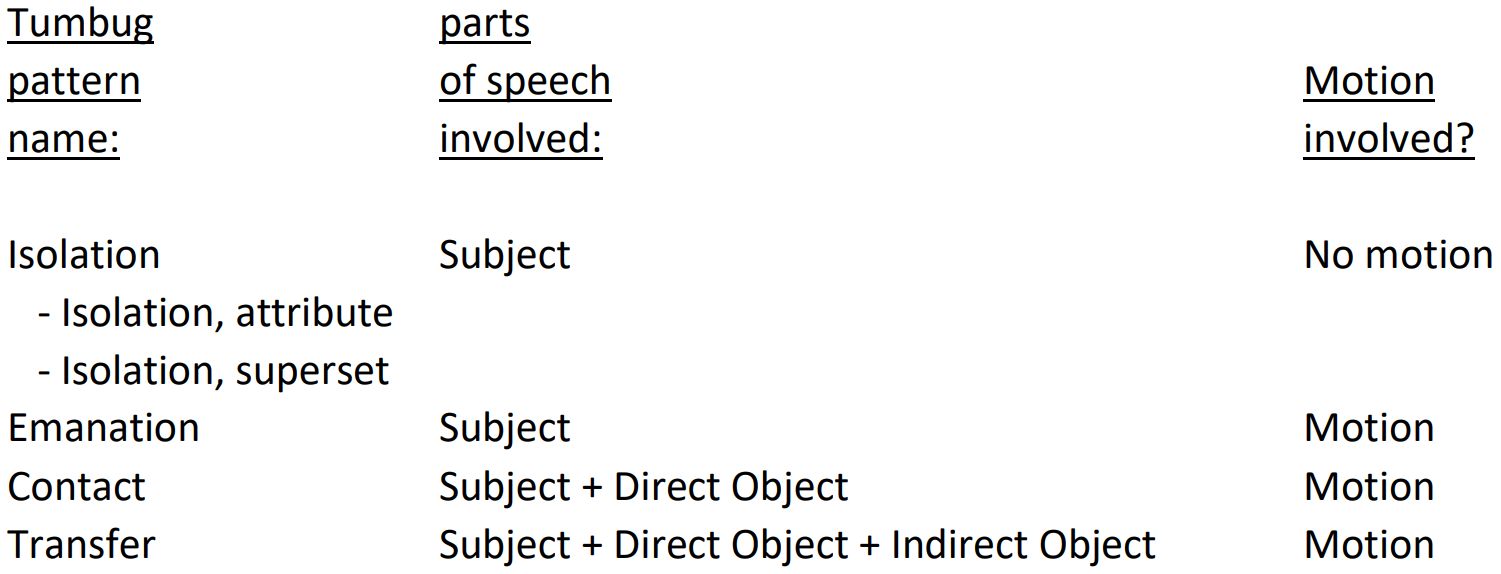}
	\caption{A summary of Tumbug patterns, with regard to Parts of Speech and motion.}
	\label{fig:kolln-patterns-summary-as-tumbug-list-snap}
	\end{center}
\end{figure}

The above heuristics were a lot of generalization for the sole purpose of simplifying human grammar, but this practice could be justified if it sufficiently improves the understanding of grammar, and/or if the simpler patterns are more frequently encountered. The names of these most basic visual sentence patterns are given by the author in Figure~\ref{fig:kolln-patterns-summary-as-tumbug-list-snap}.

An interesting and little-known fact is that Tumbug's Transfer pattern does not exist in Spanish in the same way that it exists in English. For example, the English sentence "The students gave their professor their homework" would need to be grammatically rearranged to a Spanish grammatical pattern with an extra preposition, such as "The students gave their homework to their professor" ("Los estudiantes le dieron sus tareas a su profesor"). This fact illustrates that slight variations in the number of sentence pattern types occur between major languages, but that Tumbug representations remains unchanged between such syntactical rearrangements.

\begin{figure}
	\begin{center}
	\includegraphics[width=0.35\textwidth]{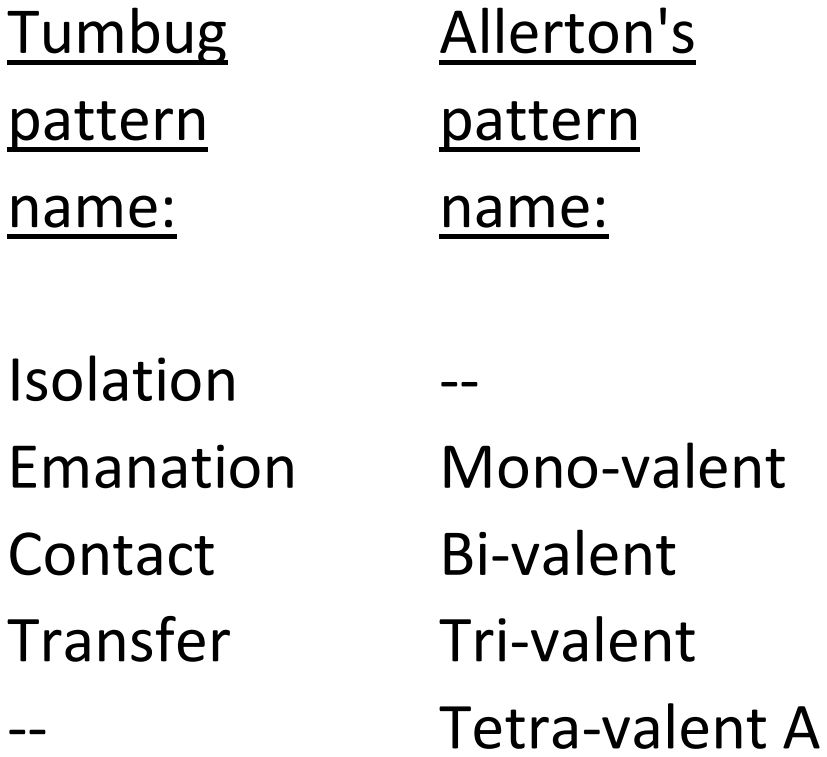}
	\caption{Three of Allerton's valency categories exactly match three of the four basic sentence patterns of Tumbug. Allerton's tetra-valent category of sentence patterns involves four objects, which is beyond the number of Tumbug patterns discussed in detail in this document.}
	\label{fig:kolln-valency-snap}
	\end{center}
\end{figure}

Three of the most basic Tumbug patterns (E, C, T) match three of the English verb valency patterns listed by D.J. Allerton (Allerton 2006, pp. 170-171), which demonstrates that Tumbug's natural grammatical patterns have already been discovered earlier in a different context, therefore Tumbug is already mostly supported by existing grammar theory. These matches are shown in Figure~\ref{fig:kolln-valency-snap}.

\textit{Side conjecture: Human parsing of a sentence, whether that input is written or spoken, may involve in part the operation of matching the grammatical pattern of the input to a small set of known, allowable grammatical patterns. This matching would be a very fast operation because of the small number of possible grammatical patterns involved, each of which has only small structures involved.}

Various problems with the Kolln criteria for classifying English sentences often become evident, and are sometimes noticed by English teachers, therefore further detailed comparison between Tumbug patterns versus Kolln patterns is probably not worthwhile.

\subsubsection{Allerton sentence patterns}

\begin{figure}
	\begin{center}
	\includegraphics[width=0.90\textwidth]{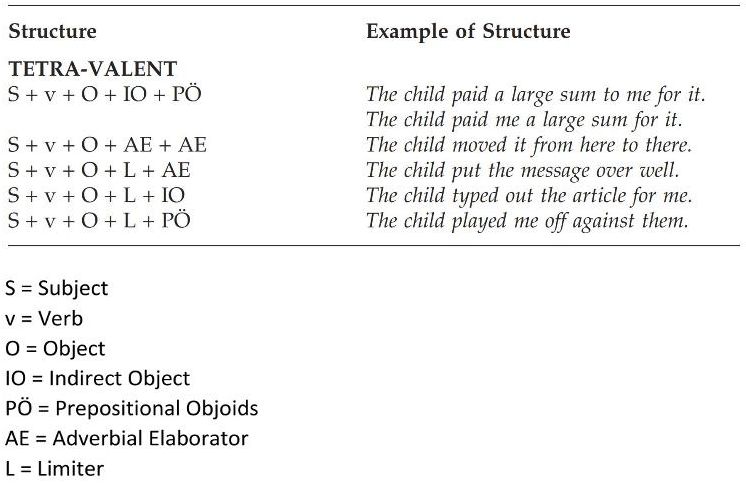}
	\caption{Allerton's tetra-valent categories with examples and abbreviation key, all described with text. (Source: Allerton 2006, p. 171.)}
	\label{fig:allerton-abbreviations-and-examples}
	\end{center}
\end{figure}

Since Allerton mentioned a valency category that is not part of Kolln's patterns or Tumbug's five mentioned patterns, namely the tetra-valent category, this section briefly considers that category and its relationship to Tumbug. The tetra-valent categories are shown as text in Figure~\ref{fig:allerton-abbreviations-and-examples}.

\begin{figure}
	\begin{center}
	\includegraphics[width=0.65\textwidth]{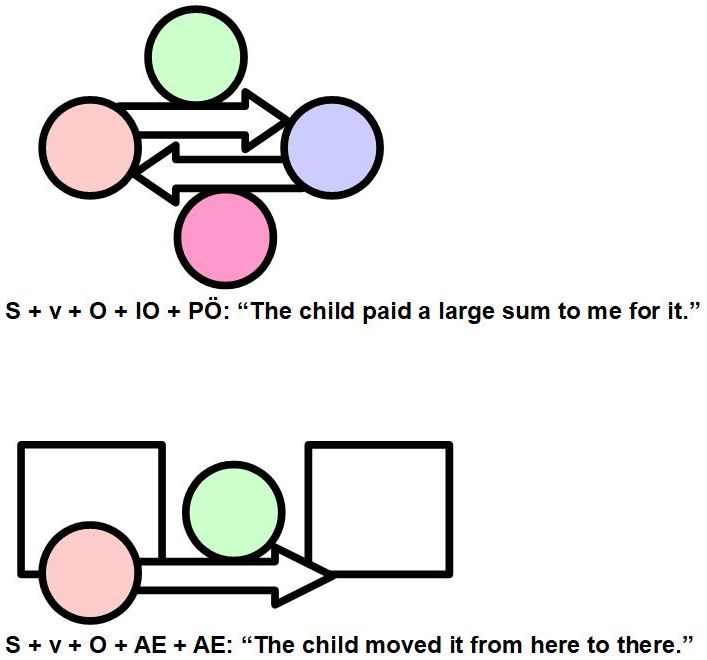}
	\caption{The first two tetra-valent examples of Allerton, all described with Tumbug images.}
	\label{fig:allerton-as-diagram1}
	\end{center}
\end{figure}

Allerton lists five examples of the tetra-valent category. Of those five examples, only one involves a fourth true object. Incidentally, Tumbug calls this sentence pattern "Swap" since it generalizes to any type of trade due to two people + two exchanged objects. The other Allerton examples show either two locations, which are not true objects, or phrasal verbs, each of which can be reduced to one verb. The first two examples in this category are shown in Figure~\ref{fig:allerton-as-diagram1}.

The tetra-valent (= Swap) patterns suggest that there does not exist any upper limit to the number of sentence patterns that may exist. This is correct, and this is partly why Tumbug was developed: Tumbug is \textit{model-based}, which means that any number of objects and actions can be combined in any way, which results in a system that can be arbitrarily complicated but will still describable by a visual model. Some examples of higher valency are:

\begin{itemize}
	\item
		swap (4-valent): 2 people + 2 objects
	\item
		escrow (4-valent): 3 parties + 1 property
	\item
		nuclear fusion (4-valent): 2 incoming atoms + 2 outgoing atoms
	\item
		attacked pinned piece in chess (4-valent): 4 pieces
	\item
		discovered check whose uncovering piece wins a piece in chess (4-valent): 4 pieces
	\item
		swap of two properties through one escrow (5-valent): 2 properties + 3 parties
\end{itemize}

\subsection{Object-attribute-value (= OAV) triples}

\begin{figure}
	\begin{center}
	\includegraphics[width=0.50\textwidth]{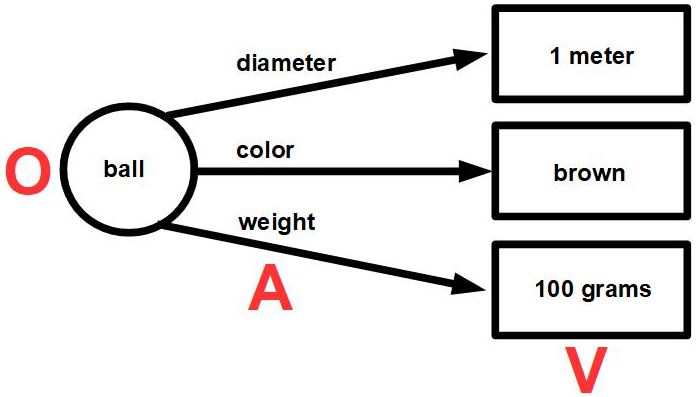}
	\caption{The traditional way to diagram an OAV triple.}
	\label{fig:oav-traditional}
	\end{center}
\end{figure}

\begin{figure}
	\begin{center}
	\includegraphics[width=0.40\textwidth]{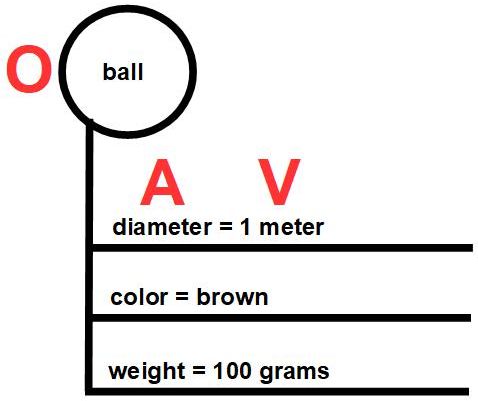}
	\caption{The Tumbug way to diagram an OAV triple.}
	\label{fig:oav-tumbug}
	\end{center}
\end{figure}

Object-attribute-value (OAV) representation is relatively well-known in computer science, and is usually called an "OAV triple," or sometimes "OAV triplet." The traditional OAV way of diagramming OAV triples is shown in Figure~\ref{fig:oav-traditional}, and the equivalent Tumbug way is shown in Figure~\ref{fig:oav-tumbug}.

\begin{figure}
	\begin{center}
	\includegraphics[width=0.65\textwidth]{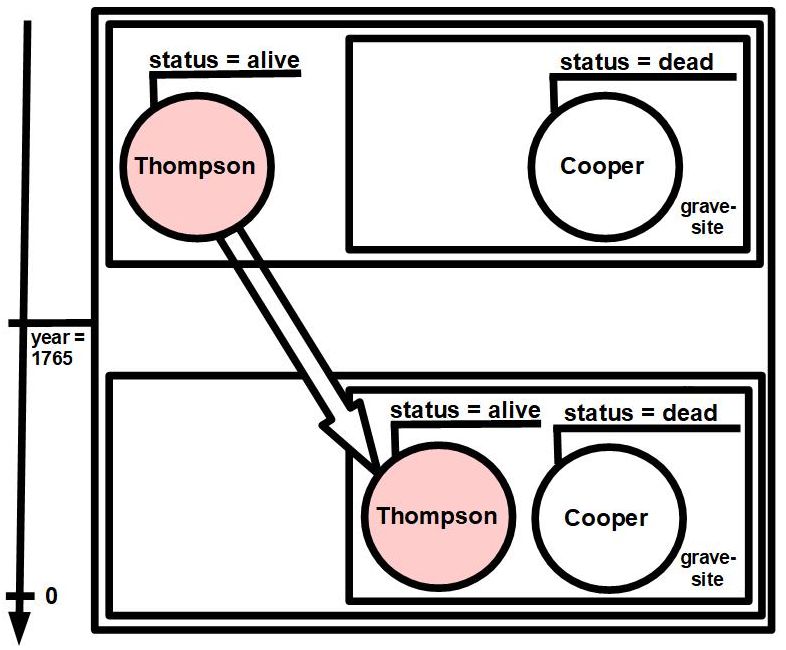}
	\caption{[78] Tumbug for "In 1765, Thompson went to Cooper's grave." This is an SOAV diagram because objects (O, e.g. "Thompson"), attributes (A, e.g., "status"), and values (V, e.g., "alive") are integrated into a larger system (S), where the larger system is the entire Tumbug diagram shown above for this English sentence.}
	\label{fig:oav-078-instantiation}
	\end{center}
\end{figure}

\begin{figure}
	\begin{center}
	\includegraphics[width=0.55\textwidth]{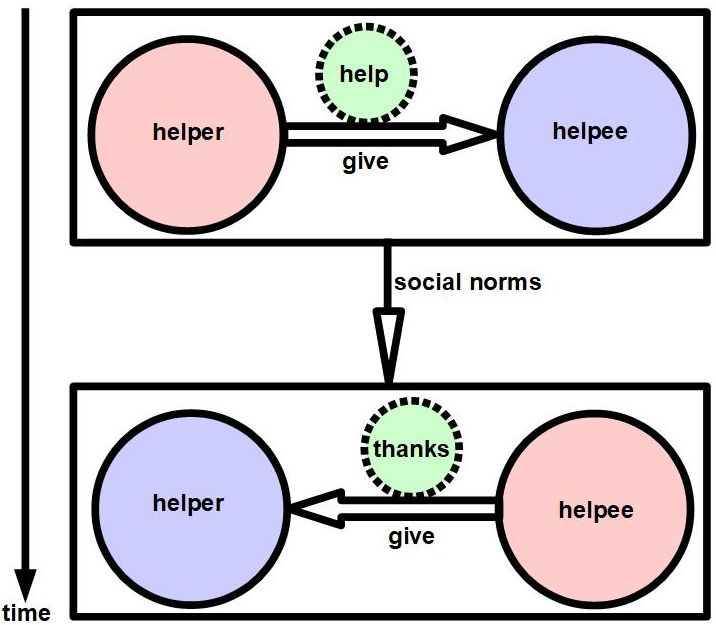}
	\caption{[3] Tumbug for "Per social norms, if a helper helps a helpee, then the helpee will thank the helper." This is an SO diagram because no attributes (A) or values (V) are shown in this system (S) that contains only objects (O) without Attribute Lines.}
	\label{fig:oav-003-given}
	\end{center}
\end{figure}

Tumbug uses OAV terminology and its meaning verbatim, although Tumbug sometimes extends this terminology and the associated diagrams. For example, Tumbug sometimes adds a novel extension to OAV called "system-object-attribute-value" (SOAV), shown in Figure~\ref{fig:oav-078-instantiation}, and Tumbug sometimes splits the OAV concept into the "subject-object" (SO) concept, an example of which is shown in Figure~\ref{fig:oav-003-given}.

\subsection{Vector graphics editors}

\begin{figure}
	\begin{center}
	\includegraphics[width=0.50\textwidth]{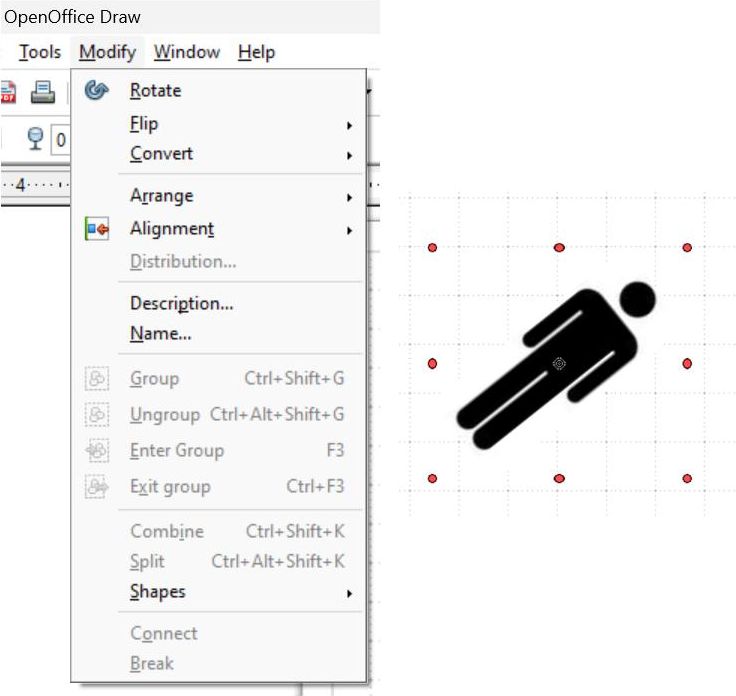}
	\caption{An OpenOffice Draw menu after an image has been selected. Note the affine transformations of Rotate and Flip (= reflection) at the start of the Modify menu.}
	\label{fig:graphics-rotation}
	\end{center}
\end{figure}

There exists a basic similarity between Tumbug and vector graphics editors in that both systems are typically used to represent real-world systems with diagrams. Both involve images in motion: Tumbug has arrows in its diagrams that imply motion, and vector graphics editors allow users to actually drag such diagrams across the computer screen in real time via a mouse button. Both systems manipulate images with respect to their shapes (circles, squares, rectangles, lines, arrows, etc.), motions (translation, rotation, scaling, reflection), shape distortions (shearing, etc.), and their attributes (color, texture, border, etc.). All the aforementioned listed motions are called affine transformations, which in 2D using homogeneous coordinates can be done via a single multiplication of a 3x3 transformation matrix with a 3x1 vector of the object's coordinates. Figure~\ref{fig:graphics-rotation} shows an example of one vector graphics editor, OpenOffice Draw, where the user in the process of rotating an icon.

\begin{figure}
	\begin{center}
	\includegraphics[width=1.00\textwidth]{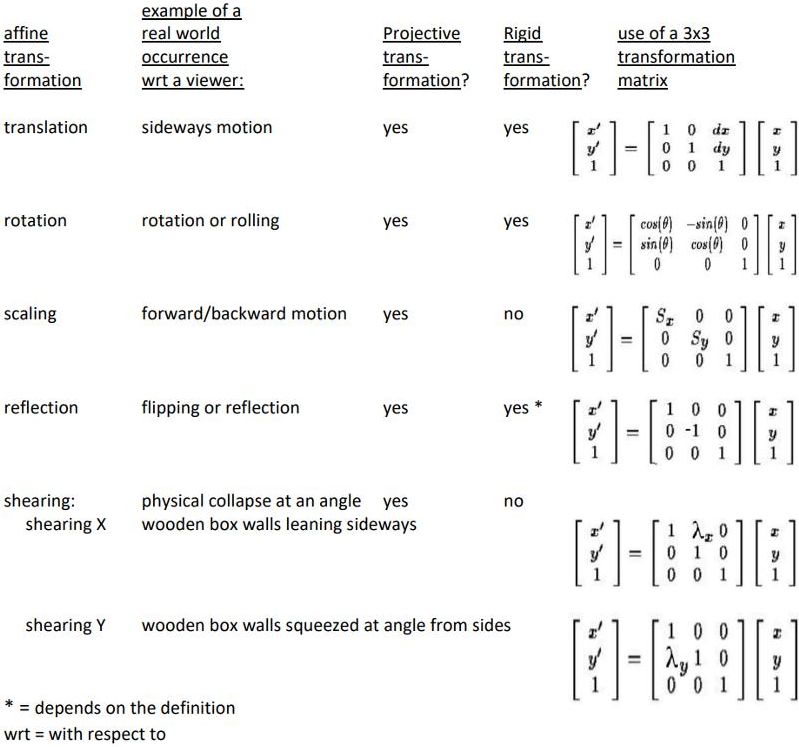}
	\caption{A summary of affine transformations and how they relate to the real world.}
	\label{fig:graphics-affine}
	\end{center}
\end{figure}

It should not be surprising to realize that the operations done by vector graphics editors are the same operations that humans see performed in the real world on physical objects, since moved diagrams are usually used to represent some motion aspect of the real world in some useful manner to humans. This is summarized in Figure~\ref{fig:graphics-affine}, assuming that the occurrence is happening with respect to a given viewer.

\begin{figure}
	\begin{center}
	\includegraphics[width=0.50\textwidth]{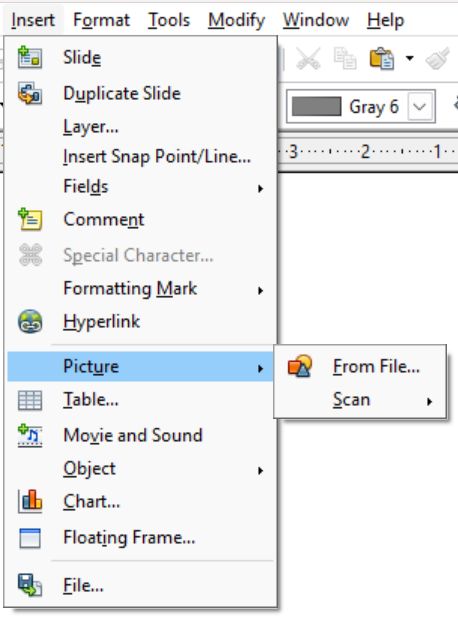}
	\caption{OpenOffice Draw implements the object creation operation by "Insert $\mid$ Picture."}
	\label{fig:graphics-insert-picture}
	\end{center}
\end{figure}

The sole visual operations of a vector graphics editor that tend not to correspond to anything in the real world are the abilities to create and destroy objects instantly. (Spontaneous creation of virtual particles of the real world is a possibility in theoretical physics, however.) A vector graphics editor might have an "Insert $\mid$ Picture" menu option to implement the creation operation of an object (see Figure~\ref{fig:graphics-insert-picture}, and the deletion operation might be done by selecting the object and hitting the "delete" key on the keyboard. It might be said that the human brain uses an unknown piece of software that is a combination of vector graphics editor + simulator + video recorder.

\begin{figure}
	\begin{center}
	\includegraphics[width=0.50\textwidth]{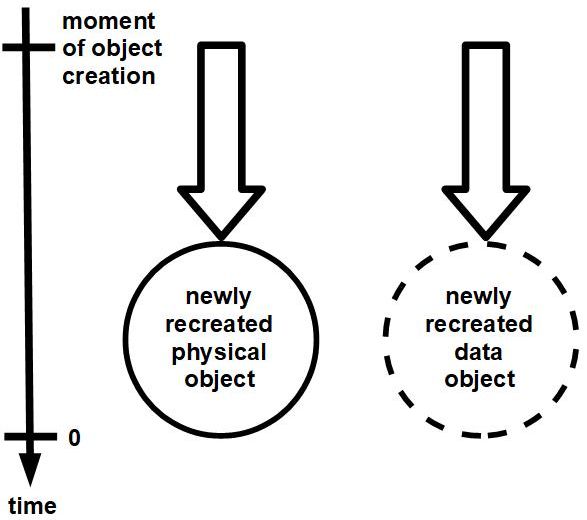}
	\caption{Creation of objects out of nothing can be represented in Tumbug by showing no earlier object history.}
	\label{fig:graphics-creation}
	\end{center}
\end{figure}

Creation of objects out of nothing can be represented easily by Tumbug, as well, since when the history of the object is viewed with help from a timeline, suddenly the object is question ceases to exist before a certain point in its history. If a Tumbug diagrammer desires to be more explicit, the "Does Not Exist" value could be placed at the specific time and specific place that an object appeared, as shown in Figure~\ref{fig:graphics-creation}. Conclusion: Tumbug can represent any primary operation that can be performed by any typical vector graphics editor.

\begin{figure}
	\begin{center}
	\includegraphics[width=0.50\textwidth]{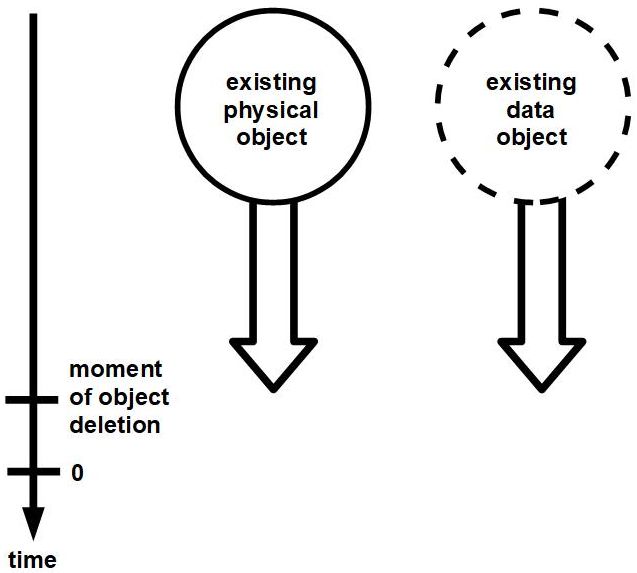}
	\caption{Deletion of objects can be represented in Tumbug by showing no later object history.}
	\label{fig:graphics-deletion}
	\end{center}
\end{figure}

The same statements are true for object deletion. See Figure~\ref{fig:graphics-deletion}.

\subsection{Semantic networks}

Semantic networks are a fairly well-known KRM in computer science, one that uses nodes and links, and therefore bears a superficial resemblance to Tumbug.  A well-known article by Woods (1975) discusses a large number of shortcomings of semantic networks. Tumbug addresses most or all these shortcomings. Full coverage of a comparison and semantic networks would be too lengthy for this document, however, so only a sampling of Woods' criticisms are given here, with each criticism followed by the Tumbug solution.

\subsubsection{Kinship relations}

With regard to data bases and semantic networks, the sentence "Harry is John's uncle" is problematic because:

\begin{enumerate}
	\item
		The term "uncle" is partially ambiguous because it could refer to either the mother's side of the family or the father's side of a family.
	\item
		Even if the ambiguity is removed, the uncle kinship relation involves comparing nodes that are not directly connected in a kinship tree.
	\item
		Even if the applicable nodes are explicitly connected with an "uncle" link, it is too impractical to store all possible uncle relationships explicitly, due to combinatorial explosion of all possible matches.
\end{enumerate}

\begin{figure}
	\begin{center}
	\includegraphics[width=0.75\textwidth]{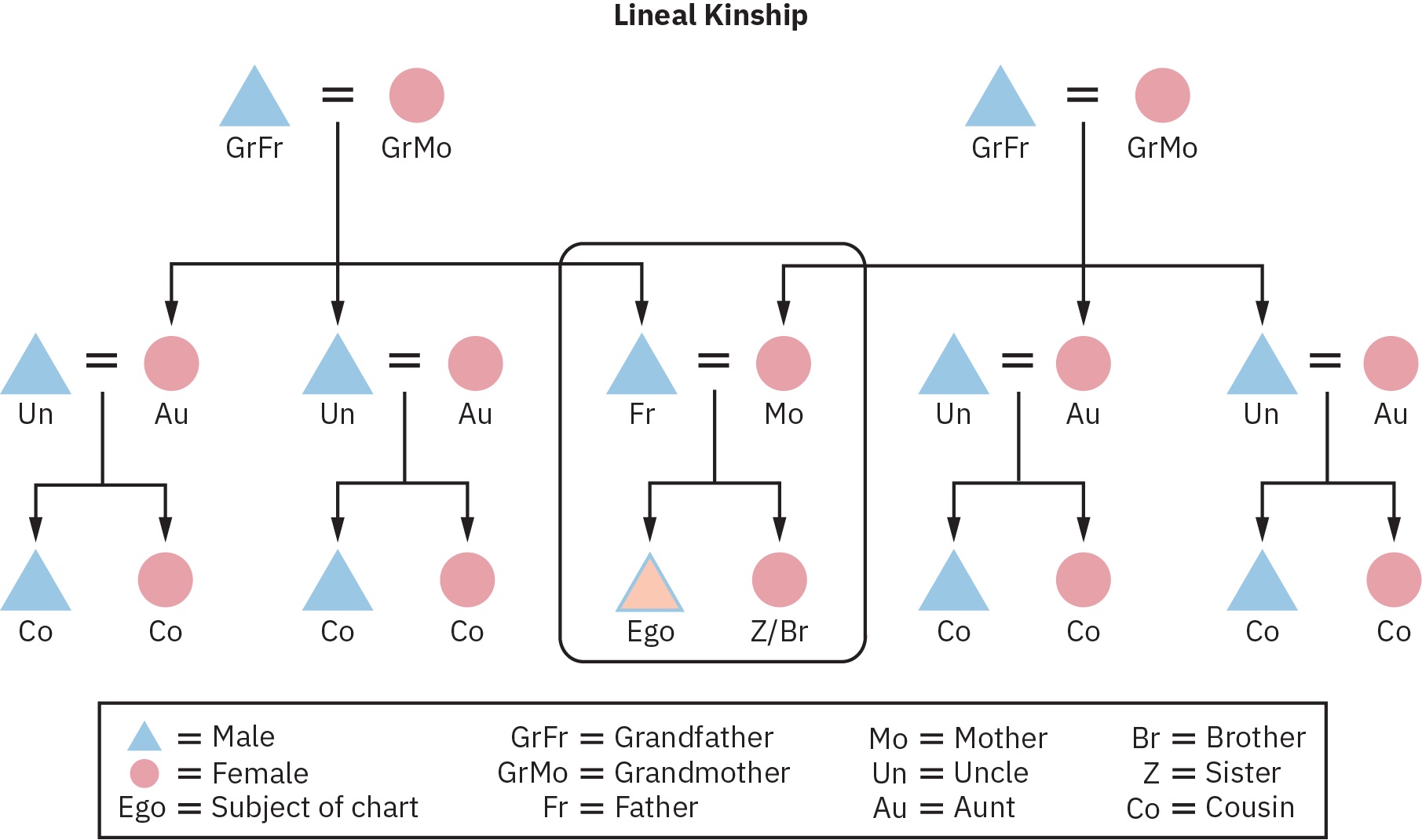}
	\caption{A standard (lineal) kinship diagram. (Source: “Eskimo Kinship Chart” by Fred the Oyster/Wikimedia Commons, CC0.)}
	\label{fig:sn-lineal-kinship}
	\end{center}
\end{figure}

\begin{figure}
	\begin{center}
	\includegraphics[width=0.75\textwidth]{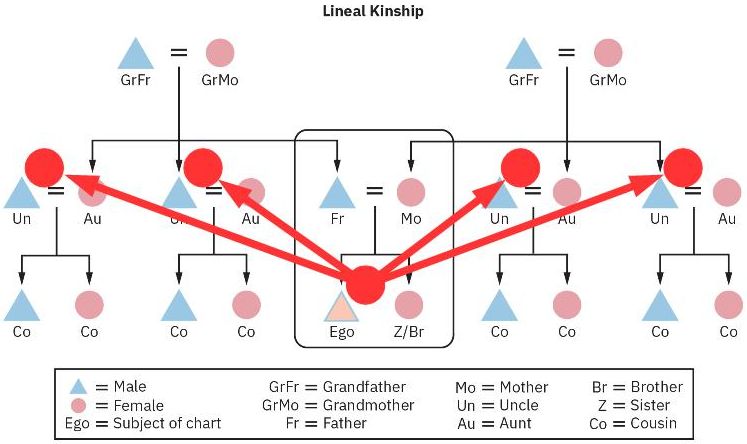}
	\caption{A kinship diagram that flags the possible uncles of the shown ego.}
	\label{fig:sn-uncle-using-jpg-markers}
	\end{center}
\end{figure}

\begin{figure}
	\begin{center}
	\includegraphics[width=0.75\textwidth]{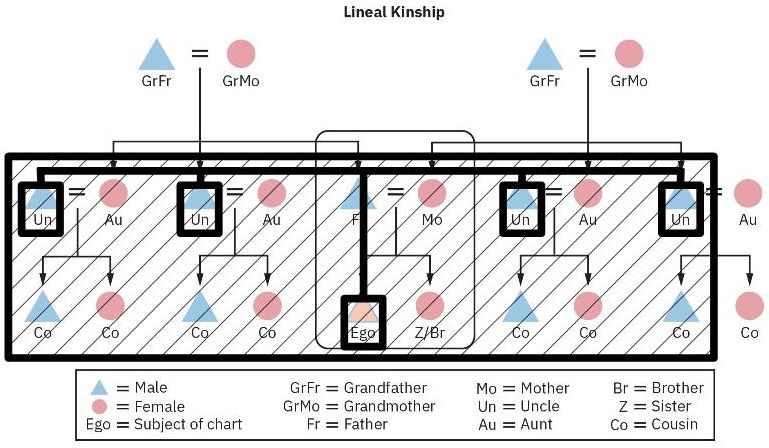}
	\caption{The ego-uncle relationship can be thought of a spatial template that can be moved to start at any ego node.}
	\label{fig:sn-uncle-using-jpg-template}
	\end{center}
\end{figure}

\begin{figure}
	\begin{center}
	\includegraphics[width=0.75\textwidth]{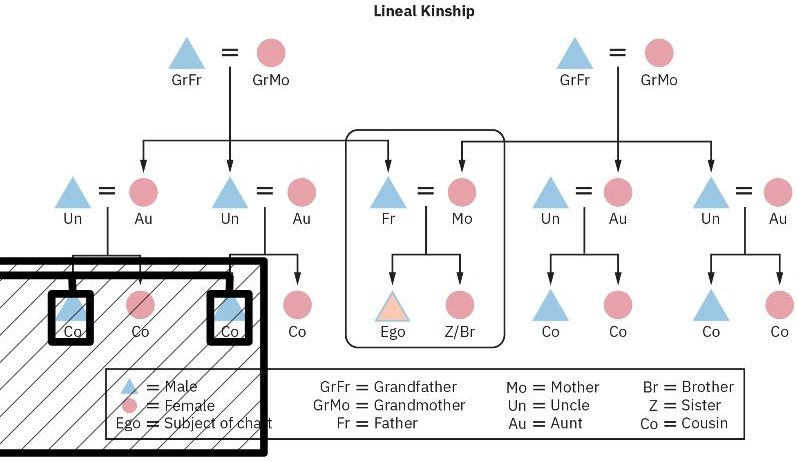}
	\caption{The ego-uncle relationship template moved to a different ego node (off the left-hand edge of the chart).}
	\label{fig:sn-uncle-using-jpg-template-2nd-position}
	\end{center}
\end{figure}

See Figure~\ref{fig:sn-lineal-kinship} for a standard kinship tree to see the nodes and links that are involved in this problem. If a specific ego (= the person under consideration when describing that person's kinship relationships) is chosen, then the ego and the two choices of uncle are marked with a 0D Marker as in Figure~\ref{fig:sn-uncle-using-jpg-markers}. Tumbug solves these issues by the following solutions, respectively:

\begin{enumerate}
	\item
		Either use an "OR" to disjunct the mother's side of the family with the father's side of a family, or combine both sides of the family into a more generic group called "parents' siblings" and use the male members of that group.
	\item
		Use a 0D Marker to flag the ego and uncle group, then use a relocatable template that highlights any two nodes, based on the ego at one end and the uncle distance-and-position at the other end as in Figure~\ref{fig:sn-uncle-using-jpg-template} and Figure~\ref{fig:sn-uncle-using-jpg-template-2nd-position}. (Again, one of the main goals of Tumbug is to represent everything visually, as much as practical.)
	\item
		Store a specific ego-uncle relationship only when that specific relationship has been needed to be considered by the system when answering a query.
\end{enumerate}

\textit{Further development anticipated: The uncle kinship relationship template shown is rigid and specific to the tree layout shown, not flexible with respect to arbitrary tree layout or unpredictable counts or unpredictable branches. The conversion of such a rigid to a flexible template was not attempted.}

\subsubsection{Specific objects versus generic objects}

Woods wrote (Woods 1975, p. 22): "For example, if I create a node and establish two links from it, one labeled SUPERC and pointing to the "concept" TELEPHONE and another labeled MOD and pointing to the "concept" BLACK, what do I mean this node to represent? Do I intend it to stand for the "concept" of a black telephone, or perhaps I mean it to assert a relationship between the concepts of telephone and blackness--i.e., that telephones are black (all telephones?, some telephones?). When one devises a semantic network notation, it is necessary not only to specify the types of nodes and links that can be used and the rules for their possible combinations (the syntax of the network notation) but also to specify the import of the various types of links and structures--what is meant by them (the semantics of the network notation)."

\begin{figure}
	\begin{center}
	\includegraphics[width=0.50\textwidth]{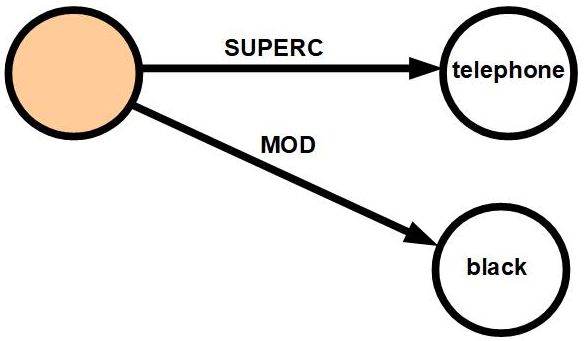}
	\caption{The SUPERC semantic network example described by Woods.}
	\label{fig:sn-superc}
	\end{center}
\end{figure}

First, presumably SUPERC = superclass, which would be clearer in modern times due to the modern popularity of object-oriented programming. Woods' described semantic network diagram is shown in Figure~\ref{fig:sn-superc}.

\begin{figure}
	\begin{center}
	\includegraphics[width=0.50\textwidth]{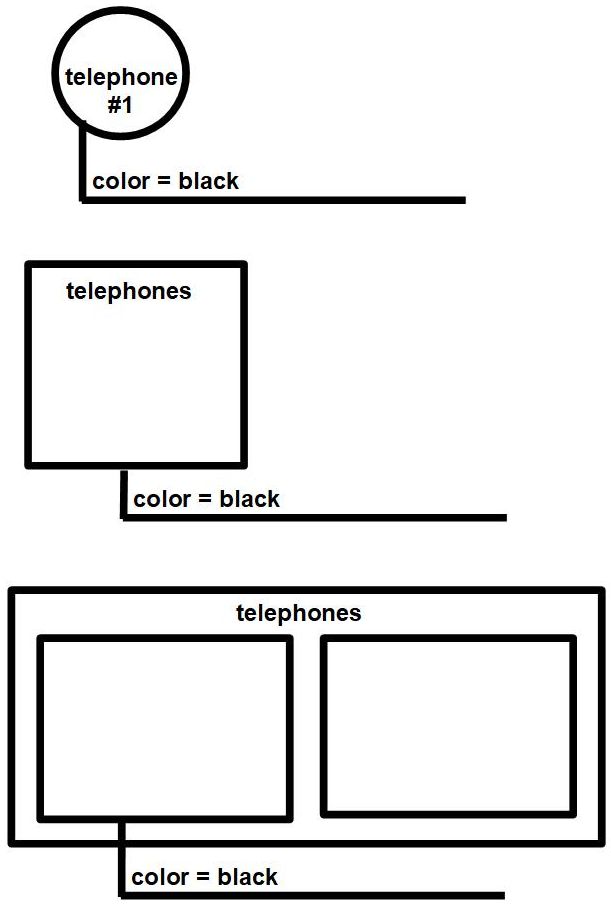}
	\caption{Woods' three interpretations of the SUPERC example, as represented in Tumbug. Top: "The particular telephone in question (= telephone \#1) is black." Middle: "All telephones are black." Bottom: "Some telephones are black."}
	\label{fig:sn-superc-interpretations}
	\end{center}
\end{figure}

The three interpretations that Wood mentioned are shown in Figure~\ref{fig:sn-superc-interpretations}. Tumbug represents a superclass using the Venn diagram convention of drawing a region around the region in question, where the larger region represents the superclass. Tumbug represents each of these three interpretations differently, by virtue of making attribute-value pairs distinctly different from Object Circle labels and Location Box labels, and by virtue of making objects and classes distinctly different via Object Circles and Aggregation Boxes, and by virtue of allowing Aggregation Boxes to span arbitrary regions, where each Aggregation Box may have its own unique label. Woods is correct about the drawbacks of semantic networks, but clearly Tumbug has much more representative power than semantic networks. Even object-oriented programming clearly has more representative power than semantic networks. Semantic networks are clearly becoming old-fashioned due to their inherent limitations.

\subsubsection{Intensions versus extensions}

Woods wrote (1975, pp. 22-23): "Basically a predicate such as the English word "red" has associated with it two possible conceptual things which could be related to its meaning in the intuitive sense. One of these is the set of all red things--this is called the \underline{extension} of the predicate. The other concept is an abstract entity which in some sense characterizes what it \underline{means} to be red, it is the notion of \underline{redness} which may or may not be true of a given object; this is called the \underline{intension} of the predicate."

\begin{figure}
	\begin{center}
	\includegraphics[width=0.75\textwidth]{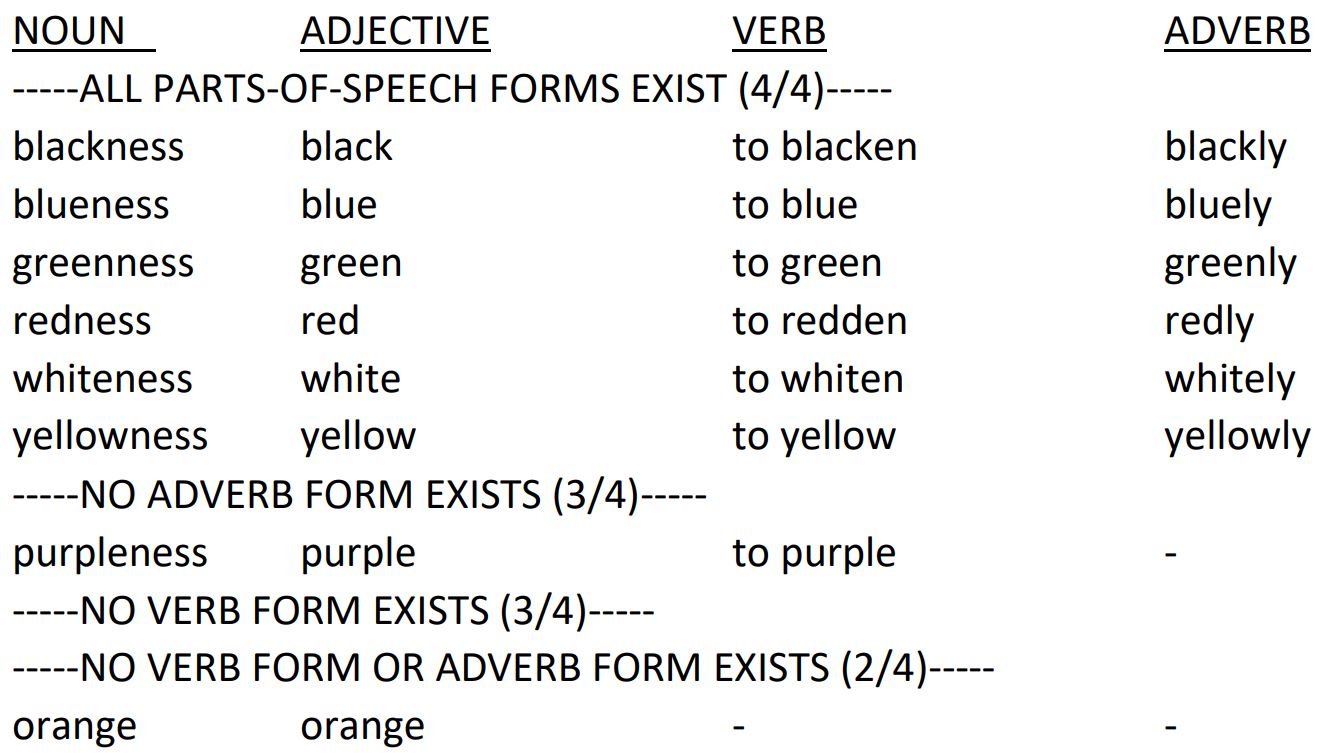}
	\caption{The concept of color, like most concepts, can appear as different Parts of Speech.}
	\label{fig:thingification-pos-color-table-snap}
	\end{center}
\end{figure}

This confusion was already addressed in this document by the insight that a concept can arbitrarily change its Part of Speech merely by moving the node that holds that concept to a different part of the OAVC diagram of Figure~\ref{fig:thingification-pos}. Therefore "red" as an adjective differs from "red" as a noun in a diagrammatically simple way in Tumbug. Therefore "extension" is "red" as an adjective, "intension" is "red" as a noun. Such Parts of Speech differences were shown in Figure~\ref{fig:thingification-pos-color-table-snap}, and are shown again here in Figure~\ref{fig:thingification-pos-color-table-snap} with only words for colors.

\section{Applications of Tumbug}

\subsection{Natural language translation}

Below is the translation of a single English sentence to two different foreign languages, where each foreign language is from a different branch of the Indo-European languages: French, from the Romance languages, and German, from the Germanic languages. Three steps are shown in the translation to Tumbug: (1) translation, (2) analysis, (3) Tumbug representation, then a general discussion of the process and its nuances.

\subsubsection{Conversion from a natural language to Tumbug}

\setlength\parindent{0pt}

\textbf{1. Translation}

\begin{figure}
	\begin{center}
	\includegraphics[width=0.75\textwidth]{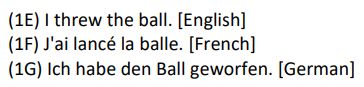}
	\caption{An English sentence translated into French and German.}
	\label{fig:translation-threw-ball-1-translation-snap}
	\end{center}
\end{figure}

See Figure~\ref{fig:translation-threw-ball-1-translation-snap}.\\

\textbf{2. Analysis}

\begin{figure}
	\begin{center}
	\includegraphics[width=0.75\textwidth]{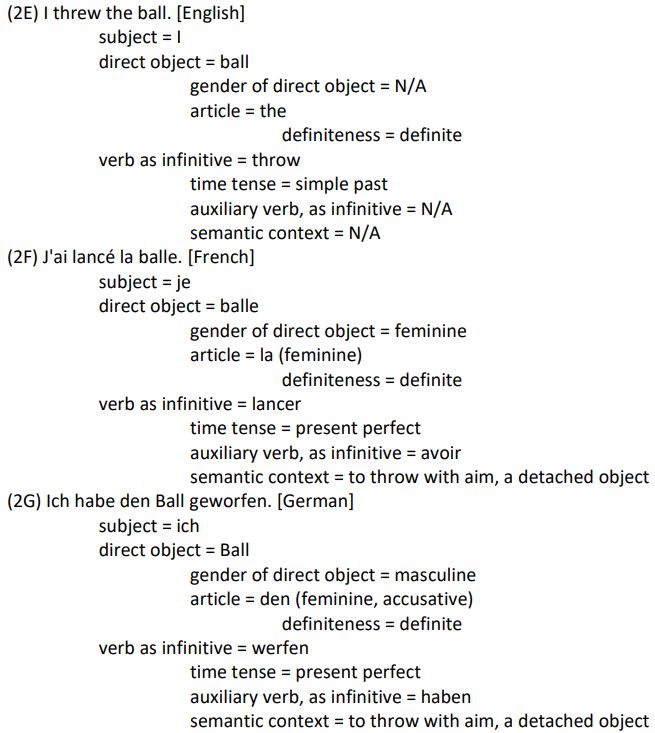}
	\caption{Analysis of the sentence in the three translations.}
	\label{fig:translation-threw-ball-2-analysis-snap}
	\end{center}
\end{figure}

See Figure~\ref{fig:translation-threw-ball-2-analysis-snap}.\\

\textbf{3. Tumbug representation}

The following abbreviations appear in the diagrams:\\

DC = Don't Care\\
DNE = Does Not Exist\\

\begin{figure}
	\begin{center}
	\includegraphics[width=0.65\textwidth]{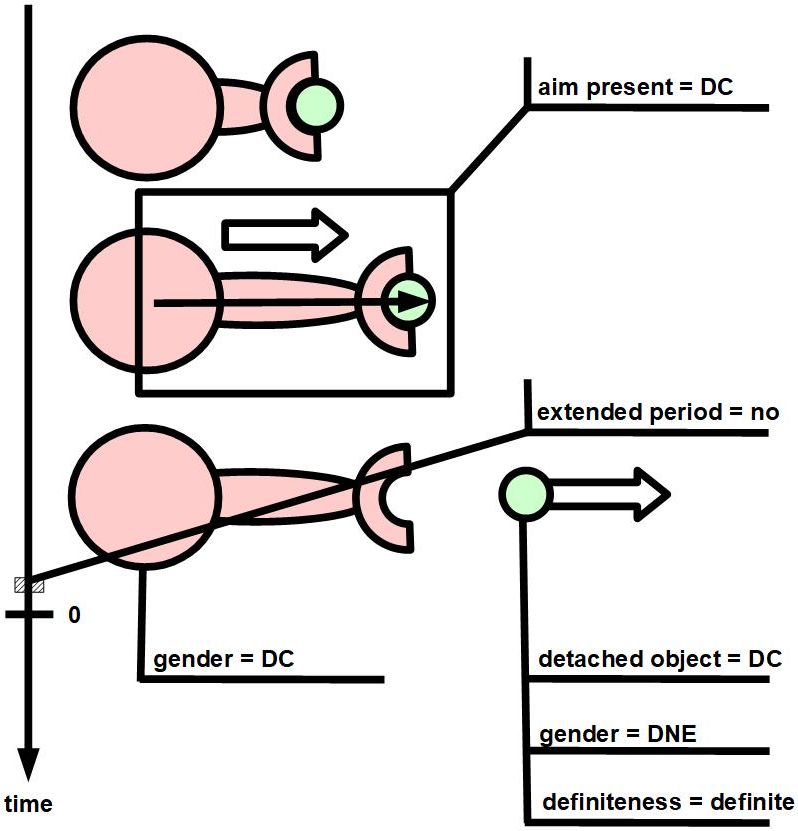}
	\caption{Tumbug for "I threw the ball" in English.}
	\label{fig:translation-transformation-e}
	\end{center}
\end{figure}

For the English sentence, see Figure~\ref{fig:translation-transformation-e}.

\begin{figure}
	\begin{center}
	\includegraphics[width=0.65\textwidth]{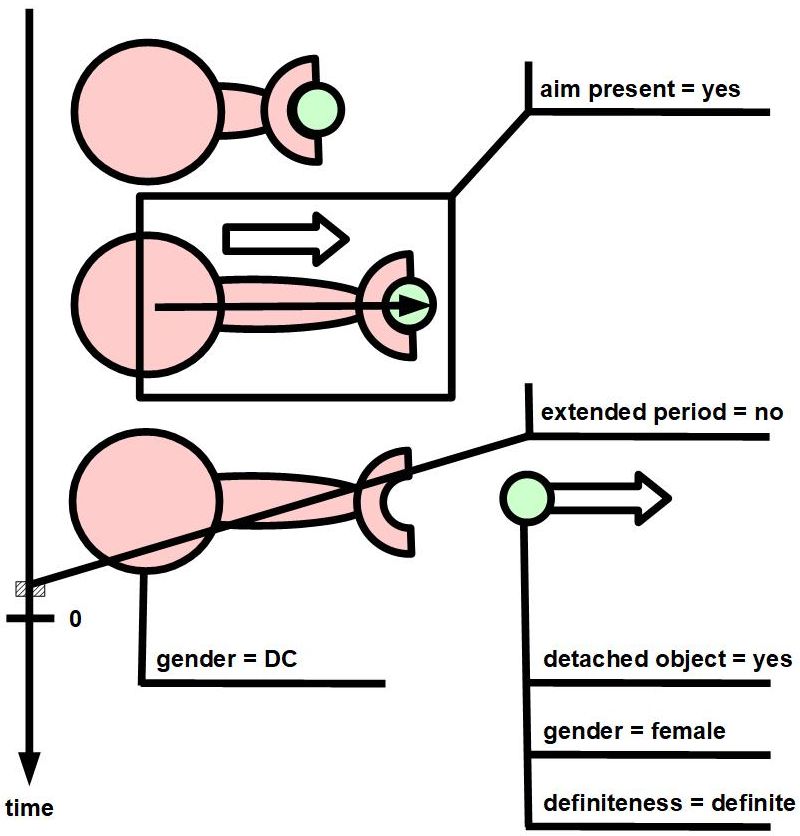}
	\caption{Tumbug for "I threw the ball" in French.}
	\label{fig:translation-transformation-f}
	\end{center}
\end{figure}

For the French translation, see Figure~\ref{fig:translation-transformation-f}.
 
\begin{figure}
	\begin{center}
	\includegraphics[width=0.65\textwidth]{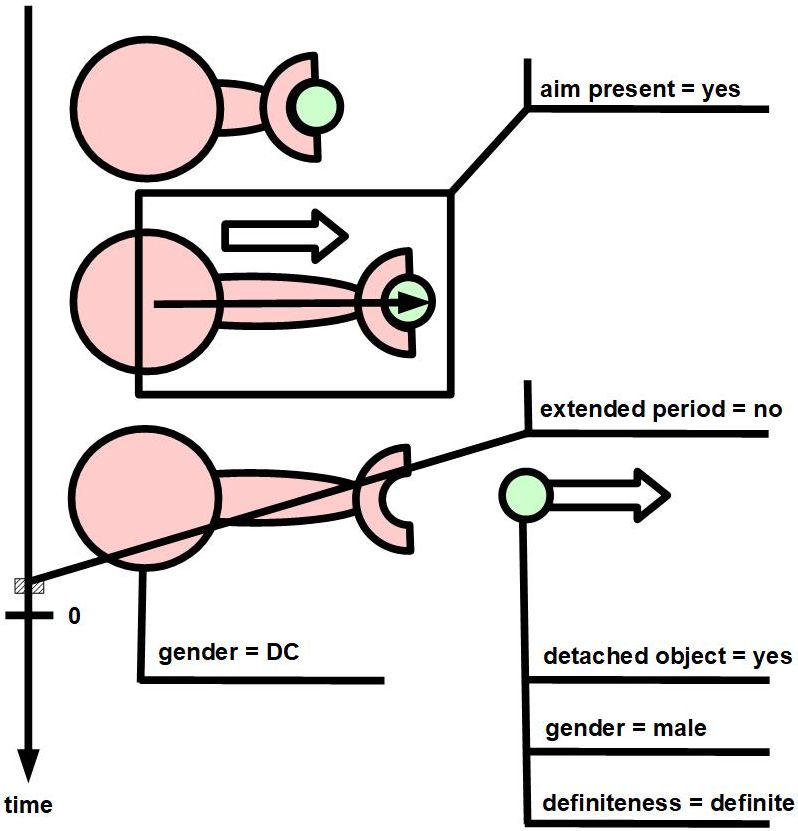}
	\caption{Tumbug for "I threw the ball" in German.}
	\label{fig:translation-transformation-g}
	\end{center}
\end{figure}

For the German translation, see Figure~\ref{fig:translation-transformation-g}.

\setlength\parindent{24pt}

\subsubsection{Discussion of language translation}

\textbf{1. Small mismatches}

\begin{figure}
	\begin{center}
	\includegraphics[width=1.00\textwidth]{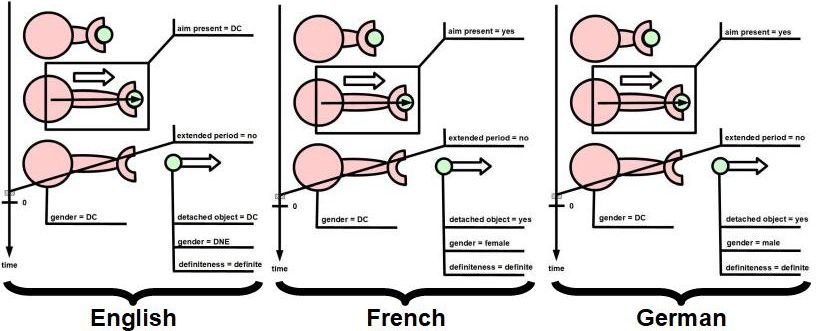}
	\caption{Comparison of Tumbug representations of the same sentence across three languages.}
	\label{fig:translation-all-3-transformations}
	\end{center}
\end{figure}

Evidence that Tumbug is nearly a universal representation between these three languages is that the iconic Tumbug structures for the same sentence are absolutely identical, as seen in Figure~\ref{fig:translation-all-3-transformations}.

\begin{figure}
	\begin{center}
	\includegraphics[width=1.00\textwidth]{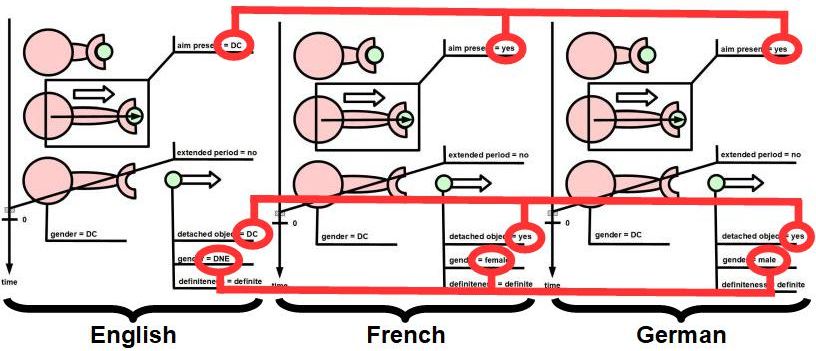}
	\caption{Emphasized differences between Tumbug representations of the same sentence across three languages.}
	\label{fig:translation-all-3-transformations-ANN}
	\end{center}
\end{figure}

There are a few subtle mismatches, however, all in the attribute values, and all due to some languages being more specific than others. These mismatches of attribute values are marked in Figure~\ref{fig:translation-all-3-transformations-ANN}.

Pairs of minor mismatches across languages in Tumbug of this example are marked with arcs.

The mismatches are the following:

\begin{enumerate}
	\item
		Regarding aim: French and German care about the exact meaning of "to throw" since each of those languages would use a different verb if the meaning were slightly different. For example, French would use "jeter" if no aim or detached object were involved, and German would use "abgeben" (or "ausgeben") for the same reason. English makes no such distinction, so English shows "DC" (= don't care) for the "aim present" attribute.
	\item
		Regarding attachment: French and German care whether the object being thrown is an attached object or not, such as "to throw a switch" versus "to throw a ball." A different verb would be used, depending on the attachment status.
	\item
		Regarding gender: French and German have different grammatical genders for their respective words for "ball." English has no grammatical genders so it shows "DNE" (= Does Not Exist). In this case, grammatical gender is a human-attributed quality that does not exist in inanimate objects, so in this example, object gender is irrelevant. If a human were in the place of direct object or indirect object, however, then gender could be relevant, possibly because of outfit, marriage status, voice range, or any other influences of biological gender.
\end{enumerate}

Other subtle mismatches could occur easily with different sentences or with different languages. For example "you" may have different levels of formality, such as in German and Dutch, some languages do not have grammatical genders, such as in Japanese, and some languages do not have articles, such as in Japanese and Latin.

All such mismatches are due to differing levels of details across the languages, however, none whatsoever with the structure and relationships between the top-level entities or actions after the exact word meaning has been determined. Of course a sentence could be completely ambiguous even in its own language, such as "The pot is in the pen," but that is a deficiency on the part of a speaker or the language, not a deficiency in Tumbug. Tumbug does have some potential weaknesses, but these are subtle and very unlikely to be encountered. These weaknesses are discussed later in this document.

\textbf{2. Corrections of the small mismatches}

There exist only two obvious ways to correct the aforementioned minor mismatches in the translation example: (1) require the writer/speaker to be more specific, as specific as the most specific languages that might be encountered in later translation, (2) use context to determine the most likely meaning. The first option is usually not within the reader's control, so usually context must be used.

Consider how context of a sentence can be implied by following sentences:\\
\begin{enumerate}
	\item
		Textual passage: "I saw the player trying to steal third base and he wasn't too far away. I threw the ball."\\
		Suggested context: baseball, where a ball is a critical piece of equipment that must be thrown accurately\\
		Implied aim present: TRUE\\
		Implied detached object: TRUE
	\item
		Textual passage: "I was going through the trash, pocketing anything of value and tossing aside anything else. First I found an earring, so I pocketed it. Then I found a disgustingly moldy ball of cheese. I threw the ball."\\
		Suggested context: trash, where the ball described is something the holder would toss aside quickly, to anywhere\\
		Implied aim present: FALSE\\
		Implied detached object: TRUE
\end{enumerate}

\begin{figure}
	\begin{center}
	\includegraphics[width=0.75\textwidth]{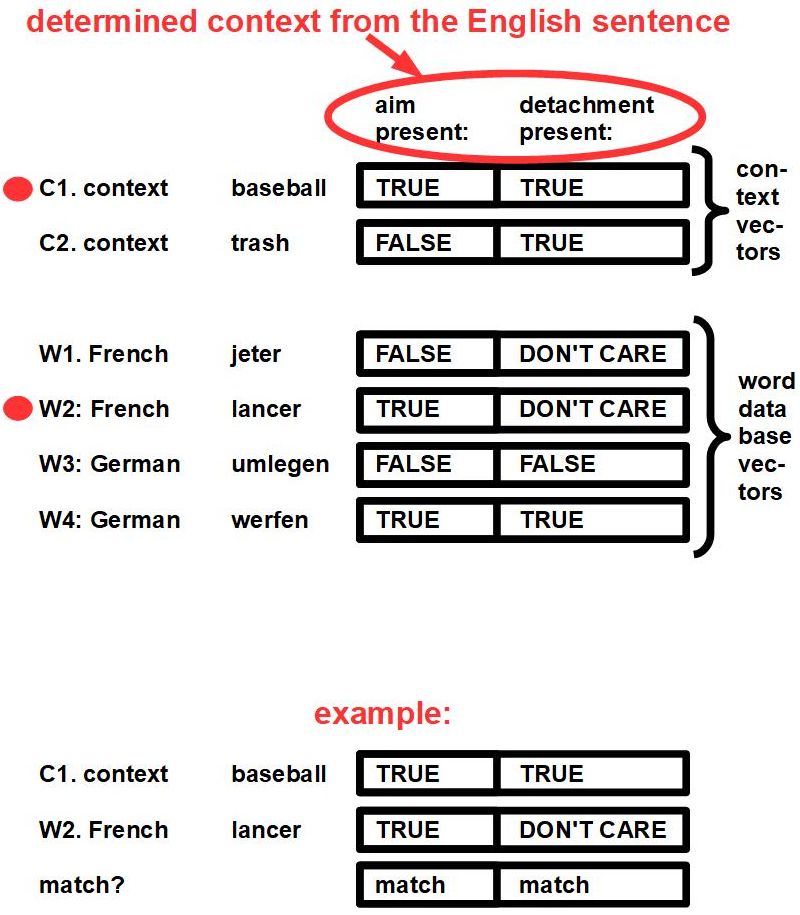}
	\caption{The one determined context (here C1) from the source language is matched against each word in a data base of candidate words in the target language, with one such candidate word (here W2) used in the example. Note that a "DON'T CARE" value will match either a "TRUE" or "FALSE" value.}
	\label{fig:translation-context-matching}
	\end{center}
\end{figure}

The contextual information in this case clarifies the two main ambiguities from before: whether there was an intended target, and whether the thrown item was attached. The ambiguity of which gender is involved depends only on language-specific grammatical rules, and not the meaning of the word after the proper word has been clarified. The result of this context determination from the English (or other language) sentence is the first step for translation, and results in essentially a word vector of TRUE-FALSE values. The second step of translation matches this word vector against a database of (typically) foreign words, as shown in Figure~\ref{fig:translation-context-matching}.

\begin{figure}
	\begin{center}
	\includegraphics[width=0.75\textwidth]{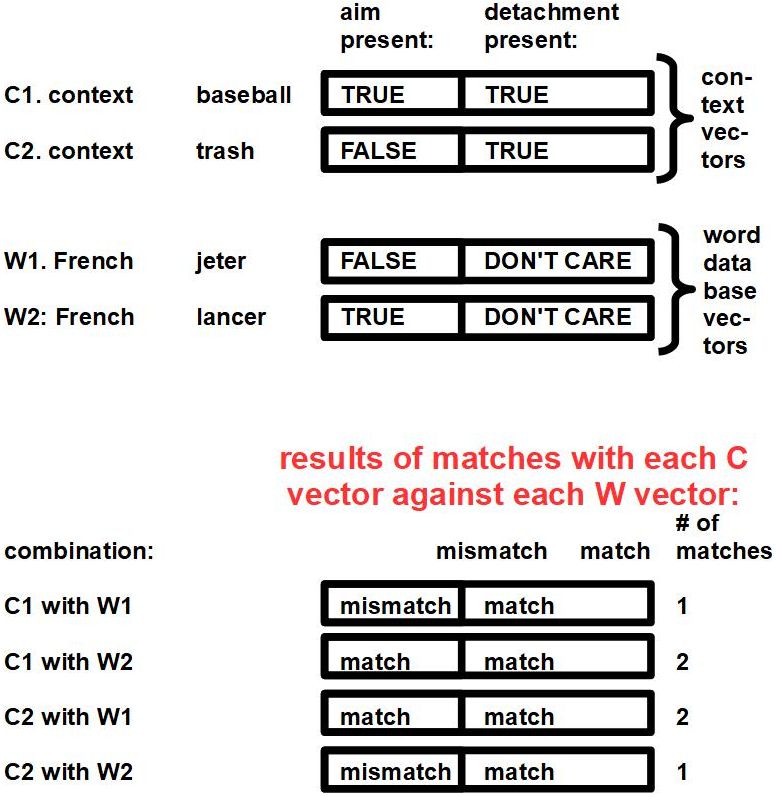}
	\caption{Each given context (here C1 and C2) is matched against each candidate word in the target language (here W1 and W2), and the count of matches across each vector's elements is shown at the right of that vector. In the context of baseball (C1) being matched against French words, the French word with the highest count is assumed to be the French word that has the closest match. The closest matching French word is therefore assumed to be the word "lancer" (W2), which has two matches, as compared to the next highest matching word, "jeter" (W1), which has only one match. Therefore in the context of baseball, the French word with the closest match to the English word "to throw" is "lancer."}
	\label{fig:translation-context-matched}
	\end{center}
\end{figure}

The third step would be to calculate which context-word combinations had the closest match. For example, in the context of baseball, a context that was determined by analysis of the the sentence just described whose source language was English, each context vector of the source language would be matched against each context vector of the target language. For two contexts from English and two candidate words from French the result would be (2x2 =) four vectors, each cell of which contains an indication as to whether there was a match or mismatch. For example, TRUE matched with FALSE = mismatch, and FALSE matched with DON'T CARE = match. Figure~\ref{fig:translation-context-matched} shows the final results. One simple metric for determining degree of match would be to simply add the number of matches. In the context of baseball (C1), the highest matching candidate French word was "lancer," which had two matches versus only one match with "jeter." Figure~\ref{fig:translation-context-matched} shows this entire matching process as described.

\subsection{Mathematics: Arithmetic}

Arithmetic in Tumbug uses a Causation Arrow to represent the operator (e.g., "+" or "-") or function (e.g., 3x, x2), and uses Data C Object Circles to represent the numbers. Numbers are not real-world objects so numbers cannot use Physical C Object Circles. This is in contrast to the way the OOP language Smalltalk handles numbers: in Smalltalk, numbers are objects, and operators are methods. For example, in Smalltalk to add two numbers, such as 1 and 2, Smalltalk would send the method "+ 2" to the number "1" (Winston 1998, p. 16).

\begin{figure}
	\begin{center}
	\includegraphics[width=0.50\textwidth]{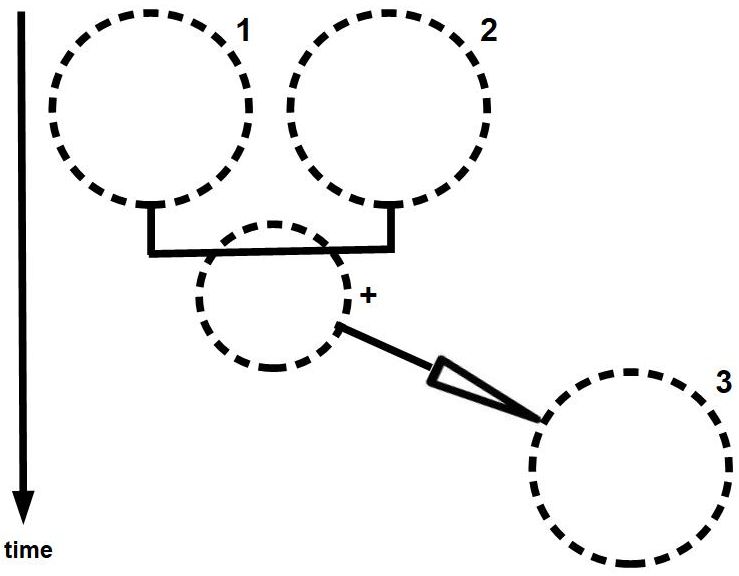}
	\caption{To add the numbers 1 and 2, Tumbug sends them as two pieces of data to a virtual operator called "+".}
	\label{fig:arithmetic-1-2-3}
	\end{center}
\end{figure}

In Tumbug, "1 + 2" would be rendered as in Figure~\ref{fig:arithmetic-1-2-3}, which translates to the verbal description "Messages 1 and 2 together are inputted to the virtual adder, which causes the number 3 to be outputted." A "virtual adder" could mean anything like a person's mind or an imaginary computer adder. If a real world computer or real-world calculator were involved then the circle around the Causation Arrow would be solid: a Physical C Object Circle. Numbers will always be virtual, so numbers are always represented by Data C Object Circles. Although computer memory can erase and overwrite numbers at will, human memory cannot: the concept of "1" and "2" are practically permanent concepts in human memory after being learned, as is their sum "3," therefore the diagram shows the "3" concept in a different location than directly below the "1" or "2" so that neither the "1" nor "2" are overwritten by Tumbug conventions. Whether the machine performing the adding is mechanical (e.g., a cash register) or electrical (e.g., a computer), any calculation it performs takes time, therefore a timeline (Time Arrow) should always be present in any Tumbug diagram that represents calculation.

\subsection{Logic: First Order Predicate Calculus}

\subsubsection{Syllogisms with Venn diagrams}

Venn diagrams are not only closed geometrical figures, each of which may contain points in arbitrary locations within any given region, but can also be considered diagrams of physical regions that may contain objects within any given region, like an aerial view of a horse corral that may contain horses. Therefore a Venn diagram can be considered only a cleaned up and abstracted aerial photograph, which conveniently bridges the gap between reality and geometrical diagrams. Therefore there is no need for Tumbug to include a separate Building Block for a Venn diagram.

\begin{figure}
	\begin{center}
	\includegraphics[width=0.50\textwidth]{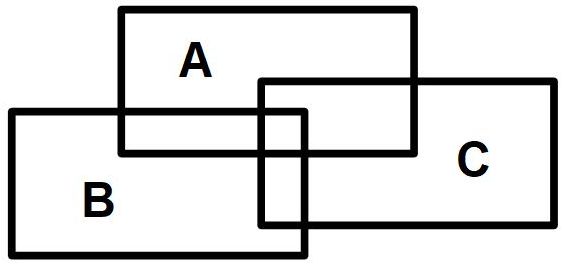}
	\caption{A Venn diagram in Tumbug uses C Aggregation Boxes instead of the familiar circles of Venn diagrams.}
	\label{fig:syllogisms-3-rectangles-abc}
	\end{center}
\end{figure}

One type of formal logic is First Order Predicate Calculus (FOPC). FOPC often uses syllogisms represented as Venn diagrams, which are used to represent sets. As noted earlier in this document, sets are very conveniently represented in Tumbug with any type of Aggregation Boxes, since order (or 2D position) is not important in a set. Therefore three intersecting sets in Tumbug would be represented as three intersecting C Aggregation Boxes, as shown in Figure~\ref{fig:syllogisms-3-rectangles-abc}. 

There exist only 24 valid types of syllogisms, and each type has been given a person's name. In this document the Socrates example has the Barbara form, the reptiles example has the Celarent form, and the rabbits example has the Darii form.

\subsubsection{Barbara form (AAA-1)}

The Socrates syllogism example in this section is the well-known syllogism:\\

\setlength\parindent{0pt}
All men are mortal.\\
Socrates is a man.\\
Therefore, Socrates is mortal.\\
\setlength\parindent{24pt}

\begin{figure}
	\begin{center}
	\includegraphics[width=0.85\textwidth]{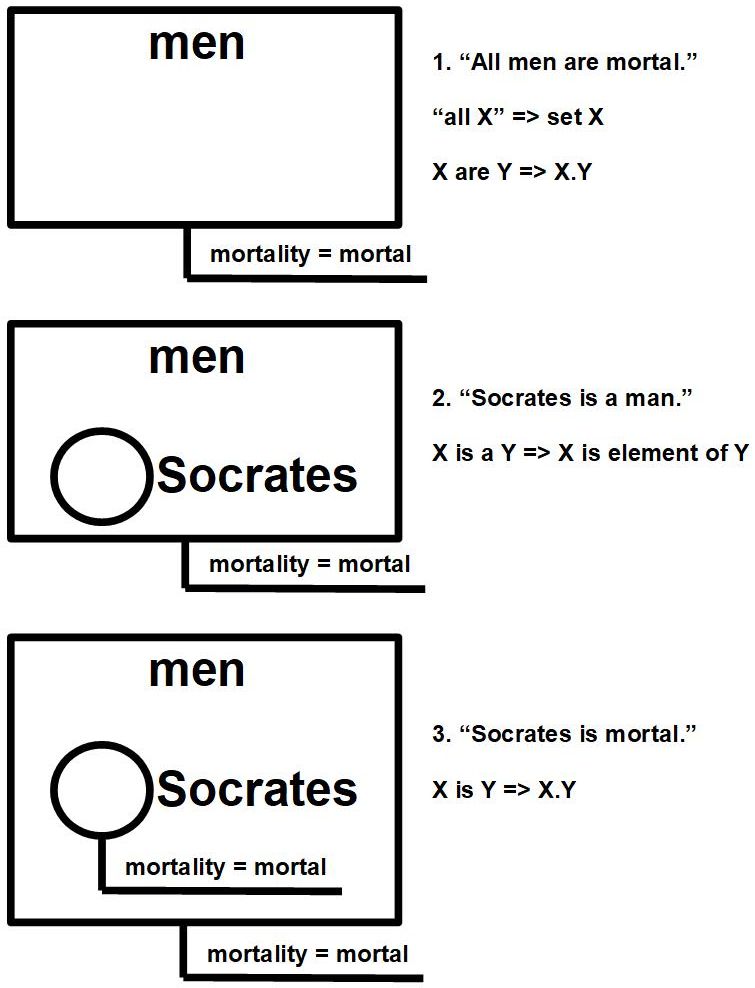}
	\caption{The three diagram creation steps that correspond to the three sentences of the syllogism.}
	\label{fig:syllogisms-socrates-steps}
	\end{center}
\end{figure}

In this syllogism the two sets involved--assuming that the information is regarded as sets whenever possible--happen to be nested and non-intersecting, as shown in Figure~\ref{fig:syllogisms-socrates-steps}. In the literature, usually even Socrates is represented as a third set, for some reason, but technically Socrates is an element of a set, not a set \textit{per se}, so is represented an object here.

\begin{figure}
	\begin{center}
	\includegraphics[width=0.25\textwidth]{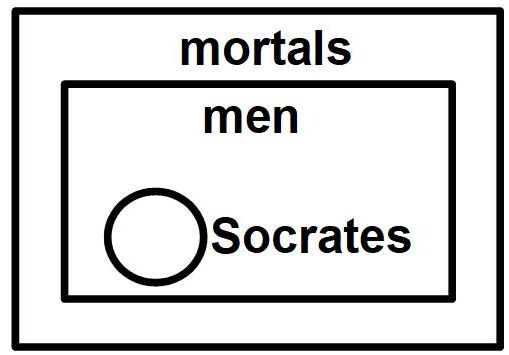}
	\caption{One way to represent the Socrates syllogism in Tumbug.}
	\label{fig:syllogisms-3-rectangle-venn-socrates-circle}
	\end{center}
\end{figure}

The way a person would likely construct a mental picture such as in the aforementioned syllogism would be incrementally as each of the three sentences is read: a single mental picture is used across all three sentences, and after each sentence is read a new piece is added to the composite picture as shown in Figure~\ref{fig:syllogisms-socrates-steps}. Note that the final composite picture does not use the "mortals" concept as a set, but rather as an attribute-value pair.

The clue that a set is involved is the verbal form "all X," such as in "all men" or "all birds" or "all even numbers." In logical calculus or math the term "for all" would be the analogous clue. The English verb "is" (or "are" or "am") when used in descriptions of the form "X is a Y" (or "X are a Y" or "I am a Y") means "X is in set Y," as in "Socrates is man" or "Tweety is a bird" or "6 is an even number."

Using these rules the aforementioned sequence of three steps results mechanically, with little or no thought, while building up the associated image on the side: (1) "All men are mortal" = "The set of men has attribute mortal" $\Rightarrow$ create set "men" and attach attribute-value "mortality = mortal." (2) "Socrates is a man" = "The object Socrates is in the set of men" $\Rightarrow$ create object "Socrates" and place it inside set "men." (3) "Socrates is mortal" = "The object Socrates has attribute mortal" $\Rightarrow$ realize that since set "men" has attribute-value "mortality = mortal" then every object in that set must have that same attribute-value, by convention.

\begin{figure}
	\begin{center}
	\includegraphics[width=0.50\textwidth]{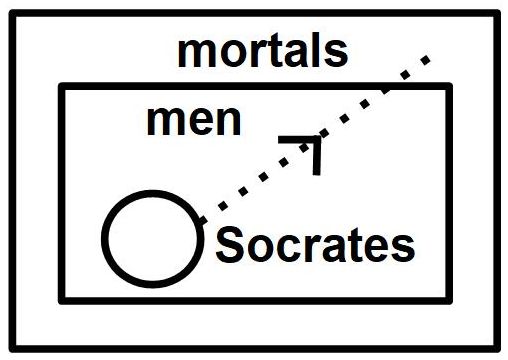}
	\caption{The insight that Socrates is mortal as represented by a Relationship Marker that directly connects the two concepts.}
	\label{fig:syllogisms-mortals-insight}
	\end{center}
\end{figure}

The last (3rd) step as represented in Figure~\ref{fig:syllogisms-mortals-insight} may seem somewhat trivial and redundant. However, if this syllogism had been represented with two sets, one set for "mortals" as is usually done, the insight of the third step could be made visible as an arrow connecting Socrates and the set "mortals." Such an insight more closely describes of people would typically think of such a new association, often as a "that's true, though I never quite thought of it that way before" reaction. This type of spatial fusion evidently happens frequently when mentally piecing together maps to plan a route. For example, if a person knows the layout of the interior of a building well, and knows the relative placement of the surrounding buildings but has always entered and exited through the same entrance of the first building, then if challenged with finding a new exit that is closer to the destination building, would likely mentally need to fuse the two separate mental maps in order to decide which exit would be closer to the destination from a global viewpoint--a shortcut that could be always be used thereafter. This same "insight arrow," called a Relationship Marker in Tumbug, is explicitly shown in all the syllogisms in this document. Note that extra connections such as these are likely responsible for increasing a person's "understanding" of a topic (Atkins 2019).

\begin{figure}
	\begin{center}
	\includegraphics[width=0.85\textwidth]{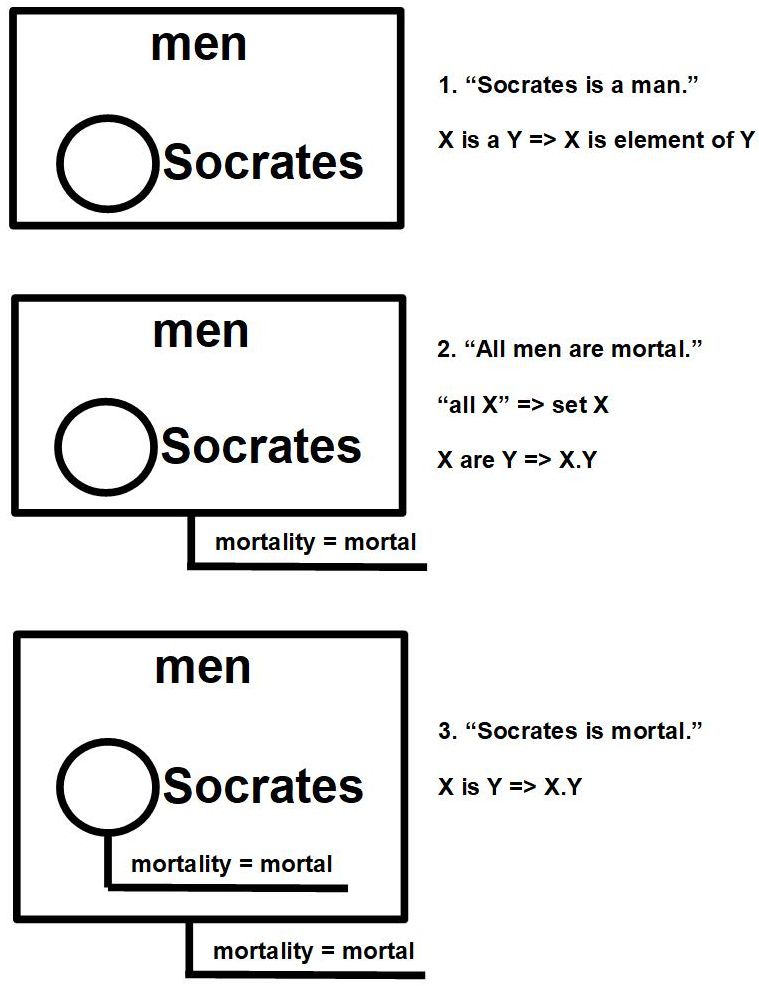}
	\caption{The three diagram creation steps that correspond to the rearranged three sentences of the syllogism.}
	\label{fig:syllogisms-socrates-steps-reordered}
	\end{center}
\end{figure}

Note that the order of the three sentences of the Socrates syllogism is not critical. For example, if the 1st and 2nd steps were swapped, the conclusion (and final diagram) would be the same, but would have been built up in a different order, as shown in the Figure~\ref{fig:syllogisms-socrates-steps-reordered}.\\

\setlength\parindent{0pt}
Socrates is a man.\\
All men are mortal.\\
Therefore, Socrates is mortal.\\
\setlength\parindent{24pt}

Insight: The described process of the brain building up composite images that represent meaning across multiple sentences may provide major clues about the brain automatically providing context for disambiguation of pronouns in subsequent sentences. In contrast, if a sentence containing an ambiguous pronoun was a sentence that was removed from its surrounding sentences then the sentence would likely lose critical information that could disambiguate that pronoun if the sentence were given to a computer to analyze.

\subsubsection{Celarent form (EAE-1)}

\begin{figure}
	\begin{center}
	\includegraphics[width=0.50\textwidth]{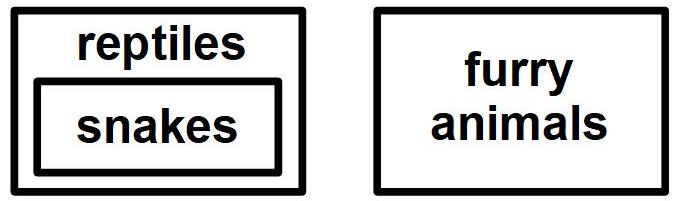}
	\caption{One way to represent the reptile syllogism in Tumbug.}
	\label{fig:syllogisms-reptiles}
	\end{center}
\end{figure}

The reptile syllogism example of this section is of the Celarent form, and is shown in Figure~\ref{fig:syllogisms-reptiles}.\\

\setlength\parindent{0pt}
No reptile has fur.\\
All snakes are reptiles.\\
Therefore, no snake has fur.\\
\setlength\parindent{24pt}

The clue that a set is involved is the verbal form "no X," such as in "no reptile" or "no man" or "no negative number." Implied is that there exists a set called "X" and no element in that set can have the properties described afterward. In logical calculus or math the term "there does not exist" would be the analogous clue. 

\begin{figure}
	\begin{center}
	\includegraphics[width=0.75\textwidth]{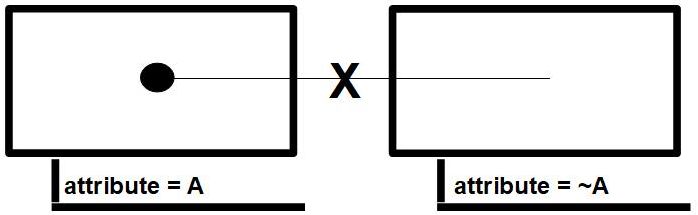}
	\caption{An element of the left set with attribute A cannot appear in the right set.}
	\label{fig:syllogisms-convention-xed-line}
	\end{center}
\end{figure}

\begin{figure}
	\begin{center}
	\includegraphics[width=0.75\textwidth]{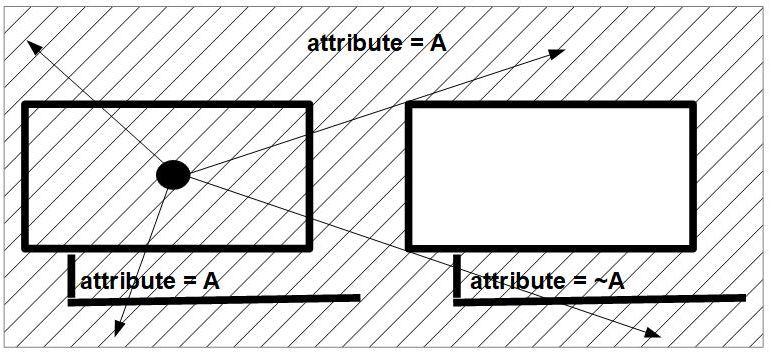}
	\caption{An element of the left set with attribute A can appear in any set or region except the white (~A) region.}
	\label{fig:syllogisms-not-a}
	\end{center}
\end{figure}

A complication arises from this example: How to represent "no" (or "not") with respect to an allowed range in Venn diagrams. Two ways are shown in Figure~\ref{fig:syllogisms-convention-xed-line} and Figure~\ref{fig:syllogisms-not-a}. 
Representation \#1: The X-ed line between the two sets means that an element from one set cannot exist in the set on the other side of the X-ed line.

\begin{figure}
	\begin{center}
	\includegraphics[width=0.75\textwidth]{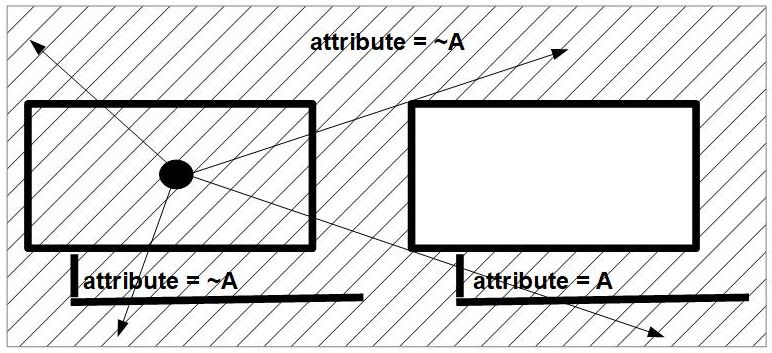}
	\caption{An element of the left set with attribute ~A can appear in any set or region except the white (A) region.
This is the diagram that applies to the reptiles example, where A = fur, ~A = NOT fur.}
	\label{fig:syllogisms-not-not-a}
	\end{center}
\end{figure}

Representation \#2: The black dot represents an element in the set on the left, and the arrows emanating from it suggest the range in which it is allowed to appear in the Venn diagram. The dot can appear in any hatched area, i.e., the dot can appear in any area except the white (unhatched) region. The arrows are shown for clarity here, but are not needed in practice since the shading already spans every region where the element may go. See Figure~\ref{fig:syllogisms-not-a} and Figure~\ref{fig:syllogisms-not-not-a}.

For Tumbug the hatched region representation would probably be clearest, since Tumbug has many lines in the diagram already, and using an X-ed line does not make it clear where the element in the left set can go, only where it cannot go.

\begin{figure}
	\begin{center}
	\includegraphics[width=0.85\textwidth]{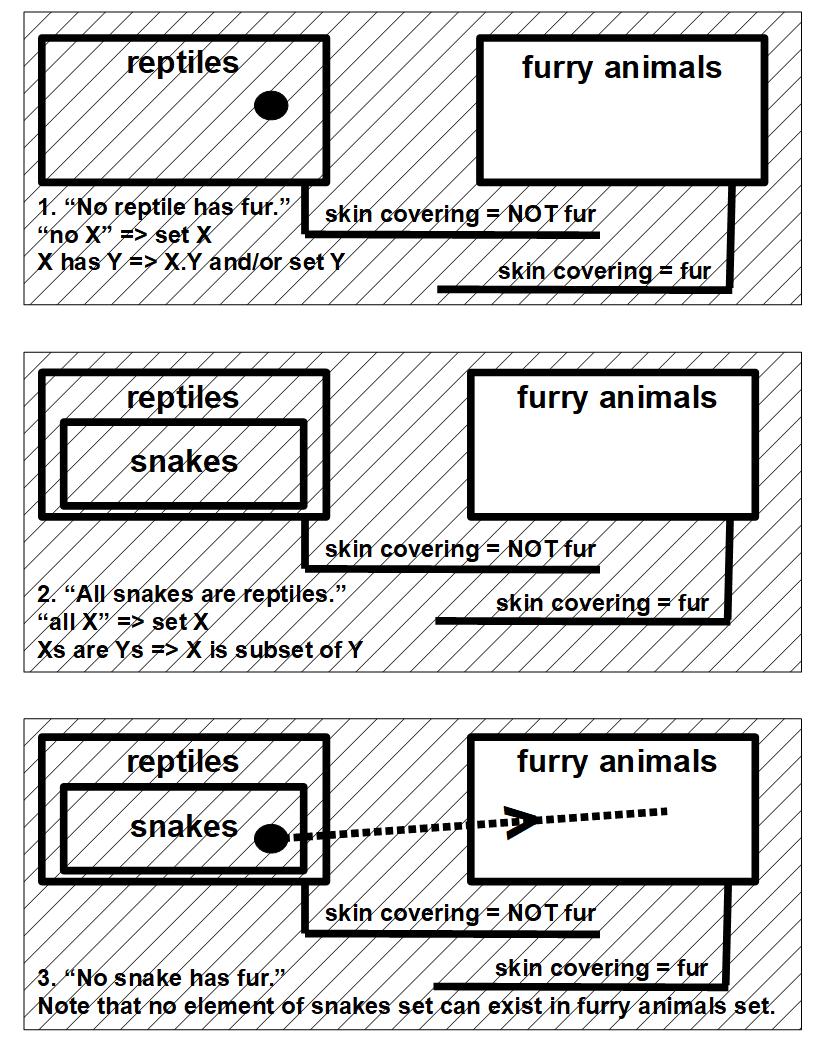}
	\caption{The three diagram creation steps that correspond to the three sentences of the syllogism.}
	\label{fig:syllogisms-reptiles-steps}
	\end{center}
\end{figure}

Using the same conventions as in the Barbara example, the following steps would be involved to create the appropriate Venn diagram, as shown in Figure~\ref{fig:syllogisms-reptiles-steps}. (1) "No reptile has fur" = "The set of reptiles and the set of furry animals cannot intersect" $\Rightarrow$ create set "reptiles" and set "furry animals" so that they are non-intersecting. (Alternative: Attach attribute-value "skin covering = NOT fur" to the reptiles set, and attach attribute-value "skin covering = fur" to the furry animals set.) (2) "All snakes are reptiles" = "The set of snakes is a subset of the set of reptiles" $\Rightarrow$ create set "snakes" and place it inside set "reptiles." (3) "No snake has fur" = "No element in the snakes set can exist in the furry animals set." Realize that since any snake must exist only in the "snakes" set, which is mutually exclusive with the "furry animals" set, this is already implied, by the shading convention.

\subsubsection{Darii form (AII-1)}

\begin{figure}
	\begin{center}
	\includegraphics[width=0.25\textwidth]{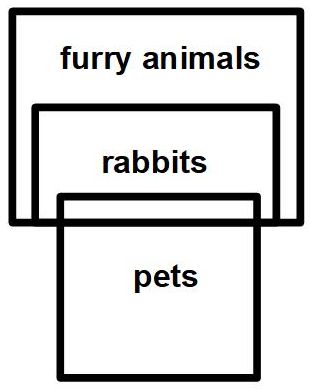}
	\caption{One way to represent the rabbits syllogism in Tumbug.}
	\label{fig:syllogisms-rabbits}
	\end{center}
\end{figure}

The rabbits syllogism example of this section is of the Darii form. Figure~\ref{fig:syllogisms-rabbits} shows this syllogism in square Venn diagram form.\\

\setlength\parindent{0pt}
All rabbits have fur.\\
Some pets are rabbits.\\
Therefore, some pets have fur.\\
\setlength\parindent{24pt}

The clue that a set is involved is the verbal form "all X," here "all rabbits."

\begin{figure}
	\begin{center}
	\includegraphics[width=0.85\textwidth]{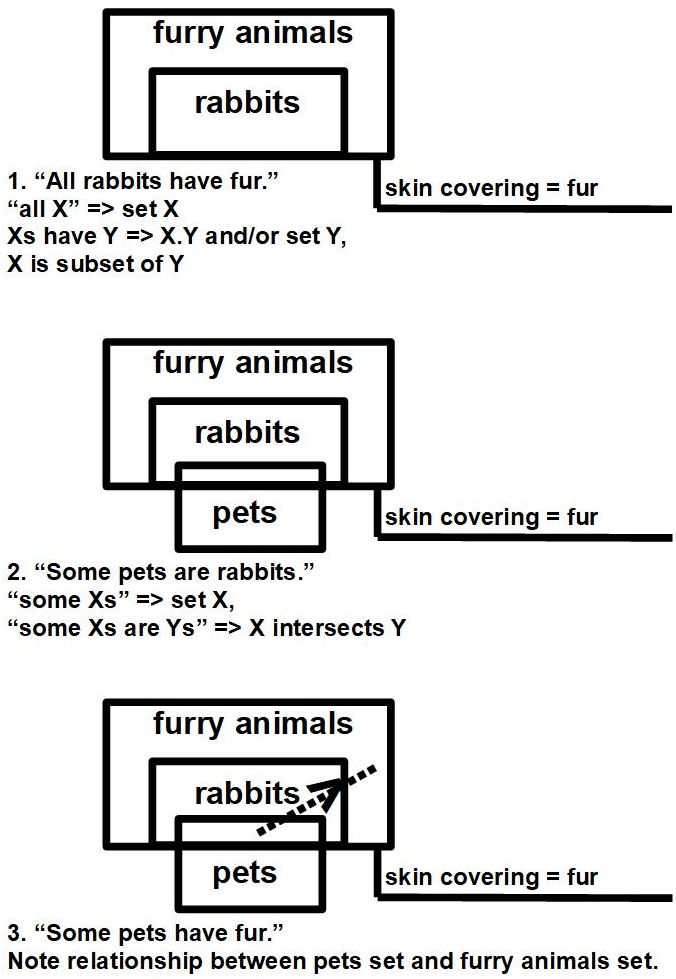}
	\caption{The three diagram creation steps that correspond to the three sentences of the syllogism.}
	\label{fig:syllogisms-rabbits-steps}
	\end{center}
\end{figure}

Using the same conventions as in the Barbara example, the following steps would be involved to create the appropriate Venn diagram, as shown in Figure~\ref{fig:syllogisms-rabbits-steps}. (1) "All rabbits have fur" = "The set of rabbits is a subset of the set of furry animals" $\Rightarrow$ create set "rabbits" and set "furry animals" so that "rabbits" is a subset of "furry animals." (Alternative: Attach attribute-value "skin covering = fur" to the rabbits set.) (2) "Some pets are rabbits" = "Some of the set of pets is inside the set of rabbits, but not necessarily inside the set of furry animals" $\Rightarrow$ create set "pets" and make it overlap with the set "furry animals" but in a way that it does not overlap into the purely "furry animals" region. (3) "Some pets have fur" = "Some part of the pets set is inside the furry animals set." Realize that part of the "pets" set exists in the "furry animals" set.

\subsection{Computer language representation}

Algorithms can be represented in Tumbug rather easily, especially from flowcharts, but also from the code itself.

Sequential steps of an algorithm are shown in Tumbug as occurring sequentially in time, along a Timeline Arrow. The rectangles along the timeline are not flowchart boxes, but rather C Aggregation Boxes. The contents of each C Aggregation Box on the left is the Tumbug representation of the corresponding statement ${s_i}$ in the code to the right, so each C Aggregation Box will typically hold connected Tumbug icons such as C Object Circles, Motion Arrows, attribute-value pairs, etc. Each red dot shown in the code samples here indicates the statement within the program that is executing, and there is a corresponding red dot beside the corresponding part of the Tumbug representation.

\textit{Caveat: Tumbug's ordinary level of representation deals with human life and human-sized objects, so using Tumbug for computer code would likely be extravagantly wasteful of space. The same is true of mathematical operations. Tumbug is similar to a very high order programming language, especially an object-oriented simulation language, so Tumbug would be most appropriate for important human-level events that lend themselves to visual simulation.}

\subsubsection{Sequential code}

\begin{figure}
	\begin{center}
	\includegraphics[width=0.15\textwidth]{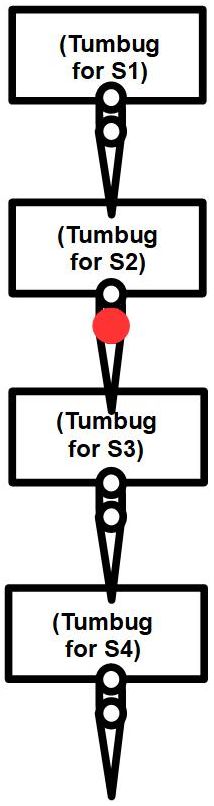}
	\caption{Sequential code in Tumbug is represented by one path composed largely of Pathway Tubes.}
	\label{fig:sequential-flowchart}
	\end{center}
\end{figure}

\begin{figure}
	\begin{center}
	\includegraphics[width=0.50\textwidth]{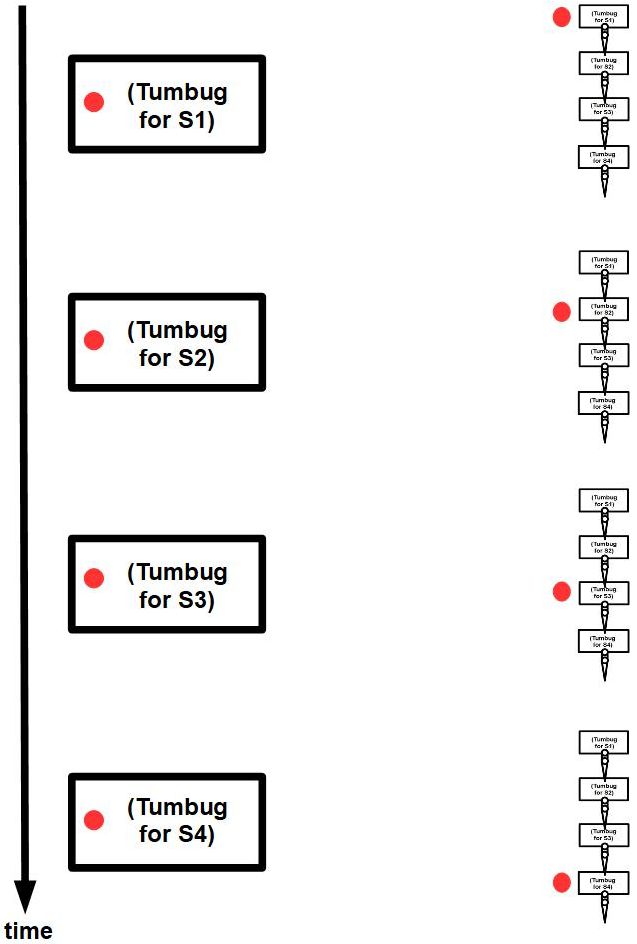}
	\caption{Trace of sequential code in Tumbug, with flowchart copied at each step on right: (S1 S2 S3 S4).}
	\label{fig:sequential-flowchart-trace}
	\end{center}
\end{figure}

Sequential code as a flowchart in Tumbug would look like Figure~\ref{fig:sequential-flowchart}. Figure~\ref{fig:sequential-flowchart-trace} of a Tumbug representation of sequential flowchart run being traced suggests why Tumbug is more applicable to machine understanding of a situation: At each step the entire piece of code is recopied, with the only change being where the current program statement is, which is indicated by the red dot. This results in the ability to see not only the current focus (i.e., the current program statement) but also its context in all the code around it (i.e., the program).

\subsubsection{Looped code}

\begin{figure}
	\begin{center}
	\includegraphics[width=0.50\textwidth]{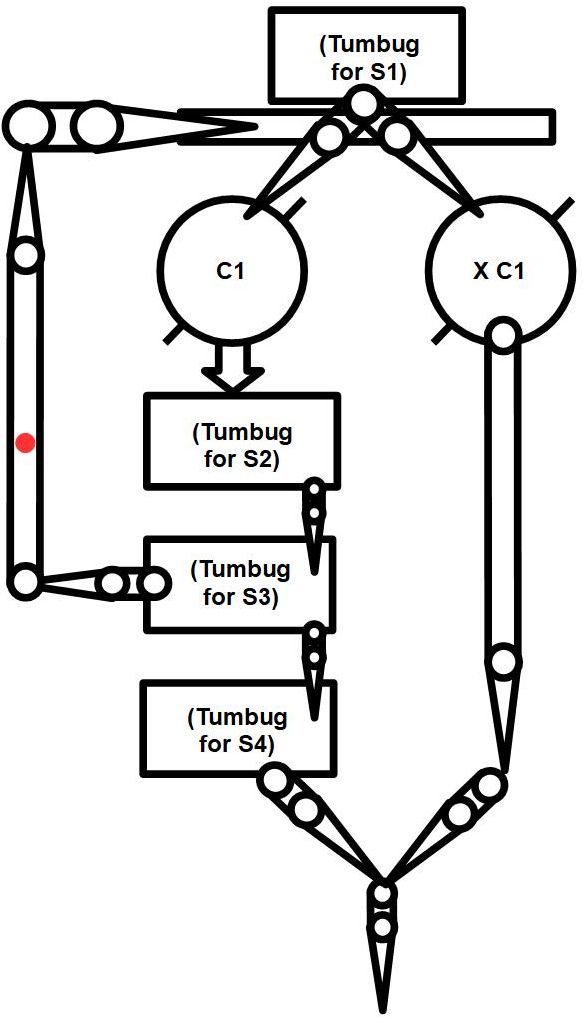}
	\caption{An IF-THEN in Tumbug is represented by two paths, the choice of which depends on one state.}
	\label{fig:loop-flowchart}
	\end{center}
\end{figure}

\begin{figure}
	\begin{center}
	\includegraphics[width=0.90\textwidth]{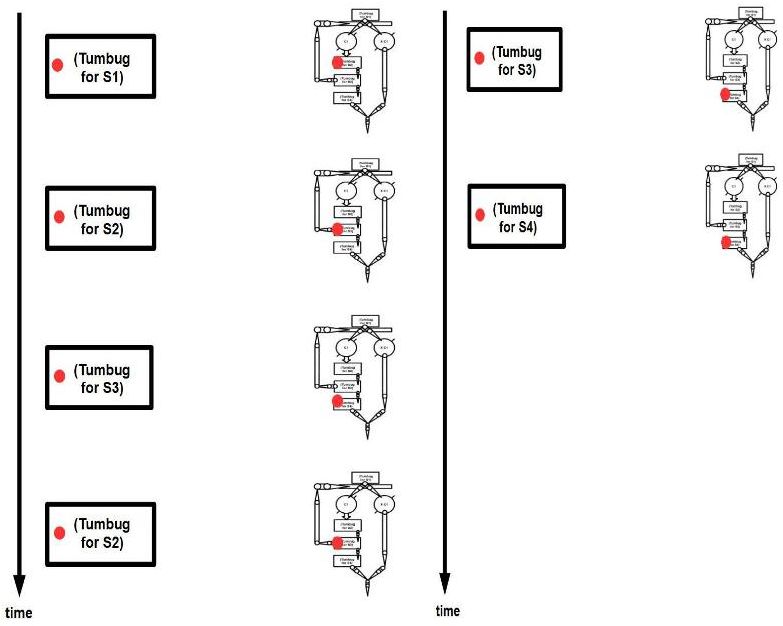}
	\caption{Trace of looped code in Tumbug, with flowchart copied at each step on right: (S1 S2 S3 S2 S3 S4).}
	\label{fig:loop-flowchart-trace-both}
	\end{center}
\end{figure}

A loop in the code would cause the collection of steps inside the loop to occur repeatedly, with each collection occurring sequentially, and the steps within each loop occurring sequentially, as well. This is shown in Figure~\ref{fig:loop-flowchart} and Figure~\ref{fig:loop-flowchart-trace-both}.

\subsubsection{Branching code}

\begin{figure}
	\begin{center}
	\includegraphics[width=0.35\textwidth]{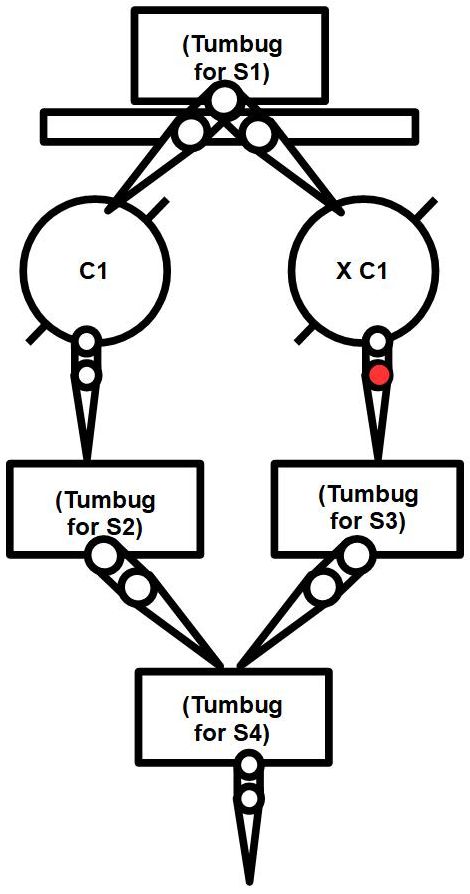}
	\caption{An IF-THEN-ELSE in Tumbug is represented by two paths, the choice of which depends of one state.}
	\label{fig:conditional-tumbug}
	\end{center}
\end{figure}

\begin{figure}
	\begin{center}
	\includegraphics[width=0.55\textwidth]{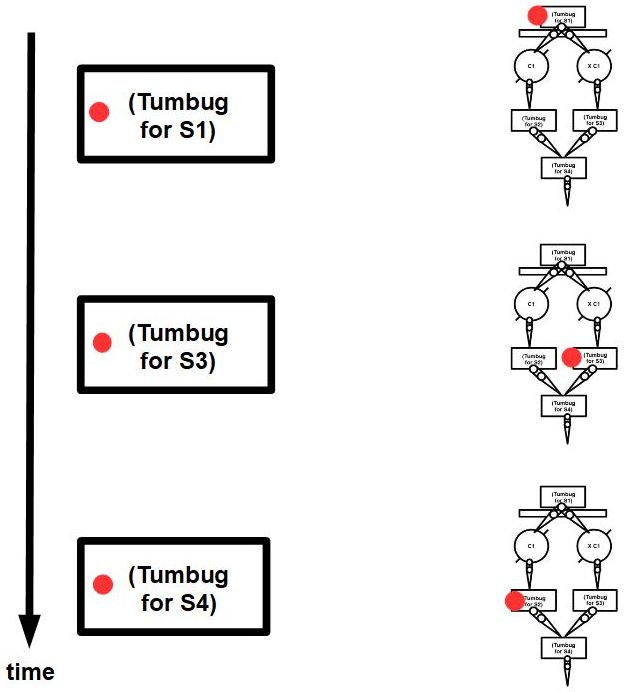}
	\caption{Trace of branching code in Tumbug, with flowchart copied at each step on right: (S1 S3 S4).}
	\label{fig:conditional-flowchart-trace}
	\end{center}
\end{figure}

A conditional statement (such as IF-THEN, IF-THEN-ELSE, or CASE) causes only one thread of execution to occur, chosen appropriately from the multiple alternatives given in the program. Tumbug uses a Split Time Arrow to represent this splitting of possibilities. The result looks very similar to a flowchart, as shown in Figure~\ref{fig:conditional-tumbug} and Figure~\ref{fig:conditional-flowchart-trace}.

.

\subsection{Natural language representation}

This section discusses the more difficult grammatical topics within English that might cause translation from English into Tumbug to be difficult. Almost all these topics are about verbs. The first section covers several of the simpler topics in a single section; each subsequent section covers one complicated topic per section.

\subsubsection{Categories of verbs}

\textbf{1. Active verbs versus stative verbs}

Active verbs are also called "dynamic verbs." Active verbs are verbs that describe obvious actions (e.g., "to throw," "to run," "to hit," "to jump," "to sing"), whereas stative verbs describe state of being (e.g., "to know," "to earn," "to love," "to want," "to own").

Sometimes a third category is included, called "event verbs." This is said to include sentences such as "Four people died in the crash" and "It’s raining again," but these examples fit well as stative verbs since "to die" is a change of state from living to dead, and "to rain" is a change of state of weather to wet precipitation.

\begin{figure}
	\begin{center}
	\includegraphics[width=0.75\textwidth]{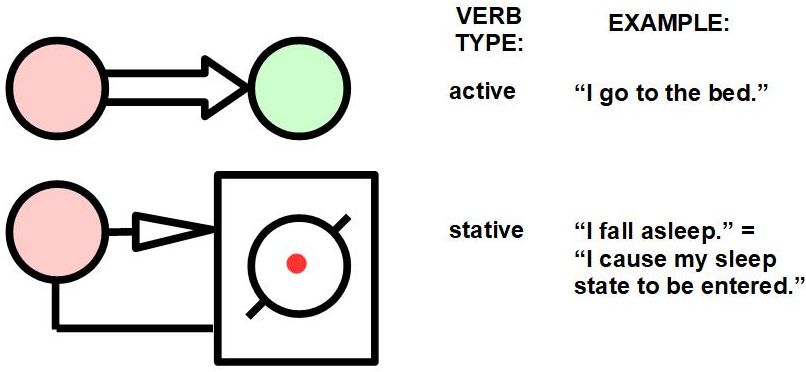}
	\caption{In Tumbug, active verbs use very different arrows than stative verbs.}
	\label{fig:verb-active-stative}
	\end{center}
\end{figure}

The differences between these categories is clearly represented in Tumbug because actions must either involve motion, whereupon a Motion Arrow and C Object Circle is used, or be caused or entered via a state diagram, whereupon a Causation Arrow and State Circle is used. If color coding is used then the difference in diagrams is even more obvious since State Circles have no color whereas C Object Circles do. See Figure~\ref{fig:verb-active-stative}.

\textbf{2. Transitive verbs versus intransitive verbs}

\begin{figure}
	\begin{center}
	\includegraphics[width=0.75\textwidth]{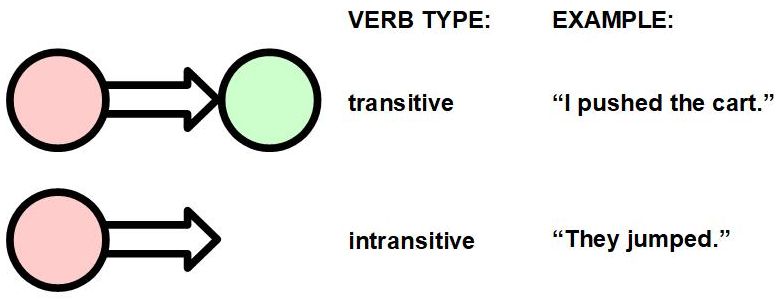}
	\caption{In Tumbug, transitive verbs use two Object Circles whereas intransitive verbs use one Object Circle.}
	\label{fig:verb-transitive-intransitive}
	\end{center}
\end{figure}

A transitive verb is one that requires a recipient object for the sentence to make sense (e.g., "I pushed the cart," "Please bring coffee," "Juan threw the ball"), in contrast to the sentence making without a recipient object (e.g., "They jumped," "The dog ran," "She arrived"). Some verbs can be either, depending on context (e.g., "Her husband shaved" versus "The barber shaved the customer"). The difference is extremely obvious in Tumbug: if only the subject's C Object Circle is present then the verb is transitive, but if the direct object's C Object Circle is present as well as the subject's C Object Circle then the verb is intransitive. See Figure~\ref{fig:verb-transitive-intransitive}.

\textbf{3. Linking verbs}

\begin{figure}
	\begin{center}
	\includegraphics[width=0.75\textwidth]{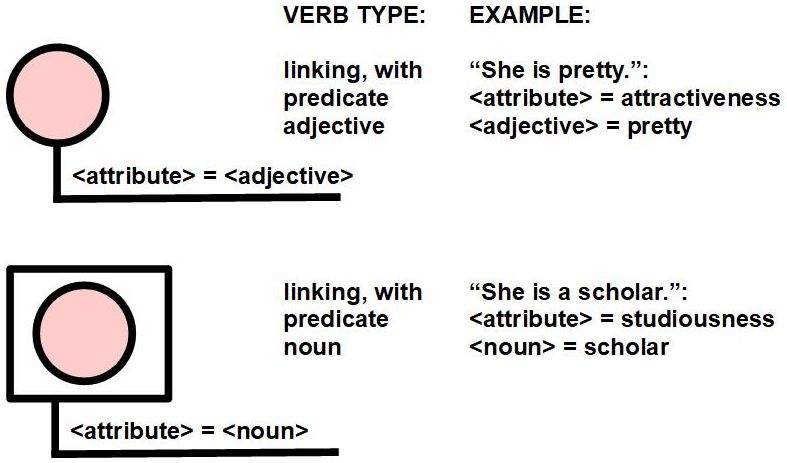}
	\caption{There exist two types of linking verbs, and neither involves an action or a causation.}
	\label{fig:verb-linking}
	\end{center}
\end{figure}

Linking verbs do not describe actions or causations of states, but rather describe an object with either a predicate adjective or a predicate noun. In Tumbug, a predicate adjective is represented by attribute-value pair that hangs off an Object Circle, and a predicate noun (also called a "predicate nominative") is represented as a superset that is represented as a C Aggregation Box that surrounds the predicate noun. The main example of a linking verb in English is "to be," which in Spanish has the two forms "ser" and "estar." Any verb that describes sensory input (e.g., "to feel $\langle$adjective$\rangle$," "to smell $\langle$adjective$\rangle$)," "to taste $\langle$adjective$\rangle$") or any verb that describes impressions (e.g., "to seem $\langle$adjective$\rangle$," "to look $\langle$adjective$\rangle$," "to become $\langle$adjective$\rangle$") is a linking verb. Tumbug inherently does not consider linking verbs to be verbs in the normal sense since Tumbug does not represent linking verbs with any type of arrow, only with Attribute Lines. See Figure~\ref{fig:verb-linking}.

\textbf{4. Helping verbs}

"Helping verbs" are also called "auxiliary verbs." A helping verb is an additional verb used with the main verb to convey some type of additional information. The types of information conveyed can be:

\begin{itemize}
	\item
		time tense
			\begin{itemize}
				\item
					HELPING VERB EXAMPLES: to be, to have, to do
				\item
					SENTENCE EXAMPLES:
					\begin{itemize}
						\item
							I am going to the zoo.
								\begin{itemize}
									\item
										helping verb = "to be"
									\item
										conveyed info = present continuous tense
								\end{itemize}
						\item
							I have waited a long time for this.
								\begin{itemize}
									\item
										helping verb = "to be"
									\item
										conveyed info = present continuous tense
								\end{itemize}
						\item
							I did not want to go home.
								\begin{itemize}
									\item
										helping verb = "to do"
									\item
										conveyed info = simple past tense
								\end{itemize}
				\end{itemize}
		\end{itemize}
	\item
		modality
			\begin{itemize}
				\item
					HELPING VERB EXAMPLES: can, could, may, might, will, would
				\item
					SENTENCE EXAMPLES:
						\begin{itemize}
							\item
								It might rain.
									\begin{itemize}
										\item
											helping verb = "might"
										\item
											conveyed info = possibility
									\end{itemize}
							\item
								Jorgen can skate backwards.
									\begin{itemize}
										\item
											helping verb = "can"
										\item
											conveyed info = (general) ability
									\end{itemize}
							\item
								I may delete this later.
									\begin{itemize}
										\item
											helping verb = "may"
										\item
											conveyed info = possibility
									\end{itemize}
						\end{itemize}
			\end{itemize}
	\item
		emphasis
			\begin{itemize}
				\item
					HELPING VERB EXAMPLES: to do
				\item
					SENTENCE EXAMPLES:
						\begin{itemize}
							\item
								I do like the book!
									\begin{itemize}
										\item
											helping verb = "to do"
										\item
											conveyed info = emphasis
									\end{itemize}
						\end{itemize}
			\end{itemize}
\end{itemize}

\begin{figure}
	\begin{center}
	\includegraphics[width=0.65\textwidth]{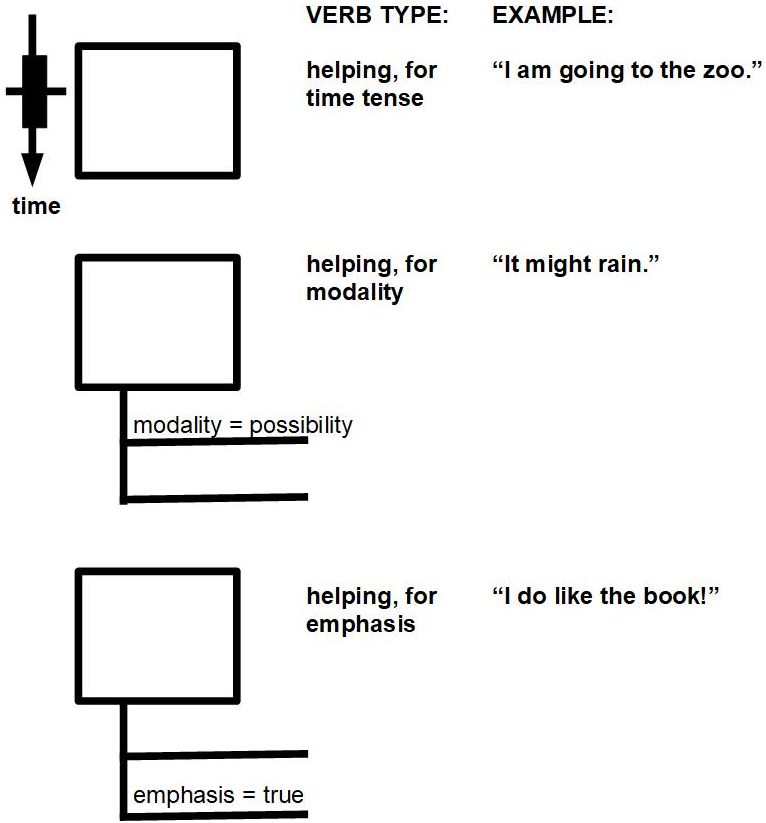}
	\caption{"Helping verbs" is a diverse class of verbs with different uses and representations in Tumbug.}
	\label{fig:verb-helping}
	\end{center}
\end{figure}

From a Tumbug perspective, "helping verb" is an ambiguous concept based on syntax, not grammar, therefore most helping verb categories can be ignored in this study. See Figure~\ref{fig:verb-helping}. However, time tenses and modal verbs are deeper grammatical topics that are discussed in detail elsewhere in this document, each in its own dedicated section.

\textbf{5. Regular verbs versus irregular verbs}

\begin{figure}
	\begin{center}
	\includegraphics[width=0.50\textwidth]{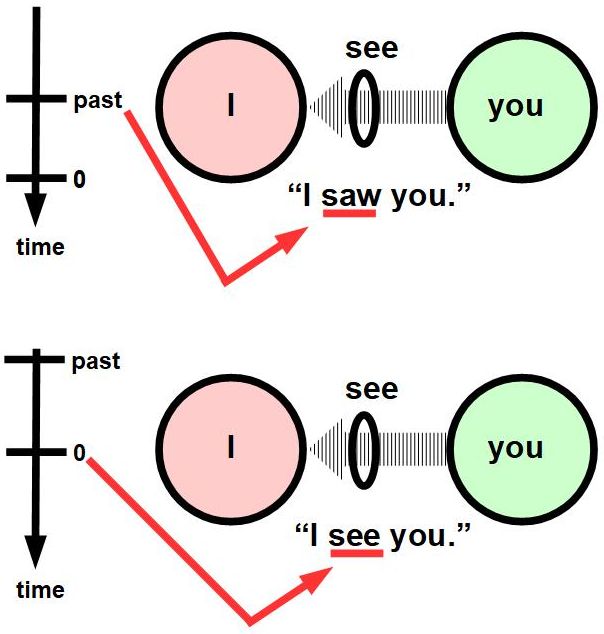}
	\caption{Tumbug ordinarily distinguishes between past and present via timeline, therefore Tumbug is not usually concerned with textual descriptions or their spelling differences since such text is redundant. Therefore the two diagrams are identical except for the positioning of times on the timeline.}
	\label{fig:verb-irregular}
	\end{center}
\end{figure}

The issue of whether a verb is regular or irregular is a syntactical issue, not a semantic issue. Since Tumbug deals with semantics, Tumbug \textit{per se} does not consider whether a verb is regular or irregular, which means for example that there will be no written distinction between the concepts of "see" and "saw": the only difference is implied by the time shown at that event on the timeline (Time Arrow) as shown in Figure~\ref{fig:verb-irregular}. The closest that Tumbug might approach this distinction is if: (1) software of some type were translating from Tumbug representation of a sentence in one language to a different language; (2) Tumbug were being used to store conjugation rules.

\textbf{6. Phrasal verbs}

\begin{figure}
	\begin{center}
	\includegraphics[width=0.60\textwidth]{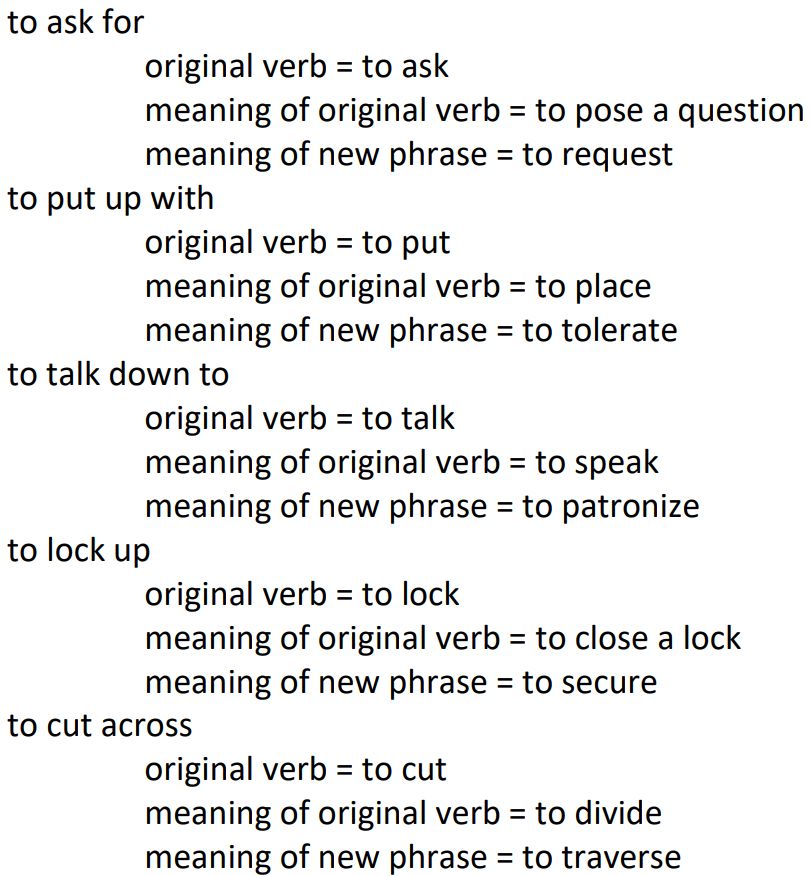}
	\caption{Examples of phrasal verbs. Tumbug is concerned only with unique concepts, not with new meanings that have been attributed to groups of existing words by humans, therefore phrasal verbs have little inherent meaning to Tumbug.}
	\label{fig:phrasal-verb-table-snap}
	\end{center}
\end{figure}

A phrasal verb is a verb that includes a preposition (e.g., for, with, to, up, across) and/or adverb to the verb, which together can create a new meaning that is completely different from the meaning of the original verb alone. Phrasal verbs are irrelevant in Tumbug, for the same reason that conjugations are irrelevant in Tumbug: these are spelling, phrasal, or syntactical differences that are independent of meaning. Some examples are listed in Figure~\ref{fig:phrasal-verb-table-snap}.

A Tumbug diagram of the original verb will usually be very different than a Tumbug diagram of the same verb made into a phrasal verb.

\textbf{7. Infinitives}

\begin{figure}
	\begin{center}
	\includegraphics[width=0.55\textwidth]{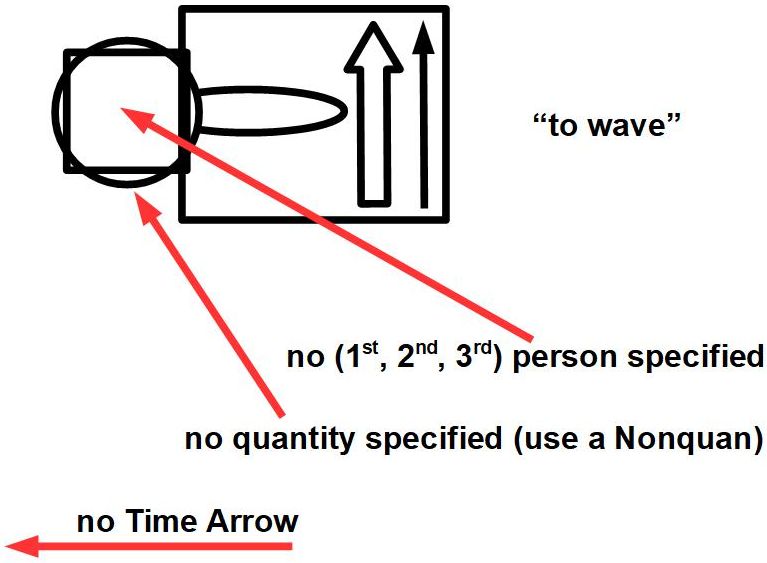}
	\caption{In Tumbug the infinitive form of a verb will show no tense (i.e., no Time Arrow), no quantity (i.e., a Nonquan is shown rather than an Object Circle or Location Box), and no person (i.e., no relative location of the agent is implied by the diagram). Here the arrows suggest an appendage moving upwards, which suggests the infinitive "to wave."}
	\label{fig:verb-infinitive}
	\end{center}
\end{figure}

As noted elsewhere, conjugations of verbs by time, number, and person are essentially meaningless in Tumbug since Tumbug fundamentally uses no text anyway; such conjugation information can only be inferred by the diagram or its attribute values, which also fundamentally use no text. Since the infinitive form of a verb is characterized by lack of time, number, and person, a generic Tumbug diagram that somehow depicts a given action alone is how an infinitive would appear in Tumbug, as shown in Figure~\ref{fig:verb-infinitive}.

\textbf{8. Reflexive verbs}

\begin{figure}
	\begin{center}
	\includegraphics[width=0.45\textwidth]{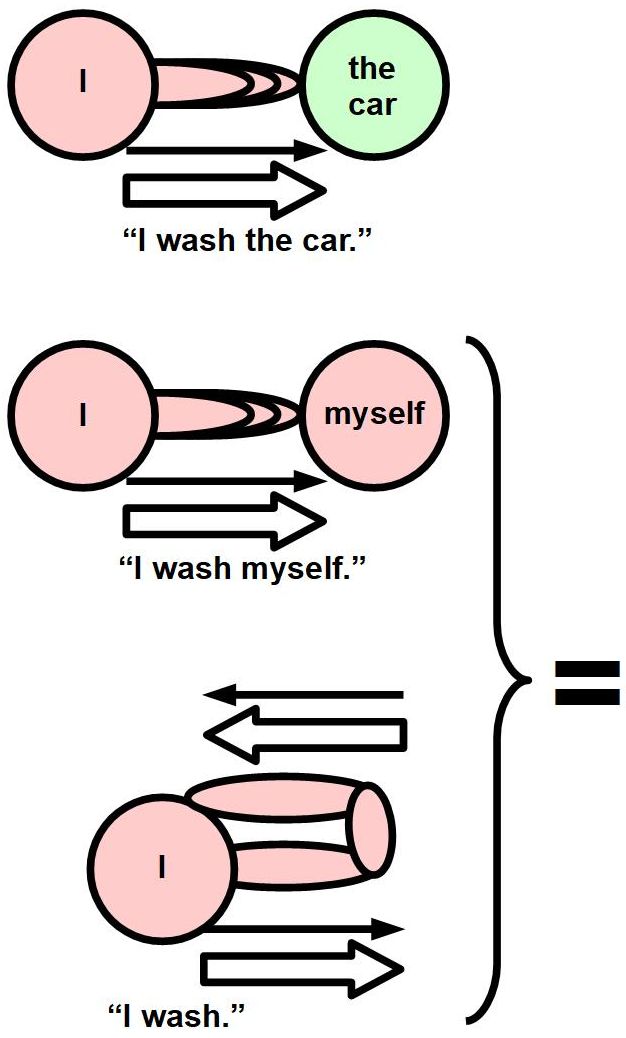}
	\caption{Top: "To wash" as a transitive verb. Bottom: "To wash oneself" or "to wash" as a reflexive verb.}
	\label{fig:verb-reflexive}
	\end{center}
\end{figure}

One definition of a "reflexive verb" is a verb whose direct object is the same as its subject. For example, a verb used with a reflexive pronoun in a sentence of the form "I $\langle$verb$\rangle$ myself" or "You $\langle$verb$\rangle$ yourself" are necessarily reflexive verbs, such as the sentence "I wash myself" as shown in Figure~\ref{fig:verb-reflexive}. Verbs that describe actions that are ordinarily done to oneself, such as "to wash" or "to shave," do not require a reflexive pronoun. In Tumbug, reflexive verbs are immediately visually recognizable because the action immediately circles back to the subject. This is a benefit since no confusion about the meaning of the sentence can occur, unlike in written English.

\subsubsection{Active tense versus passive tense}

\begin{figure}
	\begin{center}
	\includegraphics[width=0.50\textwidth]{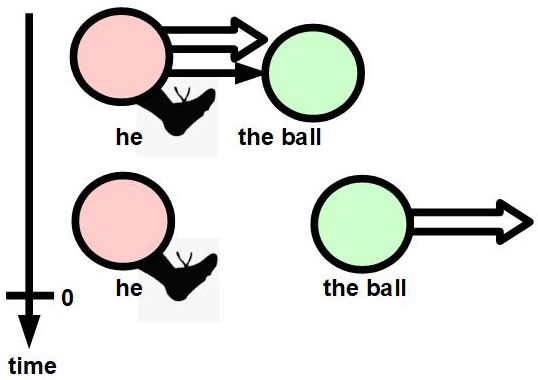}
	\caption{Tumbug for "He kicked the ball." (active tense)}
	\label{fig:active-passive-active}
	\end{center}
\end{figure}

\begin{figure}
	\begin{center}
	\includegraphics[width=0.50\textwidth]{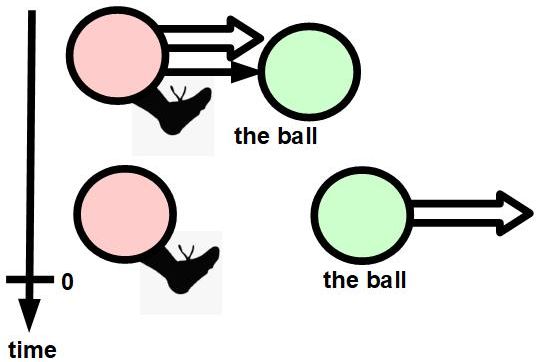}
	\caption{Tumbug for "The ball was kicked." (passive tense)}
	\label{fig:active-passive-passive}
	\end{center}
\end{figure}

\begin{figure}
	\begin{center}
	\includegraphics[width=0.50\textwidth]{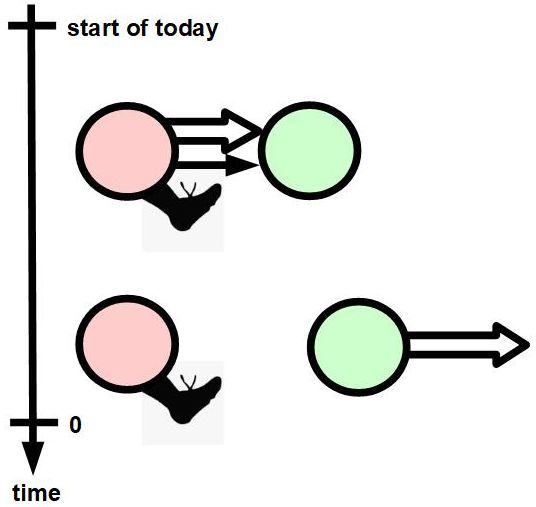}
	\caption{Tumbug for "There was some kicking today." (passive tense)}
	\label{fig:active-passive-kicking-today}
	\end{center}
\end{figure}

Marvin Minsky mentioned active tense versus active tense (Minsky 1974, p. 33) via the three example sentences in Figure~\ref{fig:active-passive-active}, Figure~\ref{fig:active-passive-passive}, Figure~\ref{fig:active-passive-kicking-today}.

Tumbug works as beautifully on passive tense as it does on active tense. If one of the objects is not identified, a completely unlabeled Object Circle can be used, or a question mark "?" can either be placed inside an Object Circle, a lone question mark can be used, or an icon can be labeled with a lone question mark can be used. The representation method that is most consistent with the spirit of Tumbug is that of showing a completely unlabeled Object Circle for a sentence's Subject because some Subject must have caused the action: a missing Subject would imply that the Subject either did not exist or that action arose out of a void, or both, which is logically impossible. In addition, the placing of an unlabeled Object Circle in the diagram makes it less likely that such passive tense wording can be used to misdirect the reader, which is often the intent of passive tense. With this Tumbug convention any unaccounted-for object must automatically be brought into the diagram with an obvious presence that is difficult to ignore. For an example of this justification, in the sentence "The ball was kicked," the Merriam-Webster definition of "to kick" is "to strike out with the foot or feet," which means that a visual representation of "to kick" absolutely must contain some reference to a foot, which in turn implies the existence of some entity that had that foot as part of its body, an entity that was capable of striking out with that foot. This undefined entity is exactly the meaning of an unlabeled Physical Object Circle, therefore such a Physical Object Circle with a foot should be used, which has the automatic side benefit of making any implied unknowns blatantly obvious instead of being hidden via language.

\subsubsection{Grammatical time aspects}

\begin{figure}
	\begin{center}
	\includegraphics[width=0.30\textwidth]{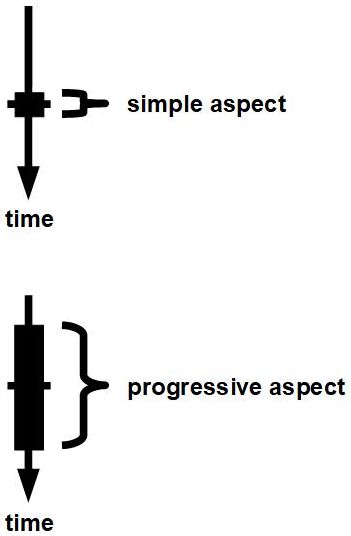}
	\caption{Aspect is largely only a matter of how long the interval of time is, over which an action happens.}
	\label{fig:aspect-summary}
	\end{center}
\end{figure}

The basic time ranges in which events are described in natural languages are, of course: (1) past, (2) present, and (3) future. Most natural languages add an additional complication to timelines called "aspect," however. The most common aspects are: (1) simple, (2) progressive, (3) perfect, (4) present progressive. "Aspect" describes whether a given event happened at a specific point in time, versus whether a given event happened over a period of time. Each of these cases is listed in Figure~\ref{fig:aspect-summary} with its overall description and associated Tumbug diagram. Note that these particular meanings of types of aspect are not always fixed--natural language is imperfect--but for any specific interpretation Tumbug will render them faithfully. For example, although "simple" aspect usually means at a single point in time, it can also refer to a cluster of points in time.

In this section the time spans that are of interest are emphasized using a thickened Time Arrow in those regions. This practice is not a fixed part of Tumbug: braces or shaded regions can accomplish the same task of indicating any specific times spans of interest.

\textbf{1. Simple aspect}

\begin{figure}
	\begin{center}
	\includegraphics[width=0.50\textwidth]{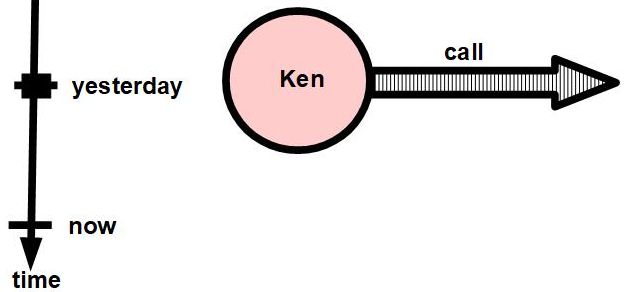}
	\caption{(in the past:) "Ken called (via an information line) yesterday."}
	\label{fig:aspect-simple-past}
	\end{center}
\end{figure}

\begin{figure}
	\begin{center}
	\includegraphics[width=0.50\textwidth]{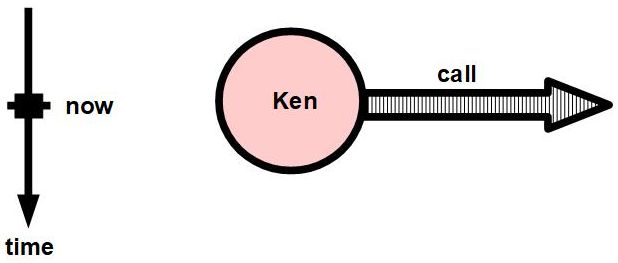}
	\caption{(in the present:) "Ken is calling (via an information line) now."}
	\label{fig:aspect-simple-present}
	\end{center}
\end{figure}

\begin{figure}
	\begin{center}
	\includegraphics[width=0.50\textwidth]{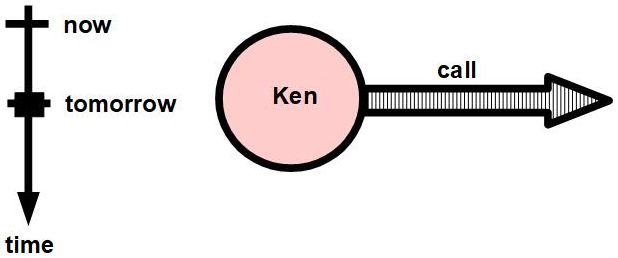}
	\caption{(in the future:) "Ken will call (via an information line) tomorrow."}
	\label{fig:aspect-simple-future}
	\end{center}
\end{figure}

The "simple" aspect means a single point in time, or sometimes a cluster of such points, shown here by the thick line that ranges over narrow period of time. See Figure~\ref{fig:aspect-simple-past}, Figure~\ref{fig:aspect-simple-present}, Figure~\ref{fig:aspect-simple-future}.
 
\textbf{2. Progressive aspect}

\begin{figure}
	\begin{center}
	\includegraphics[width=0.65\textwidth]{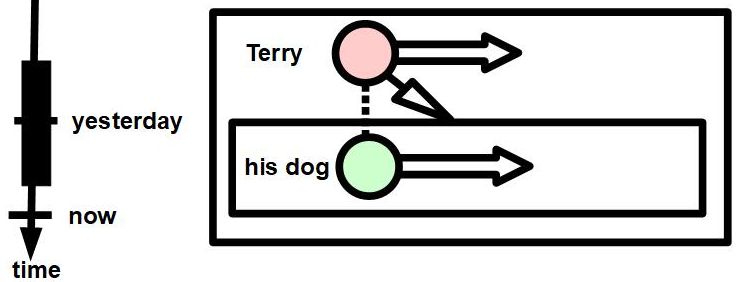}
	\caption{(in the past:) "Terry was walking his dog yesterday."}
	\label{fig:aspect-progressive-past}
	\end{center}
\end{figure}

\begin{figure}
	\begin{center}
	\includegraphics[width=0.65\textwidth]{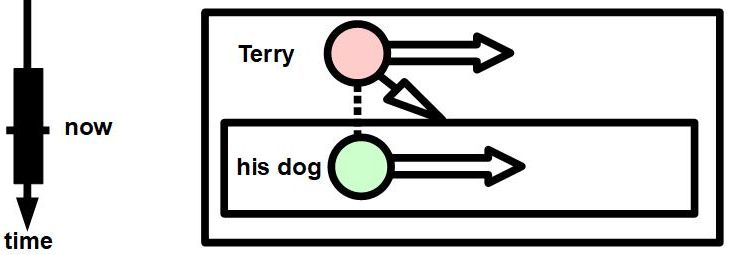}
	\caption{(in the present:) "Terry is walking his dog now."}
	\label{fig:aspect-progressive-present}
	\end{center}
\end{figure}

\begin{figure}
	\begin{center}
	\includegraphics[width=0.65\textwidth]{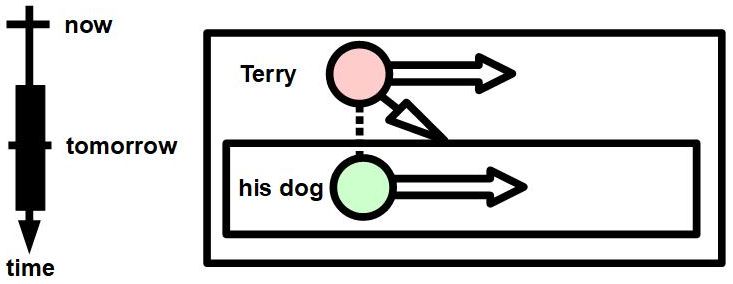}
	\caption{(in the future:) "Terry will be walking his dog tomorrow."}
	\label{fig:aspect-progressive-future}
	\end{center}
\end{figure}

The "progressive" aspect means the described action occurred over a range of time. This situation is shown here by the thick line that ranges over an extended period of time longer than a single moment. See Figure~\ref{fig:aspect-progressive-past}, Figure~\ref{fig:aspect-progressive-present}, Figure~\ref{fig:aspect-progressive-future}.

\textbf{3. Perfect aspect}

"Perfect" in grammatical context means "complete," meaning the described action has stopped. The "perfect" aspect means an ongoing action in the past was completed by a certain reference time.

\begin{figure}
	\begin{center}
	\includegraphics[width=0.50\textwidth]{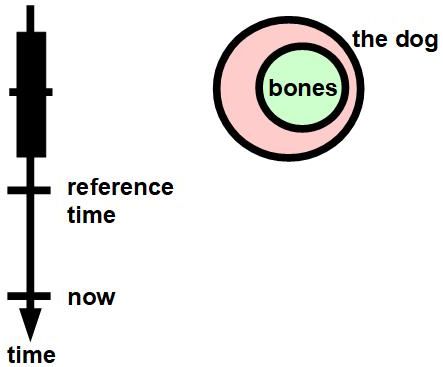}
	\caption{(perfect in the past / past perfect:) "The dog had eaten bones (by the time we got home)."}
	\label{fig:aspect-perfect-past}
	\end{center}
\end{figure}

\begin{figure}
	\begin{center}
	\includegraphics[width=0.50\textwidth]{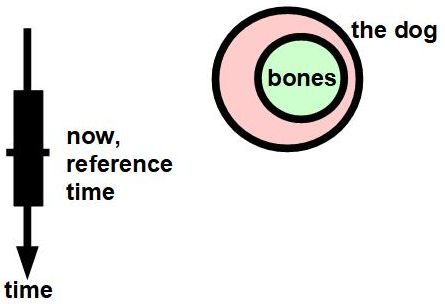}
	\caption{(perfect in the present / present perfect:) "The dog has eaten bones (by now)."}
	\label{fig:aspect-perfect-present}
	\end{center}
\end{figure}

\begin{figure}
	\begin{center}
	\includegraphics[width=0.50\textwidth]{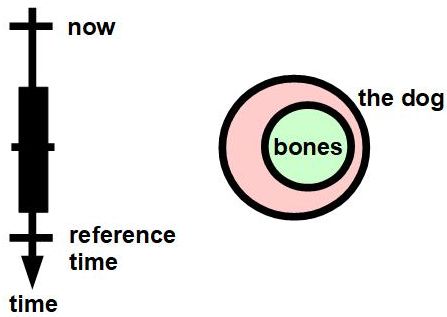}
	\caption{(perfect in the future / future perfect:) "The dog will have eaten bones (by the time we get home).}
	\label{fig:aspect-perfect-future}
	\end{center}
\end{figure}

See Figure~\ref{fig:aspect-perfect-past}, Figure~\ref{fig:aspect-perfect-present}, Figure~\ref{fig:aspect-perfect-future}.

Note that in the present perfect the reference time coincides with "now." This differs from the other two perfect tenses (past perfect and future perfect), where the reference time is always assumed to be after the time of the described action, and with both the described action and reference time on the same side of "now."

\textbf{4. Perfect-progressive aspect}

\begin{figure}
	\begin{center}
	\includegraphics[width=0.50\textwidth]{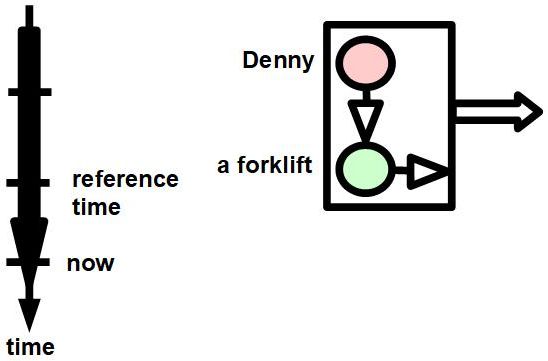}
	\caption{(in the past / past perfect progressive:) "Denny had been driving a forklift (and might have continued to do so)."}
	\label{fig:aspect-perfect-progressive-arrow-past}
	\end{center}
\end{figure}

\begin{figure}
	\begin{center}
	\includegraphics[width=0.50\textwidth]{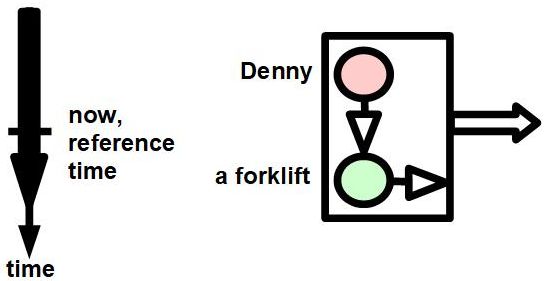}
	\caption{(in the present / present perfect progressive:) "Denny has been driving a forklift (and might still be doing so)."}
	\label{fig:aspect-perfect-progressive-arrow-present}
	\end{center}
\end{figure}

\begin{figure}
	\begin{center}
	\includegraphics[width=0.50\textwidth]{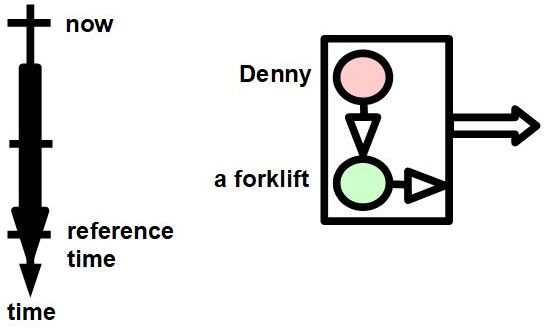}
	\caption{(in the future / future perfect progressive:) "Denny will have been driving a forklift (and might continue to do so)."}
	\label{fig:aspect-perfect-progressive-arrow-future}
	\end{center}
\end{figure}

"Perfect" in grammatical context means "complete," meaning the described action has stopped. The "progressive" aspect means the described action occurred over a range of time. The "perfect-progressive" aspect means an action over a length of time was completed. See Figure~\ref{fig:aspect-perfect-progressive-arrow-past}, Figure~\ref{fig:aspect-perfect-progressive-arrow-present}, Figure~\ref{fig:aspect-perfect-progressive-arrow-future}.

Note that English is ambiguous in this aspect since this aspect does not necessarily imply that the described action will happen after the reference time. Some English instructors say the described action is "expected to continue," others say the described action "may continue." English obviously has some flaws for accurate scientific usage. The figures show that the thick line of event duration continues (via arrow head), which implies the "expected to continue" interpretation was intended. Figure~\ref{fig:aspect-perfect-progressive-arrow-past}, Figure~\ref{fig:aspect-perfect-progressive-arrow-present}, and Figure~\ref{fig:aspect-perfect-progressive-arrow-future} are identical to the preceding figures except the event duration stops, which implies the "may continue" interpretation was intended.

\begin{figure}
	\begin{center}
	\includegraphics[width=0.60\textwidth]{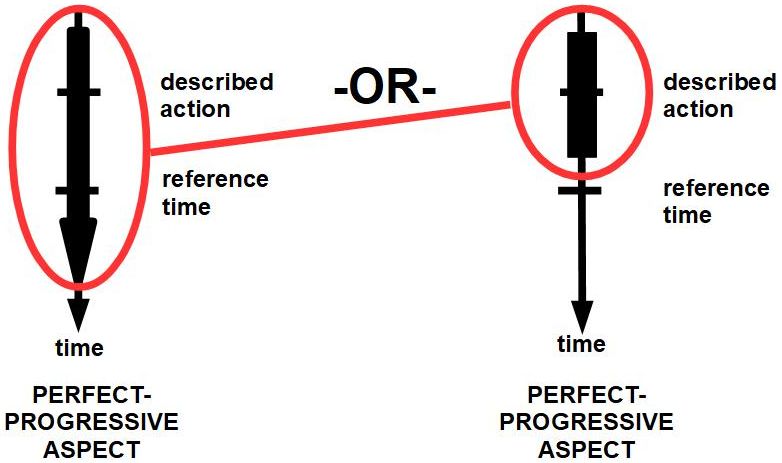}
	\caption{In unfortunate English language ambiguity, perfect-progressive aspect can mean either that the described action continued or that it stopped.}
	\label{fig:aspect-summary-alternatives-or}
	\end{center}
\end{figure}

\begin{figure}
	\begin{center}
	\includegraphics[width=0.60\textwidth]{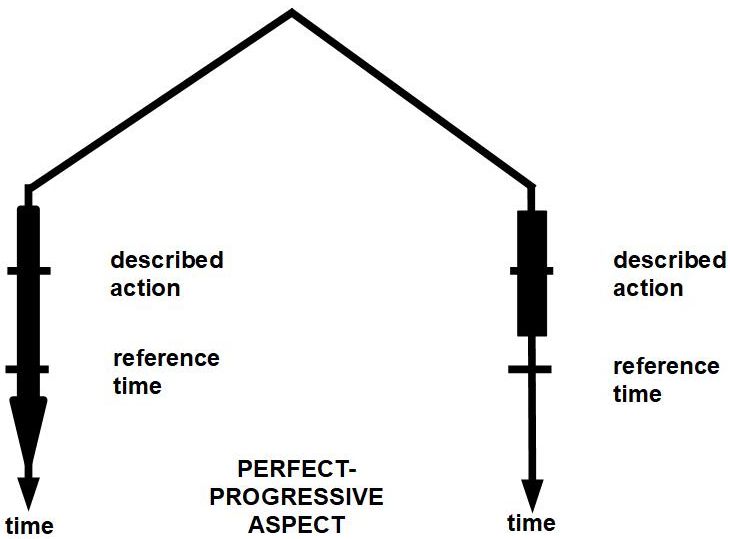}
	\caption{A Split Time Arrow can be used to include both possibilities of "perfect-progressive aspect."}
	\label{fig:aspect-summary-alternatives-split-timeline}
	\end{center}
\end{figure}

\begin{figure}
	\begin{center}
	\includegraphics[width=0.60\textwidth]{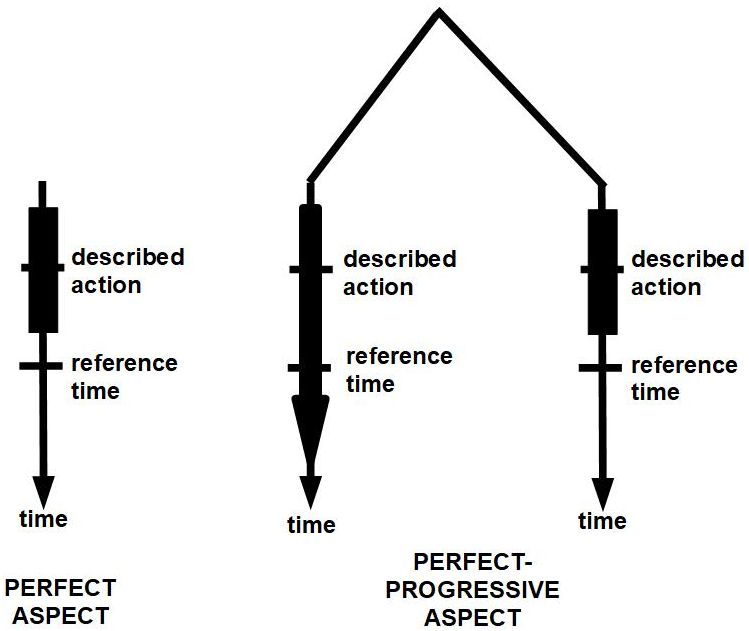}
	\caption{A comparison of the "perfect aspect" diagram with the Split Time Arrow diagram of "perfect-progressive aspect."}
	\label{fig:aspect-summary-comparison}
	\end{center}
\end{figure}

The difference between perfect aspect and progressive-perfect aspect is summarized in Figure~\ref{fig:aspect-summary-alternatives-or}, Figure~\ref{fig:aspect-summary-alternatives-split-timeline}, and Figure~\ref{fig:aspect-summary-comparison}.

\textbf{5. Discussion of grammar}

The preceding sections demonstrate that Tumbug can accurately represent all the time aspects of English grammar, so well in fact that certain ambiguities in English time aspects become very clear.

Note that Tumbug representations can be made much more thoroughly visual than they were in the aforementioned aspect diagrams. For example, instead of writing "yesterday," "now," or "tomorrow" on the timelines, any type of visual representation could be used instead, such as a clock, a calendar, numbers, or lengths of a cylinder that represents some multiple of time. Similarly, icons of humans can be used instead of C Object Circles. Also, details can always be added, such as a human holding a leash that is around the dog's neck, which would presumably represent an additional force (represented by a Force Arrow) that constrains the dog's movement. Tumbug is ultimately intended to be entirely visual, and it has very nearly that capability already if the human diagrammer has the time to fill in the visual details.

\subsubsection{Modal verbs}

\textbf{1. Lists}

\textbf{1.1. Simple lists}

The following list is likely a complete list of the modal verbs of American English and British English, in alphabetical order:

\begin{enumerate}
	\item
		be able to
	\item
		can
	\item
		could
	\item
		had best (uncommon, spoken only)
	\item
		had better (uncommon)
	\item
		have got to (uncommon)
	\item
		have to
	\item
		may
	\item
		might
	\item
		must
	\item
		needn’t (uncommon)
	\item
		ought to
	\item
		shall
	\item
		should
	\item
		will
	\item
		would
\end{enumerate}

These modal verbs precede regular verbs, and convey the following modal concepts that modify the context of the regular verbs that follow them:

\begin{enumerate}
	\item
		ability
	\item
		advice
	\item
		conditional - decomposes
	\item
		formal directive
	\item
		future
	\item
		habit in past tense
	\item
		habit, present tense
	\item
		intention
	\item
		likelihood/certainty
	\item
		not necessary
	\item
		obligation/correctness/rightness
	\item
		offer/invitation/persuasion
	\item
		permission
	\item
		(polite) request
	\item
		prediction
	\item
		suggestion
	\item
		willpower/intent (after "I" or "we" in formal British)
\end{enumerate}

Unfortunately the modal verbs listed above are highly overloaded, meaning that one modal verb can have multiple meanings from the above list. The above list is a conglomeration of the words used to describe of concept meanings given by different authors on the Internet. Below is the nearly the same list of modal verbs, with most of their possible conveyed concepts added, each concept taken from the above list.\\

\textbf{1.2. Modal verb $\Rightarrow$ modal concept}

\begin{enumerate}
	\item
		be able to
			\begin{enumerate}
				\item
					Mike is 
					\textbf{able to} solve complicated math equations. (ability)
				\item
					Will you \textbf{be able to} walk the dog this afternoon? ((polite) request)
			\end{enumerate}
	\item
		can
			\begin{enumerate}
				\item
					She \textbf{can} swim underwater. (ability)
				\item
					You \textbf{can}’t learn a new language if you don’t practice. (conditional - decomposes)
				\item
					I \textbf{can} be with you at 9 a.m. (likelihood/certainty)
				\item
					\textbf{Can} I give you a hand? (offer/invitation/persuasion)
				\item
					Yes, you \textbf{can} have an ice-cream. (permission)
				\item
					\textbf{Can} you change this for a different color, please? ((polite) request)
			\end{enumerate}
	\item
		could
			\begin{enumerate}
				\item
					He \textbf{could} run the mile in 4 minutes when he was younger. (ability)
				\item
					If I \textbf{could} be President for one day, I would change the world. (conditional - decomposes)
				\item
					I \textbf{could} make friends easily when I was younger. (habit in past tense)
				\item
					Mary \textbf{could} have been better than her sister at ballet. (likelihood/certainty)
				\item
					I \textbf{could} go for you? (offer/invitation/persuasion)
				\item
					\textbf{Could} you pass me the salt please? ((polite) request)
				\item
					\textbf{Could} I leave early today, please? (permission)
			\end{enumerate}
	\item
		had best
			\begin{enumerate}
				\item
					You\textbf{’d best} leave it till Monday. (advice, informal)
			\end{enumerate}
	\item
		had better
			\begin{enumerate}
				\item
					I \textbf{had better} not buy that coat. (advice, formal)
			\end{enumerate}
	\item
		have got to
			\begin{enumerate}
				\item
					Drivers \textbf{have got to} get a license to drive a car in the US. (necessity)
				\item
					I \textbf{have got to} be at work by 8:30 AM. (obligation/correctness/rightness)
			\end{enumerate}
	\item
		have to
			\begin{enumerate}
				\item
					We \textbf{have to} wear a uniform at work. (obligation/correctness/rightness)
			\end{enumerate}
	\item
		may
			\begin{enumerate}
				\item
					She \textbf{may} come later. (likelihood/certainty)
				\item
					You \textbf{may} go now. (permission)
				\item
					\textbf{May} I speak to Mr. Jones please? ((polite) request)
			\end{enumerate}
	\item
		might
			\begin{enumerate}
				\item
					The train \textbf{might} be delayed. (likelihood/certainty)
				\item
					\textbf{Might} I have a glass of water please? ((polite) request)
				\item
					She \textbf{might} like a jewelry box for her birthday. (suggestion)
			\end{enumerate}
	\item
		must
			\begin{enumerate}
				\item
					It's snowing, so it \textbf{must} be very cold outside. (likelihood/certainty)
				\item
					Children \textbf{must} do their homework. (obligation/correctness/rightness)
			\end{enumerate}

	\item
		needn't
			\begin{enumerate}
				\item
					You \textbf{needn't} do your homework. (not necessary)
			\end{enumerate}
	\item
		ought to
			\begin{enumerate}
				\item
					You \textbf{ought to} have your car serviced before the winter. (advice)
			\end{enumerate}
	\item
		shall
			\begin{enumerate}
				\item
					Pupils \textbf{shall} not use the main entrance. (formal directive)
				\item
					This time next week I \textbf{shall} be in Scotland. (future)
				\item
					\textbf{Shall} we tell him? (offer/invitation/persuasion)
				\item
					We \textbf{shall} not be moved! (willpower/intent)
			\end{enumerate}
	\item
		should
			\begin{enumerate}
				\item
					You \textbf{should} stop smoking. (advice)
				\item
					He \textbf{should} be here in 5 minutes. (future)
				\item
					The government \textbf{should} reduce the sales tax. (ideal/preferred)
				\item
					\textbf{Should} I turn the heating on? (offer/invitation/persuasion)
				\item
					\textbf{Should} we invite Sarah and David? (possibility)
			\end{enumerate}
	\item
		will
			\begin{enumerate}
				\item
					This year Christmas \textbf{will} fall on a Monday. (future)
				\item
					John \textbf{will} always be late! (habit, present tense)
				\item
					She \textbf{will} come tomorrow. (intention)
				\item
					You \textbf{will} need to show your boarding pass at the gate. (obligation/correctness/rightness)
				\item
					Short hair \textbf{will} suit you. (prediction)
				\item
					\textbf{Will} you take the dog for a walk please? ((polite) request)   
			\end{enumerate}
	\item
		would
			\begin{enumerate}
				\item
					Tom \textbf{would} do something like that, wouldn't he? (habit, present tense)
				\item
					When I lived in Italy, we \textbf{would} often eat in the restaurant next to my flat. (habit in past tense)
			\end{enumerate}
	\end{enumerate}

In particular, requests such as "Could you pass me the salt please?" are at the polite end of a spectrum of imperatives, the other end of which would be a stark imperative such as "Pass me the salt." Prediction also involves a spectrum of likelihood between certainty and non-certainty, such as "The total will be 10" versus "Short hair will suit you." Note that the "permission" concept of some modal verbs is analogous to having a pass, whereas the "formal directive" concept (in the negative) is the opposite, and analogous to not having a pass, therefore these two concepts are related.\\

\begin{figure}
	\begin{center}
	\includegraphics[width=0.85\textwidth]{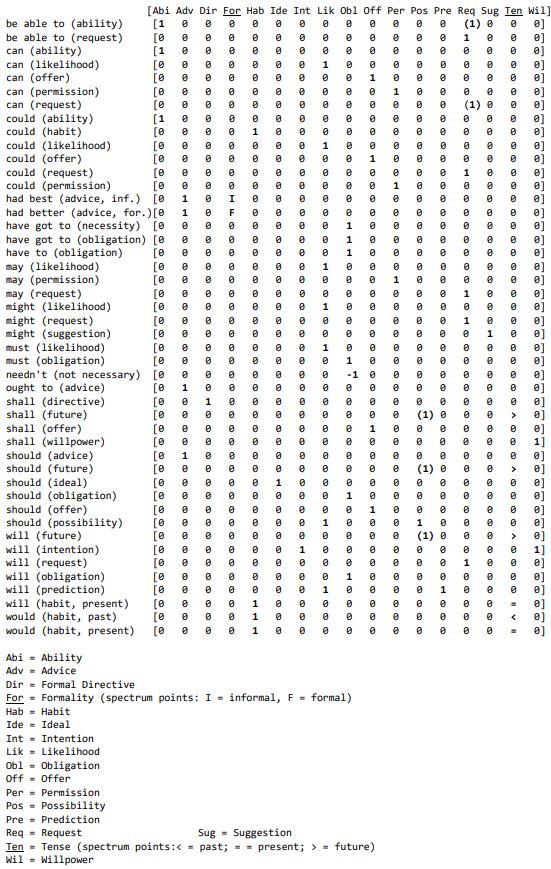}
	\caption{All the modal verbs of English and their meanings. A cell with parenthesized contents means this extra implication should exist.}
	\label{fig:modal-verb-table-snap}
	\end{center}
\end{figure}

There are some interesting or important things to know about these modal verbs: (1) Consensus among linguists is that Future Tense always implies possibility, at least if the agents are people. (2) "To be able" seems at first impression like a synonym for "can" but a little-known English grammar rule renders this first impression untrue since the generality of the event being discussed dictates which of those two modal verbs must be used. However, some people maintain that either "to be able" or "can" can be used for a (polite) request, even though politeness usage is rarely mentioned in conjunction with "to be able." The table in Figure~\ref{fig:modal-verb-table-snap} includes these extra meanings for "to be able," and includes the implication of possibility with each use of Future Tense, though with these extra meanings put in parentheses.\\

In summary, the following concept meanings were excluded for the following reasons:

\begin{itemize}
	\item
		Conditional: irrelevant since "if-then" has no effect on the concept included
	\item
		Desire: often seems implied, but not listed by anybody as implied, as in fact is not logically implied
\end{itemize}

Modal verbs may be redundant, especially in tense, due to prior coverage by Tumbug. For example, one meaning of "shall" is mere future tense: In British English "I shall" is exactly equivalent to "I will" in American English, yet Tumbug already is capable of showing future tense with a timeline so Tumbug does not need a modal verb to express future tense another way.

In this way the original list of modal concepts is reduced to the following shorter list of modal concepts:

\begin{enumerate}
	\item
		Ability
	\item
		Advice
	\item
		Formal Directive
	\item
		Formality (Formal, Informal)
	\item
		Habit (Past Tense, Present Tense)
	\item
		Ideal
	\item
		Intention
	\item
		Likelihood
	\item
		Obligation
	\item
		Offer
	\item
		Permission
	\item
		Possibility
	\item
		Prediction
	\item
		(polite) Request
	\item
		Suggestion
	\item
		Tense
	\item
		Willpower
\end{enumerate}

\textbf{2. Table of modal verbs versus meanings}

Under the assumption that the above list is definitive, a vector that contained all of these attributes would therefore capture all the nuances of every modal verb in English by inclusion or exclusion of selected modal concepts within the vector's components. For example, if the vector's components were defined as... 

[Ability, Advice, Formal Directive, Formality, Habit, Intention, Likelihood, Obligation, Offer, Permission, Request, Possibility, Prediction, Suggestion, Will]

...then the modal verbs of American English and British English would be defined as in Figure~\ref{fig:modal-verb-table-snap}.

\textbf{3. Tumbug implementation}

All of the above lists and summary table of modal verbs converted the ambiguities of modal verbs into discrete cases arranged as mathematical vectors (= each of the lists) and a matrix (= the table). Tumbug eventually must be able to visually represent any given modal verb, however, which is discussed next.

\begin{figure}
	\begin{center}
	\includegraphics[width=0.75\textwidth]{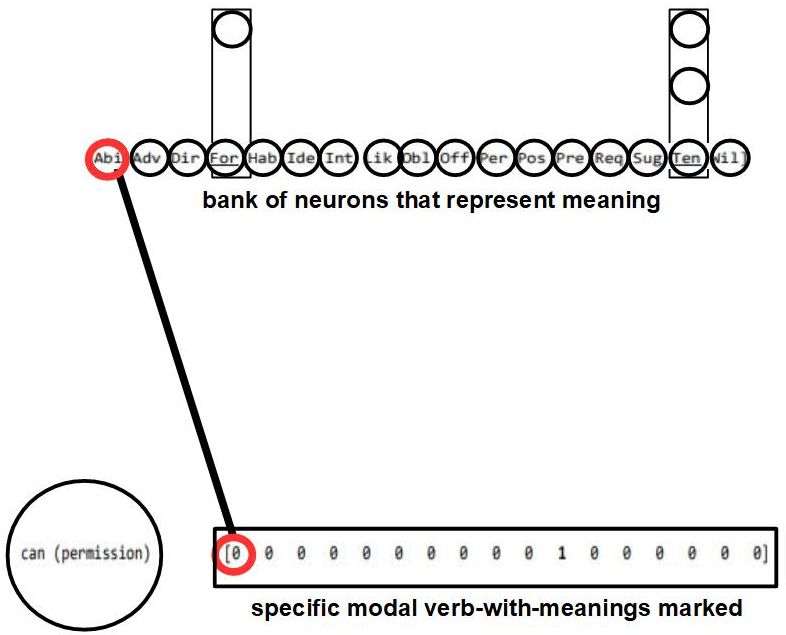}
	\caption{Tumbug implementation of modal verb-with-meaning "can (permission)."}
	\label{fig:verb-implementation-can-permission}
	\end{center}
\end{figure}

The table in Figure~\ref{fig:modal-verb-table-snap} can be regarded as a crossbar switch, where each row is a specific modal value, and each column is a modal concept. As each specific modal verb-with-meaning (row) is activated, one or more associated modal concepts (columns) become activated. If each row is implemented as an artificial neuron and each column is implemented as an artificial neuron, then the firing of one row causes the firing of one more columns as dictated by the connections in crossbar switch. The Figure~\ref{fig:verb-implementation-can-permission} shows the specific modal verb-with-meaning "can (permission)" neuron activating the modal concepts "Per" (= Permission) and "Req" (= Request) neurons. Note that a device other than an artificial neuron could be used instead, such as a light-emitting diode (LED) as a signaling device.

\begin{figure}
	\begin{center}
	\includegraphics[width=0.75\textwidth]{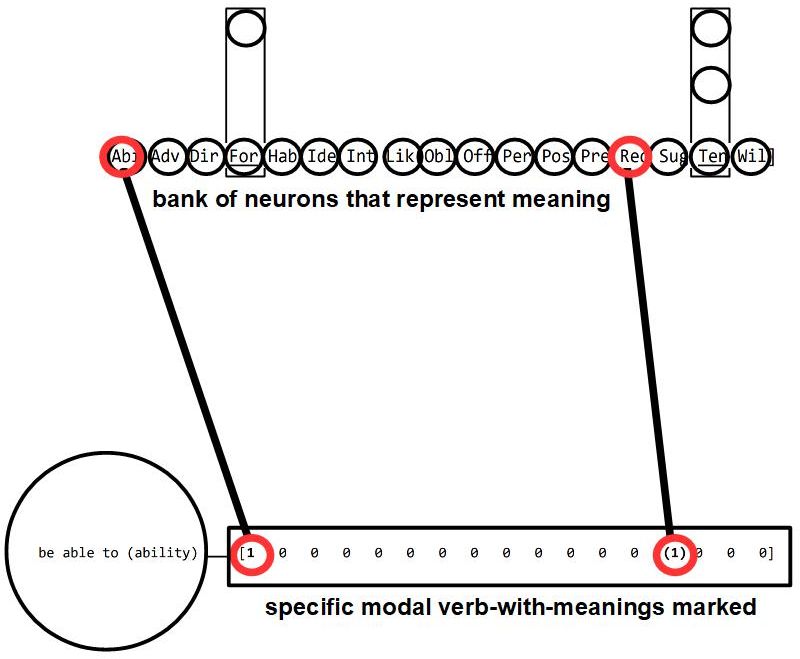}
	\caption{Tumbug implementation of modal verb-with-meaning "be able to (ability)."}
	\label{fig:verb-implementation-able-to-ability}
	\end{center}
\end{figure}

Similar Figure~\ref{fig:verb-implementation-able-to-ability} shows the specific modal verb-with-meaning "able to (ability)" neuron activating the modal concept "Abi" (= Ability) neuron and the modal concept "Req" (= Request) neuron. 

\begin{figure}
	\begin{center}
	\includegraphics[width=0.50\textwidth]{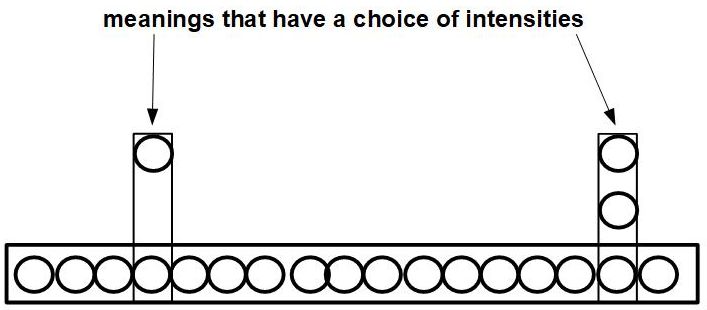}
	\caption{The skeletal template for any modal verb, which will represent a single modal verb when appropriate cells are filled in.}
	\label{fig:verb-implementation-shape-unactivated}
	\end{center}
\end{figure}

When only the column header and its choices are considered, this combination of slots appears as a pegboard-like pattern, shown in Figure~\ref{fig:verb-implementation-shape-unactivated}. Activation of a neuron within one of the vertically protruding banks of neurons could be implemented in various ways, such as: (1) using the vertical, perpendicular bank of neurons as shown in the figure; (2) omitting the vertical bank of neurons, and using a neuron firing at a different intensity in the horizontal bank of neurons, such as firing weakly to represent Informal versus firing strongly to represent Formal.

When this shape is enhanced with activated neurons in unique locations, each activated location shown here in white, a uniquely colored variation of the pegboard pattern results. Each of these unique visual patterns can be considered a visual representation of a different modal verb.

\begin{figure}
	\begin{center}
	\includegraphics[width=1.00\textwidth]{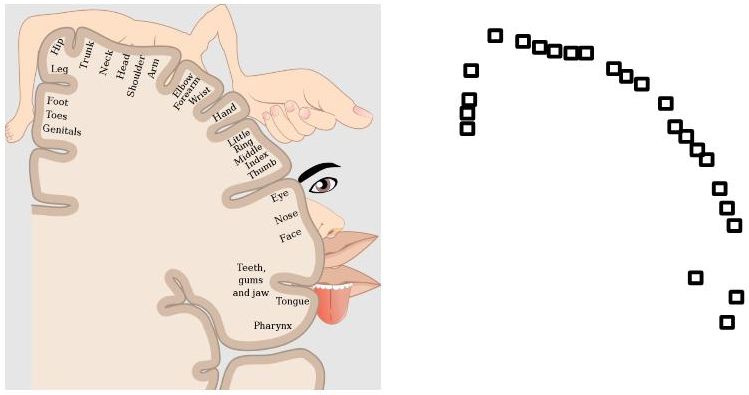}
	\caption{Left: A "body map" of a human as represented by a human's somatosensory cortex, with labeled regions. (Source: Course Hero, Inc.) Right: A Swirly Array whose cells spatially correspond to each labeled region on the left.}
	\label{fig:cortical-map-both}
	\end{center}
\end{figure}

Although such a visual pattern might seem to bear no resemblance to any image a human might associate with a given modal verb, this does not matter. The reason it does not matter is that the brain routinely uses maps that do not intuitively correspond to any map in the real world. For example, a spatial map exists across the somatosensory cortex that represents its person's body, roughly in the same shape as the body in one region (see Figure~\ref{fig:cortical-map-both}), and does (or at least could) contain a representation every part of the body that contains sensors, whether internal or external. Via this map, stimulation of a given set of neurons is inherently understood by the brain to represent signals from that corresponding part of the body, not understood as signals from neurons themselves. In fact, the phenomenon of "phantom limbs" has been hypothesized by Dr. Ronald Melzack to be caused by a "neuromatrix" that is an intermediate biological neural network that sometimes errs in the mapped locations that it forwards to the brain (Melzack and Wall 1965). There is no reason to believe that collections of neurons cannot represent nonexistent or abstract concepts such as modal verb meanings in the same way. Swirly arrays bridge the gap between smooth, irregular, biological structures, and regular, grid-like, computer data structures.

\textbf{4. WS150 example: \#31 (garden)}

"[31] There is a gap in the wall. You can see the garden through it. You can see the garden through what? POSSIBLE ANSWERS: \{the gap, the wall\}"

\begin{figure}
	\begin{center}
	\includegraphics[width=0.55\textwidth]{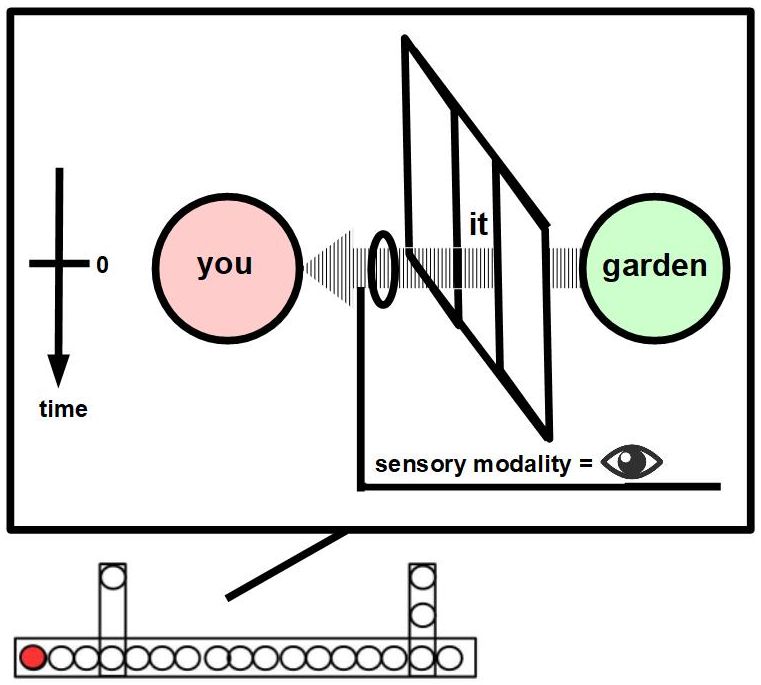}
	\caption{[31] Tumbug for "You can see the garden through it."}
	\label{fig:ws-031-can-see}
	\end{center}
\end{figure}

In Figure~\ref{fig:ws-031-can-see}, the entire diagram for "can (ability)" is used in place of text, which renders that modal verb-with-meaning a pure image. Similarly, an eye icon is used in place of the word "vision" after "sensory modality = ".

\begin{figure}
	\begin{center}
	\includegraphics[width=0.90\textwidth]{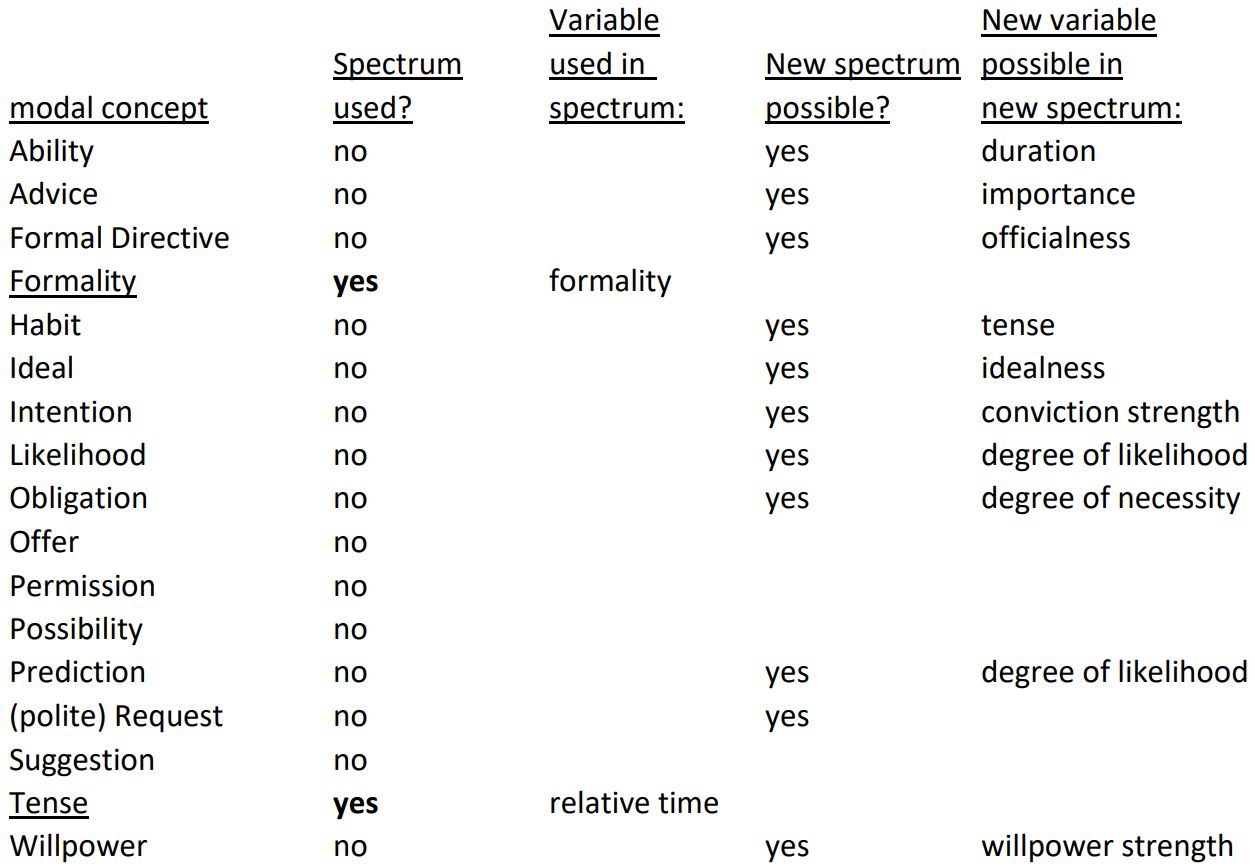}
	\caption{All modal concepts currently covered by Tumbug, with emphasis on those that may involve a spectrum of values.}
	\label{fig:modal-concepts-table-snap}
	\end{center}
\end{figure}

Figure~\ref{fig:modal-concepts-table-snap} shows which of the modal concepts use multiple choices for their attribute values, and shows additional attributes that could optionally be used in a choice-like or spectrum-like manner.

\subsubsection{Propositional attitudes}

\textbf{1. Lists}

A propositional attitude is a mental state held by an agent about a proposition. The presence of a propositional attitude in a sentence is signaled by a verb of attitude followed immediately by a clause that begins with "that," a clause that the verb governs. Some examples from WS150 are: "[41] I'm sure \textbf{that} my map will show this building..." has the verb "to be sure that" that signals belief. "[133] Dr. Adams informed Kate \textbf{that} she had cancer..." has the verb "to inform that" that signals knowledge. "[109] Bill thinks \textbf{that} calling attention to himself was rude to Bert..." has the verb "to think that..." that signals belief.

The propositional attitudes listed by Davis are (Davis 1990, ch. 8): 

\begin{enumerate}
	\item
		belief/believe/believes
	\item
		knowledge
	\item
		knowing whether and what
	\item
		perception/perceive/perceives
\end{enumerate}

Stuart J. Russell and Peter Norvig list some of these and more (Russell and Norvig 2010, p. 450):

\begin{enumerate}
	\item
		belief/believe/believes
	\item
		intends
	\item
		informs
	\item
		knows
	\item
		want/wants
\end{enumerate}

Various online sources list these:

\begin{enumerate}
	\item
		assert
	\item
		belief/believe/believes
	\item
		command
	\item
		consider
	\item
		deny
	\item
		desire
	\item
		doubt
	\item
		fear
	\item
		hope
	\item
		imagine
	\item
		intend
	\item
		judge
	\item
		know
	\item
		perception/perceive/perceives
	\item
		want/wants
	\item
		wish
\end{enumerate}

\textbf{2. Tumbug implementation}

Iconic representation of propositional attitudes is difficult because of their artificial nature and because of their quantity. Unlike emotions and modal verbs, each of which constitute a relatively small quantity of basic categories that are often grouped by similarity, there exist at least dozens of propositional attitudes of extremely varied meanings. As a result, a slightly different way of representing propositional attitudes should be used for Tumbug. This document does not recommend any specific solution in the form of a master icon that shows all propositional attitudes (similar to the Robinson Icon), though in general it is recommended that the methods of organization used earlier in this document for emotions and modal verbs be used, namely (1) generalization of meaning, (2) alphabetization, and (3) dedication of one neuron per concept. For example, the following categories appear promising, and cover all the aforementioned listed propositional attitudes:

\begin{itemize}
	\item
		emotional motivation: {fear, hope}
	\item
		general motivation: {desire, intend, want, wish}
	\item
		cognitive: {believe, consider, deny, doubt, imagine, judge, know, perceive}
	\item
		communication: {assert, inform}
	\item
		grammatical: {command}
\end{itemize}

\textit{Pro tem}, in this document propositional attitudes are represented as text, namely as Label Strings, placed at the top of a C Aggregation Box to avoid the need to label a typically short, tilted line stemming from the agent. Tumbug uses a C Aggregation Box to hold a proposition, and the C Aggregation Box is given an attribute-value pair in text that contains the propositional attitude, such as "belief = true" or "desirability = false." Note that propositional attitudes need not necessarily have discrete true or false values. For example, "believe" and "desire" can span a wide range of applicability strength, and "imagine" and "perceive" are not binary at all.

Note that proposition attitudes are typically based on a multitude of incoming data from multiple senses, therefore the specific sensory modalities involved are hidden as an irrelevant detail, and similarly Motion Arrows representing the flow of objects (such as soup, for taste) or data (such as light, for vision) are omitted when propositional attitudes are used. Instead, single Attribute Line connects a C Aggregation Box that summarizes the impression, perception, belief, knowledge, etc. that the subject has about what is happening inside the C Aggregation Box.

Also note that a propositional attitude always involves an observer who is interpreting the situation depicted in the C Aggregation Box. This is analogous to a cathected emotion, as in the Robinson Icon, meaning that the observer is investing energy in an object, idea, or person, which in the case of propositional attitudes is a situation. Therefore a propositional attitude situation diagrammed in Tumbug must always contain: (1) the observer (Object Circle, though often unlabeled), (2) the situation (C Aggregation Box), (3) the attitude (label over the C Aggregation Box), (4) the link (Attribute Line) between the observer and the situation.

\begin{figure}
	\begin{center}
	\includegraphics[width=0.60\textwidth]{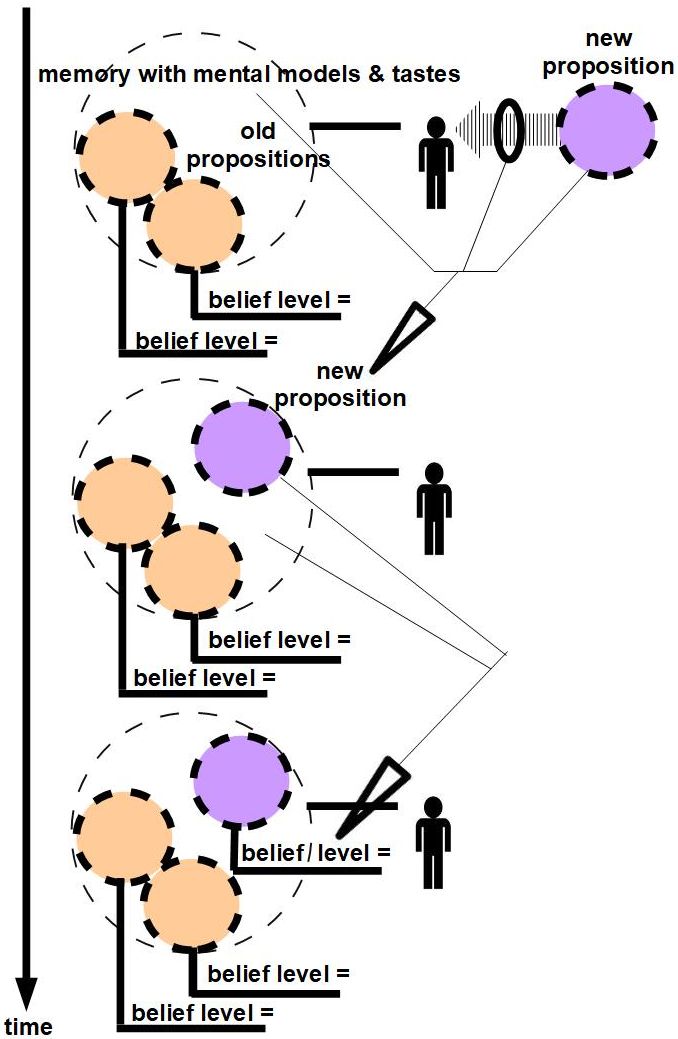}
	\caption{Belief: New proposition automatically enters memory, and is automatically assigned a belief level.}
	\label{fig:pa-belief}
	\end{center}
\end{figure}

\begin{figure}
	\begin{center}
	\includegraphics[width=0.60\textwidth]{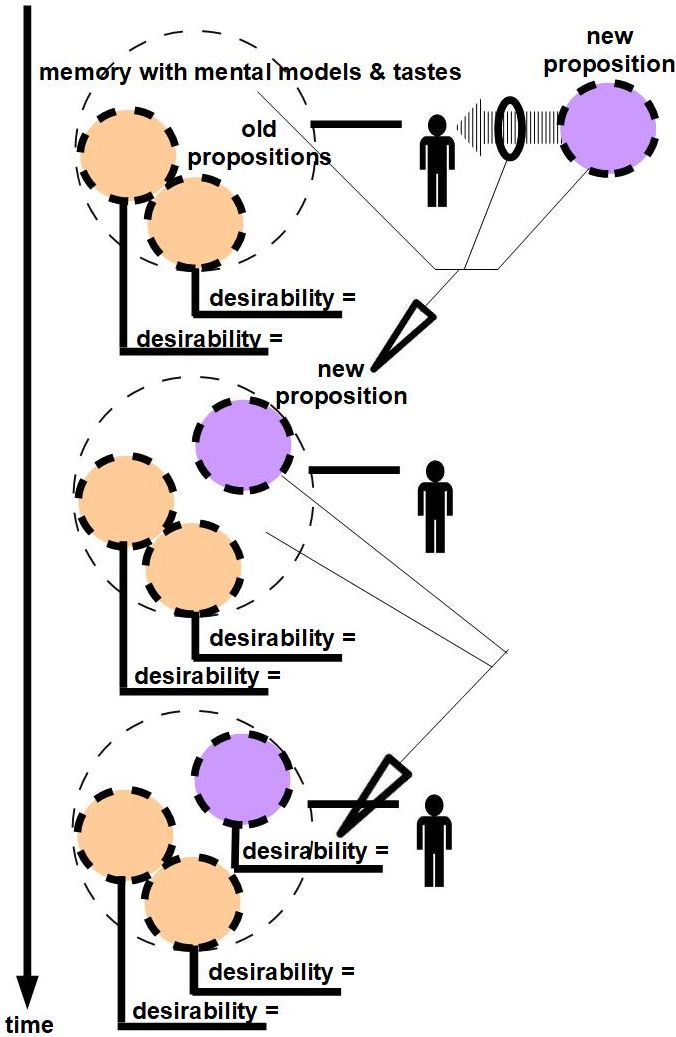}
	\caption{Desire: New proposition automatically enters memory, and is automatically assigned a desirability level.}
	\label{fig:pa-desire}
	\end{center}
\end{figure}

Figure~\ref{fig:pa-belief} and Figure~\ref{fig:pa-desire} demonstrate "belief" and "desire," two of the most common propositional attitudes in the English language. Memory is shown as a Data C Object Circle since memory is a virtual phenomenon. (If desired, memory could be shown housed inside of a skull that is represented by a Physical C Object Circle.) Memory is shown attached to the human icon with an Attribute Line, since memory is an attribute of humans. (If desired, a zoom ability could be used to position this memory inside the head of the human icon, where the skull is physically located, but the resulting diagram would be difficult to understand at a glance because of the very small sub-diagram.) Propositions are also virtual phenomena, so those are also shown as Data C Object Circles. Two causations are shown via Causation Arrows, one caused by a person's memory automatically storing a new proposition, another caused by a person's mental models automatically assigning a belief level to the newly stored proposition. Note that nearly the same situation occurs in the human memory ("memory with mental models and tastes"), regardless of which of these propositional attitudes is involved.

\textbf{3. WS150 example: \#87 (dishwasher)}

\begin{figure}
	\begin{center}
	\includegraphics[width=0.60\textwidth]{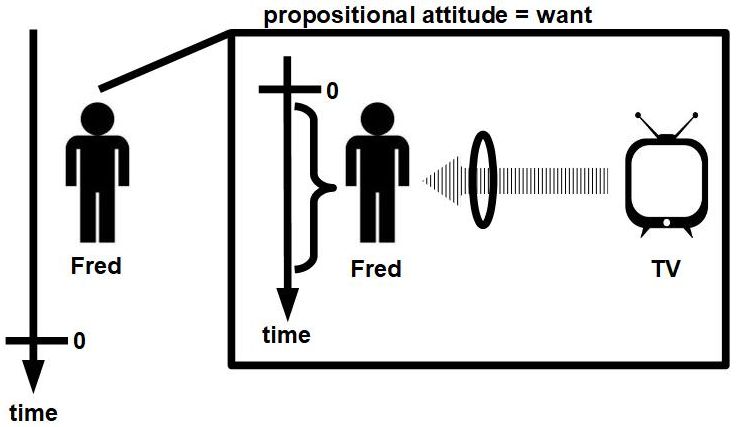}
	\caption{[87] "Fred wanted to watch TV."}
	\label{fig:ws-087-dishwasher}
	\end{center}
\end{figure}

Figure~\ref{fig:ws-087-dishwasher} shows Tumbug representation of a part of WS150 question \#87, which involves the propositional attitude "want." Note that the result resembles the method used to depict a personal fantasy that is used in Hollywood films, where the person doing the visualization is visualizing themself at an approximate time period while doing what they want to do. Often films use a wavy fluctuation of the fantasy enactment to clue the viewer that the scene is fantasy, often done as a smaller frame within the main frame, whereas in Tumbug the fantasy scenario is contained in an Aggregation Box.

\subsection{Tumbug as software}

Tumbug was intended to be a diagrammatic KRM that can be drawn in 2D, but Tumbug would greatly benefit if Tumbug were implemented as a software tool. Some desirable effects of software implementation of Tumbug would be:

\begin{itemize}
	\item
		Correlation Boxes could keep attribute values in sync, automatically, in the background.
	\item
		Motion Arrows could be eliminated because each intended moving object would actually move.
	\item
		Timelines could be eliminated because flow of time would already be present and obvious.
	\item
		Every intermediate position or state between two endpoints in time could be explicitly seen.
	\item
		Automatic storage of new experiences into memory could be implemented.
	\item
		Automatic assignment of propositional attitudes as attributes in memory could be implemented.
	\item
		Any static zoomed diagrams could be omitted because zooming would be available on demand.
	\item
		Icon libraries could be extensive and readily available, which would enhance understandability.
	\item
		Libraries of scenarios of common actions (e.g., walking, talking, buying) could save writing time.
	\item
		3D displays would be much more practical.
	\item
		Visual aids could be automatically placed on the diagram, such as to show alignment of objects, if desired.
	\item
		All diagrams created by a visual editor that are in a document could be easily updated in parallel, mostly automatically, to reflect a change in one diagram or in the text describing that diagram.
\end{itemize}

Tumbug is only a KRM and intended to be only a foundation for subsequent AGI architectures, but a software version of Tumbug that converted textual sentences into spatiotemporal icons might already have a few uses, such as the following:

\begin{itemize}
	\item
		A written story could be immediately simulated, especially when writing a movie script.
	\item
		Contradictions, especially in legal cases, might be detected via conflicting attribute values.
	\item
		An aid to search engines, by allowing objects and actions to be combined to narrow the search, or to search for a specific situation using general terms that would ordinarily return too many matches.
	\item
		An aid to foreign language translation programs.
	\item
		Children learning to read could immediately relate words to visual objects and visual motions.
\end{itemize}

\section{A few deep implications of Tumbug}

\subsection{SCOVA: an extension to OAV triples}

It is remarkable that all of Tumbug's Building Blocks generalize to only five Basic Building Blocks (BBBs):

\begin{itemize}
	\item
		(O) Object-like concepts. (objects and aggregations of objects)
	\item
		(A) Attribute-like concepts. (adjectives and adverbs)
	\item
		(V) Value-like concepts. (values, ranges of values, and wildcards)
	\item
		(C) Change-like concepts. (time, motion, causation, state changes, and functions).
	\item
		(S) System-like concepts. (systems composed of any mixed combination of the above types)
\end{itemize}

Since three of these concepts (O, A, V) are already addressed by standard OAV triples, this suggests that the OAV triple form of knowledge representation could be augmented if the other two concepts (C, S) of this document were added. With letters rearranged, this combination of concepts \{O, A, V, C, S\} can be spelled "SCOVA," which could be the name of a new KRM. The main difference between SCOVA and OOP would be that whereas change in OOP is often done by message passing between objects, change in SCOVA would be an additional component that would reduce message passing and would eliminate functions ("methods") defined inside of OOP software objects.

\begin{figure}
	\begin{center}
	\includegraphics[width=0.50\textwidth]{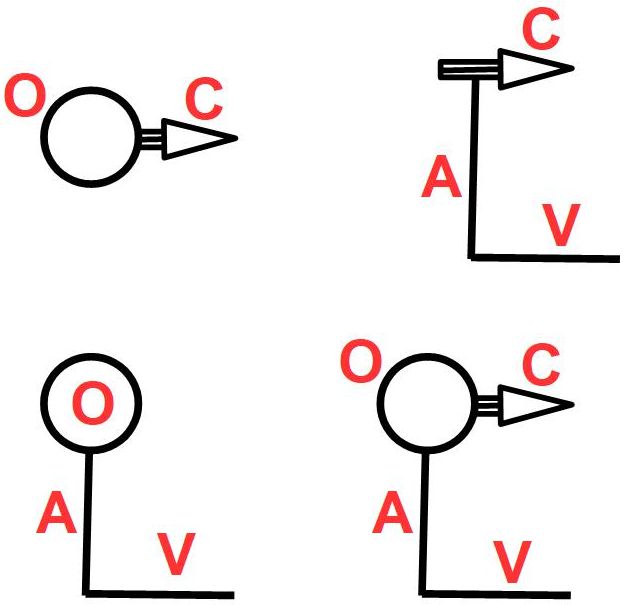}
	\caption{Some commonly mentioned, simple combinations of the SCOVA components. From left-to-right, starting on the top row, these are named in this document: OC, AVC, OAV, OAVC.}
	\label{fig:scova-combinations}
	\end{center}
\end{figure}

Some of the simplest, most common, most useful, and most meaningful combinations of these give SCOVA components mentioned in this document are shown in Figure~\ref{fig:scova-combinations}. The names used here attempt to keep the more familiar abbreviation "OAV" at the beginning of each longer abbreviation. "S" is not often used until Phase 2.

\textit{Side conjecture: Since the number of Basic Building Blocks of Tumbug, namely SCOVA, are five in number, and since the layers of the neocortex are six in number (I. molecular layer, II. external granular layer, III. external pyramidal layer, IV. internal granular layer, V. Internal pyramidal layer, and VI. multiform layer), it is conceivable that each layer of the neocortex is assigned one of the five components of SCOVA. If true, this would confirm the suspicion of Jeff Hawkins that the existence of those six layers is a major clue to intelligence (Hawkins 2004, p. 51, 69), since the implication would then be that the neocortex has somehow managed to implement what appear to be the five naturally arising building blocks of representation.}

\textit{Side conjecture: Since the number of Basic Building Blocks of Tumbug, namely SCOVA, are five in number, and since the number of main types of neocortical neurons are six in number (1. pyramidal cells, 2. fusiform cells, 3. stellate cells, 4. basket cells, 5. cells of Cajal-Retzius, and 6. cells of Martinotti), it is conceivable that each layer of the neocortex is assigned one of the five components of SCOVA.}

\subsection{Mathematical implications}

The main mathematical implication found during this study is that mathematics is a disembodied form of Tumbug.

The Correlation Boxes of Tumbug make clear the relationship between Tumbug and mathematics: mathematics is concerned only about the numerical values of attributes of objects, whereas Tumbug puts those numbers into context and gives them meaning with respect to the entire diagrammed system. This is why a single number in mathematics has no relationship to the real world unless it is associated with a variable that is defined, especially if units of measurement are supplied (e.g., centimeters, joules, years, radians, mean, etc.). This is also why mathematics is considered abstract: it strips away all meaning and structure from the real world except for numerical values. This in turn suggests that mathematics is the wrong tool for implementing AGI, since AGI must simulate the real world with all of its structures, complexities, fuzziness, and vagueness. This, in fact, is said to be exactly the reason that microworld research (Partridge 1991, p. 95) of early AI failed, which included famous systems such as the SHRDLU program (which operated in the virtual blocks world), Shakey the robot (which operated inside a special, uncluttered building), FREDDY the stationary robot (which operated on a steerable platform), the SAM program (which operated in the virtual world of textual stories), and the FRUMP program (which operated in the virtual world of wire-service news reports).

\begin{figure}
	\begin{center}
	\includegraphics[width=0.75\textwidth]{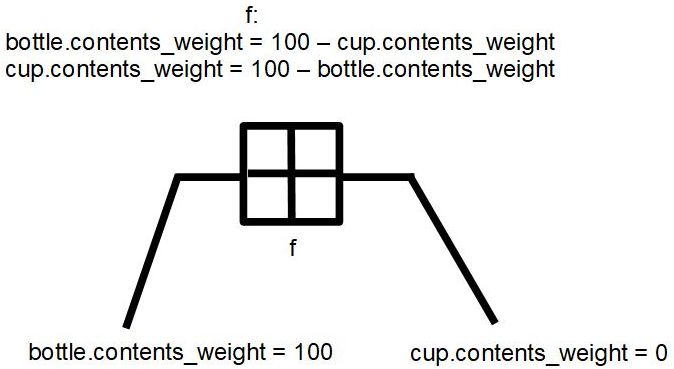}
	\caption{[24] A Tumbug diagram with a Correlation Box that has everything except the math stripped away.}
	\label{fig:icon-correlation-disembodied-math}
	\end{center}
\end{figure}

As an example, Figure~\ref{fig:icon-correlation-poured-water-abstract} is the earlier Tumbug representation of a bottle that poured water into a cup, and Figure~\ref{fig:icon-correlation-disembodied-math} shows the Tumbug equivalent of the same system with everything stripped away except for the mathematical components.

Note that any sense of time is removed (i.e., the timeline is removed), any sense of the meaning of the variables is removed (i.e., the objects and their inner partitions are removed), and the physical contact between the objects is removed (i.e., the stream of water is removed). All the mathematical system knows is the values for the variables, how to transform one variable's value into the other variable's value (via the function f), and maybe the variable names (which in the computer are ultimately only numerical memory addresses, not even text, and certainly the computer does not understand how the text relates to anything in the real world). These observations lend credence to the various claims that mathematics is the wrong tool for describing the real world, and therefore the wrong tool for AGI, e.g., (Devlin 1997, p. 283), (Wolfram 2002, p. 821).

\section{Insights from Tumbug}

\subsection{CD theory and Parts of Speech}

\begin{figure}
	\begin{center}
	\includegraphics[width=0.75\textwidth]{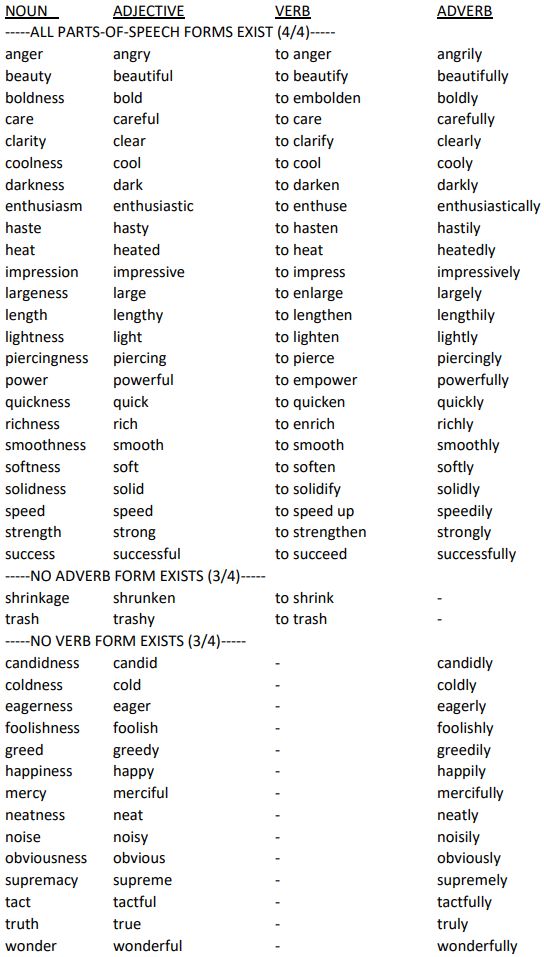}
	\caption{The rules of English often require a concept to change its concept name slightly, according to which Part of Speech it is adopting at any given moment.}
	\label{fig:thingification-pos-table-snap}
	\end{center}
\end{figure}

The core of CD theory, described earlier in this document, is Roger Schank's insight that oftentimes several similar words refer to the same concept, but the words differ only because of the Part of Speech for which they are used. (In this document on this topic, the only Parts of Speech of interest are noun, adjective, verb, and adverb.) For example, "quick" is the word form when used as an adjective, but "quickly" is the word form when used as an adverb, "quickness" is the word form when used as a noun, and "quicken" is the word form when used as a verb. Obviously the same concept and same root word are referenced, viz. "quick," but the word becomes distorted due only to English grammatical rules, a situation that is neither logical nor efficient when representing meaning in a diagrammatic form. (In speech, however, where a sentence diagram must be converted to a 1D temporal chain for audio communication, this is very efficient.) Some English words that function this way in at least three Parts of Speech are listed in the table in Figure~\ref{fig:thingification-pos-table-snap}.

\begin{figure}
	\begin{center}
	\includegraphics[width=0.50\textwidth]{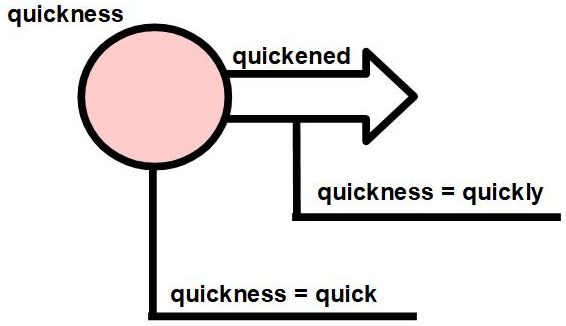}
	\caption{Tumbug for "Quick quickness quickly quickened," which is grammatically correct. Parts of Speech rules of English dictate that the concept's basic word must change form in many cases, however.}
	\label{fig:thingification-quick-quickness}
	\end{center}
\end{figure}

\begin{figure}
	\begin{center}
	\includegraphics[width=0.50\textwidth]{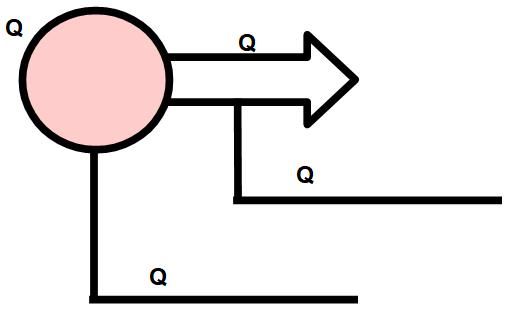}
	\caption{In a very general view, any single concept represented by variable "Q" can occupy any slot for any structure or attribute, regardless of Parts of Speech rules of English, and regardless of whether a word formally exists for every Part of Speech variation of that concept.}
	\label{fig:thingification-quick-q}
	\end{center}
\end{figure}

It is understandable that a natural language would need to modify the root word to succinctly flag its use in the sentence, otherwise a grammatically correct sentence such as "Quick quickness quickly quickened" (see Figure~\ref{fig:thingification-quick-quickness} would be rendered as "Quick quick quick quick." However, in a diagram that represents meaning, this is inefficient practice since essentially the same word appears in quadruplicate. This suggests an interesting insight about language: objects, actions, and attributes can be described by the same concept. This means that a single concept, say represented by variable Q, can appear as a description of any object, action, or attribute, as illustrated in Figure~\ref{fig:thingification-quick-q}.

\begin{figure}
	\begin{center}
	\includegraphics[width=0.75\textwidth]{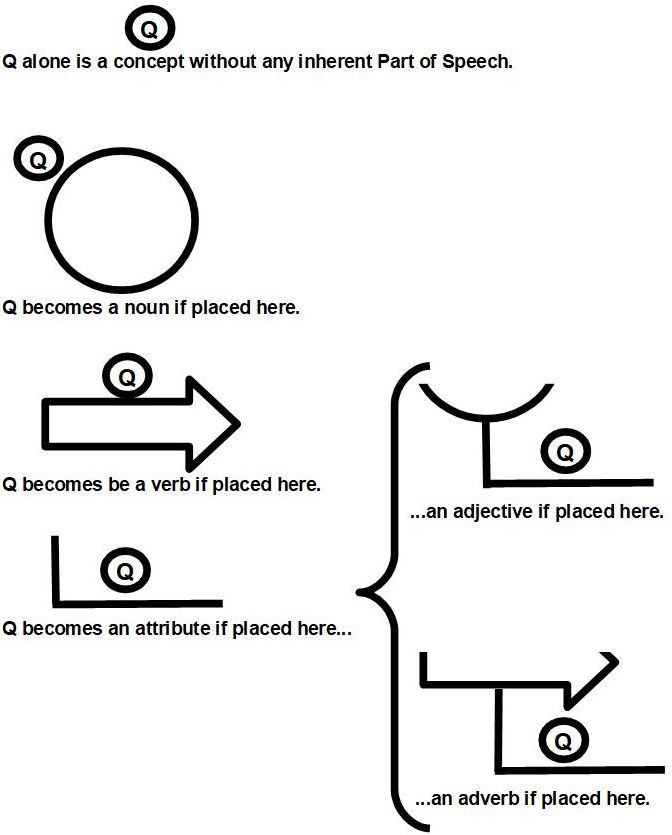}
	\caption{The Part of Speech associated with core concept Q is dependent only on the structure at which is appears.}
	\label{fig:thingification-pos}
	\end{center}
\end{figure}

The reverse situation generates another implication: The Part of Speech usage of concept Q depends only on the type of structure that Q is identifying. See Figure~\ref{fig:thingification-pos}. In particular: (1) If Q is identifying a given C Object Circle then Q must be rendered as a noun in spoken or written language. (2) If Q is identifying a given Motion Arrow, then Q must be rendered as a verb in spoken or written language. (3) If Q is identifying a given Attribute Line then Q must be rendered as an adjective or adverb, depending only on whether the Attribute Line is attached to an C Object Circle or Motion Arrow, respectively. Otherwise, Q is one of the unlabeled core concepts of which Schank described as the foundation of CD theory. 

There exists at least one caveat to the above observation, however: Since Motion Arrows are 2D rather than 1D, it must be specified whether the action described by a Motion Arrow is moving toward versus away from the state Q specified. In English the \textit{de facto} convention is clear: Motion toward state Q is always implied, never in the reverse direction. For example, the verb "to embolden" means to become bold, not to stray from being bold. Similarly, the verb "to beautify" means to make something beautiful, not to cause something to be no longer beautiful.

\section{Pros and cons of Tumbug}

\subsection{The main strengths of Tumbug}

\begin{itemize}
	\item
		Tumbug is likely the first purely visual representation ever used in computer science.\\
	\item
		Tumbug initiates a totally new, unexplored category of WS and CSR approaches.\\
	\item
		Tumbug clearly distinguishes between active tense and passive tense.\\
	\item
		Tumbug clearly distinguishes between simple aspect and progressive aspect.\\
	\item
		Tumbug clearly distinguishes between prefect and perfect-progressive aspect.\\
	\item
		Tumbug has better representational power than CD theory.\\
	\item
		Tumbug appears to implement the "universal language representation" needed in natural language processing (NLP).\\
	\item
		Tumbug closely approaches the ability of a visual simulator, therefore probably also closely approaches the way that humans think about events.\end{itemize}

\subsection{Possible weaknesses of Tumbug}

\begin{enumerate}

	\item
		It is not clear in a Tumbug diagram with a timeline which phases of the event should be documented, especially in how much detail. For example, an eating (INGEST) action for a human usually involves details such as chewing and swallowing, but these details are probably not important enough to document. As another example, reaching the top of the stair for a human usually involves traversing each step along the way, which is likely an irrelevant detail for most purposes.
	\item
		The functions currently allowed in the Correlation Boxes are not required to be invertible functions. For example, the allowed function y = $x^{2}$ over the real numbers is not invertible because then x = $\pm \sqrt{y}$. This allowance is currently assumed to be necessary since general intelligence would need to be able to access the total picture of how a function (or system) behaves, and would presumably be able to compensate for the many-to-one maps, or to decide on its own which values to accept from multiple possibilities. However, some drawbacks of allowing arbitrary functions are: (1) increased complexity of the mapping, (2) lack of clear-cut direction of correlation (viz., positive or negative), (3) possible inability of the system to make rapid, accurate guesses about the type of correlation involved between newly encountered variables because of the extra caution the system must exercise in order to remain general, (4) possible inability of the system to fill in missing functions by unsupervised examination of its own newly stored knowledge.
	\item
		There can probably never be a canonical form of every sentence that is to be rendered in Tumbug. Even if the shortcut notations, such as for motion or neglect of a C Aggregation Box, are ignored, there still may not be a clear-cut standard form for some reason. There may be, but this is not clear. A convincing argument that canonical forms are difficult for any KRM to produce is given by William A. Woods (Woods 1975, pp. 16-21).
	\item
		Several problematic situations occur when attempting to represent situations spatially. For example, how would one diagram "a small cluster of points in time" along a timeline without showing an exact number of points in the representation, and without showing their exact temporal order? Or how would one diagram a list of unsorted numbers in a generic way, without showing a specific number of numbers, or any of their values? How would one show in a generic way that a given set of jigsaw puzzle pieces needed to be placed together tightly but without forcing, and without showing the number of puzzle pieces or their shapes? The author has solved a number of such representational problems, therefore Tumbug can represent at least some such complications, but that topic is out of scope for this document. More generally, such problems qualify as representation of mathematical concepts, which has been barely addressed in this study.
	\item
		Some Tumbug icons are difficult to generalize easily from 1D to 2D. For example, a 1D sound broadcasting line is difficult to draw as fanning out into 2D space from a point. Similarly, a moving appendage such as an arm is difficult to describe with straight arrows, since most likely the appendage will swing on joints, which implies usage of arrows swinging through an angle. Most such problems could be resolved sufficiently with a software implementation of Tumbug without too much difficulty, but such software does not exist yet.
	\item
		The differences and overlap of action verbs and stative verbs is not completely clear. For example, the verb "to eat" would probably be most accurately classified as an action verb since eating involves food moving to the mouth, the motions of chewing and swallowing, and the food moving to the stomach. However, from a wider perspective the detailed motions of eating are not very important. For example, some animals such as starfish eat by egesting their stomachs around their victims instead of using their mouths to hold and chew food, so overall the act of eating is to change the state of the eater and food so that the food becomes located inside the eater for nutrition purposes. Also, a phrase such as "the photocopier ate my document" is immediately understandable, despite not involving any living organism, mouth, or stomach. The author's prediction is that Marvin Minsky may have been correct: it is very possible the brain uses several types of KRMs, including motion diagrams, state diagrams, and verbatim images, and that the brain links these representations in a coordinated way through learning.
\end{enumerate}

\subsection{Programs writing programs}

\begin{figure}
	\begin{center}
	\includegraphics[width=0.75\textwidth]{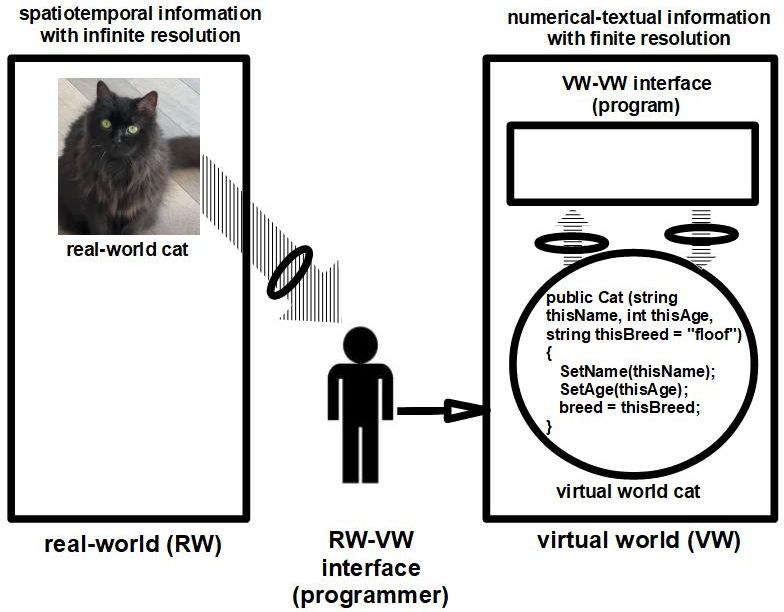}
	\caption{Programs struggle to write other programs partly due to their lost spatiotemporal information. As a result, programs need a programmer to cause a program to exist in the virtual world, based on the programmer's ability to understand the real world well enough to abstract it and to represent that abstraction.}
	\label{fig:self-programming-programmer}
	\end{center}
\end{figure}

A KRM like Tumbug is probably the most promising approach to provide a future means for machines to write computer programs or self-modifying code because Tumbug's ability to store spatiotemporal images verbatim reduces information lost in the transition from the real world to the virtual world. Currently, the real world consists of spatiotemporal images that need to be converted to a numerical-textual format that the virtual world of digital computers can store and manipulate easily, which typically requires time to be discretized, images to be discretized, details to be omitted that initially seemed irrelevant to the programmer (but that may be deemed important later), shape information to be lost, and cause-effect information (that only a human would understand well) to be lost, as shown in Figure~\ref{fig:self-programming-programmer}. Currently, computers cannot truly understand the real world, even when outfitted with video cameras, radar, and lidar. Even with multisensory input, computers cannot make commonsense conclusions about occluded objects or about interactions between physical objects. Computers also suffer additional problems from multisensory integration.

\begin{figure}
	\begin{center}
	\includegraphics[width=0.75\textwidth]{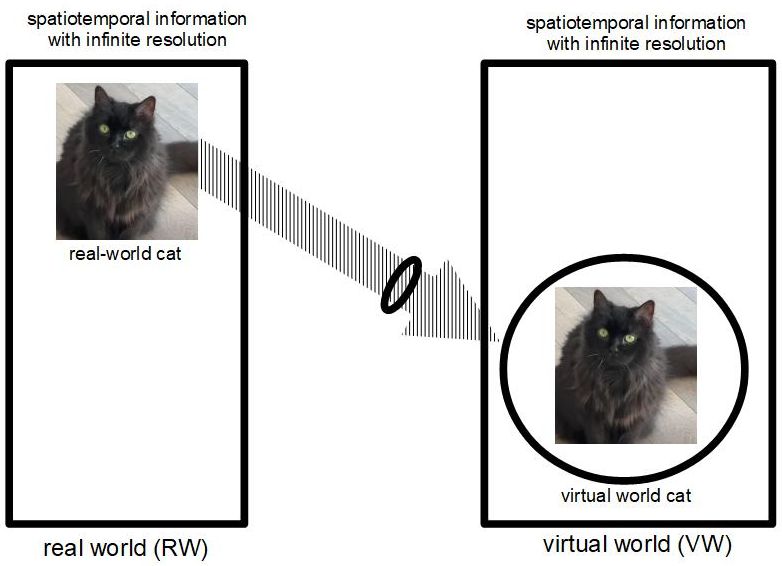}
	\caption{Programs whose representation matched that of the real world would not need a programmer.}
	\label{fig:self-programming-lossless}
	\end{center}
\end{figure}

\begin{figure}
	\begin{center}
	\includegraphics[width=0.75\textwidth]{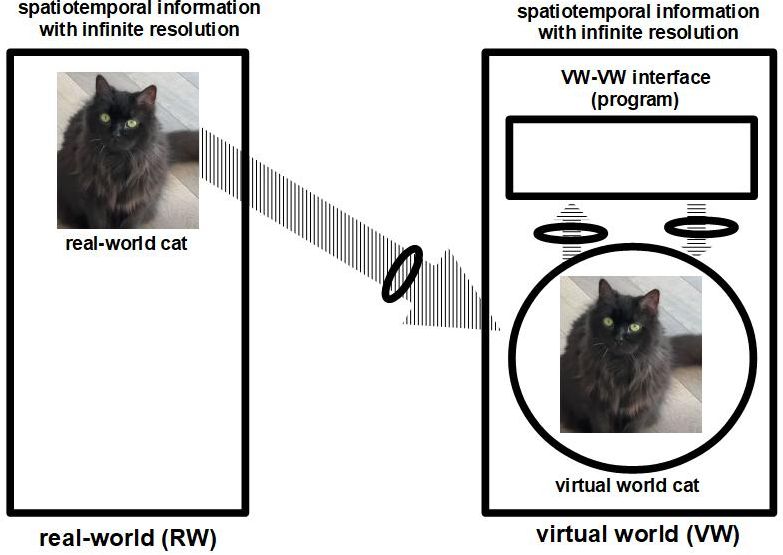}
	\caption{A virtual world that matched the real world could host a program to write other programs.}
	\label{fig:self-programming-programmerless}
	\end{center}
\end{figure}

In contrast, a machine that could store information in the same spatiotemporal organization as the real world would automatically eliminate the programmer, at least as a data type translator, as shown in Figure~\ref{fig:self-programming-lossless}. That would free the programmer to focus on writing behavioral programs instead of also needing to convert between data types. The language of the real world obviously consists of spatiotemporal images across various sensory modalities, not abstractions like numbers, text, and algorithms. An immediate consequence of having such an internal representation that matched the representation of the real world is that presumably a program could then write other programs as shown in Figure~\ref{fig:self-programming-programmerless}, which could be considered the ultimate goal of computer programming.

\section{Discussion of Tumbug}

The two main categories of approaches to ANI are symbolic AI and subsymbolic AI. Symbolic methods refer to text and are typically implemented with rule-based expert systems, logic programming, and semantic nets. Subsymbolic methods refer to non-text are usually implemented with neural networks. Tumbug is closer to a symbolic approach, but uses icons instead of symbols. Since Tumbug's icons have great freedom of placement, unlike text, Tumbug is more flexible than text, and therefore more powerful than text for representation. Thus Tumbug represents a new category of approach that has not been pursued, which in itself seems promising.

Was the design of Tumbug intended to imply that the brain is literally moving around circular icons? No. The circular icons are merely the simplest pictorial concept of a generic "thing." In practice all that would be needed to implement a circular icon would be a fuzzy cluster of one to several neurons whose summarized attributes include coordinates of the thing they represent, combined with a possibly crude method of distinguishing one location (or state) from another. In this sense, Tumbug circular icons could be considered part of a fuzzy system, as in fuzzy logic, where the exact size, shape, boundaries, and curvature of the icons are irrelevant.

Tumbug can be considered a fuzzy system, but the fuzziness is created through simpler diagrams, not membership functions. In a sense, $\langle$percentage of fuzziness$\rangle$ = 1 - $\langle$percentage of detail present in a diagram$\rangle$).

Is not the design of Tumbug's representation of modal verbs and emotions spatially arbitrary? Yes, but so are biological systems. The same patch of skin with pressure sensors can be placed anywhere on the surface of the body, but in one position it could represent pressure on a foot whereas in another position it could represent pressure on the neck. In biological systems the only difference in meaning is based on that patch's location on the body map that is stored in the cortex. In a computer, labels substitute for locations in such a body map. For virtual concepts such as modal verbs, such a design suggests that humans can physically "feel" the difference between two different modal verb meanings using the same hardware that humans use to feel the difference between two physical body locations.

The late Marvin Minsky may or may not have agreed with the approach being used by Tumbug. Although Minsky strongly advocated emphasis on KRMs as the foundation of AGI, his recommended approach was to \textit{combine existing} KRMs and to focus on the management system that would coordinate those KRMs (GBH Archives 1990), whereas the Tumbug approach has so far been to find a single, universal KRM that could represent every possible concept, and to eschew the unnatural, nonvisual KRMs that have been developed so far for use only in digital computers. In Pei Wang's terminology, Minsky's approach may have tended toward the hybrid approach whereas the Tumbug approach definitely takes the unified approach. However, Minsky's opinion still retains its merit: note that Tumbug has naturally gravitated toward appending specific computer science KRMs such as SCOVA + state diagram as in Figure~\ref{fig:cdt-atrans-mtrans-with-states-ANN2}, and state diagram + image as in Figure~\ref{fig:icon-state-traffic-light-american-with-object}. If such lower level KRMs are what Minsky envisioned, then Tumbug is already tending toward Minsky's envisioned "missing link" of AGI.

\textit{Conjecture: Humans have a part of the brain that allows humans to feel virtual concepts in the same way that a part of the brain feels physical sensations. This conjecture fits Jeff Hawkins' mentioned observation that the same cortical structure is used throughout the brain, regardless of sensory modality. Tumbug suggests that this observation can be generalized to state that the same cortical structure might be used for virtual concepts as is used for sensory modality, i.e., that understanding a virtual concept is architecturally equivalent to understanding a sensory modality.}

\section{Summary}

The main premise of this project is that if the proper KRM is used then the most difficult parts of AGI system design are automatically and greatly simplified. The proposed KRM for that purpose is a visual, iconic system called Tumbug that was designed by the author. Tumbug is the research topic of this Phase 1, which is part of a larger research project that aims at producing full AGI. The full envisioned research project has five phases that are collectively called The Visualizer Project. Older but unpublished results from initial excursions into Phase 2 and Phase 3 algorithms appear to confirm the premise that Tumbug is a promising foundation for AGI. Tumbug consists of about 30 Building Blocks (icons) that generalize to only five Basic Building Blocks, which turn out to be essentially the three components of OAV triples, namely Object (O), Attribute (A), and Value (V), plus two additional components called Change (C) and System (S). Change is a general concept that includes concepts such as time, motion, force, and cause-and-effect, and System is any legally assembled combination of Tumbug's Building Blocks.

By using a single pictorial KRM that consists of only about 30 Building Blocks, Tumbug can represent an extremely wide span of topics in mathematics, logic, algorithms, physics, human motivations, and human activities. Tumbug is a radically different type of KRM in that Tumbug can be made fully visual without use of text or numbers, which sets Tumbug apart from all other AI approaches such as expert systems, neural networks (whose units require labeling with text), cellular automata (whose units require labeling with text), and any variations of these. In short, Tumbug can probably visually represent any topic or concept that the average person is likely to encounter. This unified representation method could allow various computer systems, whether based on natural language, data base records, numbers, computer code, knowledge graphs, or images, to be integrated without the need for conversion at their interfaces. Several examples taken from sentences of the Winograd Schema are shown. For additional practical applications, Tumbug seems to be particularly well-suited for natural language translation, and possibly also for universal computer code representation in the style of .NET.

Tumbug appears to be an improvement on the Conceptual Dependency KRM of Roger Shank in that Tumbug's Building Blocks are far more universal and general, such as objects, motions, and attributes rather than detailed and largely human-based actions such as ingest, expel, and grasp. Tumbug could be the basis of "The Language of Thought" that Jerry Fodor conjectured in 1975, a hypothesized language often called "Mentalese."

\section{Future Research Plan}

\subsection{The Visualizer Project}

A "visualizer" is defined here as a type of processor whose native KRM is imageoids, in contrast to a "computer," whose native KRM is numbers. Therefore computers and visualizers are different processing architectures that each have their own unique niche of problems that they can solve particularly competently. 

\begin{figure}
	\begin{center}
	\includegraphics[width=0.95\textwidth]{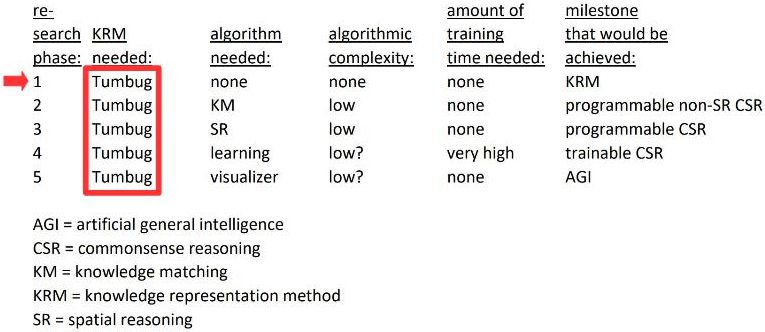}
	\caption{The research plan. This document completes Phase 1. The developed KRM from this phase (viz., Tumbug) will be used in all expected future phases of this research.}
	\label{fig:decomposition-csr-list}
	\end{center}
\end{figure}

This article completes and documents Phase 1 of a 5-phase research project called the Visualizer Project that aims to produce AGI in the form of a design for a visualizer. Tumbug is the foundation of all the phases that are expected to follow. The author believes that the design of a \textit{bona fide} AGI system will be produced by the end of the fifth phase, though probably not yet coded at that time. Figure~\ref{fig:decomposition-csr-list} shows all the anticipated phases of the Visualizer Project.

One inherent limitation of Phase 1 is it is difficult to state when a collection of Building Blocks is "complete" because there is no theoretical limit to the number of Tumbug Building Blocks, only a practical limit, usually based on the frequency of usage within a given domain, and also there has not been a need in Phase 1 to fully explore some recently introduced Building Blocks. A few Tumbug Building Blocks were introduced in this document largely because they are expected to be needed sometime in future, especially for spatial reasoning (especially Correlation Boxes and 1D Markers), for reducing textual descriptions from Tumbug (especially Value Bars), for coding implementation (especially 0D Markers and Zoom Boxes), for a learning algorithm (especially Data Set Boxes), or for mathematical theorems (especially C-A Aggregation Boxes), but so far there has not been sufficient need to justify the time required to thoroughly explore those constructs.

It is expected that the need for several new Building Blocks will arise from the need to represent certain concepts from mathematics, especially from calculus (concepts such as "continuous" and "limit"), group theory, topology (concepts such as "connectedness"), probability (concepts such as "likelihood," "heuristics," "permutation," and "combination"), statistics (concepts such as "distribution"), algebra (concepts such as "invertible function"), and geometry (concepts such as "alignment," "area," "volume," and "dimension"), as well as from puzzle solving (concepts such as "ordered," "random," "path," and "to flush fit") and science/engineering topics in general (concepts such as "density," "cluster," "complexity," "independent," "coupled," and "representation").


\subsection{Roadmap of the next phases}

The Visualizer Project is not just a theoretical conjecture with no results and no motivating clues. The following subsections describe existing documentation, existing discoveries, and expectations of the next phases.

\subsubsection{Phase 2 (non-spatial reasoning)}

The basic algorithm for Phase 2 was already developed and documented in 2022 (Atkins 2022) with several complete examples diagrammed step-by-step during the CSR solution process. That article was not accepted for publication, however. In that year the author believed that the Tumbug-SOAV approach was already capable of solving 82\% of WS150 problems, namely all of WS150's Non-spatial Reasoning Problems. Now that an extensive Tumbug introduction exists in the form of this Phase 1 document, presumably the older article describing the algorithm needs only to be rewritten with finer details, preferably also with more examples, without needing to include a description of Tumbug as well, in order to complete Phase 2.

\subsubsection{Phase 3 (spatial reasoning)}

A few Phase 3 results already exist, as well. The 2022 Tumbug article already mentioned the discovery that two of the nine Spatial Reasoning Non-Algorithmic problems from WS150 appear to be solvable by a more generalized version of Tumbug. If this generalization can be generalized even further, then the 82\% solvability of WS150 problems by Tumbug would increase to at least 97\%, probably to 100\%. An article documenting the more general algorithm would likely use the same organization as a newly revised Phase 2 article.

\subsubsection{Phase 4 (learning)}

It is likely that the needed learning algorithm will ultimately be based on the Descriptive Boxes described in this document. A bridge needs to be made from the real world, currently representable in Tumbug by Verbatim Boxes, to the semi-computerized format of Descriptive Boxes, in order for Tumbug to work directly with real-world data without the need of a programmer. The author has already developed visual versions of several spatial prepositions such as "near," "above," "to," and "inside," so this phase definitely has the potential to move quickly, as well.

\subsubsection{Phase 5 (visualizer)}

A visualizer will need several additional components and techniques even if trainable CSR were produced. The author has already published one of these techniques and has publicly discussed a second technique that is believed to be needed. Additional components of a visualizer were already described in a research proposal written by the author in 2014, but that proposal was not awarded.
	
\section{References}

\setlength\parindent{0pt}

Allerton, D.J. 2006. "Verbs and their Satellites." In \textit{The Handbook of English Linguistics}, edited by Bas Aarts and April McMahon, 146-179. Malden, MA: Blackwell Publishing.\\

Atkins, Mark. 2000. "S-96: A Semantic Net Implemented With Synchronized Neurons for Binding and Inferencing." Ph.D. dissertation, Florida Tech, Melbourne, Florida.\\

Atkins, Mark. 2019. "Two Approaches Toward Graphical Definitions of Knowledge and Wisdom." DaKM 2020 conference, Vienna, Austria.\\

Atkins, Mark A. 2022. "A promising visual approach to solution of 82\% of Winograd Schema problems via Tumbug Visual Grammar." Unpublished.\\

Atkins, Mark A. 2023. \textit{Important things to know before learning any foreign language}. Unpublished.\\

Baez, Albert V. 1967. \textit{The New College Physics: A Spiral Approach}. San Francisco: W. H. Freeman and Company.\\

Bailey, James. 1996. \textit{After Thought: The Computer Challenge to Human Intelligence}. New York, NY: BasicBooks.\\

Beckmann, Petr. 1971. \textit{A History of Pi}. New York: The Golem Press.\\

Berthoz, Alain. 2000. \textit{The Brain's Sense of Movement}. Cambridge, Massachusetts: Harvard University Press.\\

Brooks, Rodney A. 1992. "Intelligence without representation." In \textit{Foundations of Artificial Intelligence}, edited by David Kirsh, 139-159. Cambridge, Massachusetts: The MIT Press.\\

Cercone, Nick, and Gordon McCalla. 1987. "What is Knowledge Representation?" In \textit{The Knowledge Frontier: Essays in the Representation of Knowledge}, edited by Nick Cercone and Gordon McCalla, 1-43. New York: Springer-Verlag.\\

Chang, Le, and Doris Y. Tsao. 2017. "The code for facial identity in the primate brain." Cell. 2017 June 01; 169(6): 1013–1028.e14.\\

Chubb, Charles. 1995. "Motion Perception." In \textit{Early Vision and Beyond}, edited by Thomas V. Papathomas, 109-112. Cambridge, Massachusetts: The MIT Press.\\

Coppin, Ben. 2004. \textit{Artificial Intelligence Illuminated}. Sudbury, Massachusetts: Jones and Bartlett Publishers.\\

Davis, Ernest. 1990. \textit{Representations of Commonsense Knowledge}. San Mateo, California: Morgan Kaufmann Publishers.\\

Davis, Ernest. 2018. Collection of Winograd Schemas.\\

Davis, Ernest, and Gary Marcus. 2015. "Commonsense reasoning and commonsense knowledge in artificial intelligence." Communications of the ACM, 58(9), 92-103. https://doi.org/10.1145/2701413.\\

Devlin, Keith. 1997. \textit{Goodbye, Descartes: The End of Logic and the Search for a New Cosmology of the Mind}. New York: John Wiley \& Sons.\\

Dreyfus, Herbert L. 1979. \textit{What Computers Can't Do, Revised Edition: The Limits of Artificial Intelligence}. New York, N.Y.: Harper \& Row, Publishers.\\

Eade, James. 1996. \textit{Chess For Dummies}. Foster City, CA: IDG Books Worldwide.\\

Fischler, Martin A., and Oscar Firschein. 1987. \textit{Intelligence: The Eye, the Brain, and the Computer. Reading, Massachusetts: Addison-Wesley Publishing Company}.\\

GBH Archives. 1990. “The Machine That Changed The World: Interview with Marvin Minsky.”\\

Gleick, James. 1987. \textit{Chaos: Making a New Science}. New York, New York: Viking Penguin.\\

Gosling, James, Bill Joy, Guy Steele, and Gilad Bracha. 1996. \textit{The Java Language Specification, Second Edition}. Boston: Addison-Wesley.\\

Harris, Laurence R., and Michael R.M. Jenkin. 1997. "Computational and psychophysical mechanisms of visual coding." In \textit{Computational and Psychophysical Mechanisms of Visual Coding}, edited by Michael Jenkin and Laurence Harris, 1-19. New York, NY: Cambridge University Press.\\

Haugeland, John. 1985. \textit{Artificial Intelligence: The Very Idea}. Cambridge, Massachusetts: The MIT Press.\\

Hawkins, Jeff. 2004. \textit{On Intelligence}. New York: Times Books.\\

Hofstadter, Douglas R. 1979. \textit{Gödel, Escher, Bach: an Eternal Golden Braid}. New York: Basic Books.\\

Hogan, James P. 1997. \textit{Mind Matters: Exploring the World of Artificial Intelligence}. New York: The Ballantine Publishing Group.\\

Huang, Bin, Siao Tang, Guangyao Shen, Guohao Li, Xin Wang, and Wenwu Zhu. 2020. "Commonsense Learning: An Indispensable Path towards Human-centric Multimedia." HuMA’20, October 12, 2020, Seattle, WA, USA.\\

Jenkin, Heather L. 1997. "A historical review of the imagery debate." In \textit{Computational and Psychophysical Mechanisms of Visual Coding}, edited by Michael Jenkin and Laurence Harris, 268-295. New York, NY: Cambridge University Press.\\

Kaku, Michio. 2011. \textit{Physics of the Future: How Science Will Shape Human Destiny and Our Daily Lives By the Year 2100}. New York: Doubleday.\\

Khinchin, A. Ya. 1964. \textit{Continued fractions}. Chicago: University of Chicago Press.\\

Kocijan, Vid, Ernest Davis, Thomas Lukasiewicz, Gary Marcus, and Leora Morgenstern. 2022. "The Defeat of the Winograd Schema Challenge." https://arxiv.org/abs/2201.02387 (accessed September 3, 2022)\\

Kolln, Martha, and Robert Funk. 2006. \textit{Understanding English Grammar, Seventh Edition}. New York: Pearson Education.\\

Kosslyn, Stephen M., William L. Thompson, and Giorgio Ganis. 2006. \textit{The Case for Mental Imagery}. New York: Oxford University Press.\\

Kurzweil, Raymond. 1990. \textit{The Age of Intelligent Machines}. Cambridge, Massachusetts: Massachusetts Institute of Technology.\\

Kurzweil, Ray. 1999. \textit{The Age of Spiritual Machines: When Computers Exceed Human Intelligence}. New York, New York: Viking Penguin.\\

Lettvin, J. Y. H. R. Maturana, W. S. McCulloch, and W. H. Pitts. 1959. "What the Frog's Eye Tells the Frog's Brain." Proceedings of the IRE (Volume: 47, Issue: 11, November 1959).\\

Luger, George F., and William A. Stubblefield. 1998. \textit{Artificial Intelligence: Structures and Strategies for Complex Problem Solving, Third Edition}. Reading, MA: Addison Wesley Longman.\\

Lytinen, Steven L. 1992. "Conceptual Dependency and its Descendants." Computers \& Mathematics with Applications, Vol. 23, No. 2-5, pp. 51-73. Great Britain: Pergamon Press.\\

MacLean, Paul D. 1990. \textit{The Triune Brain in Evolution}. New York: Plenum Press.\\

Maor, Eli. 1994. \textit{e: The Story of a Number}. Princeton, New Jersey: Princeton University Press.\\

Markman, Arthur B. 1999. \textit{Knowledge Representation}. Mahwah, New Jersey: Lawrence Erlbaum Associates.\\

Marr, David. 1982. \textit{Vision: A Computational Investigation into the Human Representation and Processing of Visual Information}. Cambridge, Massachusetts: The MIT Press.\\

Melzack, Ronald, and Patrick D. Wall. 1965. "Pain Mechanisms: A New Theory: A gate control system modulates sensory input from the skin before it evokes pain perception and response." Science, Vol. 150, Issue 3699, 19 Nov 1965, pp. 971-979, DOI: 10.1126/science.150.3699.97.\\

Minsky, Marvin. 1974. A Framework for Representing Knowledge. Memo No. 306.\\

Minsky, Marvin. 1986. \textit{The Society of Mind}. New York: Simon and Schuster.\\

Newell, Allen. 1990. \textit{Unified Theories of Cognition}. Cambridge, Massachusetts: Harvard University Press.\\

O'Neill, Gerard K. 1981. \textit{2081: A Hopeful View of the Human Future}. New York: Simon and Schuster.\\

Orban, Guy A. 2008. "Higher Order Visual Processing in Macaque Extrastriate Cortex." \textit{Physiological Reviews}. January 2008, pp. 59-89.\\

Partridge, Derek. 1991. \textit{A New Guide to Artificial Intelligence}. Norwood, New Jersey: Ablex Publishing Corporation.\\

Pezzulo, Giovanni. 2007. "Anticipation and Future-Oriented Capabilities in Natural and Artificial Cognition." In \textit{50 Years of Artificial Intelligence: Essays Dedicated to the 50th Anniversary of Artificial Intelligence}, edited by Max Lungarella, Fumiya Iida, Josh Bongard, and Rolf Pfeifer, 257-270. Berlin, Germany: Springer-Verlag.\\

Reichgelt, Han. 1991. \textit{Knowledge Representation: An AI Perspective}. Norwood, New Jersey: Ablex Publishing Corporation.\\

Reisberg, Daniel, and Fridrike Heuer. 2005. "Visuospatial Images." In \textit{The Cambridge Handbook of Visuospatial Thinking}, edited by Priti Shah and Akira Miyake, 35-80. New York, NY: Cambridge University Press.\\

Rieger, Chuck. 1975. "The Commonsense Algorithm as a Basis for Computer Models of Human Memory, Inference, Belief and Contextual Language Comprehension." Theoretical Issues in Natural Language Processing.\\

Robinson, David L. 2009. "Brain function, emotional experience and personality." The Netherlands Journal of Psychology. pp. 152–167.\\

Rorvig, Mordechai. 2021. "Supersized AI". \textit{New Scientist}. October 9, 2021, pp. 37-40.\\

Russell, Stuart J., and Peter Norvig. 2010. \textit{Artificial Intelligence: A Modern Approach, Third Edition}. Upper Saddle River, New Jersey: Prentice Hall.\\

Sabbatini, Renato M.E. 1998. "The mind, artificial intelligence and emotions: Interview with Marvin Minsky."\\

Schank, Roger C. 1972. "Conceptual Dependency: A Theory of Natural Language Understanding." Cognitive Psychology 3, 552-631.\\

Schank, Roger C. 1975. "The Primitive ACTs of Conceptual Dependency." In \textit{Theoretical Issues in Natural Language Processing}.\\

Schank, Roger C. 1976. "The Role of Memory in Language Processing." In \textit{The Structure of Human Memory}, edited by Charles N. Cofer, 162-189. San Francisco: W. H. Freeman and Company.\\

Shastri, Lokendra, and Venkat Ajjanagadde. 1990. “From simple associations to systemic reasoning: A connectionist representation of rules, variables and dynamic bindings.” Behavioral and Brain Sciences, 16(3), January 1990, pp. 417–494.\\

Simpson, Patrick K. 1990. "Neural Networks for Sonar Signal Processing," In \textit{Handbook of Neural Computing Applications}, edited by Alianna Maren, Craig Harston, and Robert M. Pap, 319-335. San Diego, California: Academic Press.\\

Sowa, John F. 2000. \textit{Knowledge Representation}. Pacific Grove, CA: Brooks Cole Publishing Co.\\

Tveter, Donald R. 1998. \textit{The Pattern Recognition Basis of Artificial Intelligence}. Los Alamitos, California: The Institute of Electrical and Electronics Engineers.\\

Winston, Patrick Henry. 1998. \textit{On To Smalltalk}. Reading, Massachusetts: Addison-Wesley.\\

Wolfram, Stephen. 2002. \textit{A New Kind of Science}. Champaign, IL: Wolfram Media.\\

Woods, W.A. 1975. What's in a link: Foundations for Semantic Networks. Bolt Beranek and Newman.\\

Woods, William A. 1986. "Important Issues in Knowledge Representation." Proceedings of the IEEE, 74(10), October 1986, pp. 1322-1334.

\setlength\parindent{24pt}

\end{document}